Handbook *of*

# Rough Set Extensions

## and Uncertainty Models

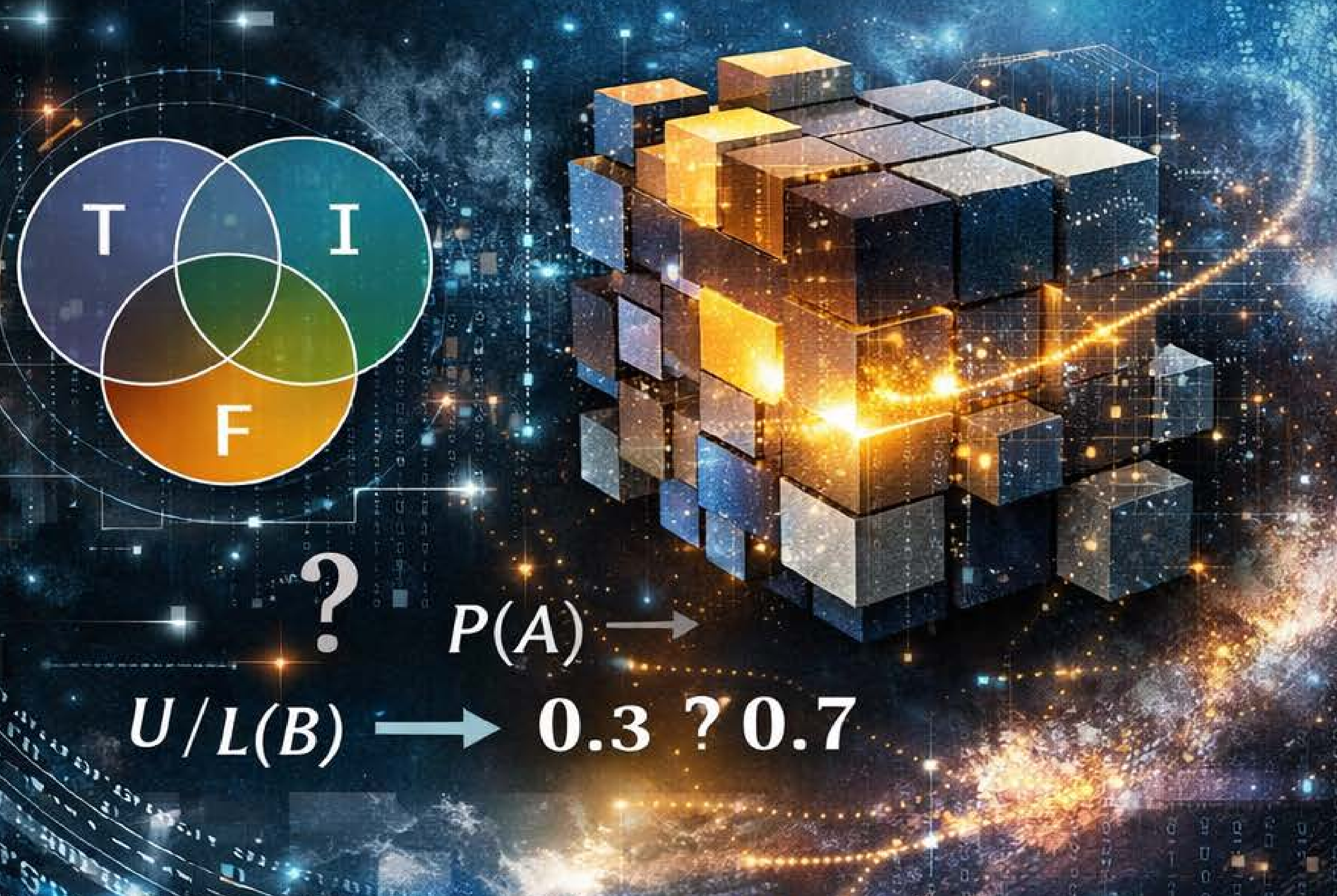

**Takaaki Fujita**
**Florentin Smarandache**

**Takaaki Fujita, Florentin Smarandache**

# Handbook of Rough Set Extensions and Uncertainty Models

*Editor:*

***Peer-Reviewers:***

**Maikel Leyva-Vázquez**
Universidad de Guayaquil, Guayas, ECUADOR
maikel.leyvav@ug.edu.ec

**Victor Christianto**
Malang Institute of Agriculture (IPM), Malang, INDONESIA
victorchristianto@gmail.com

# Contents in this book

The remainder of this book is organized as follows.

# Chapter 1

# Introduction

## 1.1 Uncertain Theory

Classical (crisp) set theory provides a precise and widely used language for formal reasoning and mathematical modeling [1]. Over the past decades, many generalized set frameworks have been introduced to represent uncertainty and vagueness, including fuzzy sets [2], intuitionistic fuzzy sets [3], hesitant fuzzy sets [4], picture fuzzy sets [5], single-valued neutrosophic sets [6, 7], quadripartitioned neutrosophic sets [8], pentapartitioned neutrosophic sets [9], double-valued neutrosophic sets [10], hesitant neutrosophic sets [11], rough sets [12,13], plithogenic sets [14,15], and soft sets [16, 17].

A fuzzy set assigns to each element $x$ a single membership grade $\mu(x) \in [0, 1]$, thereby capturing gradual inclusion rather than a sharp yes/no decision [2,18]. Neutrosophic sets extend this viewpoint by associating three (generally independent) degrees $T(x), I(x), F(x) \in [0, 1]$, interpreted as truth, indeterminacy, and falsity, respectively [6,19]. Because these models encode uncertainty more flexibly than crisp sets, they have been applied widely, for example in decision-making [20], and neural networks [21, 22].

## 1.2 Rough Set Theory

Rough set theory models uncertainty by approximating concepts with lower and upper sets induced by indiscernibility relations in data tables [12, 23]. Rough set research provides rigorous tools to model vagueness from indiscernibility, enabling data-driven approximations [24], feature reduction [25, 26], rule induction, and decision supports [27, 28], machine-learning [29], neural network [30, 31], engineering [32, 33], chemistry [34, 35], diagnostics [36], and explainability applications. For reference, a concise comparison between classical (crisp) sets and Pawlak rough sets is provided in Table 1.1.

Table 1.1: Concise comparison of classical (crisp) sets and Pawlak rough sets.

| Aspect | Classical (crisp) set | Rough set (Pawlak approximation) |
|---|---|---|
| Basic data | Universe $U$ and an exact subset $A \subseteq U$ | Approximation space $(U, R)$ where $R$ is an equivalence (indiscernibility) relation; target $Y \subseteq U$ |
| Representation of a concept | Single set $A$ (exact concept) | Pair $(\underline{Y}, \overline{Y})$ (lower/upper approximations) |
| Membership semantics | Binary: $x \in A$ or $x \notin A$ | Definite: $x \in \underline{Y}$ iff $[x]_R \subseteq Y$; Possible: $x \in \overline{Y}$ iff $[x]_R \cap Y \neq \varnothing$ |
| Uncertainty source | No intrinsic uncertainty (perfect discernibility assumed) | Indiscernibility of objects under available attributes / data tables |
| Boundary / vagueness | No intrinsic boundary region | Boundary $\mathrm{BND}(Y) = \overline{Y} \setminus \underline{Y}$ captures the vague/undecidable part |
| Definability criterion | Always exact as given | $Y$ is crisp/definable w.r.t. $R$ iff $\underline{Y} = \overline{Y}$ (equivalently, $Y$ is a union of $R$-classes) |
| Typical analysis outputs | Exact set operations, predicates, counts | Approximations/regions; often used for feature reduction and rule induction |

## 1.3 Our Contributions

In light of these developments, research on rough set theory remains important. Moreover, because a large number of papers on rough sets and their extensions continue to appear, survey-style works play an increasingly valuable role in organizing and clarifying the landscape. Motivated by this need, this book provides a survey-style overview of rough set theory and its major developments, with an emphasis on how rough approximations support data-driven tasks.

More concretely, the book is organized as follows:

- **Types of rough sets (Chapter 2).** We systematically summarize a broad spectrum of rough-set extensions, giving concise definitions and brief discussions to clarify how each model generalizes Pawlak's original framework.

- **Uncertain rough sets (Chapter 3).** We review representative hybridizations that incorporate graded or multi-component uncertainty, including fuzzy, intuitionistic fuzzy, neutrosophic, plithogenic, and related uncertain rough-set formulations.

- **Related concepts (Chapter 4).** We collect adjacent structures and viewpoints—such as rough graphs, rough topological spaces, rough groups, rough matroids, and soft rough graphs—to help readers connect rough approximations with other mathematical frameworks.

Overall, our goal is to provide a compact reference that helps readers navigate the rapidly expanding rough-set literature and quickly locate suitable models for theoretical study and practical applications. For reference, Table 1.2 presents an at-a-glance taxonomy for rough-set–based models.

In addition, for reference, Table 1.3 provides a practical selection guide on which rough-set family to use under typical data conditions.

Table 1.2: At-a-glance taxonomy for rough-set–based models (granulation, value semantics, outputs, and typical uses).

| **Axis** | **Main options (keywords)** | **Meaning / typical choice criteria** |
|---|---|---|
| **Granulation** | *equivalence*, *tolerance*, *covering*, *neighborhood*, *probabilistic* | Defines the indiscernibility/approximation mechanism. Equivalence (partition, crisp indiscernibility); Tolerance (similarity-based, reflexive/symmetric); Covering (non-partition granules, overlaps); Neighborhood (metric/graph/kNN style, local granules); Probabilistic (variable precision, error-tolerant approximations). |
| **Value semantics** | *crisp*, *fuzzy*, *intuitionistic fuzzy*, *neutrosophic*, *plithogenic*, *structure-valued* | Specifies how membership/uncertainty is represented. Crisp: $\{0,1\}$; Fuzzy: $\mu \in [0,1]$; Intuitionistic fuzzy: $(\mu, \nu)$ with hesitation; Neutrosophic: $(T, I, F)$ with explicit indeterminacy; Plithogenic: attribute-wise appurtenance + contradiction degree; Structure-valued: intervals/vectors/distributions/sets as labels. |
| **Outputs** | $(\underline{X}, \overline{X})$, *regions*, *reduct*, *rules* | What the model produces. Approximations: lower/upper $(\underline{X}, \overline{X})$; Regions: POS/BND/NEG (positive/boundary/negative); Reduct: feature/attribute reduction preserving discernibility; Rules: decision/association rules (often interpretable). |
| **Representative uses** | *feature selection*, *rule induction*, *ranking*, *streaming*, *classification* | Common application targets. Feature selection via reducts; rule induction for interpretable decision support; Ranking via dominance/ordering-aware rough sets; Streaming/incremental updates for dynamic data; Classification via approximations/regions/rules. |

Please note that this book is a survey that focuses primarily on theoretical aspects. We hope that future work by domain experts will advance algorithm design and other methodological developments, as well as practical studies using machine learning and related techniques.

Table 1.3: Practical selection guide: which rough-set family to use under typical data conditions.

| Data condition / requirement | Recommended direction (granulation / values / outputs) |
|---|---|
| Strict indiscernibility; categorical attributes; classical setting | equivalence + crisp; outputs: $(\underline{X}, \overline{X})$, regions; reduct for feature selection. |
| Similarity/approximate matching; noise; continuous attributes | tolerance or neighborhood; values: crisp/fuzzy; outputs: approximations, rules; optionally probabilistic for error tolerance. |
| Overlapping groups; multiple granular views; rule mining under overlaps | covering; values: crisp/fuzzy; outputs: regions + rules (covering-based rule induction). |
| Uncertainty beyond fuzziness (hesitation / inconsistency / indeterminacy) | choose value semantics: intuitionistic fuzzy / neutrosophic / plithogenic; keep granulation as tolerance/covering/neighborhood as needed. |
| Ranking/ordered decision criteria (e.g., credit scoring, risk assessment) | dominance / order-aware rough sets (often probabilistic/variable precision variants); outputs: ranking + rules. |
| Dynamic / streaming / incremental updates | neighborhood/covering with incremental approximations; outputs: incremental reduct / online rule updates. |

## Abstract

Rough set theory models uncertainty by approximating target concepts via *lower* and *upper* sets induced by indiscernibility (or more general granulation) relations in data tables. This perspective captures vagueness caused by limited observational resolution and supports *set-theoretic* reasoning about what can be determined with certainty versus what remains only possible.

The present book is written as a *model map*: rather than developing a single algorithmic pipeline in depth, we provide a systematic survey of the main rough-set paradigms and their extension routes. Concretely, we organize representative variants according to (i) the underlying granulation mechanism (e.g., equivalence-, tolerance-, covering-, neighborhood-, and probabilistic-based approximations) and (ii) the uncertainty semantics attached to data and relations (e.g., crisp, fuzzy, intuitionistic fuzzy, neutrosophic, and plithogenic settings), and we explain how each choice changes the form of approximations and the interpretation of boundary regions. Throughout, small illustrative examples are used to clarify modeling intent and typical use-cases in classification and decision support.

Finally, we note an important scope clarification: if the purpose of this book is limited to a *model map*, then the Abstract/Introduction should not lead readers to expect that *feature reduction* and *rule induction* are primary goals. Those topics are central in the rough-set literature, but here they are discussed mainly as motivating applications and as pointers to the broader ecosystem; the main objective is to align the book's stated aim with its actual focus on surveying and positioning rough-set models and extensions.

# Chapter 2

# Types of Rough Set

As types of rough sets, a wide variety of extended rough-set models have been proposed. In this chapter, we provide a survey-style introduction and brief discussion of these extensions.

## 2.1 Rough Set

Rough set theory approximates a target set via lower and upper approximations induced by equivalence classes, modeling vagueness and uncertainty [12, 37].

**Definition 2.1.1** (Rough Set Approximation)**.** [38] [12, 37] Let $X$ be a finite universe and let

$$R \subseteq X \times X$$

be an equivalence relation, whose equivalence classes are written $[x]_R$ for each $x \in X$. For any subset $Y \subseteq X$, define:

$$\underline{Y} = \{\, x \in X \mid [x]_R \subseteq Y \},\qquad \overline{Y} = \{\, x \in X \mid [x]_R \cap Y \neq \varnothing \}.$$

Here $\underline{Y}$ collects all elements whose entire indiscernibility class lies inside $Y$ (those that *definitely* belong), while $\overline{Y}$ gathers elements whose class meets $Y$ nontrivially (those that *possibly* belong). The pair $(\underline{Y}, \overline{Y})$ is called the *rough approximation* of $Y$, and satisfies

$$\underline{Y} \subseteq Y \subseteq \overline{Y}.$$

**Example 2.1.2** (Medical triage with incomplete symptom information)**.** Let $U = \{p_1, p_2, p_3, p_4, p_5, p_6\}$ be a set of patients. Consider two easily observed symptoms (condition attributes):

$$C = \{\mathsf{Fever}, \mathsf{Cough}\},$$

where $\mathsf{Fever} \in \{\mathsf{High}, \mathsf{Normal}\}$ and $\mathsf{Cough} \in \{\mathsf{Yes}, \mathsf{No}\}$. Assume the recorded symptom table is:

| Patient | Fever | Cough | Diagnosis (ground truth) |
|---|---|---|---|
| $p_1$ | High | Yes | Flu |
| $p_2$ | High | Yes | Cold |
| $p_3$ | High | No | Flu |
| $p_4$ | High | No | Flu |
| $p_5$ | Normal | No | Healthy |
| $p_6$ | Normal | No | Healthy |

Define the indiscernibility relation $R = \mathrm{IND}(C)$ by

$$(x, y) \in R \iff x \text{ and } y \text{ have identical values of } \mathsf{Fever} \text{ and } \mathsf{Cough}.$$

Then the equivalence classes are

$$[p_1]_R = \{p_1, p_2\}, \qquad [p_3]_R = \{p_3, p_4\}, \qquad [p_5]_R = \{p_5, p_6\}.$$

Let the target concept be the set of influenza cases

$$Y := \{p \in U \mid p \text{ is diagnosed as } \mathsf{Flu}\} = \{p_1, p_3, p_4\}.$$

The Pawlak lower and upper approximations of $Y$ are

$$\underline{Y} = \{x \in U \mid [x]_R \subseteq Y\} = \{p_3, p_4\}, \qquad \overline{Y} = \{x \in U \mid [x]_R \cap Y \neq \varnothing\} = \{p_1, p_2, p_3, p_4\}.$$

Hence the boundary region is

$$\mathrm{BND}(Y) = \overline{Y} \setminus \underline{Y} = \{p_1, p_2\},$$

which reflects the real-life ambiguity: patients with High fever and Yes cough cannot be classified with certainty as Flu using only these two symptoms.

## 2.2 Generalized rough sets

Generalized rough sets extend Pawlak approximations by replacing equivalence relations with broader relations or granulations, yielding flexible lower–upper approximation operators [39–43].

**Definition 2.2.1** (Generalized rough set (relation-based)). Let $U$ and $W$ be nonempty universes and let $R \subseteq U \times W$ be an arbitrary binary relation. For each $x \in U$, define the successor neighborhood (image) of $x$ by

$$R(x) := \{\, y \in W \mid (x, y) \in R \,\} \subseteq W.$$

For any target set $A \subseteq W$, the *lower* and *upper* approximations of $A$ with respect to $(U, W, R)$ are defined by

$$\underline{R}(A) := \{\, x \in U \mid R(x) \subseteq A \,\}, \qquad \overline{R}(A) := \{\, x \in U \mid R(x) \cap A \neq \varnothing \,\}.$$

The ordered pair

$$\big(\underline{R}(A), \overline{R}(A)\big)$$

is called the *generalized rough set* of $A$ (or the $R$-rough set induced by $A$).

**Example 2.2.2** (Eco-conscious customers via a customer–product relation). Let $U$ be a set of customers and $W$ a set of products:

$$U = \{\mathrm{Alice}, \mathrm{Bob}, \mathrm{Carol}, \mathrm{Dan}\}, \qquad W = \{p_1, p_2, p_3, p_4\}.$$

Interpret the binary relation $R \subseteq U \times W$ as *"customer $x$ purchased product $y$ during the last month"*. Assume the recorded purchases are:

$$\begin{aligned} R(\mathrm{Alice}) &= \{p_1, p_2\}, \\ R(\mathrm{Bob}) &= \{p_2, p_3\}, \\ R(\mathrm{Carol}) &= \{p_3\}, \\ R(\mathrm{Dan}) &= \{p_2\}. \end{aligned}$$

| Aspect | Pawlak rough set (classical) | Generalized rough set (relation-based) |
|---|---|---|
| Universes | Single universe $U$ | Two universes $U$ (objects) and $W$ (attributes/targets) |
| Underlying relation | Equivalence relation $E \subseteq U \times U$ (indiscernibility) | Arbitrary binary relation $R \subseteq U \times W$ |
| Neighborhood of $x$ | Equivalence class $[x]_E = \{u \in U : (x,u) \in E\}$ | Successor neighborhood $R(x) = \{y \in W : (x,y) \in R\}$ |
| Target concept | $A \subseteq U$ | $A \subseteq W$ |
| Lower approximation | $\underline{E}(A) = \{x \in U : [x]_E \subseteq A\}$ | $\underline{R}(A) = \{x \in U : R(x) \subseteq A\}$ |
| Upper approximation | $\overline{E}(A) = \{x \in U : [x]_E \cap A \neq \varnothing\}$ | $\overline{R}(A) = \{x \in U : R(x) \cap A \neq \varnothing\}$ |
| Boundary region | $\mathrm{BND}_E(A) = \overline{E}(A) \setminus \underline{E}(A)$ | $\mathrm{BND}_R(A) = \overline{R}(A) \setminus \underline{R}(A)$ |
| Exactness / definability | $A$ is exact iff $\underline{E}(A) = \overline{E}(A)$ (i.e., $A$ is a union of $E$-classes) | $A$ is exact (w.r.t. $(U,W,R)$) iff $\underline{R}(A) = \overline{R}(A)$ |
| Expressiveness | Models uncertainty from indiscernibility (partition of $U$) | Models uncertainty from general relational links (may be non-symmetric, non-transitive, non-reflexive; cross-universe) |
| Special-case relation | — | If $U = W$ and $R = E$ is an equivalence relation, then $R(x) = [x]_E$ and $(\underline{R}(A), \overline{R}(A)) = (\underline{E}(A), \overline{E}(A))$ |

Table 2.1: Concise comparison of Pawlak rough sets and relation-based generalized rough sets.

Let the target set of *eco-labeled products* be

$$A = \{p_1, p_2\} \subseteq W.$$

Then the generalized (relation-based) lower and upper approximations of $A$ are

$$\underline{R}(A) = \{x \in U \mid R(x) \subseteq A\} = \{\text{Alice}, \text{Dan}\},$$

$$\overline{R}(A) = \{x \in U \mid R(x) \cap A \neq \varnothing\} = \{\text{Alice}, \text{Bob}, \text{Dan}\}.$$

Hence, Alice and Dan are *definitely eco-consumers* (all their purchases are eco-labeled), Bob is *possibly eco-consumer* (at least one eco-labeled purchase), and Carol lies in the *negative region* (no eco-labeled purchase).

For reference, a concise comparison between Pawlak rough sets and relation-based generalized rough sets is provided in Table 2.1.

## 2.3 HyperRough Set

The *HyperRough Set* extends rough set theory by incorporating multiple attributes. Its formal definition is given below [44].

**Definition 2.3.1** (HyperRough Set). [44] Let $X$ be a nonempty finite universe, and let $T_1, T_2, \ldots, T_n$ be $n$ distinct attributes with corresponding domains $J_1, J_2, \ldots, J_n$. Define the Cartesian product

$$J = J_1 \times J_2 \times \cdots \times J_n.$$

Let $R \subseteq X \times X$ be an equivalence relation on $X$, with $[x]_R$ denoting the equivalence class of $x$. A *HyperRough Set* over $X$ is a pair $(F, J)$, where:

- $F : J \to \mathcal{P}(X)$ is a mapping that assigns to each attribute value combination $a = (a_1, a_2, \ldots, a_n) \in J$ a subset $F(a) \subseteq X$.

- For each $a \in J$, the rough set approximations of $F(a)$ are defined as

$$\underline{F(a)} = \{x \in X \mid [x]_R \subseteq F(a)\}, \quad \overline{F(a)} = \{x \in X \mid [x]_R \cap F(a) \neq \emptyset\}.$$

Here, $\underline{F(a)}$ comprises all elements whose equivalence classes are completely contained within $F(a)$, while $\overline{F(a)}$ contains elements whose equivalence classes intersect $F(a)$. Additionally, the following properties hold for all $a \in J$:

- $\underline{F(a)} \subseteq \overline{F(a)}$.

- If $F(a) = \varnothing$, then $\underline{F(a)} = \overline{F(a)} = \varnothing$.

- If $F(a) = X$, then $\underline{F(a)} = \overline{F(a)} = X$.

**Example 2.3.2** (Loan screening with partially observed applicant profiles)**.** Let

$$X = \{x_1, x_2, x_3, x_4, x_5, x_6\}$$

be a set of loan applicants. Consider three attributes

$$T_1 = \mathsf{Employment} \in J_1 := \{\mathsf{Stable}, \mathsf{Unstable}\},$$

$$T_2 = \mathsf{Credit} \in J_2 := \{\mathsf{Good}, \mathsf{Bad}\},$$

$$T_3 = \mathsf{Income} \in J_3 := \{\mathsf{High}, \mathsf{Low}\},$$

so that

$$J = J_1 \times J_2 \times J_3.$$

Assume the (true) applicant table is

| Applicant | Employment | Credit | Income |
|---|---|---|---|
| $x_1$ | Stable | Good | High |
| $x_2$ | Stable | Good | Low |
| $x_3$ | Stable | Bad | High |
| $x_4$ | Stable | Bad | Low |
| $x_5$ | Unstable | Good | Low |
| $x_6$ | Unstable | Bad | Low |

Define the HyperRough mapping $F : J \to \mathcal{P}(X)$ by

$$F(e, c, i) := \{\, x \in X \mid (\mathsf{Employment}(x), \mathsf{Credit}(x), \mathsf{Income}(x)) = (e, c, i) \,\}.$$

For instance,

$$F(\mathsf{Stable}, \mathsf{Good}, \mathsf{High}) = \{x_1\}, \qquad F(\mathsf{Stable}, \mathsf{Good}, \mathsf{Low}) = \{x_2\}.$$

In practice, a bank may only observe $\mathsf{Employment}$ and $\mathsf{Credit}$ at the first stage, while $\mathsf{Income}$ is verified later. Model this limited observability by the equivalence relation $R \subseteq X \times X$:

$$(x, y) \in R \iff \mathsf{Employment}(x) = \mathsf{Employment}(y) \text{ and } \mathsf{Credit}(x) = \mathsf{Credit}(y).$$

Then

$$[x_1]_R = [x_2]_R = \{x_1, x_2\}, \quad [x_3]_R = [x_4]_R = \{x_3, x_4\}, \quad [x_5]_R = \{x_5\}, \quad [x_6]_R = \{x_6\}.$$

Now consider the attribute-combination $a = (\mathsf{Stable}, \mathsf{Good}, \mathsf{High})$. Its rough approximations are

$$\underline{F(a)} = \{\, x \in X \mid [x]_R \subseteq F(a) \,\} = \varnothing, \qquad \overline{F(a)} = \{\, x \in X \mid [x]_R \cap F(a) \neq \varnothing \,\} = \{x_1, x_2\}.$$

Interpretation: using only the coarse information $(\mathsf{Employment}, \mathsf{Credit})$, the bank cannot *definitely* identify a "Stable–Good–High" applicant (lower approximation is empty), but it can identify those who are *possibly* in that profile (upper approximation contains $x_1$ and $x_2$), since $x_1$ and $x_2$ are indiscernible at this stage.

By contrast, for $a' = (\mathsf{Unstable}, \mathsf{Good}, \mathsf{Low})$ we have $F(a') = \{x_5\}$ and

$$\underline{F(a')} = \overline{F(a')} = \{x_5\},$$

because $[x_5]_R = \{x_5\}$ is a singleton granule under the observed attributes.

## 2.4 $(m, n)$-SuperHyperRough Set

A (m,n)-SuperHyperRough set maps mth-iterated subsets to nth-iterated subsets, using lifted relations to form rough lower/upper approximations within hierarchical power-set levels [45]. Let $X$ be a nonempty finite universe and let

$$R \subseteq X \times X$$

be an equivalence relation on $X$. We write $[x]_R = \{\, y \in X \mid (x, y) \in R\}$ for the $R$-equivalence class of $x$. For each $k \geq 0$, define the iterated power set

$$P^0(X) = X, \quad P^{k+1}(X) = \mathcal{P}\big(P^k(X)\big).$$

We will lift $R$ to an equivalence $R^k$ on $P^k(X)$ and then define $(m, n)$-SuperHyperRough Sets.

**Definition 2.4.1** (Lifted Relation $R^k$)**.** Define recursively for each $k \geq 0$:

$$R^0 = R \subseteq X \times X,$$

and for $k \geq 1$,

$$R^k \subseteq P^k(X) \times P^k(X)$$

by declaring, for $A, B \in P^k(X)$,

$$A \, R^k \, B \iff \big(\forall\, a \in A \;\exists\, b \in B : \; (a, b) \in R^{k-1}\big) \;\wedge\; \big(\forall\, b \in B \;\exists\, a \in A : \; (a, b) \in R^{k-1}\big).$$

**Definition 2.4.2** ($(m,n)$-SuperHyperRough Set)**.** Fix integers $m, n \geq 0$. An $(m,n)$-*SuperHyperRough Set* on $(X,R)$ is a function

$$F: P^m(X) \longrightarrow P^n(X).$$

For each $A \in P^m(X)$, set $C = F(A) \in P^n(X)$. Its *lower* and *upper* approximations in $P^{n-1}(X)$ are

$$\underline{C} = \{ B \in P^{n-1}(X) \mid [B]_{R^{n-1}} \subseteq C \}, \qquad \overline{C} = \{ B \in P^{n-1}(X) \mid [B]_{R^{n-1}} \cap C \neq \emptyset \},$$

where $[B]_{R^{n-1}} = \{ D \in P^{n-1}(X) \mid B\, R^{n-1}\, D \}$. Thus each $A$ yields the rough pair $\big(\underline{F(A)},\ \overline{F(A)}\big)$.

**Example 2.4.3** ((1,2)-SuperHyperRough set for grocery-bundle recommendation)**.** Let $X$ be a finite set of products

$$X = \{m_o, m_r, b_w, b_g\},$$

where $m_o$ denotes organic milk, $m_r$ regular milk, $b_w$ white bread, and $b_g$ whole-grain bread.

Define an equivalence relation $R \subseteq X \times X$ expressing *category indiscernibility*:

$$x\, R\, y \quad \Longleftrightarrow \quad \big(x, y \in \{m_o, m_r\}\big) \text{ or } \big(x, y \in \{b_w, b_g\}\big).$$

Hence the $R$-classes are

$$[m_o]_R = [m_r]_R = \{m_o, m_r\}, \qquad [b_w]_R = [b_g]_R = \{b_w, b_g\}.$$

Fix $(m,n) = (1,2)$. Then $P^1(X) = \mathcal{P}(X)$ and $P^2(X) = \mathcal{P}(\mathcal{P}(X))$. We interpret $A \in P^1(X)$ as a *shopping basket*, and $F(A) \in P^2(X)$ as a *collection of suggested bundles* (each suggested bundle is itself a subset of $X$).

Consider the basket

$$A = \{m_o, b_g\} \in P^1(X) \quad \text{(organic milk + whole-grain bread).}$$

Suppose the recommender proposes a small list of alternatives at $A$:

$$C := F(A) = \big\{\{m_o, b_g\}, \{m_r, b_g\}, \{m_o, b_w\}\big\} \in P^2(X).$$

Lift $R$ to an equivalence $R^1$ on $P^1(X) = \mathcal{P}(X)$ (cf. Definition 2.4.1) by requiring mutual category-wise match: for $B_1, B_2 \subseteq X$,

$$B_1\, R^1\, B_2 \quad \Longleftrightarrow \quad \Big(\forall u \in B_1\ \exists v \in B_2 :\ u\, R\, v\Big)\ \wedge\ \Big(\forall v \in B_2\ \exists u \in B_1 :\ u\, R\, v\Big).$$

In particular, the $R^1$-class of the "milk + bread" bundle $\{m_o, b_g\}$ is

$$E := [\{m_o, b_g\}]_{R^1} = \big\{\{m_o, b_g\}, \{m_r, b_g\}, \{m_o, b_w\}, \{m_r, b_w\}\big\},$$

because any bundle consisting of one milk and one bread is indistinguishable at the category level.

We now form the rough approximations of $C \subseteq P^1(X)$ with respect to $R^1$:

$$\underline{C} = \{ B \in P^1(X) \mid [B]_{R^1} \subseteq C \}, \qquad \overline{C} = \{ B \in P^1(X) \mid [B]_{R^1} \cap C \neq \varnothing \}.$$

| Aspect | Rough set (Pawlak) | $(m, n)$-SuperHyperRough set [45] |
|---|---|---|
| Base universe | Single universe $X$ | Same base $X$, but works on iterated levels $P^k(X)$ |
| Underlying relation | Equivalence $R \subseteq X \times X$ (indiscernibility on objects) | Lifted equivalences $R^k$ on $P^k(X)$, defined recursively from $R$ |
| Basic neighborhood | $[x]_R \subseteq X$ for $x \in X$ | $[B]_{R^k} \subseteq P^k(X)$ for $B \in P^k(X)$ |
| Target to approximate | A fixed set $A \subseteq X$ | For each input $A \in P^m(X)$, a level-$n$ target $C = F(A) \in P^n(X)$ |
| What is being modeled | Uncertain classification of *individuals* in $X$ | Uncertainty for *higher-order / hierarchical objects* (sets of sets, etc.) |
| Lower approximation | $\underline{A} = \{x \in X \mid [x]_R \subseteq A\} \subseteq X$ | $\underline{C} = \{B \in P^{n-1}(X) \mid [B]_{R^{n-1}} \subseteq C\} \subseteq P^{n-1}(X)$ |
| Upper approximation | $\overline{A} = \{x \in X \mid [x]_R \cap A \neq \varnothing\} \subseteq X$ | $\overline{C} = \{B \in P^{n-1}(X) \mid [B]_{R^{n-1}} \cap C \neq \varnothing\} \subseteq P^{n-1}(X)$ |
| Boundary region | $\mathrm{BND}(A) = \overline{A} \setminus \underline{A}$ | $\mathrm{BND}(C) = \overline{C} \setminus \underline{C}$ (at level $P^{n-1}(X)$) |
| Input–output form | Approximates one target set $A$ (static) | A function $F : P^m(X) \to P^n(X)$ (context-dependent: each input $A$ yields a rough pair) |
| Reduction to Pawlak case | — | If $n = 1$ and $F$ returns a subset of $X$, then approximations are taken in $P^0(X) = X$ using $R^0 = R$ (Pawlak-type behavior). |

Table 2.2: Concise comparison of Pawlak rough sets and $(m, n)$-SuperHyperRough sets.

Since $E \nsubseteq C$ (the bundle $\{m_r, b_w\}$ is not included in the recommendation list), no element of $E$ is *certainly* recommended; in particular,

$$\underline{C} \cap E = \varnothing.$$

On the other hand, for every $B \in E$ we have $[B]_{R^1} = E$ and $E \cap C \neq \varnothing$, so

$$E \subseteq \overline{C}.$$

Thus, under the coarse indiscernibility "same category," the system can only assert that *some* milk–bread combination is recommended (membership in the upper approximation), while it cannot certify *which specific variant* is recommended with certainty (the lower approximation is empty on $E$).

For reference, Table 2.2 provides a concise comparison of Pawlak rough sets and $(m, n)$-SuperHyperRough sets.

## 2.5 MultiRough Set

MultiRough set assigns for each equivalence relation in a family Pawlak lower and upper approximations of a subset simultaneously indexed [44].

**Definition 2.5.1** (MultiRough set)**.** Let $X$ be a nonempty finite universe and let $\mathcal{I}$ be a nonempty finite index set. For each $i \in \mathcal{I}$, let

$$R_i \subseteq X \times X$$

be an equivalence relation, and write the $R_i$-equivalence class of $x \in X$ as

$$[x]_{R_i} := \{ y \in X \mid (x, y) \in R_i \}.$$

For any $Y \subseteq X$, define the (Pawlak) lower and upper approximations with respect to $R_i$ by

$$\underline{Y}^{\,i} := \{ x \in X \mid [x]_{R_i} \subseteq Y \}, \qquad \overline{Y}^{\,i} := \{ x \in X \mid [x]_{R_i} \cap Y \neq \varnothing \}.$$

A *MultiRough set* of $Y$ (with respect to the family $\{R_i\}_{i\in\mathcal{I}}$) is the indexed family

$$\mathrm{MR}_{\mathcal{I}}(Y) := \big( (\underline{Y}^{\,i}, \overline{Y}^{\,i}) \big)_{i\in\mathcal{I}} \in \Big(\mathcal{P}(X) \times \mathcal{P}(X)\Big)^{\mathcal{I}}.$$

**Remark 2.5.2.** For every $i \in \mathcal{I}$ we have $\underline{Y}^{\,i} \subseteq Y \subseteq \overline{Y}^{\,i}$, hence each component $(\underline{Y}^{\,i}, \overline{Y}^{\,i})$ is an ordinary rough approximation pair. If $|\mathcal{I}| = 1$, then $\mathrm{MR}_{\mathcal{I}}(Y)$ reduces to the classical rough approximation of $Y$. In an information system $(X, A)$, a common choice is $R_i = \mathrm{IND}(B_i)$ induced by attribute subsets $B_i \subseteq A$, so that $\mathrm{MR}_{\mathcal{I}}(Y)$ records multiple attribute-granular rough views of the same target set $Y$.

**Example 2.5.3** (MultiRough set in hospital triage under multiple information sources)**.** Let $X = \{p_1, p_2, p_3, p_4, p_5, p_6\}$ be a set of patients arriving at an emergency department. Let the target concept be

$$Y := \{p \in X \mid p \text{ truly requires ICU-level care}\} = \{p_1, p_3, p_4\}.$$

We consider two different (but common) ways of grouping patients into indiscernibility classes, leading to two equivalence relations $R_1$ and $R_2$ on $X$.

**(i) Coarse triage-vital grouping.** Define $R_1$ by equality of the pair *(oxygen saturation category, fever category)*:

$$p\, R_1\, q \iff (\mathsf{O2Cat}(p), \mathsf{FeverCat}(p)) = (\mathsf{O2Cat}(q), \mathsf{FeverCat}(q)).$$

Assume this yields the partition

$$[p_1]_{R_1} = [p_2]_{R_1} = \{p_1, p_2\}, \qquad [p_3]_{R_1} = [p_4]_{R_1} = \{p_3, p_4\}, \qquad [p_5]_{R_1} = [p_6]_{R_1} = \{p_5, p_6\}.$$

Then the Pawlak approximations of $Y$ w.r.t. $R_1$ are

$$\underline{Y}^{\,1} = \{x \in X \mid [x]_{R_1} \subseteq Y\} = \{p_3, p_4\}, \qquad \overline{Y}^{\,1} = \{x \in X \mid [x]_{R_1} \cap Y \neq \varnothing\} = \{p_1, p_2, p_3, p_4\}.$$

**(ii) Imaging+lab grouping.** Define $R_2$ by equality of the pair *(CT finding category, inflammation-marker category)*:

$$p\, R_2\, q \iff (\mathsf{CTCat}(p), \mathsf{InflamCat}(p)) = (\mathsf{CTCat}(q), \mathsf{InflamCat}(q)).$$

Assume this yields the partition

$$[p_1]_{R_2} = [p_3]_{R_2} = \{p_1, p_3\}, \qquad [p_2]_{R_2} = [p_4]_{R_2} = \{p_2, p_4\}, \qquad [p_5]_{R_2} = [p_6]_{R_2} = \{p_5, p_6\}.$$

Then the Pawlak approximations of $Y$ w.r.t. $R_2$ are

$$\underline{Y}^2 = \{x \in X \mid [x]_{R_2} \subseteq Y\} = \{p_1, p_3\}, \qquad \overline{Y}^2 = \{x \in X \mid [x]_{R_2} \cap Y \neq \varnothing\} = \{p_1, p_2, p_3, p_4\}.$$

Hence the MultiRough set of $Y$ with respect to $\mathcal{I} = \{1, 2\}$ is

$$\mathrm{MR}_{\mathcal{I}}(Y) = \big((\underline{Y}^1, \overline{Y}^1), (\underline{Y}^2, \overline{Y}^2)\big).$$

Interpretation: under vital-sign granulation, $p_3, p_4$ are *definitely* ICU-cases, whereas under imaging+lab granulation, $p_1, p_3$ are *definitely* ICU-cases; both views agree on the *possible* ICU-cases $\{p_1, p_2, p_3, p_4\}$.

**Proposition 2.5.4** (Monotonicity (componentwise))**.** *Fix $i \in \mathcal{I}$. If $Y \subseteq Z \subseteq X$, then*

$$\underline{Y}^i \subseteq \underline{Z}^i \qquad \textit{and} \qquad \overline{Y}^i \subseteq \overline{Z}^i.$$

*Consequently,* $\mathrm{MR}_{\mathcal{I}}(Y)$ *is monotone in $Y$ componentwise.*

*Proof.* Fix $i$ and assume $Y \subseteq Z$. If $x \in \underline{Y}^i$ then $[x]_{R_i} \subseteq Y \subseteq Z$, hence $x \in \underline{Z}^i$. If $x \in \overline{Y}^i$ then $[x]_{R_i} \cap Y \neq \varnothing$, and since $Y \subseteq Z$ we also have $[x]_{R_i} \cap Z \neq \varnothing$, hence $x \in \overline{Z}^i$. □

An Iterated MultiRough Set repeatedly applies multiple rough approximations under several equivalence relations, producing a recursively nested, indexed family of lower–upper approximation pairs.

**Definition 2.5.5** (Iterated MultiRough types)**.** Define a hierarchy of sets $\{\mathcal{T}_k(X)\}_{k \geq 0}$ recursively by

$$\mathcal{T}_0(X) := \mathcal{P}(X), \qquad \mathcal{T}_{k+1}(X) := \big(\mathcal{T}_k(X) \times \mathcal{T}_k(X)\big)^{\mathcal{I}} \quad (k \geq 0).$$

Thus, an element of $\mathcal{T}_{k+1}(X)$ is a function

$$\mathcal{I} \to \mathcal{T}_k(X) \times \mathcal{T}_k(X), \qquad i \longmapsto (A_i^-, A_i^+),$$

so $\mathcal{T}_{k+1}(X)$ may be viewed as an $\mathcal{I}$-indexed family of "rough pairs" at level $k$.

**Definition 2.5.6** (Iterated MultiRough set)**.** For each $k \geq 0$, define a map

$$\mathrm{IMR}_{\mathcal{I}}^{(k)} : \mathcal{P}(X) \longrightarrow \mathcal{T}_k(X)$$

recursively by

$$\mathrm{IMR}_{\mathcal{I}}^{(0)}(Y) := Y \in \mathcal{P}(X),$$

and for $k \geq 0$,

$$\mathrm{IMR}_{\mathcal{I}}^{(k+1)}(Y) := \Big(\mathrm{IMR}_{\mathcal{I}}^{(k)}(\underline{Y}^i)\ ,\ \mathrm{IMR}_{\mathcal{I}}^{(k)}(\overline{Y}^i)\Big)_{i \in \mathcal{I}} \ \in\ \big(\mathcal{T}_k(X) \times \mathcal{T}_k(X)\big)^{\mathcal{I}}.$$

We call $\mathrm{IMR}_{\mathcal{I}}^{(k)}(Y)$ the *iterated MultiRough set of depth $k$ of $Y$*. In particular, $\mathrm{IMR}_{\mathcal{I}}^{(1)}(Y) = \mathrm{MR}_{\mathcal{I}}(Y)$ is the ordinary MultiRough set.

**Theorem 2.5.7** (Well-definedness of iterated MultiRough sets). *For every integer $k \geq 0$ and every $Y \subseteq X$, the value $\mathrm{IMR}_{\mathcal{I}}^{(k)}(Y)$ is well-defined and satisfies*

$$\mathrm{IMR}_{\mathcal{I}}^{(k)}(Y) \in \mathcal{T}_k(X).$$

*Equivalently, the recursion in Definition 2.5.6 defines a total function $\mathrm{IMR}_{\mathcal{I}}^{(k)} : \mathcal{P}(X) \to \mathcal{T}_k(X)$ for each $k$.*

*Proof.* We proceed by induction on $k$.

*Base case $k = 0$.* For any $Y \subseteq X$, $\mathrm{IMR}_{\mathcal{I}}^{(0)}(Y) = Y \in \mathcal{P}(X) = \mathcal{T}_0(X)$ by definition.

*Inductive step.* Assume for some $k \geq 0$ that $\mathrm{IMR}_{\mathcal{I}}^{(k)}$ is well-defined and that $\mathrm{IMR}_{\mathcal{I}}^{(k)}(Z) \in \mathcal{T}_k(X)$ holds for all $Z \subseteq X$. Fix $Y \subseteq X$. For each $i \in \mathcal{I}$, the Pawlak approximations $\underline{Y}^{\,i}$ and $\overline{Y}^{\,i}$ are subsets of $X$ by construction, hence $\underline{Y}^{\,i}, \overline{Y}^{\,i} \in \mathcal{P}(X)$. Therefore, the induction hypothesis implies that

$$\mathrm{IMR}_{\mathcal{I}}^{(k)}\big(\underline{Y}^{\,i}\big) \in \mathcal{T}_k(X) \quad \text{and} \quad \mathrm{IMR}_{\mathcal{I}}^{(k)}\big(\overline{Y}^{\,i}\big) \in \mathcal{T}_k(X) \qquad (i \in \mathcal{I}).$$

Consequently, for each $i \in \mathcal{I}$ the ordered pair

$$\Big(\mathrm{IMR}_{\mathcal{I}}^{(k)}\big(\underline{Y}^{\,i}\big),\ \mathrm{IMR}_{\mathcal{I}}^{(k)}\big(\overline{Y}^{\,i}\big)\Big) \in \mathcal{T}_k(X) \times \mathcal{T}_k(X)$$

is well-defined, and hence the $\mathcal{I}$-indexed family of these pairs lies in

$$\big(\mathcal{T}_k(X) \times \mathcal{T}_k(X)\big)^{\mathcal{I}} = \mathcal{T}_{k+1}(X).$$

By Definition 2.5.6, this family is exactly $\mathrm{IMR}_{\mathcal{I}}^{(k+1)}(Y)$. Thus $\mathrm{IMR}_{\mathcal{I}}^{(k+1)}(Y) \in \mathcal{T}_{k+1}(X)$, and the recursion defines a total function at level $k + 1$.

This completes the induction. □

## 2.6 Weighted Rough Set

Weighted rough sets attach weights to attributes (or objects) and incorporate these weights into approximation operators, so that imbalanced data and heterogeneous attribute importance can be handled in a controlled manner [46–49]. As a related line of research, *weighted fuzzy rough sets* have also been studied [50, 51].

**Definition 2.6.1** (Information system). [52] An *information system* is a quadruple

$$S = (U, A, \{V_a\}_{a \in A}, \{f_a\}_{a \in A}),$$

where:

- $U$ is a finite, nonempty set of objects (e.g., patients, loan applicants, or regions).

- $A$ is a finite set of attributes.
- For each $a \in A$, $V_a$ is a nonempty set called the *domain* of $a$ (e.g., $V_{\text{Fever}} = \{\text{High}, \text{Normal}\}$).
- For each $a \in A$, $f_a : U \to V_a$ assigns to each object $x \in U$ an attribute value $f_a(x) \in V_a$.

Often, $A$ is partitioned into condition attributes $C$ and a (distinguished) decision attribute $D$, and we write

$$S = (U, C \cup \{D\}).$$

For example, in a medical diagnosis system, $C$ may contain symptoms (Fever, Cough, Fatigue), while $D$ represents the diagnosis (e.g., Flu vs. NonFlu).

**Definition 2.6.2** (Weighted rough set). [52] Let $S = (U, C \cup \{D\})$ be an information system.

**(i) Indiscernibility and classical lower approximation.** For any $B \subseteq C$, the *indiscernibility relation* induced by $B$ is

$$(x, y) \in \text{IND}(B) \iff \forall a \in B,\ f_a(x) = f_a(y).$$

It partitions $U$ into equivalence classes $[x]_B$. For a target set $X \subseteq U$, the Pawlak lower approximation is

$$\underline{X}^B := \{x \in U \mid [x]_B \subseteq X\}.$$

**(ii) Positive region and dependency degree.** Let $U/\,\text{IND}(D)$ denote the partition of $U$ into decision classes. The *positive region* of $D$ with respect to $B$ is

$$\text{POS}_B(D) := \bigcup \Big\{ G \in U/\,\text{IND}(B) \ \Big|\ \exists\, H \in U/\,\text{IND}(D) \text{ with } G \subseteq H \Big\}.$$

The *dependency degree* of $D$ on $B$ is

$$\gamma_B := \frac{|\,\text{POS}_B(D)|}{|U|} \in [0, 1].$$

**(iii) Attribute significance and weights.** For $a \in B$, the *significance* of $a$ (relative to $B$) is

$$\theta(a) := \gamma_B - \gamma_{B \setminus \{a\}} \quad (\geq 0).$$

Assuming $\sum_{b \in B} \theta(b) > 0$, define the normalized weight of $a$ by

$$w(a) := \frac{\theta(a)}{\sum_{b \in B} \theta(b)}.$$

Then $w(a) \geq 0$ and $\sum_{a \in B} w(a) = 1$.

**(iv) Weighted lower approximation (attribute-wise weighted inclusion).** For each single attribute $a \in B$, write $[x]_a := [x]_{\{a\}}$ for the granule induced by $\{a\}$. Fix a threshold $\alpha \in [0, 1]$ and define the score

$$\sigma_w(x; X, B) := \sum_{a \in B} w(a)\, I\big([x]_a \subseteq X\big),$$

where $I(\varphi) = 1$ if $\varphi$ is true and $I(\varphi) = 0$ otherwise. The *weighted lower approximation* of $X$ with respect to $(B, w)$ is

$$\underline{B}_w(X) := \{\, x \in U \mid \sigma_w(x; X, B) \geq \alpha \,\}.$$

**Remark.** If one replaces $I([x]_a \subseteq X)$ by $I([x]_B \subseteq X)$ inside the sum, then the sum becomes $I([x]_B \subseteq X) \sum_{a \in B} w(a) = I([x]_B \subseteq X)$, and the weights have no effect. The attribute-wise formulation above avoids this degeneracy and realizes a genuine weighted inclusion.

**Example 2.6.3** (Medical triage as a weighted rough set)**.** Let $U = \{p_1, p_2, p_3, p_4, p_5, p_6\}$ be a set of patients and consider $S = (U, C \cup \{D\})$ with condition attributes $C = \{F, Cg, Fa\}$, where $F$ = Fever, $Cg$ = Cough, $Fa$ = Fatigue, and $D \in \{\text{Flu}, \text{NonFlu}\}$.

The observed data are:

| Patient | $F$ | $Cg$ | $Fa$ | $D$ |
|---|---|---|---|---|
| $p_1$ | $H$ | $Y$ | $Y$ | Flu |
| $p_2$ | $H$ | $Y$ | $Y$ | Flu |
| $p_3$ | $H$ | $Y$ | $N$ | Flu |
| $p_4$ | $H$ | $N$ | $Y$ | NonFlu |
| $p_5$ | $N$ | $N$ | $N$ | NonFlu |
| $p_6$ | $N$ | $N$ | $N$ | NonFlu |

Fix $B = C$. The $\text{IND}(B)$-granules are

$$[p_1]_B = \{p_1, p_2\}, \quad [p_3]_B = \{p_3\}, \quad [p_4]_B = \{p_4\}, \quad [p_5]_B = \{p_5, p_6\}.$$

Each granule is decision-pure, hence $\text{POS}_B(D) = U$ and $\gamma_B = 1$.

Remove one attribute at a time:

- For $B \setminus \{F\} = \{Cg, Fa\}$, the induced granules remain decision-pure, hence $\gamma_{B\setminus\{F\}} = 1$.
- For $B\setminus\{Cg\} = \{F, Fa\}$, the granule $\{p_1, p_2, p_4\}$ (same $(F, Fa) = (H, Y)$) mixes Flu/NonFlu, so only $\{p_3\}$ and $\{p_5, p_6\}$ belong to the positive region. Thus $|\,\text{POS}_{B\setminus\{Cg\}}(D)| = 3$ and $\gamma_{B\setminus\{Cg\}} = \frac{3}{6} = \frac{1}{2}$.
- For $B \setminus \{Fa\} = \{F, Cg\}$, the induced granules are decision-pure again, hence $\gamma_{B\setminus\{Fa\}} = 1$.

Therefore,

$$\theta(F) = 0, \qquad \theta(Cg) = \frac{1}{2}, \qquad \theta(Fa) = 0,$$

so the normalized weights are

$$w(Cg) = 1, \qquad w(F) = w(Fa) = 0.$$

Let the target concept be the set of flu patients $X = \{p_1, p_2, p_3\}$ and take $\alpha = 1$. Since only $Cg$ has nonzero weight, the score reduces to

$$\sigma_w(x; X, B) = I\big([x]_{Cg} \subseteq X\big),$$

where the $Cg$-granules are

$$[p_1]_{Cg} = [p_2]_{Cg} = [p_3]_{Cg} = \{p_1, p_2, p_3\} \subseteq X, \qquad [p_4]_{Cg} = [p_5]_{Cg} = [p_6]_{Cg} = \{p_4, p_5, p_6\} \nsubseteq X.$$

Hence,

$$\underline{B}_w(X) = \{p_1, p_2, p_3\}.$$

Interpretation: in this dataset, cough carries the full weight, and the weighted rough model selects exactly those patients whose cough-granule is certainly contained in the flu concept.

## 2.7 Neighborhood Rough Set

Neighborhood rough sets approximate a target set using neighborhood granules induced by a relation, enabling non-equivalence and covering-based models flexibly [53–56].

**Definition 2.7.1** (Neighborhood approximation space and neighborhood rough approximations)**.** Let $U$ be a nonempty universe. A *neighborhood approximation space* is a pair $(U, N)$, where

$$N : U \to \mathcal{P}(U), \qquad x \mapsto N(x),$$

assigns to each $x \in U$ a (nonempty) neighborhood $N(x) \subseteq U$. Typical choices include:

(i) *Relation-induced neighborhoods:* for a binary relation $S \subseteq U \times U$, $N_S(x) := \{\, y \in U \mid (x, y) \in S \,\}$.

(ii) *Metric neighborhoods:* for a (pseudo-)metric $d$ on $U$ and $\delta \geq 0$, $N_\delta(x) := \{\, y \in U \mid d(x, y) \leq \delta \,\}$.

For any $X \subseteq U$, the *neighborhood lower* and *neighborhood upper* approximations of $X$ (with respect to $N$) are defined by

$$\underline{\mathrm{apr}}_N(X) \;:=\; \{\, x \in U \mid N(x) \subseteq X \,\}, \qquad \overline{\mathrm{apr}}_N(X) \;:=\; \{\, x \in U \mid N(x) \cap X \neq \varnothing \,\}.$$

The *neighborhood rough set* of $X$ is the pair

$$\big(\underline{\mathrm{apr}}_N(X),\; \overline{\mathrm{apr}}_N(X)\big).$$

**Example 2.7.2** (Neighborhood rough-set screening in predictive maintenance)**.** Consider a factory with seven vibration sensors

$$U = \{s_1, s_2, s_3, s_4, s_5, s_6, s_7\}.$$

Each sensor $s_i$ is described by a 2D feature vector

$$\varphi(s_i) = (\text{temperature deviation},\ \text{vibration RMS}) \in \mathbb{R}^2,$$

given by

| $s_i$ | $s_1$ | $s_2$ | $s_3$ | $s_4$ | $s_5$ | $s_6$ | $s_7$ |
|---|---|---|---|---|---|---|---|
| $\varphi(s_i)$ | $(0,0)$ | $(0.8,0.2)$ | $(1.6,0.2)$ | $(2.1,0.9)$ | $(5,5)$ | $(5.7,5.2)$ | $(6.5,5.1)$ |

and let $d(s_i, s_j) := \|\varphi(s_i) - \varphi(s_j)\|_2$ be the Euclidean distance. Fix $\delta = 1$ and define the metric neighborhoods

$$N_\delta(s) = \{t \in U \mid d(s,t) \le \delta\} \qquad (s \in U).$$

A direct computation yields the neighborhood system

$$N_\delta(s_1) = \{s_1, s_2\}, \qquad N_\delta(s_2) = \{s_1, s_2, s_3\}, \qquad N_\delta(s_3) = \{s_2, s_3, s_4\},$$
$$N_\delta(s_4) = \{s_3, s_4\}, \qquad N_\delta(s_5) = \{s_5, s_6\}, \qquad N_\delta(s_6) = \{s_5, s_6, s_7\}, \qquad N_\delta(s_7) = \{s_6, s_7\}.$$

Suppose that, after a manual inspection, the set of *confirmed faulty* sensors is

$$X = \{s_1, s_2, s_4\} \subseteq U.$$

Using neighborhood rough approximations,

$$\underline{\mathrm{apr}}_{N_\delta}(X) = \{s \in U \mid N_\delta(s) \subseteq X\}, \qquad \overline{\mathrm{apr}}_{N_\delta}(X) = \{s \in U \mid N_\delta(s) \cap X \ne \varnothing\},$$

we obtain

$$\underline{\mathrm{apr}}_{N_\delta}(X) = \{s_1\}, \qquad \overline{\mathrm{apr}}_{N_\delta}(X) = \{s_1, s_2, s_3, s_4\}.$$

Hence the regions are

$$\mathrm{POS}_{N_\delta}(X) = \{s_1\}, \qquad \mathrm{BND}_{N_\delta}(X) = \{s_2, s_3, s_4\}, \qquad \mathrm{NEG}_{N_\delta}(X) = \{s_5, s_6, s_7\}.$$

Interpretation: $s_1$ is *definitely* within the faulty cluster (its whole neighborhood is faulty), $s_2, s_3, s_4$ are *possibly* faulty (their neighborhoods touch the faulty set), and $s_5, s_6, s_7$ are *definitely not* near the faulty cluster under the tolerance $\delta = 1$.

## 2.8 Sequential Rough Set

Sequential rough sets model evolving approximations by applying successive relations or time-indexed granules, updating lower/upper regions across stages of information [57, 58].

**Definition 2.8.1** (Sequential rough approximations)**.** Let $U$ be a nonempty finite universe and let

$$\mathbf{R} = \langle R_1, R_2, \ldots, R_m \rangle$$

be an *ordered* finite family of equivalence relations on $U$ (each $R_i \subseteq U \times U$). For $x \in U$, write

$$[x]_{R_i} := \{\, y \in U \mid (x, y) \in R_i \,\}$$

for the $R_i$-equivalence class of $x$.

For each $i \in \{1, \ldots, m\}$ and each $X \subseteq U$, define the usual Pawlak lower/upper operators:

$$\underline{\mathrm{apr}}_{R_i}(X) := \{\, x \in U \mid [x]_{R_i} \subseteq X \,\}, \qquad \overline{\mathrm{apr}}_{R_i}(X) := \{\, x \in U \mid [x]_{R_i} \cap X \ne \varnothing \,\}.$$

The *sequential lower approximation* of $X$ along $\mathbf{R}$ is the iterated operator

$$\underline{\mathrm{apr}}_{\mathbf{R}}^{\mathrm{seq}}(X) \;:=\; \big(\underline{\mathrm{apr}}_{R_m} \circ \underline{\mathrm{apr}}_{R_{m-1}} \circ \cdots \circ \underline{\mathrm{apr}}_{R_1}\big)(X),$$

and the *sequential upper approximation* is defined dually by

$$\overline{\mathrm{apr}}_{\mathbf{R}}^{\mathrm{seq}}(X) \;:=\; U \setminus \underline{\mathrm{apr}}_{\mathbf{R}}^{\mathrm{seq}}(U \setminus X).$$

The *sequential rough set* of $X$ (with respect to $\mathbf{R}$) is the pair

$$\big(\underline{\mathrm{apr}}_{\mathbf{R}}^{\mathrm{seq}}(X),\; \overline{\mathrm{apr}}_{\mathbf{R}}^{\mathrm{seq}}(X)\big).$$

**Example 2.8.2** (Two-stage medical triage as a sequential rough set)**.** Let $U = \{p_1, p_2, p_3, p_4, p_5, p_6\}$ be patients arriving at an emergency department.

**Stage 1 (symptom-screen granulation).** Define an equivalence relation $R_1$ on $U$ by the partition

$$U/R_1 = \big\{\{p_1, p_2, p_3\},\; \{p_4, p_5, p_6\}\big\},$$

where $\{p_1, p_2, p_3\}$ are "high-suspicion" by symptoms and $\{p_4, p_5, p_6\}$ are "lower-suspicion".

**Stage 2 (rapid-test granulation).** Define an equivalence relation $R_2$ on $U$ by the partition

$$U/R_2 = \big\{\{p_1, p_2\},\; \{p_3, p_4\},\; \{p_5, p_6\}\big\},$$

where each block represents indistinguishability by a rapid-test pattern.

Let the target concept be the (later confirmed) infected set

$$X = \{p_1, p_2, p_3\} \subseteq U.$$

Consider the ordered family $\mathbf{R} = \langle R_1, R_2\rangle$ and the sequential lower approximation

$$\underline{\mathrm{apr}}_{\mathbf{R}}^{\mathrm{seq}}(X) := \big(\underline{\mathrm{apr}}_{R_2} \circ \underline{\mathrm{apr}}_{R_1}\big)(X).$$

First,

$$\underline{\mathrm{apr}}_{R_1}(X) = \{p_1, p_2, p_3\},$$

since the $R_1$-class $\{p_1, p_2, p_3\}$ is contained in $X$, while $\{p_4, p_5, p_6\} \not\subseteq X$. Next,

$$\underline{\mathrm{apr}}_{R_2}\big(\underline{\mathrm{apr}}_{R_1}(X)\big) = \underline{\mathrm{apr}}_{R_2}(\{p_1, p_2, p_3\}) = \{p_1, p_2\},$$

because the $R_2$-class $\{p_1, p_2\}$ is fully contained in $\{p_1, p_2, p_3\}$, but $\{p_3, p_4\}$ is not.

Hence,

$$\underline{\mathrm{apr}}_{\mathbf{R}}^{\mathrm{seq}}(X) = \{p_1, p_2\}.$$

The sequential upper approximation is defined dually by

$$\overline{\mathrm{apr}}_{\mathbf{R}}^{\mathrm{seq}}(X) := U \setminus \underline{\mathrm{apr}}_{\mathbf{R}}^{\mathrm{seq}}(U \setminus X).$$

Since $U \setminus X = \{p_4, p_5, p_6\}$,

$$\underline{\mathrm{apr}}_{R_1}(U \setminus X) = \{p_4, p_5, p_6\}, \qquad \underline{\mathrm{apr}}_{R_2}(\{p_4, p_5, p_6\}) = \{p_5, p_6\}.$$

Therefore,

$$\overline{\mathrm{apr}}_{\mathbf{R}}^{\mathrm{seq}}(X) = U \setminus \{p_5, p_6\} = \{p_1, p_2, p_3, p_4\}.$$

After two stages, $\{p_1, p_2\}$ are *definitely* infected (sequential lower), $\{p_5, p_6\}$ are *definitely not* infected (sequential negative region), and $\{p_3, p_4\}$ lie in the *boundary* where the second-stage granule $\{p_3, p_4\}$ prevents a definitive decision.

## 2.9 ContraRough Set

ContraRough sets incorporate contradiction degrees and thresholds to define lower/upper approximations, producing positive, boundary, negative regions under inconsistent evidence [59].

**Definition 2.9.1** (ContraRough Set). [59] Let $X \neq \varnothing$ be a universe and let $U \subseteq X$ be a target concept.

**(i) Contradiction kernels.** A *relation-contradiction kernel* is a map

$$c_R : X \times X \to [0, 1]$$

satisfying

$$c_R(x, x) = 0 \quad \text{(reflexivity)}, \qquad c_R(x, y) = c_R(y, x) \quad \text{(symmetry)}.$$

A *membership-contradiction kernel* for $U$ is a map

$$c_U : X \to [0, 1],$$

where $c_U(y)$ quantifies how contradictory it is to assert $y \in U$ (smaller values mean more consistent membership).

**(ii) Thresholded consistency regions.** Fix thresholds $(\alpha, \beta, \gamma) \in [0, 1]^3$ with $\beta \leq \gamma$. Define the admitted relation and its (kernel) neighborhood by

$$R(\alpha) := \{(x, y) \in X \times X \mid c_R(x, y) \leq \alpha\}, \qquad N_\alpha(x) := \{y \in X \mid c_R(x, y) \leq \alpha\}.$$

Define the *definite* and *possible* slices of $U$ by

$$U_{\text{def}}(\beta) := \{y \in X \mid c_U(y) \leq \beta\},$$

$$U_{\text{pos}}(\gamma) := \{y \in X \mid c_U(y) \leq \gamma\}.$$

**(iii) ContraRough approximations.** The *ContraRough lower* and *ContraRough upper* approximations of $U$ are

$$\underline{\text{apr}}^{\text{CR}}_{(\alpha,\beta)}(U) \;:=\; \{x \in X \mid N_\alpha(x) \subseteq U_{\text{def}}(\beta)\},$$

$$\overline{\text{apr}}^{\text{CR}}_{(\alpha,\gamma)}(U) \;:=\; \{x \in X \mid N_\alpha(x) \cap U_{\text{pos}}(\gamma) \neq \varnothing\}.$$

The induced regions are defined as

$$\text{POS}_{(\alpha,\beta)}(U) := \underline{\text{apr}}^{\text{CR}}_{(\alpha,\beta)}(U), \qquad \text{NEG}_{(\alpha,\gamma)}(U) := X \setminus \overline{\text{apr}}^{\text{CR}}_{(\alpha,\gamma)}(U),$$

$$\text{BND}_{(\alpha,\beta,\gamma)}(U) := \overline{\text{apr}}^{\text{CR}}_{(\alpha,\gamma)}(U) \setminus \underline{\text{apr}}^{\text{CR}}_{(\alpha,\beta)}(U).$$

**Example 2.9.2** (Contradictory incident reports classified by a ContraRough set)**.** We illustrate the ContraRough construction in a real-life fact-checking setting.

**Universe and target concept.** Let $X$ be a set of short incident reports about the same bridge-collapse event:

$$X = \{r_1, r_2, r_3, r_4, r_5, r_6\},$$

where $r_1$ is an official emergency bulletin, $r_5$ is a verified CCTV-based report, $r_2, r_3$ are citizen posts, and $r_4, r_6$ are unverified viral posts. Let $U \subseteq X$ denote the (unknown) concept "reliable reports".

**(i) Contradiction kernels.** Assume we have:

- a *relation-contradiction kernel* $c_R : X \times X \to [0,1]$, where $c_R(x, y)$ measures how contradictory the factual content of reports $x$ and $y$ is (e.g., extracted-claim inconsistency; 0 means fully consistent);
- a *membership-contradiction kernel* $c_U : X \to [0,1]$, where $c_U(y)$ measures how contradictory it is to assert "$y \in U$" (e.g., based on source reputation + cross-check signals; smaller is more consistent).

We set

$$c_U(r_1) = 0.05, \quad c_U(r_2) = 0.25, \quad c_U(r_3) = 0.35, \quad c_U(r_4) = 0.55, \quad c_U(r_5) = 0.08, \quad c_U(r_6) = 0.75.$$

**(ii) Thresholds and admitted neighborhoods.** Choose thresholds $(\alpha, \beta, \gamma) = (0.20, 0.10, 0.40)$ (with $\beta \leq \gamma$). Thus

$$U_{\mathrm{def}}(\beta) = \{y \in X \mid c_U(y) \leq 0.10\} = \{r_1, r_5\}, \qquad U_{\mathrm{pos}}(\gamma) = \{y \in X \mid c_U(y) \leq 0.40\} = \{r_1, r_2, r_3, r_5\}.$$

Assume the admitted relation $R(\alpha) = \{(x, y) \mid c_R(x, y) \leq 0.20\}$ yields exactly the following low-contradiction pairs (besides reflexivity):

$$c_R(r_1, r_5) = 0.15, \qquad c_R(r_2, r_3) = 0.10, \qquad c_R(r_4, r_6) = 0.05,$$

and all other distinct pairs have $c_R > 0.20$. Hence the admitted neighborhoods $N_\alpha(x) = \{y \in X \mid c_R(x, y) \leq \alpha\}$ are

$$N_\alpha(r_1) = \{r_1, r_5\}, \quad N_\alpha(r_5) = \{r_5, r_1\}, \quad N_\alpha(r_2) = \{r_2, r_3\},$$

$$N_\alpha(r_3) = \{r_3, r_2\}, \quad N_\alpha(r_4) = \{r_4, r_6\}, \quad N_\alpha(r_6) = \{r_6, r_4\}.$$

**(iii) ContraRough approximations and regions.** The ContraRough lower approximation collects reports whose *entire* admitted neighborhood lies inside the *definite* slice $U_{\mathrm{def}}(\beta)$:

$$\underline{\mathrm{apr}}^{\mathrm{CR}}_{(\alpha,\beta)}(U)$$

$$= \{x \in X \mid N_\alpha(x) \subseteq U_{\mathrm{def}}(\beta)\} = \{r_1, r_5\}.$$

The ContraRough upper approximation collects reports whose admitted neighborhood intersects the *possible* slice $U_{\mathrm{pos}}(\gamma)$:

$$\overline{\mathrm{apr}}^{\mathrm{CR}}_{(\alpha,\gamma)}(U) = \{x \in X \mid N_\alpha(x) \cap U_{\mathrm{pos}}(\gamma) \neq \varnothing\} = \{r_1, r_2, r_3, r_5\}.$$

Therefore the induced regions are

$$\mathrm{POS}_{(\alpha,\beta)}(U) = \{r_1, r_5\},$$
$$\mathrm{BND}_{(\alpha,\beta,\gamma)}(U) = \{r_2, r_3\},$$
$$\mathrm{NEG}_{(\alpha,\gamma)}(U) = \{r_4, r_6\}.$$

$r_1$ and $r_5$ are *definitely reliable* because everything they are (low-contradiction) consistent with is also definitely reliable; $r_2, r_3$ remain *borderline* (possibly reliable) due to weaker credibility signals; and $r_4, r_6$ fall into the *negative* region because their admitted neighborhoods do not touch any possibly reliable report.

## 2.10 Probabilistic Rough Set

Probabilistic rough sets define approximations via conditional probability thresholds, permitting bounded misclassification and yielding positive, boundary, negative regions for classification [60–62]. Probabilistic rough sets have attracted a substantial volume of research in recent years as well [63–65]. Related concepts include *fuzzy probabilistic rough sets* [66–68], probabilistic variable precision rough sets [69, 70], and *neutrosophic probabilistic rough sets* [71, 72].

**Definition 2.10.1** (($\alpha,\beta$)-probabilistic rough approximations / probabilistic rough set)**.** [60–62] Let $U$ be a nonempty finite universe and let $E \subseteq U \times U$ be an equivalence relation. For $x \in U$, write the granule (equivalence class)

$$[x]_E := \{\, y \in U \mid (x,y) \in E \,\}.$$

For any $A \subseteq U$, define the (rough-membership / conditional probability)

$$\mu_A(x) \;:=\; \Pr(A \mid [x]_E) \;\approx\; \frac{|A \cap [x]_E|}{|[x]_E|},$$

(where the ratio gives the usual empirical estimate under a uniform assumption on $[x]_E$).

Fix thresholds $\alpha, \beta$ with $0 \le \beta < \alpha \le 1$. The $(\alpha,\beta)$-*probabilistic lower* and *upper* approximations of $A$ are

$$\underline{\mathrm{apr}}_{(\alpha,\beta)}(A) \;:=\; \{\, x \in U \;\big|\; \Pr(A \mid [x]_E) \ge \alpha \,\},$$
$$\overline{\mathrm{apr}}_{(\alpha,\beta)}(A) \;:=\; \{\, x \in U \;\big|\; \Pr(A \mid [x]_E) > \beta \,\}.$$

Equivalently, the induced three disjoint regions are

$$\mathrm{POS}_{(\alpha,\beta)}(A) := \{x \in U \mid \Pr(A \mid [x]_E) \ge \alpha\},$$
$$\mathrm{BND}_{(\alpha,\beta)}(A) := \{x \in U \mid \beta < \Pr(A \mid [x]_E) < \alpha\},$$
$$\mathrm{NEG}_{(\alpha,\beta)}(A) := \{x \in U \mid \Pr(A \mid [x]_E) \le \beta\}.$$

The $(\alpha, \beta)$-*probabilistic rough set of* $A$ is represented by the approximation pair

$$\left(\underline{\mathrm{apr}}_{(\alpha,\beta)}(A),\ \overline{\mathrm{apr}}_{(\alpha,\beta)}(A)\right),$$

(or equivalently by the triple $(\mathrm{POS}_{(\alpha,\beta)}(A), \mathrm{BND}_{(\alpha,\beta)}(A), \mathrm{NEG}_{(\alpha,\beta)}(A))$).

If $\alpha = 1$ and $\beta = 0$, then $\underline{\mathrm{apr}}_{(1,0)}(A) = \{x \mid \Pr(A \mid [x]_E) = 1\}$ and $\overline{\mathrm{apr}}_{(1,0)}(A) = \{x \mid \Pr(A \mid [x]_E) > 0\}$, recovering the classical Pawlak lower/upper approximations.

**Example 2.10.2** (Credit approval as a probabilistic rough set)**.** Let $U$ be a finite set of loan applicants,

$$U = \{a_1, a_2, \ldots, a_{12}\}.$$

Suppose applicants are described only by a coarse attribute tuple

$$(\text{Income band},\ \text{Credit-history flag}) \in \{\mathrm{H}, \mathrm{M}, \mathrm{L}\} \times \{\mathrm{Good}, \mathrm{Bad}\},$$

and let $E \subseteq U \times U$ be the induced indiscernibility (equivalence) relation: $x\,E\,y$ iff $x$ and $y$ share the same tuple. Assume the resulting $E$-classes are

$$G_1 = \{a_1, a_2, a_3, a_4\}, \quad G_2 = \{a_5, a_6, a_7, a_8, a_9\}, \quad G_3 = \{a_{10}, a_{11}, a_{12}\},$$

so that $[x]_E \in \{G_1, G_2, G_3\}$ for all $x \in U$.

Let the target concept $A \subseteq U$ be the set of applicants who (based on historical outcomes) *repay on time*:

$$A = \{a_1, a_2, a_3, a_4, a_5, a_6, a_7\}.$$

Following the probabilistic rough-set model, we estimate the conditional probability

$$\Pr(A \mid [x]_E) \approx \frac{|A \cap [x]_E|}{|[x]_E|}, \qquad (x \in U),$$

and choose thresholds $0 \le \beta < \alpha \le 1$ to form the three regions $\mathrm{POS}_{(\alpha,\beta)}(A)$, $\mathrm{BND}_{(\alpha,\beta)}(A)$, $\mathrm{NEG}_{(\alpha,\beta)}(A)$.

Take $(\alpha, \beta) = (0.8, 0.3)$. Then, at the class level,

$$\Pr(A \mid G_1) = \frac{|A \cap G_1|}{|G_1|} = \frac{4}{4} = 1, \qquad \Pr(A \mid G_2) = \frac{3}{5} = 0.6, \qquad \Pr(A \mid G_3) = \frac{0}{3} = 0.$$

Hence,

$$\mathrm{POS}_{(0.8,0.3)}(A) = G_1 = \{a_1, a_2, a_3, a_4\},$$
$$\mathrm{BND}_{(0.8,0.3)}(A) = G_2 = \{a_5, a_6, a_7, a_8, a_9\},$$
$$\mathrm{NEG}_{(0.8,0.3)}(A) = G_3 = \{a_{10}, a_{11}, a_{12}\},$$

since $1 \ge 0.8$, $0.3 < 0.6 < 0.8$, and $0 \le 0.3$, respectively.

Therefore, the $(0.8, 0.3)$-*probabilistic rough set of* $A$ is the approximation pair

$$\left(\underline{\mathrm{apr}}_{(0.8,0.3)}(A),\ \overline{\mathrm{apr}}_{(0.8,0.3)}(A)\right) =$$

$$\left(\mathrm{POS}_{(0.8,0.3)}(A),\ \mathrm{POS}_{(0.8,0.3)}(A) \cup \mathrm{BND}_{(0.8,0.3)}(A)\right) = \left(G_1,\ G_1 \cup G_2\right).$$

Applicants in $\mathrm{POS}_{(0.8,0.3)}(A)$ can be *automatically approved* (high-confidence repayment), those in $\mathrm{NEG}_{(0.8,0.3)}(A)$ can be *automatically rejected* (high-confidence non-repayment), and those in $\mathrm{BND}_{(0.8,0.3)}(A)$ are sent to *manual review* (uncertain region).

## 2.11 IndetermRough Set

IndetermRough set introduces an explicit indeterminate region between lower and upper approximations, capturing uncertainty beyond boundary using indiscernibility relations objects [73]. Related concepts include IndetermSoft sets (Indeterminacy Soft Sets) and related indeterminacy-based soft-set models [74–76].

**Definition 2.11.1** (IndetermRough Set)**.** [73] Let $U \neq \varnothing$ be a universe. An *indeterminate (equivalence-like) relation* on $U$ is specified by a pair of binary relations

$$R_{\mathrm{def}} \subseteq R_{\mathrm{pos}} \subseteq U \times U,$$

where $(x, y) \in R_{\mathrm{def}}$ means "$x$ is *definitely* indiscernible from $y$", and $(x, y) \in R_{\mathrm{pos}}$ means "$x$ is *possibly* indiscernible from $y$" (i.e., definite-or-indeterminate membership). Assume $R_{\mathrm{def}}$ is an equivalence relation (so definite indiscernibility is consistent).

An *indeterminate subset* of $U$ is represented by a pair

$$X^* = (X_{\mathrm{def}}, X_{\mathrm{pos}}) \quad \text{with} \quad X_{\mathrm{def}} \subseteq X_{\mathrm{pos}} \subseteq U,$$

where $X_{\mathrm{def}}$ is the set of elements *definitely* in the concept and $X_{\mathrm{pos}}$ is the set of elements *possibly* in the concept.

For each $x \in U$, define the definite and possible neighborhoods

$$N_{\mathrm{def}}(x) := \{y \in U \mid (x, y) \in R_{\mathrm{def}}\}, \qquad N_{\mathrm{pos}}(x) := \{y \in U \mid (x, y) \in R_{\mathrm{pos}}\}.$$

The *indeterminate lower* and *indeterminate upper* approximations of $X^*$ are

$$\underline{X^*} := \{x \in U \mid N_{\mathrm{def}}(x) \subseteq X_{\mathrm{def}}\}, \qquad \overline{X^*} := \{x \in U \mid N_{\mathrm{pos}}(x) \cap X_{\mathrm{pos}} \neq \varnothing\}.$$

The pair $(\underline{X^*}, \overline{X^*})$ is called the *IndetermRough approximation* of $X^*$ (under $(R_{\mathrm{def}}, R_{\mathrm{pos}})$), and the triple $(X^*, \underline{X^*}, \overline{X^*})$ is referred to as an *IndetermRough Set.*

**Example 2.11.2** (Emergency-room influenza triage with an explicit indeterminate region)**.** Let $U$ be a set of six patients arriving at an emergency room:

$$U = \{p_1, p_2, p_3, p_4, p_5, p_6\}.$$

Each patient has (i) a rapid PCR outcome in {Pos, Neg, Unk} and (ii) a fever-status in {High, Low}. Assume:

| patient | $p_1$ | $p_2$ | $p_3$ | $p_4$ | $p_5$ | $p_6$ |
|---|---|---|---|---|---|---|
| PCR | Pos | Pos | Neg | Neg | Unk | Unk |
| Fever | High | High | Low | Low | High | Low |

**Definite vs. possible indiscernibility.** Define $R_{\mathrm{def}}$ by

$$(p_i, p_j) \in R_{\mathrm{def}} \iff \big(\mathrm{PCR}(p_i) = \mathrm{PCR}(p_j)\big) \wedge \big(\mathrm{Fever}(p_i) = \mathrm{Fever}(p_j)\big).$$

Then $R_{\mathrm{def}}$ is an equivalence relation whose classes are

$$[p_1]_{R_{\mathrm{def}}} = \{p_1, p_2\}, \quad [p_3]_{R_{\mathrm{def}}} = \{p_3, p_4\}, \quad [p_5]_{R_{\mathrm{def}}} = \{p_5\}, \quad [p_6]_{R_{\mathrm{def}}} = \{p_6\}.$$

Define the coarser (possible) relation $R_{\text{pos}}$ by

$$(p_i, p_j) \in R_{\text{pos}} \iff \text{Fever}(p_i) = \text{Fever}(p_j),$$

so that

$$[p_1]_{R_{\text{pos}}} = \{p_1, p_2, p_5\}, \qquad [p_3]_{R_{\text{pos}}} = \{p_3, p_4, p_6\}.$$

Clearly $R_{\text{def}} \subseteq R_{\text{pos}}$ (definite agreement implies possible agreement).

**Indeterminate target concept.** Let $X^* = (X_{\text{def}}, X_{\text{pos}})$ represent the concept "patient has influenza," where

$$X_{\text{def}} = \{p_1, p_2\} \quad \text{(PCR positive)},$$

$$X_{\text{pos}} = \{p_1, p_2, p_5\} \quad \text{(PCR positive or PCR unknown with high fever)}.$$

Hence $X_{\text{def}} \subseteq X_{\text{pos}} \subseteq U$.

**Neighborhoods and IndetermRough approximations.** Using $N_{\text{def}}(x) = [x]_{R_{\text{def}}}$ and $N_{\text{pos}}(x) = [x]_{R_{\text{pos}}}$, the IndetermRough lower and upper approximations are

$$\underline{X^*} = \{x \in U \mid N_{\text{def}}(x) \subseteq X_{\text{def}}\}, \qquad \overline{X^*} = \{x \in U \mid N_{\text{pos}}(x) \cap X_{\text{pos}} \neq \varnothing\}.$$

Compute:

$$\underline{X^*} = \{p_1, p_2\}, \qquad \overline{X^*} = \{p_1, p_2, p_5\}.$$

Therefore, the *explicit indeterminate region* is

$$\overline{X^*} \setminus \underline{X^*} = \{p_5\},$$

and the "definitely not" region is $U \setminus \overline{X^*} = \{p_3, p_4, p_6\}$.

**Interpretation (real-life decision).** Patients in $\underline{X^*}$ are treated as *confirmed influenza* (start antivirals immediately), patients in $U \setminus \overline{X^*}$ as *non-influenza*, and patients in $\overline{X^*} \setminus \underline{X^*}$ (here, $p_5$) are *indeterminate*: isolate and order confirmatory testing because the evidence is neither definite nor dismissible.

## 2.12 HesiRough Set

HesiRough set extends rough sets by modeling hesitant membership; lower/upper approximations aggregate multiple possible degrees for each object under uncertainty. Related concepts include hesitant fuzzy sets [4, 77] and hesitant neutrosophic sets [78–80], among others.

**Definition 2.12.1** (HesiRough set (hesitation-based rough set) — refined)**.** Let $X \neq \varnothing$ be a universe. We model *hesitancy* by *set-valued (hesitant) predicates*

$$h_R : X \times X \longrightarrow \mathcal{P}(\{0,1\}) \setminus \{\varnothing\}, \qquad h_U : X \longrightarrow \mathcal{P}(\{0,1\}) \setminus \{\varnothing\}.$$

For $(x, y) \in X \times X$, the value $h_R(x, y)$ is interpreted as the set of *plausible truth values* for the statement "$x$ is $R$-related to $y$", and $h_U(y)$ for "$y \in U$":

$$h_R(x, y) = \{1\} \text{ (definitely related)}, \quad h_R(x, y) = \{0\} \text{ (definitely not related)}, \quad h_R(x, y) = \{0, 1\} \text{ (hesitant/und}$$

and similarly for $h_U(y)$.

**(H0) Definite reflexivity (nondegeneracy).** Assume

$$h_R(x,x) = \{1\} \qquad (\forall x \in X).$$

This guarantees that every object is *definitely* related to itself, preventing vacuous membership in the lower approximation from empty definite neighborhoods.

Define the *definite* and *possible* parts:

$$R_{\text{def}} := \{(x,y) \in X \times X \mid h_R(x,y) = \{1\}\}, \qquad R_{\text{pos}} := \{(x,y) \in X \times X \mid 1 \in h_R(x,y)\},$$

$$U_{\text{def}} := \{y \in X \mid h_U(y) = \{1\}\}, \qquad U_{\text{pos}} := \{y \in X \mid 1 \in h_U(y)\}.$$

For each $x \in X$, define the definite/possible neighborhoods

$$N_{\text{def}}(x) := \{y \in X \mid (x,y) \in R_{\text{def}}\}, \qquad N_{\text{pos}}(x) := \{y \in X \mid (x,y) \in R_{\text{pos}}\}.$$

The triple $(X, h_R, h_U)$ is called a *HesiRough set* (or *HesiRough approximation space*) when the lower and upper approximations of the hesitant target are defined by

$$\underline{U}^H := \{x \in X \mid N_{\text{def}}(x) \subseteq U_{\text{def}}\}, \qquad \overline{U}^H := \{x \in X \mid N_{\text{pos}}(x) \cap U_{\text{pos}} \neq \varnothing\}.$$

The induced regions are

$$\text{POS}^H(U) := \underline{U}^H, \qquad \text{BND}^H(U) := \overline{U}^H \setminus \underline{U}^H, \qquad \text{NEG}^H(U) := X \setminus \overline{U}^H.$$

**Example 2.12.2** (HesiRough set for e-commerce fraud screening under hesitant evidence)**.** Let $X = \{t_1, t_2, t_3, t_4, t_5\}$ be five online transactions. We model a hesitant "fraud" target using set-valued predicates $h_R$ (transaction linkage) and $h_U$ (fraud label), as in Definition 2.12.1.

**Hesitant relation predicate $h_R$.** Assume the nondegeneracy condition (H0):

$$h_R(x,x) = \{1\} \qquad (\forall x \in X).$$

For distinct transactions, define:

$$h_R(t_1,t_2) = h_R(t_2,t_1) = \{1\} \quad \text{(same stolen card and same device ID; definite link)},$$

$$h_R(t_4,t_5) = h_R(t_5,t_4) = \{1\} \quad \text{(same legitimate subscriber account; definite link)},$$

$$h_R(t_1,t_3) = h_R(t_3,t_1) = \{0,1\}, \qquad h_R(t_2,t_3) = h_R(t_3,t_2) = \{0,1\}$$

$$\text{(shared IP/shipping region; link plausible but unconfirmed)},$$

and for all remaining unordered pairs $\{x, y\}$ not listed above, set

$$h_R(x,y) = \{0\} \quad \text{(definitely not linked)}.$$

This yields the definite/possible relations

$$R_{\text{def}} = \{(x,y) \mid h_R(x,y) = \{1\}\}, \qquad R_{\text{pos}} = \{(x,y) \mid 1 \in h_R(x,y)\},$$

so $R_{\text{def}} \subseteq R_{\text{pos}}$.

**Hesitant target predicate $h_U$.** Let

$$h_U(t_1) = \{1\}, \qquad h_U(t_2) = \{1\} \quad \text{(chargeback confirmed)},$$

$$h_U(t_3) = \{0, 1\} \quad \text{(manual review pending; fraud uncertain)},$$

$$h_U(t_4) = \{0\}, \qquad h_U(t_5) = \{0\} \quad \text{(delivered and verified; non-fraud)}.$$

Then

$$U_{\mathrm{def}} = \{y \in X \mid h_U(y) = \{1\}\} = \{t_1, t_2\}, \qquad U_{\mathrm{pos}} = \{y \in X \mid 1 \in h_U(y)\} = \{t_1, t_2, t_3\}.$$

**Definite/possible neighborhoods.** From the above,

$$N_{\mathrm{def}}(t_1) = N_{\mathrm{def}}(t_2) = \{t_1, t_2\}, \quad N_{\mathrm{def}}(t_3) = \{t_3\}, \quad N_{\mathrm{def}}(t_4) = N_{\mathrm{def}}(t_5) = \{t_4, t_5\},$$

$$N_{\mathrm{pos}}(t_1) = N_{\mathrm{pos}}(t_2) = N_{\mathrm{pos}}(t_3) = \{t_1, t_2, t_3\}, \quad N_{\mathrm{pos}}(t_4) = N_{\mathrm{pos}}(t_5) = \{t_4, t_5\}.$$

**HesiRough approximations and regions.** The HesiRough lower/upper approximations are

$$\underline{U}^H = \{x \in X \mid N_{\mathrm{def}}(x) \subseteq U_{\mathrm{def}}\} = \{t_1, t_2\},$$

$$\overline{U}^H = \{x \in X \mid N_{\mathrm{pos}}(x) \cap U_{\mathrm{pos}} \neq \varnothing\} = \{t_1, t_2, t_3\}.$$

Hence the induced regions are

$$\mathrm{POS}^H(U) = \{t_1, t_2\}, \qquad \mathrm{BND}^H(U) = \{t_3\}, \qquad \mathrm{NEG}^H(U) = \{t_4, t_5\}.$$

Interpretation: $t_1, t_2$ are *definitely* fraudulent (their definite neighborhoods stay within confirmed fraud), $t_3$ is *boundary* (only possible linkage/label evidence), and $t_4, t_5$ are *negative* since even their possible neighborhoods do not intersect the possibly-fraudulent set.

**Theorem 2.12.3** (Well-definedness and basic inclusions)**.** *Under the assumptions of Definition 2.12.1, the sets $\underline{U}^H$ and $\overline{U}^H$ are well-defined subsets of $X$. Moreover,*

$$\underline{U}^H \subseteq \overline{U}^H \subseteq X,$$

*and the regions $\mathrm{POS}^H(U)$, $\mathrm{BND}^H(U)$, and $\mathrm{NEG}^H(U)$ are well-defined and form a disjoint cover of $X$:*

$$X = \mathrm{POS}^H(U) \,\dot{\cup}\, \mathrm{BND}^H(U) \,\dot{\cup}\, \mathrm{NEG}^H(U).$$

*Proof. (Well-definedness).* Since $h_R$ and $h_U$ are ordinary functions with codomain $\mathcal{P}(\{0,1\}) \setminus \{\varnothing\}$, the conditions $h_R(x, y) = \{1\}$ and $1 \in h_R(x, y)$ define subsets $R_{\mathrm{def}}, R_{\mathrm{pos}} \subseteq X \times X$, and $h_U(y) = \{1\}$ and $1 \in h_U(y)$ define subsets $U_{\mathrm{def}}, U_{\mathrm{pos}} \subseteq X$. Hence for each $x \in X$, the neighborhoods $N_{\mathrm{def}}(x), N_{\mathrm{pos}}(x) \subseteq X$ are well-defined. Therefore the defining predicates

$$x \in \underline{U}^H \iff N_{\mathrm{def}}(x) \subseteq U_{\mathrm{def}}, \qquad x \in \overline{U}^H \iff N_{\mathrm{pos}}(x) \cap U_{\mathrm{pos}} \neq \varnothing$$

are meaningful, so $\underline{U}^H, \overline{U}^H \subseteq X$ are well-defined.

*(Inclusion $\underline{U}^H \subseteq \overline{U}^H$).* First note that $R_{\mathrm{def}} \subseteq R_{\mathrm{pos}}$ and $U_{\mathrm{def}} \subseteq U_{\mathrm{pos}}$, hence $N_{\mathrm{def}}(x) \subseteq N_{\mathrm{pos}}(x)$ for all $x \in X$. Let $x \in \underline{U}^H$. Then $N_{\mathrm{def}}(x) \subseteq U_{\mathrm{def}}$. By (H0), $(x, x) \in R_{\mathrm{def}}$, so $x \in N_{\mathrm{def}}(x)$, hence $x \in U_{\text{def}} \subseteq U_{\text{pos}}$. Also $(x, x) \in R_{\text{def}} \subseteq R_{\text{pos}}$ implies $x \in N_{\text{pos}}(x)$. Therefore $x \in N_{\text{pos}}(x) \cap U_{\text{pos}}$, so $N_{\text{pos}}(x) \cap U_{\text{pos}} \neq \varnothing$ and thus $x \in \overline{U}^{H}$.

*(Region decomposition).* By definition,

$$\text{POS}^{H}(U) = \underline{U}^{H}, \quad \text{BND}^{H}(U) = \overline{U}^{H} \setminus \underline{U}^{H}, \quad \text{NEG}^{H}(U) = X \setminus \overline{U}^{H}.$$

These are well-defined. Using $\underline{U}^{H} \subseteq \overline{U}^{H}$, the three sets are pairwise disjoint. Finally,

$$\underline{U}^{H} \ \cup \ (\overline{U}^{H} \setminus \underline{U}^{H}) \ \cup \ (X \setminus \overline{U}^{H}) = X,$$

so they form a disjoint cover of $X$. □

## 2.13 GraphicRough Set

Graphic rough sets define lower/upper approximations of vertex subsets using graph neighborhoods or reachability, capturing uncertainty in network classification tasks [81].

**Definition 2.13.1** (GraphicRough Set induced by an attribute graph)**.** [81] Let $U$ be a nonempty finite universe of objects and let $V$ be a finite set of attributes. Let $G = (V, E)$ be an (undirected) graph on $V$ describing interrelationships among attributes. Assume that to each attribute $v \in V$ we associate an equivalence relation $R_v \subseteq U \times U$ (indiscernibility with respect to $v$).

Let $\mathsf{Sub}(G)$ denote the family of all subgraphs $H = (V_H, E_H)$ of $G$ (with $V_H \subseteq V$ and $E_H \subseteq E \cap \binom{V_H}{2}$). For each $H \in \mathsf{Sub}(G)$ define the combined equivalence relation

$$R_H \ := \bigcap_{v \in V_H} R_v.$$

For any target set $X \subseteq U$, define the $H$-lower and $H$-upper approximations by

$$\underline{X}_H \ := \ \{\, x \in U \mid [x]_{R_H} \subseteq X \,\}, \qquad \overline{X}_H \ := \ \{\, x \in U \mid [x]_{R_H} \cap X \neq \varnothing \,\},$$

where $[x]_{R_H} := \{\, y \in U \mid (x, y) \in R_H \,\}$. The *GraphicRough Set* of $X$ (induced by $G$) is the mapping

$$F_X : \mathsf{Sub}(G) \longrightarrow \mathcal{P}(U) \times \mathcal{P}(U), \qquad F_X(H) := \big(\underline{X}_H, \overline{X}_H\big).$$

**Example 2.13.2** (Credit screening as a GraphicRough Set)**.** Consider a small loan–application scenario.

**Objects.** Let

$$U = \{p_1, p_2, p_3, p_4, p_5, p_6\}$$

be six loan applicants.

**Attributes and their dependency graph.** Let the attribute set be

$$V = \{\text{Inc}, \text{Debt}, \text{Cred}\},$$

where Inc = income band, Debt = debt level, and Cred = credit history band. Assume the attribute graph $G = (V, E)$ is the path

$$E = \{\{\text{Inc}, \text{Debt}\}, \{\text{Debt}, \text{Cred}\}\},$$

encoding that Debt interacts with both income and credit in the screening process.

**Attribute values (coarse categories).**

| Applicant | Inc | Debt | Cred |
|---|---|---|---|
| $p_1$ | $H$ | $L$ | $G$ |
| $p_2$ | $H$ | $L$ | $F$ |
| $p_3$ | $H$ | $H$ | $P$ |
| $p_4$ | $L$ | $H$ | $P$ |
| $p_5$ | $L$ | $L$ | $F$ |
| $p_6$ | $L$ | $L$ | $G$ |

**Equivalence relations per attribute.** For each $v \in V$, define $R_v \subseteq U \times U$ by

$$(p_i, p_j) \in R_v \iff p_i \text{ and } p_j \text{ have the same value on attribute } v.$$

For instance, $R_{\text{Debt}}$ has two classes $\{p_1, p_2, p_5, p_6\}$ (low debt) and $\{p_3, p_4\}$ (high debt).

**Target concept.** Let

$$X = \{p_1, p_2, p_6\} \subseteq U$$

be the set of applicants the bank intends to *approve* (based on additional external checks).

**GraphicRough approximations under subgraphs.** Let $H$ be the subgraph induced by $\{\text{Inc}, \text{Debt}\}$ (so $R_H = R_{\text{Inc}} \cap R_{\text{Debt}}$). Then the $R_H$-classes are determined by the pair (Inc, Debt):

$$\{p_1, p_2\}, \ \{p_3\}, \ \{p_4\}, \ \{p_5, p_6\}.$$

Hence,

$$\underline{X}_H = \{p \in U \mid [p]_{R_H} \subseteq X\} = \{p_1, p_2\},$$

because $[p_1]_{R_H} = [p_2]_{R_H} = \{p_1, p_2\} \subseteq X$, while $[p_6]_{R_H} = \{p_5, p_6\} \nsubseteq X$. Moreover,

$$\overline{X}_H = \{p \in U \mid [p]_{R_H} \cap X \neq \varnothing\} = \{p_1, p_2, p_5, p_6\},$$

since the class $\{p_5, p_6\}$ intersects $X$ via $p_6$.

Now take the full-attribute subgraph $H' := G$ with $V_{H'} = \{\text{Inc}, \text{Debt}, \text{Cred}\}$. Then $R_{H'} = R_{\text{Inc}} \cap R_{\text{Debt}} \cap R_{\text{Cred}}$ yields singleton classes in this toy data, so the approximation becomes crisp:

$$\underline{X}_{H'} = X = \overline{X}_{H'}.$$

**Interpretation.** The GraphicRough Set map

$$F_X : \mathsf{Sub}(G) \to \mathcal{P}(U) \times \mathcal{P}(U), \qquad F_X(H) = (\underline{X}_H, \overline{X}_H),$$

encodes how the "definitely-approve" and "possibly-approve" applicants change as we move across different attribute subgraphs (i.e., different dependency-aware combinations of attributes).

## 2.14 ClusterRough Set

Cluster rough set uses clustering-induced granules instead of equivalence classes; lower/upper approximations aggregate clusters fully inside or intersecting target subset [81].

**Definition 2.14.1** (ClusterRough Set induced by a clustering of attributes)**.** [81] Let $U$ be a nonempty finite universe, $V$ a finite attribute set, and $\{R_v\}_{v\in V}$ equivalence relations on $U$ as above. Let $\mathcal{C} = \{C_1, \ldots, C_k\}$ be a partition (clustering) of $V$ into nonempty clusters.

For each cluster $C_j \in \mathcal{C}$ define the combined equivalence relation

$$R_{C_j} := \bigcap_{v\in C_j} R_v.$$

Given $X \subseteq U$, define the cluster-wise lower/upper approximations by

$$\underline{X}_{C_j} := \{ x \in U \mid [x]_{R_{C_j}} \subseteq X \}, \qquad \overline{X}_{C_j} := \{ x \in U \mid [x]_{R_{C_j}} \cap X \neq \varnothing \}.$$

The *ClusterRough Set* of $X$ (with respect to $\mathcal{C}$) is the mapping

$$G_X : \mathcal{C} \longrightarrow \mathcal{P}(U) \times \mathcal{P}(U), \qquad G_X(C_j) := (\underline{X}_{C_j}, \overline{X}_{C_j}).$$

**Example 2.14.2** (Credit-risk assessment via clustered attributes)**.** ClusterRough sets replace single-attribute granules by clusters of attributes, using $R_{C_j} = \bigcap_{v\in C_j} R_v$ and cluster-wise Pawlak lower/upper approximations $(\underline{X}_{C_j}, \overline{X}_{C_j})$ for each cluster $C_j$ (see the formal definition in [81]).

Let $U = \{a_1, a_2, a_3, a_4, a_5, a_6\}$ be six loan applicants. Consider four discretized attributes

$$V = \{\mathsf{Inc}, \mathsf{Debt}, \mathsf{Late}, \mathsf{Score}\},$$

where $\mathsf{Inc} \in \{H, L\}$ (high/low income), $\mathsf{Debt} \in \{H, L\}$ (high/low debt ratio), $\mathsf{Late} \in \{Y, N\}$ (recent late payments yes/no), and $\mathsf{Score} \in \{G, P\}$ (good/poor credit score band). Each attribute $v \in V$ induces an equivalence relation

$$x\, R_v\, y \iff f_v(x) = f_v(y),$$

where $f_v : U \to \mathrm{Dom}(v)$ records the (discretized) value of applicant $x$ on attribute $v$.

Assume the applicants have the following profiles:

| | Inc | Debt | Late | Score |
|---|---|---|---|---|
| $a_1$ | $H$ | $L$ | $N$ | $G$ |
| $a_2$ | $H$ | $L$ | $N$ | $G$ |
| $a_3$ | $H$ | $H$ | $Y$ | $P$ |
| $a_4$ | $L$ | $H$ | $Y$ | $P$ |
| $a_5$ | $L$ | $L$ | $N$ | $G$ |
| $a_6$ | $L$ | $L$ | $Y$ | $P$ |

We cluster the attributes into two groups:

$$\mathcal{C} = \{C_1, C_2\}, \qquad C_1 = \{\mathsf{Inc}, \mathsf{Debt}\} \text{ (financial cluster)}, \quad C_2 = \{\mathsf{Late}, \mathsf{Score}\} \text{ (credit-history cluster)}.$$

For each cluster $C_j \in \mathcal{C}$ define $R_{C_j} = \bigcap_{v \in C_j} R_v$.

Let $X \subseteq U$ be the target set of *high default-risk* applicants:

$$X = \{a_3, a_4, a_6\}.$$

**(1) Financial cluster $C_1 = \{\mathsf{Inc}, \mathsf{Debt}\}$.** The $R_{C_1}$-equivalence classes (same income bucket and debt bucket) are

$$[a_1]_{R_{C_1}} = \{a_1, a_2\}, \quad [a_3]_{R_{C_1}} = \{a_3\}, \quad [a_4]_{R_{C_1}} = \{a_4\}, \quad [a_5]_{R_{C_1}} = \{a_5, a_6\}.$$

Hence the cluster-wise approximations are

$$\underline{X}_{C_1} = \{x \in U \mid [x]_{R_{C_1}} \subseteq X\} = \{a_3, a_4\},$$

$$\overline{X}_{C_1} = \{x \in U \mid [x]_{R_{C_1}} \cap X \neq \varnothing\} = \{a_3, a_4, a_5, a_6\}.$$

Interpretation: using only financial attributes, $a_6$ is *not* certainly high-risk because it shares the same financial profile with $a_5 \notin X$ (a boundary effect).

**(2) Credit-history cluster $C_2 = \{\mathsf{Late}, \mathsf{Score}\}$.** The $R_{C_2}$-equivalence classes (same late-payment and score band) are

$$[a_1]_{R_{C_2}} = \{a_1, a_2, a_5\} \quad (\mathsf{Late} = N,\ \mathsf{Score} = G), \qquad [a_3]_{R_{C_2}} = \{a_3, a_4, a_6\} \quad (\mathsf{Late} = Y,\ \mathsf{Score} = P).$$

Therefore

$$\underline{X}_{C_2} = \{a_3, a_4, a_6\} = X, \qquad \overline{X}_{C_2} = \{a_3, a_4, a_6\} = X.$$

Interpretation: the credit-history cluster perfectly isolates the high-risk group in this toy dataset.

Finally, the ClusterRough representation of $X$ is the mapping

$$G_X : \mathcal{C} \to \mathcal{P}(U) \times \mathcal{P}(U), \qquad G_X(C_1) = (\{a_3, a_4\}, \{a_3, a_4, a_5, a_6\}), \quad G_X(C_2) = (X, X).$$

## 2.15 Multipolar Rough Set

A multipolar rough set applies Pawlak approximations to several overlapping poles, yielding an m-tuple of lower–upper pairs under one relation. Related concepts with a similar structural flavor include *multipolar fuzzy sets* [82], *multipolar neutrosophic sets* [83–85], and *multipolar plithogenic sets* [86].

**Definition 2.15.1** (Multipolar Rough Set)**.** Let $U \neq \varnothing$ be a universe and let $R \subseteq U \times U$ be an equivalence relation. Fix an integer $m \geq 2$ and let $X_1, \ldots, X_m \subseteq U$ be $m$ (not necessarily disjoint) subsets, interpreted as $m$ distinct evaluative "poles". For each $i \in \{1, \ldots, m\}$, define the Pawlak approximations

$$\underline{X_i} := \underline{R}(X_i), \qquad \overline{X_i} := \overline{R}(X_i).$$

The *multipolar rough set* determined by $(X_1, \ldots, X_m)$ (under $R$) is the $m$-tuple

$$\mathrm{MRS}(X_1, \ldots, X_m) := \big((\underline{X_1}, \overline{X_1}),\ (\underline{X_2}, \overline{X_2}),\ \ldots,\ (\underline{X_m}, \overline{X_m})\big).$$

**Example 2.15.2** (Multipolar rough set in clinical triage (differential diagnosis))**.** Consider an emergency clinic that performs a fast *initial screening* for each arriving patient. Let

$$U = \{p_1, p_2, \dots, p_n\}$$

be the set of patients seen in one day, and let the screening record be the binary feature vector

$$\varphi(p) := \big(\mathrm{Fever}(p),\ \mathrm{Cough}(p),\ \mathrm{SpO_2Low}(p),\ \mathrm{Travel}(p)\big) \in \{0,1\}^4.$$

Define an indiscernibility relation (equivalence relation) on $U$ by

$$(p,q) \in R \iff \varphi(p) = \varphi(q),$$

so that $[p]_R$ collects all patients with the same observable screening pattern.

In practice, clinicians may attach multiple *simultaneous* tentative labels (a differential diagnosis), so we consider several possibly overlapping "poles":

$$X_1 := \{\text{patients suspected of influenza}\},$$

$$X_2 := \{\text{patients suspected of COVID-19}\},$$

$$X_3 := \{\text{patients suspected of bacterial pneumonia}\}.$$

Overlaps $X_i \cap X_j \neq \varnothing$ naturally occur because the same patient can be suspected of multiple conditions before confirmatory tests.

For each pole $X_i$, compute Pawlak lower/upper approximations under the *same* $R$:

$$\underline{X_i} := \underline{R}(X_i) = \{p \in U \mid [p]_R \subseteq X_i\},$$

$$\overline{X_i} := \overline{R}(X_i) = \{p \in U \mid [p]_R \cap X_i \neq \varnothing\}.$$

Interpretation:

- $\underline{X_i}$ are patients *definitely* in pole $i$ given screening granules (everyone with the same screening pattern is labeled $i$).

- $\overline{X_i}$ are patients *possibly* in pole $i$ (at least one patient with the same pattern is labeled $i$).

Thus the clinical triage output can be represented as the 3-polar rough object

$$\mathrm{MRS}(X_1, X_2, X_3) = \big((\underline{X_1}, \overline{X_1}),\ (\underline{X_2}, \overline{X_2}),\ (\underline{X_3}, \overline{X_3})\big),$$

which simultaneously captures "definitely/possibly influenza", "definitely/possibly COVID-19", and "definitely/possibly pneumonia" under the same screening-based indiscernibility.

## 2.16 Bipartite Rough Set

Bipartite rough sets use a bipartite relation between two universes; approximations derive from cross-neighborhoods, enabling explicit granular two-sided uncertainty analysis.

**Definition 2.16.1** (Bipartite Rough Set)**.** Let $U \neq \varnothing$ be a universe and let $R \subseteq U \times U$ be an equivalence relation. Let $A$ and $B$ be two disjoint sets of attribute values (two "parts") and set the parameter domain

$$J \ := \ A \times B.$$

A *bipartite rough set* (under $R$) is a pair $(F, J)$ where $F : J \to \mathcal{P}(U)$ is a mapping. For each $(a, b) \in J$, define the lower and upper approximations of $F(a, b)$ by

$$\underline{F}(a, b) \ := \ \underline{R}\big(F(a, b)\big),$$

$$\overline{F}(a, b) \ := \ \overline{R}\big(F(a, b)\big).$$

Thus, each parameter $(a, b)$ induces the rough approximation pair $\big(\underline{F}(a, b), \overline{F}(a, b)\big)$.

**Example 2.16.2** (Bipartite rough customer segments for a credit-card campaign)**.** Let $U$ be the set of credit-card applicants in a given month. Let $C = \{\mathsf{IncomeBracket}, \mathsf{EmploymentType}, \mathsf{PastDefaultFlag}\}$ be condition attributes, and define the indiscernibility (equivalence) relation $R \subseteq U \times U$ by

$$(x, y) \in R \iff \forall a \in C,\ f_a(x) = f_a(y),$$

so that $[x]_R$ collects applicants with the same risk-profile summary.

Let the two disjoint "parts" of attribute values be

$$A = \{\mathsf{Young}, \mathsf{Middle}, \mathsf{Senior}\} \quad \text{(age group)},$$

$$B = \{\mathsf{Urban}, \mathsf{Suburban}, \mathsf{Rural}\} \quad \text{(residential zone)},$$

and set $J := A \times B$.

For each $(a, b) \in J$, define $F(a, b) \subseteq U$ as the set of applicants whose recorded age group is $a$, whose zone is $b$, and who clicked the bank's "premium card" offer (a behavioral signal of interest). Then $(F, J)$ is a bipartite rough set (under $R$), and for each $(a, b) \in J$ we form

$$\underline{F}(a, b) \ := \ \underline{R}\big(F(a, b)\big) = \{x \in U \mid [x]_R \subseteq F(a, b)\},$$

$$\overline{F}(a, b) \ := \ \overline{R}\big(F(a, b)\big) = \{x \in U \mid [x]_R \cap F(a, b) \neq \varnothing\}.$$

Interpretation: $\underline{F}(a, b)$ contains applicants *definitely* belonging to segment $(a, b)$ (all applicants with the same risk-profile also show interest and fall in the same part-values), whereas $\overline{F}(a, b)$ contains applicants who *possibly* belong to $(a, b)$. In practice, the bank can auto-target $\underline{F}(a, b)$ for a low-risk marketing action, and send the boundary $\overline{F}(a, b) \setminus \underline{F}(a, b)$ to manual review or to a softer offer.

## 2.17 TreeRough Set

A *TreeRough set* extends classical rough set approximations by indexing them with a fixed hierarchical (tree-structured) attribute system [44]. Intuitively, instead of working with a single indiscernibility relation, we consider a family of equivalence relations attached to the nodes of an attribute tree. A related concept with a similar hierarchical structure is the *TreeSoft set* (and its variants) [87–89]. The formal definition is given below.

**Definition 2.17.1** (TreeRough set)**.** Let $U$ be a nonempty (finite) universe, and let $\mathrm{Tree}(A)$ be a fixed rooted tree whose nodes represent attributes. Denote by $V(\mathrm{Tree}(A))$ the set of all nodes of $\mathrm{Tree}(A)$. Assume that each node $a \in V(\mathrm{Tree}(A))$ is equipped with an equivalence relation

$$R_a \subseteq U \times U,$$

and write the $R_a$-equivalence class of $x \in U$ as

$$[x]_{R_a} := \{ y \in U \mid (x, y) \in R_a \}.$$

For any subset $X \subseteq U$ and any node $a \in V(\mathrm{Tree}(A))$, define the lower and upper approximations of $X$ with respect to $R_a$ by

$$\underline{X}_a := \{ x \in U \mid [x]_{R_a} \subseteq X \}, \qquad \overline{X}_a := \{ x \in U \mid [x]_{R_a} \cap X \neq \varnothing \}.$$

The *TreeRough set* (tree-indexed rough approximation) of $X$ is the collection

$$\mathsf{TR}(X) := \left\{ (\underline{X}_a, \overline{X}_a) \;\middle|\; a \in V(\mathrm{Tree}(A)) \right\}.$$

**Example 2.17.2** (TreeRough set for hierarchical medical triage)**.** Let $U = \{p_1, p_2, p_3, p_4, p_5, p_6\}$ be a set of patients. Consider a rooted attribute tree $\mathrm{Tree}(A)$ whose nodes represent clinical attributes at different granularities:

$$\text{Clinical (root)} \longrightarrow \begin{cases} a_S = \text{Symptoms}, \\ a_I = \text{Imaging}. \end{cases}$$

We define, for each node $a \in V(\mathrm{Tree}(A))$, an equivalence relation $R_a \subseteq U \times U$ (so that $(x, y) \in R_a$ means that $x$ and $y$ are indiscernible with respect to the attribute(s) at $a$), as in the TreeRough framework.

**Data.** Assume the following observed values:

| patient | Fever | Cough | X-ray |
|---|---|---|---|
| $p_1$ | $H$ | $Y$ | $I$ |
| $p_2$ | $H$ | $Y$ | $I$ |
| $p_3$ | $N$ | $Y$ | $I$ |
| $p_4$ | $H$ | $N$ | $C$ |
| $p_5$ | $N$ | $Y$ | $C$ |
| $p_6$ | $N$ | $N$ | $C$ |

where $H/N$ means high/normal, $Y/N$ means yes/no, and $I/C$ means infiltrate/clear.

**Node relations.** Define $R_{a_S}$ by equality of the symptom pair (Fever, Cough), and define $R_{a_I}$ by equality of the X-ray result:

$$(x, y) \in R_{a_S} \iff (\text{Fever}(x), \text{Cough}(x)) = (\text{Fever}(y), \text{Cough}(y)),$$

$$(x, y) \in R_{a_I} \iff \text{Xray}(x) = \text{Xray}(y).$$

Hence the corresponding equivalence classes are

$$[p_1]_{R_{a_S}} = [p_2]_{R_{a_S}} = \{p_1, p_2\}, \quad [p_3]_{R_{a_S}} = [p_5]_{R_{a_S}} = \{p_3, p_5\}, \quad [p_4]_{R_{a_S}} = \{p_4\}, \quad [p_6]_{R_{a_S}} = \{p_6\},$$

and

$$[p_1]_{R_{a_I}} = [p_2]_{R_{a_I}} = [p_3]_{R_{a_I}} = \{p_1, p_2, p_3\}, \qquad [p_4]_{R_{a_I}} = [p_5]_{R_{a_I}} = [p_6]_{R_{a_I}} = \{p_4, p_5, p_6\}.$$

**Target concept.** Let $X \subseteq U$ be the set of patients who truly have pneumonia:

$$X = \{p_1, p_2, p_3\}.$$

**Tree-indexed rough approximations.** At the node $a_S$ (Symptoms), the Pawlak lower/upper approximations are

$$\underline{X}_{a_S} = \{x \in U \mid [x]_{R_{a_S}} \subseteq X\} = \{p_1, p_2\}, \qquad \overline{X}_{a_S} = \{x \in U \mid [x]_{R_{a_S}} \cap X \neq \varnothing\} = \{p_1, p_2, p_3, p_5\}.$$

Thus $p_5$ lies in the boundary at the symptom level (shares symptoms with $p_3$ but is not pneumonia).

At the node $a_I$ (Imaging), we obtain

$$\underline{X}_{a_I} = \{x \in U \mid [x]_{R_{a_I}} \subseteq X\} = \{p_1, p_2, p_3\}, \qquad \overline{X}_{a_I} = \{x \in U \mid [x]_{R_{a_I}} \cap X \neq \varnothing\} = \{p_1, p_2, p_3\}.$$

So the imaging node yields an exact (crisp) description of $X$ in this toy dataset.

**Interpretation.** The TreeRough set $\mathsf{TR}(X)$ collects these approximation pairs for *each* node in the attribute tree, enabling a hierarchical view: coarse screening at $a_S$ and refined certainty at $a_I$.

## 2.18 ForestRough Set

Forest rough set computes multiple lower/upper approximations across a forest of attribute trees, aggregating tree-wise rough descriptions for robust decisions. A related concept with a similar hierarchical structure is the *ForestSoft set* (and its variants) [90, 91].

**Definition 2.18.1** (ForestRough Set)**.** Let $U$ be a nonempty universe. Let $T$ be an index set and, for each $t \in T$, let $\mathsf{Tree}(A^{(t)})$ be an attribute tree. Define the (disjoint) forest of attributes by

$$\mathsf{Forest}(\{A^{(t)}\}_{t \in T}) := \bigsqcup_{t \in T} \mathsf{Tree}(A^{(t)}),$$

and assume that each node (attribute) $a \in \mathsf{Forest}(\{A^{(t)}\}_{t\in T})$ is equipped with an equivalence relation $R_a \subseteq U \times U$ on $U$.

For $x \in U$, write $[x]_{R_a} := \{\, y \in U \mid (x,y) \in R_a \,\}$. For any $X \subseteq U$ and any attribute-node $a \in \mathsf{Forest}(\{A^{(t)}\}_{t\in T})$, define the (lower/upper) rough approximations of $X$ w.r.t. $R_a$ by

$$\underline{X}_a \ := \ \{\, x \in U \mid [x]_{R_a} \subseteq X \,\}, \qquad \overline{X}_a \ := \ \{\, x \in U \mid [x]_{R_a} \cap X \neq \varnothing \,\}.$$

The *ForestRough Set* induced by the forest $\mathsf{Forest}(\{A^{(t)}\}_{t\in T})$ is the mapping

$$\mathcal{FR} : \mathcal{P}(U) \longrightarrow \mathcal{P}\big(\mathcal{P}(U) \times \mathcal{P}(U)\big), \qquad \mathcal{FR}(X) \ := \ \big\{\, (\underline{X}_a, \overline{X}_a) \ \big|\ a \in \mathsf{Forest}(\{A^{(t)}\}_{t\in T}) \,\big\}.$$

Equivalently, if each tree yields the TreeRough collection

$$\mathcal{TR}_t(X) := \{\, (\underline{X}_a, \overline{X}_a) \mid a \in \mathsf{Tree}(A^{(t)}) \,\},$$

then

$$\mathcal{FR}(X) = \bigcup_{t\in T} \mathcal{TR}_t(X).$$

**Example 2.18.2** (ForestRough set for hospital triage with multiple attribute trees)**.** Consider an emergency department (ED) triage task. Let $U$ be the finite set of patients who arrived during a fixed period (e.g., one month). We model the available clinical information by a *forest* of attribute trees, as in the ForestRough construction.

**Universe and target concept.** Let

$$U = \{\text{patients in the ED dataset}\}, \qquad X \subseteq U$$

where $X$ is the set of patients who were later confirmed (after full workup) to have *bacterial pneumonia requiring antibiotics.* At triage time, $X$ is not directly observable, so we approximate it.

**A forest of attribute trees.** Let $T := \{\text{sym}, \text{lab}, \text{vit}\}$ index three *separate* attribute trees:

- $\mathsf{Tree}(A^{(\text{sym})})$: symptoms hierarchy (e.g., respiratory $\to$ cough/dyspnea, systemic $\to$ chills, etc.),
- $\mathsf{Tree}(A^{(\text{lab})})$: laboratory hierarchy (e.g., inflammation $\to$ CRP/WBC/procalcitonin bins),
- $\mathsf{Tree}(A^{(\text{vit})})$: vital-signs hierarchy (e.g., temperature/$SpO_2$/respiratory-rate bins).

Form the disjoint forest of attributes

$$\mathsf{Forest}\big(\{A^{(t)}\}_{t\in T}\big) = \bigsqcup_{t\in T} \mathsf{Tree}(A^{(t)}).$$

**Equivalence relations at nodes.** For each attribute-node $a$ in the forest, define an equivalence relation $R_a \subseteq U \times U$ by

$$(x, y) \in R_a \iff x \text{ and } y \text{ fall in the same discretized category at node } a \text{ (same bin)}.$$

For instance, at the node "high fever" in $\mathsf{Tree}(A^{(\mathrm{vit})})$, two patients are $R_a$-equivalent if their temperatures both lie in the bin $[38.5°\mathrm{C}, \infty)$.

**ForestRough approximations (triage interpretation).** For each node $a \in \mathsf{Forest}(\{A^{(t)}\}_{t\in T})$, compute

$$\underline{X}_a = \{x \in U \mid [x]_{R_a} \subseteq X\}, \qquad \overline{X}_a = \{x \in U \mid [x]_{R_a} \cap X \neq \varnothing\}.$$

Interpretation:

- $x \in \underline{X}_a$ means that, *within the granule defined by* $a$, every patient indiscernible from $x$ (under $R_a$) ended up in $X$; thus $x$ is *definitely high-risk* under that attribute resolution.
- $x \in \overline{X}_a \setminus \underline{X}_a$ means the evidence at node $a$ is *ambiguous* (boundary under $a$).
- $x \notin \overline{X}_a$ means *no* patient in $x$'s $R_a$-granule belongs to $X$, so $x$ is *definitely not* in $X$ under that node.

The ForestRough description aggregates these rough views across *all* nodes in *all* trees:

$$\mathcal{FR}(X) = \big\{(\underline{X}_a, \overline{X}_a) \mid a \in \mathsf{Forest}(\{A^{(t)}\}_{t\in T})\big\} = \bigcup_{t\in T} \mathcal{TR}_t(X),$$

so triage decisions can be justified at different clinical granularities (symptoms vs labs vs vitals) and at multiple levels inside each hierarchy.

## 2.19 Dynamic Rough Set

Dynamic rough sets model time-varying knowledge by allowing the underlying approximation space (and optionally the target concept) to evolve over time [92–95].

**Definition 2.19.1** (Dynamic approximation space and Dynamic Rough Set)**.** Let $U$ be a nonempty (typically finite) universe and let $\mathbb{T}$ be a time index set (e.g., $\mathbb{T} = \mathbb{N}$ or $\{1, 2, \dots, T\}$). A *dynamic approximation space* is a family

$$\mathcal{A} = \big\{(U, R_t)\big\}_{t\in\mathbb{T}},$$

where for each $t \in \mathbb{T}$, $R_t \subseteq U \times U$ is an equivalence relation. For $x \in U$, write

$$[x]_{R_t} := \{\, y \in U \mid (x, y) \in R_t \,\}$$

for the $R_t$-equivalence class of $x$ at time $t$.

**(A) Fixed concept, evolving knowledge.** Given a fixed target concept $X \subseteq U$, the *time-t lower* and *time-t upper* approximations of $X$ are

$$\underline{\mathrm{apr}}_t(X) := \{\, x \in U \mid [x]_{R_t} \subseteq X \,\}, \qquad \overline{\mathrm{apr}}_t(X) := \{\, x \in U \mid [x]_{R_t} \cap X \neq \varnothing \,\}.$$

The *dynamic rough set* of $X$ (with respect to $\mathcal{A}$) is the time-indexed family

$$\mathsf{DRS}_{\mathcal{A}}(X) := \left\{\left(\underline{\mathrm{apr}}_t(X),\, \overline{\mathrm{apr}}_t(X)\right)\right\}_{t\in\mathbb{T}}.$$

Equivalently, one may regard $\mathsf{DRS}_{\mathcal{A}}(X)$ as a mapping $t \mapsto (\underline{\mathrm{apr}}_t(X), \overline{\mathrm{apr}}_t(X))$.

**(B) Evolving concept (optional generalization).** If the target concept itself varies with time, i.e., $X_t \subseteq U$ for each $t \in \mathbb{T}$, define

$$\mathsf{DRS}_{\mathcal{A}}(X_\bullet) := \left\{\left(\underline{\mathrm{apr}}_t(X_t),\, \overline{\mathrm{apr}}_t(X_t)\right)\right\}_{t\in\mathbb{T}}.$$

In either case, the induced *positive*, *boundary*, and *negative* regions at time $t$ are

$$\mathrm{POS}_t(X) := \underline{\mathrm{apr}}_t(X), \qquad \mathrm{BND}_t(X) := \overline{\mathrm{apr}}_t(X) \setminus \underline{\mathrm{apr}}_t(X), \qquad \mathrm{NEG}_t(X) := U \setminus \overline{\mathrm{apr}}_t(X).$$

**Remark 2.19.2** (Dynamic information systems (common source of $R_t$))**.** Often $R_t$ is induced by a time-dependent information system $S_t = (U, AT, f_t)$. For a fixed attribute subset $P \subseteq AT$, one sets

$$(x, y) \in \mathrm{IND}_t(P) \quad\Longleftrightarrow\quad f_t(x, a) = f_t(y, a) \text{ for all } a \in P,$$

and uses $R_t = \mathrm{IND}_t(P)$ in the above definition.

**Example 2.19.3** (Weekly merchant-risk monitoring as a Dynamic Rough Set)**.** A payment processor reassesses merchants every week as new chargeback data arrive. Let

$$U = \{m_1, m_2, m_3, m_4, m_5\} \qquad \text{and} \qquad \mathbb{T} = \{1, 2\}$$

represent five merchants observed at week $t = 1$ and week $t = 2$.

**Time-dependent knowledge (equivalence) relations.** At each week $t$, the processor discretizes (bins) observable features such as

$$P = \{\mathsf{country}, \mathsf{industry}, \mathsf{chargeback_bucket}\}$$

into categorical values, and records them via a feature map $f_t : U \to V$ (where $V$ is the finite product of the bins). Define the indiscernibility (equivalence) relation

$$(x, y) \in R_t \quad\Longleftrightarrow\quad f_t(x) = f_t(y) \qquad (x, y \in U).$$

Thus, $(U, R_t)_{t\in\mathbb{T}}$ is a dynamic approximation space.

**Target concept.** Let $X \subseteq U$ be the set of merchants confirmed as *high-risk* by investigators:

$$X = \{m_1, m_4\}.$$

**Week 1 (coarse evidence).** Suppose the week-1 bins are coarse, yielding the partition

$$[m_1]_{R_1} = [m_2]_{R_1} = [m_3]_{R_1} = \{m_1, m_2, m_3\}, \qquad [m_4]_{R_1} = [m_5]_{R_1} = \{m_4, m_5\}.$$

Then the time-1 lower/upper approximations of $X$ are

$$\underline{\mathrm{apr}}_1(X) = \{x \in U \mid [x]_{R_1} \subseteq X\} = \varnothing, \qquad \overline{\mathrm{apr}}_1(X) = \{x \in U \mid [x]_{R_1} \cap X \neq \varnothing\} = U.$$

Interpretation: with coarse features, no merchant is *definitely* high-risk, while all are *possibly* high-risk because each coarse class contains at least one confirmed high-risk merchant.

**Week 2 (refined evidence).** After an additional week, more data (e.g., a refined chargeback bucket) splits the classes:

$$[m_1]_{R_2} = \{m_1\}, \quad [m_2]_{R_2} = [m_3]_{R_2} = \{m_2, m_3\}, \quad [m_4]_{R_2} = \{m_4\}, \quad [m_5]_{R_2} = \{m_5\}.$$

Hence

$$\underline{\mathrm{apr}}_2(X) = \{m_1, m_4\}, \qquad \overline{\mathrm{apr}}_2(X) = \{m_1, m_4\}.$$

Interpretation: once the knowledge granules become finer, $m_1$ and $m_4$ become *definitely* high-risk, and $m_2, m_3, m_5$ become *definitely not* high-risk (at week 2).

**Dynamic rough set view.** The dynamic rough set of $X$ over $\mathbb{T} = \{1, 2\}$ is the time-indexed family

$$\mathsf{DRS}(X) = \Big\{ \big(\underline{\mathrm{apr}}_1(X), \overline{\mathrm{apr}}_1(X)\big), \big(\underline{\mathrm{apr}}_2(X), \overline{\mathrm{apr}}_2(X)\big) \Big\} = \Big\{ (\varnothing, U), (\{m_1, m_4\}, \{m_1, m_4\}) \Big\}.$$

## 2.20 L-valued rough sets

L-valued rough sets replace $[0, 1]$ with a lattice L for membership, defining approximations via L-relations and residuated operations systematically throughout [96–99]. As a concept with a similar structure, *lattice-valued fuzzy sets* are also well known [100–102].

**Definition 2.20.1** ($GL$-quantale (degree structure))**.** Let $(L, \wedge, \vee, 0, 1)$ be a complete lattice, and let $\odot : L \times L \to L$ be a commutative, associative binary operation such that $\alpha \odot 1 = \alpha$ for all $\alpha \in L$, and $\alpha \odot \bigvee_{j \in J} \beta_j = \bigvee_{j \in J}(\alpha \odot \beta_j)$ for all $\alpha \in L$ and families $\{\beta_j\}_{j \in J} \subseteq L$. Define the residuated implication $\Rightarrow: L \times L \to L$ by

$$\alpha \Rightarrow \beta \ := \ \bigvee \{\gamma \in L \mid \alpha \odot \gamma \leq \beta\}.$$

We call $(L, \odot)$ a *$GL$-quantale* if it additionally satisfies

$$\alpha \wedge \beta \ = \ \alpha \odot (\alpha \Rightarrow \beta) \qquad (\alpha, \beta \in L).$$

**Definition 2.20.2** ($L$-universe and $L$-powerset)**.** Let $X$ be a nonempty set and let $U : X \to L$ be a fixed $L$-set (called an *$L$-universe* on $X$). An $L$-set $Q : X \to L$ is called an *$L$-subset in $U$* if $Q(x) \leq U(x)$ for all $x \in X$. The family of all $L$-subsets of $U$ is denoted by

$$P(U) \ := \ \{ \, Q \in L^X \mid Q \leq U \, \},$$

and is called the *$L$-powerset* of $U$.

**Definition 2.20.3** ($L$-valued approximation space)**.** Let $U$ be an $L$-universe on $X$. A mapping $R : X \times X \to L$ is called an *$L$-valued relation on $U$* if

$$R(x, y) \; \le \; U(x) \wedge U(y) \qquad (x, y \in X).$$

The pair $(U, R)$ is called an *$L$-valued approximation space.*

**Definition 2.20.4** ($L$-valued rough approximation operators and $L$-valued rough set)**.** Let $(U, R)$ be an $L$-valued approximation space and let $Q \in P(U)$. Define mappings (operators) $R_*, R^* : P(U) \to P(U)$ by, for each $x \in X$,

$$R_*(Q)(x) \; := \; \bigwedge_{y \in X} \Big( U(x) \odot \big( R(y, x) \Rightarrow Q(y) \big) \Big),$$

$$R^*(Q)(x) \; := \; \bigvee_{y \in X} \Big( R(y, x) \odot \big( U(y) \Rightarrow Q(y) \big) \Big).$$

Then $R_*(Q)$ is called the *$L$-valued lower rough approximation* of $Q$, $R^*(Q)$ is called the *$L$-valued upper rough approximation* of $Q$, and the pair

$$\big( R_*(Q),\; R^*(Q) \big)$$

is called the *$L$-valued rough set (rough approximation) of* $Q$ in $(U, R)$.

**Example 2.20.5** ($L$-valued rough set for linguistic credit-risk screening)**.** A bank performs an early credit-risk triage using qualitative grades rather than precise probabilities.

**(1) Degree lattice.** Let

$$L = \Big\{ 0, \tfrac{1}{2}, 1 \Big\} \quad \text{with} \quad 0 \le \tfrac{1}{2} \le 1,$$

interpreted as $\{\text{Low}, \text{Medium}, \text{High}\}$. Use the Gödel (min) product

$$\alpha \odot \beta := \min\{\alpha, \beta\},$$

and Gödel residuum

$$\alpha \Rightarrow \beta := \begin{cases} 1, & \alpha \le \beta, \\ \beta, & \alpha > \beta. \end{cases}$$

**(2) Universe and an $L$-valued relation (similarity).** Let $X = \{u_1, u_2, u_3\}$ be three loan applicants. Take the $L$-universe $U : X \to L$ to be constant $U(u_i) = 1$ (all applicants are fully present).

Define an $L$-valued similarity relation $R : X \times X \to L$ by

| $R(\cdot, \cdot)$ | $u_1$ | $u_2$ | $u_3$ |
|---|---|---|---|
| $u_1$ | 1 | $\frac{1}{2}$ | 0 |
| $u_2$ | $\frac{1}{2}$ | 1 | $\frac{1}{2}$ |
| $u_3$ | 0 | $\frac{1}{2}$ | 1 |

where $R(u_i, u_j) = 1$ means “very similar credit profiles”, $R(u_i, u_j) = \frac{1}{2}$ means “moderately similar”, and $R(u_i, u_j) = 0$ means “dissimilar”.

**(3) $L$-subset (linguistic risk concept).** Let $Q \in L^X$ represent the analysts' preliminary (linguistic) judgment of the concept "likely to default":

$$Q(u_1) = \tfrac{1}{2}, \qquad Q(u_2) = 1, \qquad Q(u_3) = 0.$$

**(4) $L$-valued lower/upper approximations.** With $U(\cdot) = 1$, the general operators reduce to

$$R_*(Q)(x) = \bigwedge_{y \in X} \big(R(y,x) \Rightarrow Q(y)\big), \qquad R^*(Q)(x) = \bigvee_{y \in X} \big(R(y,x) \odot Q(y)\big).$$

A direct computation yields:

| $x$ | $R_*(Q)(x)$ | $R^*(Q)(x)$ |
|---|---|---|
| $u_1$ | $\frac{1}{2}$ | $\frac{1}{2}$ |
| $u_2$ | 0 | 1 |
| $u_3$ | 0 | $\frac{1}{2}$ |

**Interpretation.** Applicant $u_1$ is *definitely* medium-risk ($\frac{1}{2}$) given the similarity structure. Applicant $u_2$ is *possibly* high-risk (upper $= 1$) but not definitely so (lower $= 0$), reflecting conflicting evidence from similar medium/low-risk profiles. Applicant $u_3$ is not definitely risky (lower $= 0$) but is possibly medium-risk (upper $= \frac{1}{2}$) because $u_3$ is moderately similar to the high-risk applicant $u_2$.

## 2.21 Graded Rough Set

Graded rough sets relax strict inclusion by requiring each equivalence class overlap a target set by at least $k$ elements [103–106].

**Definition 2.21.1** (Graded rough approximations and graded rough set)**.** Let $U$ be a nonempty finite universe and let $R \subseteq U \times U$ be an equivalence relation. For $x \in U$, write

$$[x]_R := \{\, y \in U \mid (x,y) \in R \,\}$$

for the $R$-equivalence class (granule) of $x$. Fix a nonnegative integer $k \in \mathbb{N}_0$, called the *grade*.

For any $A \subseteq U$, the *grade-$k$ $R$ upper* and *grade-$k$ $R$ lower* approximations of $A$ are defined by

$$\overline{\mathrm{apr}}_R^{\,k}(A) \;:=\; \bigcup \Big\{\, [x]_R \;\Big|\; |[x]_R \cap A| > k \,\Big\},$$

$$\underline{\mathrm{apr}}_R^{\,k}(A) \;:=\; \bigcup \Big\{\, [x]_R \;\Big|\; |[x]_R \setminus A| \le k \,\Big\} \;=\; \bigcup \Big\{\, [x]_R \;\Big|\; |[x]_R| - |[x]_R \cap A| \le k \,\Big\}.$$

The pair $\big(\underline{\mathrm{apr}}_R^{\,k}(A), \overline{\mathrm{apr}}_R^{\,k}(A)\big)$ is called the *graded (grade-k) rough approximation* of $A$. If $\underline{\mathrm{apr}}_R^{\,k}(A) = \overline{\mathrm{apr}}_R^{\,k}(A)$, then $A$ is *$R$-definable by grade $k$*; otherwise, $A$ is called a *graded (grade-k) rough set* (with respect to $R$).

**Example 2.21.2** (Graded rough set in clinical triage (allowing up to one exception))**.** Consider an emergency department that groups patients by a coarse symptom profile (e.g., *high fever & cough*, *moderate fever & myalgia*, *low fever/no cough*). Let

$$U = \{p_1, p_2, p_3, p_4, p_5, p_6, p_7, p_8, p_9\}$$

be the patients arriving today, and let $R$ be the equivalence relation

$$(p_i, p_j) \in R \quad \Longleftrightarrow \quad p_i \text{ and } p_j \text{ have the same symptom profile.}$$

Assume the induced equivalence classes are

$$C_1 = \{p_1, p_2, p_3\} \text{ (high fever \& cough)},$$
$$C_2 = \{p_4, p_5, p_6, p_7\} \text{ (moderate fever \& myalgia)},$$
$$C_3 = \{p_8, p_9\} \text{ (low fever/no cough)}.$$

Let $A \subseteq U$ be the set of patients who truly have influenza (confirmed later by PCR):

$$A = \{p_1, p_2, p_3, p_4, p_5\}.$$

Fix grade $k = 1$ (we tolerate at most one "exception" in a class when declaring *definite* membership). Then the grade-1 lower approximation is

$$\underline{\mathrm{apr}}_R^1(A)$$
$$= \bigcup \Big\{ [x]_R \;\Big|\; |[x]_R \setminus A| \leq 1 \Big\} = C_1,$$

because $|C_1 \setminus A| = 0 \leq 1$ but $|C_2 \setminus A| = 2 > 1$ and $|C_3 \setminus A| = 2 > 1$. The grade-1 upper approximation is

$$\overline{\mathrm{apr}}_R^1(A)$$
$$= \bigcup \Big\{ [x]_R \;\Big|\; |[x]_R \cap A| > 1 \Big\} = C_1 \cup C_2,$$

because $|C_1 \cap A| = 3 > 1$ and $|C_2 \cap A| = 2 > 1$, while $|C_3 \cap A| = 0 \not> 1$.

Hence, the induced regions are:

$$\mathrm{POS}^{(k=1)}(A) = C_1 = \{p_1, p_2, p_3\},$$
$$\mathrm{BND}^{(k=1)}(A) = C_2 = \{p_4, p_5, p_6, p_7\},$$
$$\mathrm{NEG}^{(k=1)}(A) = C_3 = \{p_8, p_9\}.$$

Interpretation: the symptom class $C_1$ is "definitely flu" up to one exception (here, none), $C_2$ is "possibly flu" (mixed outcomes), and $C_3$ is "definitely not flu".

## 2.22 Linguistic Rough Set

Linguistic rough sets approximate fuzzy concepts using linguistic quantifiers and summaries, producing lower/upper linguistic approximations that match reasoning under uncertainty [107].

**Definition 2.22.1** (Linguistic approximation space (LAS))**.** Let $U = \{x_1, \ldots, x_m\}$ be a nonempty finite set of objects. Let $L = \{s_0, s_1, \ldots, s_g\}$ be a finite totally ordered set of linguistic labels (with $s_0 \prec s_1 \prec \cdots \prec s_g$), where $s_0$ plays the role of the least label. Let $C = \{C_1, \ldots, C_n\}$ be a family of *linguistic concepts*, where each $C_j$ is a mapping

$$C_j : U \to L.$$

The triple $\langle U, C, L \rangle$ is called a *linguistic approximation space* (LAS).

Throughout, for $V, W : U \to L$ we use the pointwise lattice operations

$$(V \wedge W)(x) := \min_{\prec}\{V(x), W(x)\}, \qquad (V \vee W)(x) := \max_{\prec}\{V(x), W(x)\} \qquad (x \in U),$$

and for $K \subseteq C$ (resp. $L' \subseteq C$) we write

$$\bigwedge_{C_j \in K} C_j, \qquad \bigvee_{C_j \in L'} C_j$$

for the iterated pointwise $\wedge$ and $\vee$.

**Definition 2.22.2** (Support and inclusion degree)**.** For a concept $V : U \to L$, define its support by

$$\mathrm{supp}(V) := \{x \in U \mid V(x) \succ s_0\}, \qquad |V| := |\mathrm{supp}(V)|.$$

For two concepts $V, W : U \to L$ with $|V| > 0$, the *(linguistic) inclusion degree* of $V$ in $W$ is

$$D(W, V) \;:=\; \frac{\big|\{x \in \mathrm{supp}(V) \mid V(x) \preceq W(x)\}\big|}{|V|} \;\in [0, 1].$$

**Definition 2.22.3** (Degree of certainty for approximating a decision concept)**.** Let $\langle U, C, L\rangle$ be a LAS and let $Y : U \to L$ be a (linguistic) decision concept. Define

$$k_L \;:=\; D\left(Y, \bigwedge_{C_j \in C} C_j\right), \qquad k_U \;:=\; D\left(\bigvee_{C_j \in C} C_j,\; Y\right), \qquad k \;:=\; \min\{k_L, k_U\}.$$

**Definition 2.22.4** ($k$-approximability and linguistic rough approximations)**.** Let $\langle U, C, L\rangle$ be a LAS, let $Y : U \to L$, and fix $k \in [0, 1]$. Define two families of attribute-subsets

$$P(Y) \;:=\; \left\{K \subseteq C \;\middle|\; D\left(Y, \bigwedge_{C_j \in K} C_j\right) \geq k\right\},$$

$$Q(Y) \;:=\; \left\{L' \subseteq C \;\middle|\; D\left(\bigvee_{C_j \in L'} C_j,\; Y\right) \geq k\right\}.$$

We say that $Y$ is *$k$-approximable (in $\langle U, C, L\rangle$)* if $P(Y) \neq \varnothing$ and $Q(Y) \neq \varnothing$.

Assume $Y$ is $k$-approximable. Choose $K^* \in P(Y)$ and $L^* \in Q(Y)$ such that

$$\left|\mathrm{supp}\Big(\bigvee_{C_j \in L^*} C_j\Big) \setminus \mathrm{supp}\Big(\bigwedge_{C_j \in K^*} C_j\Big)\right|$$

is minimized among all pairs $(K, L') \in P(Y) \times Q(Y)$. Define the *linguistic rough lower* and *linguistic rough upper* approximations of $Y$ by

$$\underline{Y}^{k} \;:=\; \bigwedge_{C_j \in K^*} C_j, \qquad \overline{Y}^{k} \;:=\; \bigvee_{C_j \in L^*} C_j.$$

The resulting *linguistic rough set* (LRS) of $Y$ (at level $k$) is the pair

$$\mathcal{LRS}_k(Y) \;:=\; \big(\underline{Y}^{k},\; \overline{Y}^{k}\big).$$

**Example 2.22.5** (Hotel-review Linguistic Rough Set (real-life example))**.** Consider a small hotel-recommendation task where review summaries are expressed by linguistic labels.

**Objects (hotels).** Let

$$U = \{h_1, h_2, h_3, h_4\}.$$

**Linguistic label set.** Let

$$L = \{s_0, s_1, s_2, s_3, s_4\}, \qquad s_0 \prec s_1 \prec s_2 \prec s_3 \prec s_4,$$

interpreted as

$$s_0 = \text{very low},\ s_1 = \text{low},\ s_2 = \text{medium},\ s_3 = \text{high},\ s_4 = \text{very high}.$$

**Linguistic concepts (attributes).** Let $C = \{C_1, C_2, C_3\}$ where $C_1$ = Cleanliness, $C_2$ = Service, $C_3$ = Location, each $C_j : U \to L$. Assume the following linguistic evaluations (e.g., obtained by aggregating text reviews):

| | $h_1$ | $h_2$ | $h_3$ | $h_4$ |
|---|---|---|---|---|
| $C_1$ (Cleanliness) | $s_3$ | $s_2$ | $s_1$ | $s_3$ |
| $C_2$ (Service) | $s_4$ | $s_3$ | $s_2$ | $s_2$ |
| $C_3$ (Location) | $s_3$ | $s_2$ | $s_1$ | $s_4$ |

**Decision concept.** Let $Y : U \to L$ denote the linguistic decision "Overall recommended":

$$Y(h_1) = s_3, \quad Y(h_2) = s_2, \quad Y(h_3) = s_1, \quad Y(h_4) = s_3.$$

**Compute a $k$-level linguistic rough approximation.** Using pointwise operations

$$(V \wedge W)(x) = \min_{\prec}\{V(x), W(x)\}, \qquad (V \vee W)(x) = \max_{\prec}\{V(x), W(x)\},$$

take

$$K^* = \{C_1, C_2\}, \qquad L^* = \{C_2, C_3\}.$$

Then the candidate lower/upper linguistic approximations are

$$\underline{Y}^k := \bigwedge_{C_j \in K^*} C_j = \min_{\prec}\{C_1, C_2\}, \qquad \overline{Y}^k := \bigvee_{C_j \in L^*} C_j = \max_{\prec}\{C_2, C_3\}.$$

Concretely,

$$\underline{Y}^k(h_1) = s_3,\ \underline{Y}^k(h_2) = s_2,\ \underline{Y}^k(h_3) = s_1,\ \underline{Y}^k(h_4) = s_2,$$
$$\overline{Y}^k(h_1) = s_4,\ \overline{Y}^k(h_2) = s_3,\ \overline{Y}^k(h_3) = s_2,\ \overline{Y}^k(h_4) = s_4.$$

**Interpretation.** The lower approximation $\underline{Y}^k$ represents hotels that are *definitely* recommended at level $k$ based on conservative aggregation of key attributes (here, cleanliness and service), while the upper approximation $\overline{Y}^k$ represents hotels that are *possibly* recommended at level $k$ based on optimistic aggregation (here, service or location).

Hence, the linguistic rough set of the decision concept $Y$ (at level $k$) is

$$\mathcal{LRS}_k(Y) = (\underline{Y}^k, \overline{Y}^k).$$

## 2.23 Weak Rough Set

A weak rough set is any pair of subsets (lower, upper) with lower contained in upper, representing certainty and possibility [108, 109].

**Definition 2.23.1** (Weak rough set)**.** [109] Let $U$ be a nonempty universe. A *weak rough set* over $U$ is an ordered pair

$$A = (A_L, A_U), \qquad A_L, A_U \subseteq U, \qquad A_L \subseteq A_U,$$

where $A_L$ is called the *lower approximation* (definite part) and $A_U$ is called the *upper approximation* (possible part).

Equivalently, a point $x \in U$ is interpreted as

$$x \in A_L \Rightarrow x \text{ definitely belongs to } A,$$

$$x \in A_U \setminus A_L \Rightarrow x \text{ is undecidable for } A,$$

$$x \notin A_U \Rightarrow x \text{ definitely does not belong to } A.$$

The *positive*, *boundary*, and *negative* regions of $A$ are defined by

$$\mathrm{POS}(A) := A_L, \qquad \mathrm{BND}(A) := A_U \setminus A_L, \qquad \mathrm{NEG}(A) := U \setminus A_U.$$

**Definition 2.23.2** (Basic operations on weak rough sets)**.** Let $A = (A_L, A_U)$ and $B = (B_L, B_U)$ be weak rough sets over $U$. Define

$$A \cup B := (A_L \cup B_L,\ A_U \cup B_U), \qquad A \cap B := (A_L \cap B_L,\ A_U \cap B_U),$$

$$A^c := (U \setminus A_U,\ U \setminus A_L),$$

and the (componentwise) inclusion order

$$A \subseteq B \iff A_L \subseteq B_L \text{ and } A_U \subseteq B_U.$$

These operations preserve the constraint "lower $\subseteq$ upper", hence are well-defined on weak rough sets.

**Remark 2.23.3** (Relationship to Pawlak rough sets)**.** Given an approximation space $(U, R)$ (typically $R$ is an equivalence relation) and any $X \subseteq U$, the Pawlak rough approximation pair $(\underline{R}(X), \overline{R}(X))$ is a weak rough set. Thus, *Pawlak rough sets are special cases of weak rough sets* in which the pair $(A_L, A_U)$ is induced by a specific information relation $R$ (and hence is constrained by that $R$).

Table 2.3: Concise comparison between Pawlak rough sets and weak rough sets.

| **Aspect** | **Pawlak rough set** | **Weak rough set** |
|---|---|---|
| Primitive data | Approximation space $(U, R)$ (usually $R$ an equivalence) plus target set $X \subseteq U$. | Only a pair $(A_L, A_U)$ of subsets of $U$ with $A_L \subseteq A_U$. |
| How (lower, upper) arise | Derived from $R$ via $\underline{R}(X) = \{x \mid [x]_R \subseteq X\}$ and $\overline{R}(X) = \{x \mid [x]_R \cap X \neq \emptyset\}$. | Chosen/estimated directly; no requirement that it comes from any indiscernibility/information relation. |
| Admissible pairs | Constrained by the granulation induced by $R$ (e.g., unions of $R$-classes). | Any pair of subsets satisfying $A_L \subseteq A_U$ (no granulation constraint). |
| Regions / semantics | Positive/boundary/negative regions determined by $(\underline{R}(X), \overline{R}(X))$. | Same semantics using $\mathrm{POS}(A) = A_L$, $\mathrm{BND}(A) = A_U \setminus A_L$, $\mathrm{NEG}(A) = U \setminus A_U$. |
| Set operations | Often studied via approximation operators induced by $R$. | Naturally closed under componentwise $\cup, \cap$ and complement $(U \setminus A_U,\ U \setminus A_L)$. |

## 2.24 Decision-Theoretic Rough sets

Decision theoretic rough sets classify objects using Bayesian expected risk and loss, yielding positive, boundary, negative regions with optimal actions [110–113]. Decision-theoretic rough sets have also seen a large number of research publications in recent years [114–117]. Multigranulation decision-theoretic rough sets [118–120] and multi-class decision-theoretic rough sets [121] are also known as related concepts.

**Definition 2.24.1** (Decision-Theoretic Rough Set (DTRS))**.** [110–113] Let $U$ be a nonempty finite universe and let $E \subseteq U \times U$ be an equivalence relation. For $x \in U$, write

$$[x]_E := \{\, y \in U \mid (x, y) \in E \,\}$$

for the $E$-equivalence class (information granule) of $x$. Fix a target concept $C \subseteq U$ and denote its complement by $C^c := U \setminus C$.

**(1) Conditional probability.** Define the conditional probability that granule $[x]_E$ belongs to $C$ by

$$p(x) \ := \ \mathrm{Prob}(C \mid [x]_E) \in [0, 1].$$

In the standard finite and uniform setting, one often uses the empirical estimate

$$p(x) \ := \ \frac{|[x]_E \cap C|}{|[x]_E|}.$$

**(2) Actions, states, and losses.** Consider three actions

$$\mathsf{Act} := \{\mathsf{Act}_P, \mathsf{Act}_B, \mathsf{Act}_N\},$$

interpreted as *accept* ($\mathsf{Act}_P$), *defer* ($\mathsf{Act}_B$), and *reject* ($\mathsf{Act}_N$) the hypothesis "$x \in C$". Let the set of states be $\Omega := \{C, C^c\}$ and let

$$\lambda : \mathsf{Act} \times \Omega \to \mathbb{R}_{\geq 0}$$

be a loss (cost) function. Write $\lambda_{iP} := \lambda(\mathsf{Act}_i, C)$ and $\lambda_{iN} := \lambda(\mathsf{Act}_i, C^c)$ for $i \in \{P, B, N\}$.

**(3) Conditional risks (expected losses).** For each $x \in U$ and action $\mathsf{Act}_i \in \mathsf{Act}$, define the conditional risk

$$\mathrm{Risk}(\mathsf{Act}_i \mid x) \;:=\; \lambda_{iP}\, p(x) \;+\; \lambda_{iN}\, \big(1 - p(x)\big), \qquad i \in \{P, B, N\}.$$

**(4) Minimum-risk decision rule and induced regions.** Assume the standard cost ordering

$$\lambda_{PP} \le \lambda_{BP} < \lambda_{NP}, \qquad \lambda_{NN} \le \lambda_{BN} < \lambda_{PN},$$

which expresses that (i) accepting is cheapest when $x \in C$ and rejecting is most costly then, and (ii) rejecting is cheapest when $x \notin C$ and accepting is most costly then. Define the probability thresholds

$$\alpha \;:=\; \frac{\lambda_{PN} - \lambda_{BN}}{(\lambda_{PN} - \lambda_{BN}) + (\lambda_{BP} - \lambda_{PP})}, \qquad \beta \;:=\; \frac{\lambda_{BN} - \lambda_{NN}}{(\lambda_{BN} - \lambda_{NN}) + (\lambda_{NP} - \lambda_{BP})}.$$

(Under the above ordering, one has $0 \le \beta < \alpha \le 1$.)

Then the DTRS three-way decision regions for $C$ are

$$\mathrm{POS}_{(\alpha,\beta)}(C) := \{\, x \in U \mid p(x) \ge \alpha \,\}, \quad \mathrm{NEG}_{(\alpha,\beta)}(C) := \{\, x \in U \mid p(x) \le \beta \,\},$$

$$\mathrm{BND}_{(\alpha,\beta)}(C) := U \setminus \big(\mathrm{POS}_{(\alpha,\beta)}(C) \cup \mathrm{NEG}_{(\alpha,\beta)}(C)\big).$$

Equivalently, the minimum-risk rule is:

$$p(x) \ge \alpha \Rightarrow \text{choose } \mathsf{Act}_P, \qquad p(x) \le \beta \Rightarrow \text{choose } \mathsf{Act}_N, \qquad \beta < p(x) < \alpha \Rightarrow \text{choose } \mathsf{Act}_B.$$

**(5) DTRS approximations.** Define the decision-theoretic lower and upper approximations of $C$ by

$$\underline{\mathrm{apr}}_{(\alpha,\beta)}(C) := \mathrm{POS}_{(\alpha,\beta)}(C), \qquad \overline{\mathrm{apr}}_{(\alpha,\beta)}(C) := U \setminus \mathrm{NEG}_{(\alpha,\beta)}(C) = \mathrm{POS}_{(\alpha,\beta)}(C) \cup \mathrm{BND}_{(\alpha,\beta)}(C).$$

The pair $\big(\underline{\mathrm{apr}}_{(\alpha,\beta)}(C), \overline{\mathrm{apr}}_{(\alpha,\beta)}(C)\big)$ is called the *decision-theoretic rough approximation* of $C$ (with respect to $E$ and $\lambda$).

**Example 2.24.2** (Real-life DTRS: credit approval with three-way decisions)**.** Consider a bank that screens loan applicants. Let $U$ be the set of applicants in the current month. Define an equivalence relation $E$ by *information granules*: two applicants are $E$-equivalent if they share the same discretized profile (e.g., income-band, employment-type, credit-score-band, and debt-ratio-band).

Let the target concept be

$$C := \{\text{applicants who will } \textit{not} \text{ default within 12 months}\}, \qquad C^c := \{\text{default}\}.$$

For an applicant $x \in U$, estimate

$$p(x) = \mathrm{Prob}(C \mid [x]_E) \approx \frac{|[x]_E \cap C|}{|[x]_E|}$$

from historical outcomes of past applicants in the same granule.

The bank uses three actions:

$$\mathsf{Act}_P = \text{approve}, \qquad \mathsf{Act}_B = \text{manual review / request more documents}, \qquad \mathsf{Act}_N = \text{reject}.$$

A simple loss model (in monetary units) is:

$$\lambda_{PP} = 0 \quad \text{(normal servicing cost)}, \qquad \lambda_{PN} = 100 \quad \text{(expected loss if default after approval)},$$

$$\lambda_{BP} = 5 \quad \text{(review cost if actually safe)}, \qquad \lambda_{BN} = 10 \quad \text{(review cost even if risky)},$$

$$\lambda_{NP} = 30 \quad \text{(opportunity cost of rejecting a safe applicant)}, \qquad \lambda_{NN} = 0 \quad \text{(correct rejection)}.$$

Then the thresholds in Definition (DTRS) are

$$\alpha = \frac{\lambda_{PN} - \lambda_{BN}}{(\lambda_{PN} - \lambda_{BN}) + (\lambda_{BP} - \lambda_{PP})} = \frac{100 - 10}{(100 - 10) + (5 - 0)} = \frac{90}{95} \approx 0.947,$$

$$\beta = \frac{\lambda_{BN} - \lambda_{NN}}{(\lambda_{BN} - \lambda_{NN}) + (\lambda_{NP} - \lambda_{BP})} = \frac{10 - 0}{(10 - 0) + (30 - 5)} = \frac{10}{35} \approx 0.286.$$

Hence the three-way decision rule becomes:

$$p(x) \geq 0.947 \Rightarrow \text{approve } (\mathsf{Act}_P),$$

$$p(x) \leq 0.286 \Rightarrow \text{reject } (\mathsf{Act}_N),$$

$$0.286 < p(x) < 0.947 \Rightarrow \text{defer } (\mathsf{Act}_B).$$

Interpreting the regions,

$$\mathrm{POS}_{(\alpha,\beta)}(C) = \{\text{high-confidence safe applicants}\},$$

$$\mathrm{NEG}_{(\alpha,\beta)}(C) = \{\text{high-confidence risky applicants}\},$$

$$\mathrm{BND}_{(\alpha,\beta)}(C) = \{\text{uncertain applicants sent to manual review}\}.$$

This is a typical operational setting where DTRS produces an *approve / reject / review* policy from granulated empirical probabilities and asymmetric costs.

## 2.25 Type-$n$ Rough Set

A Type-$n$ rough set is an $n$-level hierarchical construction in which parameters are mapped recursively to lower-level rough sets, ultimately terminating in the classical Pawlak lower–upper approximations of subsets of the universe. Related layered frameworks include Type-$n$ fuzzy sets [122, 123], Type-$n$ neutrosophic sets [124–126], and Type-$n$ soft sets [127].

**Definition 2.25.1** (Type-$n$ rough set)**.** Let $\mathsf{PA} = (U, E, \rho)$ be a parameterized approximation space. Define recursively the collections $\Sigma^{(n)}(U, E, \rho)$ of *type-n rough sets* as follows.

(1) $\Sigma^{(1)}(U, E, \rho)$ is the collection of all type-1 rough sets $\mathsf{RS}^{(1)}(X, B)$.

(2) For $n \geq 2$, a *type-n rough set* (briefly, $\mathbf{T}n\mathbf{RS}$) over $\mathsf{PA}$ is a pair $(\mathcal{F}^{(n)}, A_n)$ where $A_n \subseteq E$ is a nonempty primary parameter set and

$$\mathcal{F}^{(n)} : A_n \to \Sigma^{(n-1)}(U, E, \rho), \qquad a \longmapsto \mathcal{F}^{(n)}(a) = \big(\mathcal{F}_a^{(n-1)},\, L_a\big),$$

such that $L_a \subseteq E$ is nonempty for every $a \in A_n$. The union $\bigcup_{a \in A_n} L_a$ is called the *underlying parameter set* of the type-$n$ rough set.

**Example 2.25.2** (Real-life Type-$n$ rough set: multi-stage medical triage under layered uncertainty)**.** Let $U$ be a finite set of patients arriving at an emergency department. Let $E$ be a finite set of clinical *parameters* (features/tests), e.g.,

$$E = \{\mathsf{AgeBand}, \mathsf{SpO_2\text{-}Band}, \mathsf{TempBand}, \mathsf{CRP\text{-}Band}, \mathsf{CT\text{-}Finding}, \mathsf{ComorbidityScore}, \dots\}.$$

Assume $\rho$ associates to each $B \subseteq E$ an indiscernibility relation on $U$: patients $x, y$ are $\rho(B)$-equivalent if they share the same values on all parameters in $B$.

Fix a target concept $X \subseteq U$:

$$X := \{\text{patients who truly have a severe condition requiring admission}\}.$$

**Type-1 (single-parameter-set) rough assessment.** For a chosen parameter set $B \subseteq E$ (e.g. $B = \{\mathsf{SpO_2\text{-}Band}, \mathsf{TempBand}\}$), the type-1 rough set $\mathsf{RS}^{(1)}(X, B)$ gives a *certain-admit* lower region and a *possible-admit* upper region, based only on quick vitals.

**Type-2 (primary parameter selects a protocol, then type-1 inside).** Let the *primary* parameter set be

$$A_2 = \{\mathsf{Protocol}\} \subseteq E,$$

where $\mathsf{Protocol}$ takes values such as $\mathsf{Respiratory}$, $\mathsf{Cardiac}$, $\mathsf{Sepsis}$. Define $\mathcal{F}^{(2)} : A_2 \to \Sigma^{(1)}(U, E, \rho)$ by mapping the single primary choice $a = \mathsf{Protocol}$ to a type-1 rough set whose secondary parameter set depends on the protocol:

$$\mathcal{F}^{(2)}(\mathsf{Protocol}) = \big(\mathsf{RS}^{(1)}(X, L_{\mathsf{Protocol}}),\, L_{\mathsf{Protocol}}\big),$$

with, for instance,

$$L_{\mathsf{Respiratory}} = \{\mathsf{SpO_2\text{-}Band}, \mathsf{CT\text{-}Finding}\}, \quad L_{\mathsf{Sepsis}} = \{\mathsf{TempBand}, \mathsf{CRP\text{-}Band}\}.$$

Thus a type-2 rough set represents: "choose the clinical pathway, then approximate $X$ using the tests relevant to that pathway."

**Type-3 (add a further layer: resource level).** Introduce a higher-level primary parameter

$$A_3 = \{\mathsf{ResourceLevel}\} \subseteq E,$$

with values such as $\mathsf{Low}$ (limited tests available) and $\mathsf{High}$ (full labs/imaging). Define $\mathcal{F}^{(3)}$ : $A_3 \to \Sigma^{(2)}(U, E, \rho)$ by

$$\mathcal{F}^{(3)}(\mathsf{ResourceLevel}) = \big(\mathcal{F}^{(2)}_{\mathsf{ResourceLevel}},\, L_{\mathsf{ResourceLevel}}\big),$$

where each $\mathcal{F}^{(2)}_{\mathsf{ResourceLevel}}$ is itself a type-2 assignment that changes which protocol-specific parameter sets $L_{\mathsf{Protocol}}$ are admissible (e.g. under $\mathsf{Low}$, exclude $\mathsf{CT\text{-}Finding}$ and rely on vitals/labs only).

By iterating this construction, a Type-$n$ rough set models *layered* real clinical decision-making: policy/constraints (resources) $\to$ pathway selection $\to$ test selection $\to$ rough approximation of "needs admission," with each layer encoded by a primary parameter set and a map into the previous rough-set type.

## 2.26 Dominance-based Rough set

Dominance-based rough sets handle ordered criteria using dominance relations, approximating upward/downward decision-class unions with lower/upper sets consistent with preferences principle [128–133]. Related concepts, such as fuzzy dominance-based rough sets, are also known [134, 135].

**Definition 2.26.1** (Decision table with preference-ordered criteria)**.** [128–130] A *(multiple-criteria) decision table* is a tuple

$$\mathsf{S} = (U,\, C \cup D,\, V,\, f),$$

where $U$ is a nonempty finite set of objects, $C$ is a finite set of *condition attributes* (assumed to be criteria), $D$ is a finite set of *decision attributes*, $V = \bigsqcup_{q \in C \cup D} V_q$ is the family of attribute domains, and $f : U \times (C \cup D) \to V$ is the information function with $f(x, q) \in V_q$.

For each criterion $q \in C$, let $\succeq_q$ be a (weak) preference relation on $U$ such that $x \succeq_q y$ means "$x$ is at least as good as $y$ with respect to $q$". Assume that the decision attribute(s) induce a partition

$$\mathcal{C} = \{\mathcal{C}_t \mid t \in T\}, \qquad T = \{1, \ldots, n\},$$

and that the classes are *preference-ordered* (higher index $t$ means a better class).

**Definition 2.26.2** (Upward / downward unions)**.** For each $t \in \{1, \ldots, n\}$, define the *upward union* and *downward union* of decision classes by

$$\mathcal{C}_t^{\geq} := \bigcup_{s \geq t} \mathcal{C}_s, \qquad \mathcal{C}_t^{\leq} := \bigcup_{s \leq t} \mathcal{C}_s.$$

Thus $x \in \mathcal{C}_t^{\geq}$ means "$x$ belongs to at least class $\mathcal{C}_t$", and $x \in \mathcal{C}_t^{\leq}$ means "$x$ belongs to at most class $\mathcal{C}_t$".

**Definition 2.26.3** (Dominance relation and dominance cones)**.** Let $P \subseteq C$ be a nonempty set of criteria. The *dominance relation* induced by $P$ is

$$x\, D_P\, y \iff (\forall q \in P)\; x \succeq_q y.$$

For $x \in U$, define the $P$*-dominating* and $P$*-dominated* sets (dominance cones) by

$$D_P^{+}(x) := \{\, y \in U \mid y\, D_P\, x \,\}, \qquad D_P^{-}(x) := \{\, y \in U \mid x\, D_P\, y \,\}.$$

**Definition 2.26.4** (Dominance-based rough approximations (DRSA))**.** Fix $P \subseteq C$ and consider the family of upward and downward unions $\{\mathcal{C}_t^{\geq}, \mathcal{C}_t^{\leq}\}_{t=1}^{n}$.

**(1) Lower approximations.** For $t = 1, \ldots, n$, the $P$*-lower approximation* of $\mathcal{C}_t^{\geq}$ and $\mathcal{C}_t^{\leq}$ are

$$\underline{P}(\mathcal{C}_t^{\geq}) := \{\, x \in \mathcal{C}_t^{\geq} \mid D_P^{+}(x) \subseteq \mathcal{C}_t^{\geq} \,\}, \qquad \underline{P}(\mathcal{C}_t^{\leq}) := \{\, x \in \mathcal{C}_t^{\leq} \mid D_P^{-}(x) \subseteq \mathcal{C}_t^{\leq} \,\}.$$

These are the objects that belong to the corresponding union *without ambiguity* under the dominance principle.

**(2) Upper approximations (by complementarity).** For $t = 2, \ldots, n$ and $t = 1, \ldots, n-1$, respectively, define

$$\overline{P}(\mathcal{C}_t^{\geq}) \;:=\; U \setminus \underline{P}(\mathcal{C}_{t-1}^{\leq}), \qquad \overline{P}(\mathcal{C}_t^{\leq}) \;:=\; U \setminus \underline{P}(\mathcal{C}_{t+1}^{\geq}).$$

**(3) Boundary regions.** The *P-boundary (doubtful) regions* are

$$\mathrm{Bn}_P(\mathcal{C}_t^{\geq}) \;:=\; \overline{P}(\mathcal{C}_t^{\geq}) \setminus \underline{P}(\mathcal{C}_t^{\geq}), \qquad \mathrm{Bn}_P(\mathcal{C}_t^{\leq}) \;:=\; \overline{P}(\mathcal{C}_t^{\leq}) \setminus \underline{P}(\mathcal{C}_t^{\leq}).$$

**Example 2.26.5** (Real-life DRSA: ranking loan applicants under monotone criteria)**.** Let $U$ be a set of loan applicants. Consider an ordinal decision attribute with $n = 3$ ordered classes

$$\mathcal{C}_1 \prec \mathcal{C}_2 \prec \mathcal{C}_3,$$

where $\mathcal{C}_1 =$ *reject*, $\mathcal{C}_2 =$ *manual review*, and $\mathcal{C}_3 =$ *approve*. Let $C$ be a set of evaluation criteria and choose a monotone subset

$$P = \{\mathsf{Income}, \mathsf{CreditScore}, \mathsf{DTI}\} \subseteq C,$$

where higher Income and CreditScore are better, and lower DTI is better (so we transform it to a benefit form, e.g. $\mathsf{Affordability} := -\mathsf{DTI}$).

Define the *dominance relation* induced by $P$ by

$$x\, D_P\, y \quad\Longleftrightarrow\quad g(x) \geq g(y) \text{ for every benefit-type criterion } g \in P.$$

Let the (forward/backward) dominance cones be

$$D_P^+(x) := \{\, y \in U \mid y\, D_P\, x \,\}, \qquad D_P^-(x) := \{\, y \in U \mid x\, D_P\, y \,\}.$$

Form the upward and downward unions

$$\mathcal{C}_t^{\geq} := \bigcup_{s \geq t} \mathcal{C}_s, \qquad \mathcal{C}_t^{\leq} := \bigcup_{s \leq t} \mathcal{C}_s, \qquad (t = 1, 2, 3).$$

**Interpretation via DRSA approximations.**

- $x \in \underline{P}(\mathcal{C}_3^{\geq})$ means: every applicant who dominates $x$ (is at least as good on all $P$) is still in $\mathcal{C}_3^{\geq} = \mathcal{C}_3$, so $x$ is *certainly approvable* under the monotonicity principle.
- $x \in \underline{P}(\mathcal{C}_2^{\geq})$ means: everyone dominating $x$ is at least in $\mathcal{C}_2$, so $x$ is *certainly not rejectable* (i.e. belongs safely to *review-or-approve*).
- $x \in \mathrm{Bn}_P(\mathcal{C}_3^{\geq})$ means: $x$ is *possibly approvable* but not certain, so it falls naturally into *manual review* due to ambiguous dominance evidence.

Thus DRSA implements a realistic credit policy: decisions respect monotone preferences (better profiles should not receive worse decisions), while boundary regions identify applicants requiring additional checks.

## 2.27 Triangular rough set

A triangular rough set represents vague assessments by a triplet $(a, b, c)$, yielding piecewise-linear membership and defuzzification via mean value directly [136]. Related concepts with similar structure include *triangular fuzzy sets* [137–139] and *triangular neutrosophic sets* [140, 141].

**Definition 2.27.1** (Triangular rough set)**.** [136] Let $X \subseteq \mathbb{R}$ be a universe of discourse (e.g., a rating scale). A *triangular rough set* on $X$ is specified by a triple

$$A = (a, b, c) \in \mathbb{R}^3 \quad \text{with} \quad a \le b \le c \quad \text{and} \quad [a, c] \cap X \neq \emptyset,$$

together with the (triangular) membership function $\mu_A : X \to [0, 1]$ defined by

$$\mu_A(x) := \begin{cases} 0, & x < a, \\ \dfrac{x-a}{b-a}, & a \le x \le b \text{ and } a < b, \\ 1, & x = b, \\ \dfrac{c-x}{c-b}, & b \le x \le c \text{ and } b < c, \\ 0, & x > c, \end{cases}$$

with the usual endpoint conventions in the degenerate cases: if $a = b$ then $\mu_A(a) = 1$ and the rising branch is omitted; if $b = c$ then $\mu_A(c) = 1$ and the falling branch is omitted. A common crisp representative (defuzzification) value of $A$ is the centroid

$$\text{def}(A) \;:=\; \frac{a+b+c}{3}.$$

**Example 2.27.2** (Real-life example: customer satisfaction rating as a triangular rough set)**.** Let $X = \{1, 2, 3, 4, 5\} \subseteq \mathbb{R}$ be a 5-point customer-satisfaction (CSAT) scale. Suppose a product manager summarizes the (vague) assessment "the satisfaction is around 4, but could be as low as 3 and as high as 5" by the triangular rough set

$$A = (a, b, c) = (3, 4, 5).$$

Then the membership degrees on $X$ are

$$\mu_A(1) = 0, \quad \mu_A(2) = 0, \quad \mu_A(3) = 0, \quad \mu_A(4) = 1, \quad \mu_A(5) = 0.$$

The corresponding crisp representative (centroid) value is

$$\text{def}(A) = \frac{3+4+5}{3} = 4,$$

so the single-number summary is 4/5 while retaining the explicit uncertainty range $[3, 5]$ encoded by $(a, b, c)$.

## 2.28 Game-theoretic rough sets

Game-theoretic rough sets treat approximation regions or probabilistic thresholds as players, using payoff-based competition/cooperation to iteratively learn effective parameters automatically [62, 142–144]. Game-theoretic rough sets have also been widely studied in recent years, partly due to their ease of application [145–148].

**Definition 2.28.1** (Decision-theoretic three-way rough approximations)**.** [62, 142] Let $U$ be a finite universe, $R \subseteq U \times U$ an equivalence relation, and $X \subseteq U$ a target concept. For $x \in U$, write $[x]_R$ for the $R$-equivalence class (granule) of $x$, and set

$$p(x) \;:=\; \mathrm{Prob}\big(X \mid [x]_R\big) \in [0,1].$$

Consider three classification actions

$$\mathsf{Act} = \{a_P, a_B, a_N\},$$

interpreted as deciding *positive* $\mathrm{POS}(X)$, *boundary* $\mathrm{BND}(X)$, and *negative* $\mathrm{NEG}(X)$, respectively. Let the set of states be $\Omega = \{X, X^c\}$ and let $\lambda_{ij} \geq 0$ denote the loss incurred by taking action $a_i \in \mathsf{Act}$ when the true state is $j \in \Omega$. Define the (conditional) risk of action $a_i$ at $x$ by

$$\mathrm{Risk}(a_i \mid x) := \lambda_{iX}\, p(x) + \lambda_{iX^c}\,(1 - p(x)).$$

Assume the standard ordering of losses (so that $a_P$ is best when $x \in X$, $a_N$ is best when $x \notin X$, and $a_B$ is an intermediate/defer action), e.g.

$$\lambda_{PX} \leq \lambda_{BX} < \lambda_{NX},$$

$$\lambda_{NX^c} \leq \lambda_{BX^c} < \lambda_{PX^c}.$$

Then comparing $\mathrm{Risk}(a_P \mid x)$ vs. $\mathrm{Risk}(a_B \mid x)$ and $\mathrm{Risk}(a_N \mid x)$ vs. $\mathrm{Risk}(a_B \mid x)$ yields thresholds

$$\alpha := \frac{\lambda_{PX^c} - \lambda_{BX^c}}{(\lambda_{PX^c} - \lambda_{BX^c}) + (\lambda_{BX} - \lambda_{PX})},$$

$$\beta := \frac{\lambda_{BX^c} - \lambda_{NX^c}}{(\lambda_{BX^c} - \lambda_{NX^c}) + (\lambda_{NX} - \lambda_{BX})},$$

with $\beta < \alpha$, and the three-way decision regions

$$\mathrm{POS}_{\alpha,\beta}(X) = \{x \in U : p(x) \geq \alpha\}, \quad \mathrm{NEG}_{\alpha,\beta}(X) = \{x \in U : p(x) \leq \beta\},$$

$$\mathrm{BND}_{\alpha,\beta}(X) = U \setminus (\mathrm{POS}_{\alpha,\beta}(X) \cup \mathrm{NEG}_{\alpha,\beta}(X)).$$

The induced lower/upper approximations are

$$\underline{X}_{\alpha,\beta} := \mathrm{POS}_{\alpha,\beta}(X),$$

$$\overline{X}_{\alpha,\beta} := \mathrm{POS}_{\alpha,\beta}(X) \cup \mathrm{BND}_{\alpha,\beta}(X) = U \setminus \mathrm{NEG}_{\alpha,\beta}(X).$$

**Definition 2.28.2** (Game-theoretic rough set (GTRS) model)**.** Fix a decision-theoretic setting as in Definition 2.28.1. A *game-theoretic rough set model* is specified by a (normal-form) game

$$\mathcal{G} = (N, \{S_i\}_{i \in N}, \{u_i\}_{i \in N}),$$

together with an *interpretation map* that turns each strategy profile into a three-way approximation of $X$.

**(i) Players.)** $N$ is a finite set of players. Typical choices are:

- *Parameter game:* $N = \{\alpha, \beta\}$ (players represent the probabilistic thresholds);
- *Measure game:* $N = \{\mathsf{Acc}, \mathsf{Prec}\}$ (players represent chosen approximation measures).

**(ii) Strategies and induced approximations.)** For each $i \in N$, $S_i$ is a finite set of strategies (actions). Each profile $s = (s_i)_{i\in N} \in \prod_{i\in N} S_i$ determines an updated loss table (equivalently, updated risks), hence updated thresholds

$$(\alpha_s, \beta_s) \quad \text{and thus} \quad (\mathrm{POS}_s(X), \mathrm{BND}_s(X), \mathrm{NEG}_s(X)) := \big(\mathrm{POS}_{\alpha_s,\beta_s}(X), \mathrm{BND}_{\alpha_s,\beta_s}(X), \mathrm{NEG}_{\alpha_s,\beta_s}(X)\big).$$

(Concretely, a strategy typically corresponds to increasing/decreasing selected losses or directly increasing/decreasing $\alpha$ or $\beta$ by a small step, then recomputing the regions.)

**(iii) Payoffs.)** Each payoff function

$$u_i : \prod_{j\in N} S_j \to \mathbb{R}$$

quantifies the utility of the resulting approximation for player $i$. A generic and mathematically clean choice is:

$$u_i(s) \;:=\; m_i\big(\mathrm{POS}_s(X), \mathrm{BND}_s(X), \mathrm{NEG}_s(X)\big) \;-\; m_i\big(\mathrm{POS}_{s^{(0)}}(X), \mathrm{BND}_{s^{(0)}}(X), \mathrm{NEG}_{s^{(0)}}(X)\big),$$

where $m_i$ is the player's objective measure (e.g., boundary-size reduction, accuracy improvement, precision gain), and $s^{(0)}$ is a fixed baseline profile (e.g., the current system configuration).

**(iv) Equilibrium and the GTRS approximation.)** A profile $s^* \in \prod_{i\in N} S_i$ is a (pure) Nash equilibrium if

$$u_i(s^*) \;\geq\; u_i(s_i, s^*_{-i}) \qquad (\forall\, i \in N,\; \forall\, s_i \in S_i),$$

where $s^*_{-i}$ denotes the strategies of all players except $i$. The *game-theoretic rough approximation* of $X$ (under $\mathcal{G}$) is the three-way approximation induced by an equilibrium profile $s^*$, namely

$$(\underline{X}^{\mathrm{GTRS}}, \overline{X}^{\mathrm{GTRS}}, \mathrm{BND}^{\mathrm{GTRS}}(X)) := \big(\mathrm{POS}_{s^*}(X),\; U \setminus \mathrm{NEG}_{s^*}(X),\; \mathrm{BND}_{s^*}(X)\big).$$

**Example 2.28.3** (Real-life GTRS: tuning an e-mail spam filter via a precision–recall game)**.** Let $U$ be a finite set of e-mails arriving in one day. Let $X \subseteq U$ be the (unknown) set of truly spam e-mails. Assume that a baseline classifier assigns each $x \in U$ a spam score $p(x) \in [0, 1]$ (estimated probability that $x \in X$), and that actions are the three-way decisions

$$\mathsf{Act}_P = \text{“auto-block”},$$

$$\mathsf{Act}_B = \text{“quarantine / human review”},$$

$$\mathsf{Act}_N = \text{“deliver”}.$$

Given thresholds $(\alpha, \beta)$ with $0 \leq \beta < \alpha \leq 1$, the induced three-way regions are

$$\mathrm{POS}_{\alpha,\beta}(X) = \{x \in U : p(x) \geq \alpha\},$$

$$\mathrm{NEG}_{\alpha,\beta}(X) = \{x \in U : p(x) \leq \beta\},$$

$$\mathrm{BND}_{\alpha,\beta}(X) = U \setminus (\mathrm{POS}_{\alpha,\beta}(X) \cup \mathrm{NEG}_{\alpha,\beta}(X)).$$

**Players (a measure game).** Let $N = \{\mathsf{Prec}, \mathsf{Rec}\}$, where $\mathsf{Prec}$ represents the product team that wants to minimize false positives (maximize precision), and $\mathsf{Rec}$ represents the security team that wants to catch as much spam as possible (maximize recall).

**Strategies.** Fix small step sizes $\delta_\alpha, \delta_\beta > 0$. Let each player choose a discrete adjustment:

$$S_{\mathsf{Prec}} = \{\uparrow \alpha,\ \downarrow \alpha\},$$

$$S_{\mathsf{Rec}} = \{\uparrow \beta,\ \downarrow \beta\},$$

where, for a profile $s = (s_{\mathsf{Prec}}, s_{\mathsf{Rec}})$, the updated thresholds are

$$\alpha_s = \alpha_0 + \Delta_\alpha(s_{\mathsf{Prec}}), \qquad \beta_s = \beta_0 + \Delta_\beta(s_{\mathsf{Rec}}),$$

with $\Delta_\alpha(\uparrow \alpha) = +\delta_\alpha$, $\Delta_\alpha(\downarrow \alpha) = -\delta_\alpha$, $\Delta_\beta(\uparrow \beta) = +\delta_\beta$, and $\Delta_\beta(\downarrow \beta) = -\delta_\beta$, projected to $[0,1]$ and maintaining $\beta_s < \alpha_s$.

**Payoffs (data-driven).** Using a labeled validation subset $U_{\mathrm{val}} \subseteq U$ (obtained from user reports and audits), define the realized precision and recall at $(\alpha_s, \beta_s)$ by

$$\mathrm{Prec}(s) := \frac{|\,\mathrm{POS}_s(X) \cap X|}{|\,\mathrm{POS}_s(X)| + \varepsilon}, \qquad \mathrm{Rec}(s) := \frac{|\,\mathrm{POS}_s(X) \cap X|}{|X| + \varepsilon},$$

with a tiny $\varepsilon > 0$ to avoid division by zero. Let the payoffs be improvements over a baseline profile $s^{(0)}$:

$$u_{\mathsf{Prec}}(s) := \mathrm{Prec}(s) - \mathrm{Prec}(s^{(0)}), \qquad u_{\mathsf{Rec}}(s) := \mathrm{Rec}(s) - \mathrm{Rec}(s^{(0)}).$$

Then $\mathcal{G} = (N, \{S_i\}_{i\in N}, \{u_i\}_{i\in N})$ together with the interpretation $s \mapsto (\mathrm{POS}_s(X), \mathrm{BND}_s(X), \mathrm{NEG}_s(X))$ is a concrete GTRS instance.

**GTRS outcome.** A (pure) Nash equilibrium $s^*$ is a stable operating point where neither team can improve its objective by unilaterally changing the threshold it controls. The resulting game-theoretic rough approximation is

$$\underline{X}^{\mathrm{GTRS}} = \mathrm{POS}_{s^*}(X), \qquad \overline{X}^{\mathrm{GTRS}} = U \setminus \mathrm{NEG}_{s^*}(X), \qquad \mathrm{BND}^{\mathrm{GTRS}}(X) = \mathrm{BND}_{s^*}(X),$$

interpreted as *auto-block spam*, *possible spam*, and *safe e-mails*, respectively.

## 2.29 Variable precision rough set

VPRS generalizes rough sets by allowing a controlled misclassification rate $\beta$, defining flexible lower and upper approximations for concepts datasets [149–153]. Variable precision rough sets, like other rough-set models, have also been extensively studied [154–157].

**Definition 2.29.1** (Misclassification rate and $\beta$-inclusion)**.** Let $U$ be a nonempty finite universe and let $R \subseteq U \times U$ be an equivalence relation. For $x \in U$, write

$$[x]_R := \{\, y \in U \mid (x, y) \in R \,\}$$

for the $R$-equivalence class of $x$. For nonempty $A \subseteq U$ and any $X \subseteq U$, define the *relative misclassification rate* of $A$ with respect to $X$ by

$$\mathrm{err}(A, X) \ := \ \frac{|A \setminus X|}{|A|} \ = \ 1 - \frac{|A \cap X|}{|A|}.$$

Fix a precision parameter $\beta \in [0, \frac{1}{2})$. We say that $A$ is *$\beta$-included* in $X$, and write $A \subseteq_\beta X$, if

$$\mathrm{err}(A, X) \le \beta \quad \Longleftrightarrow \quad \frac{|A \cap X|}{|A|} \ge 1 - \beta.$$

**Definition 2.29.2** (Variable precision rough approximations (VPRS)). Let $(U, R)$ be as in Definition 2.29.1, let $X \subseteq U$, and fix $\beta \in [0, \frac{1}{2})$. The *$\beta$-lower approximation* and *$\beta$-upper approximation* of $X$ are defined by

$$\underline{\mathrm{apr}}_\beta(X) := \{ x \in U \mid [x]_R \subseteq_\beta X \}$$

$$= \Big\{ x \in U \;\Big|\; \frac{|[x]_R \cap X|}{|[x]_R|} \geq 1 - \beta \Big\},$$

$$\overline{\mathrm{apr}}_\beta(X) :=$$

$$U \setminus \underline{\mathrm{apr}}_\beta(U \setminus X) = \Big\{ x \in U \;\Big|\; [x]_R \not\subseteq_\beta (U \setminus X) \Big\}.$$

The induced *positive*, *negative*, and *boundary* regions are

$$\mathrm{POS}_\beta(X) := \underline{\mathrm{apr}}_\beta(X),$$

$$\mathrm{NEG}_\beta(X) := U \setminus \overline{\mathrm{apr}}_\beta(X) = \underline{\mathrm{apr}}_\beta(U \setminus X),$$

$$\mathrm{BND}_\beta(X) := \overline{\mathrm{apr}}_\beta(X) \setminus \underline{\mathrm{apr}}_\beta(X).$$

**Remark 2.29.3.** If $\beta = 0$, then $A \subseteq_0 X$ is equivalent to $A \subseteq X$, hence $\underline{\mathrm{apr}}_0(X)$ and $\overline{\mathrm{apr}}_0(X)$ reduce to the classical Pawlak lower and upper approximations based on $R$.

**Example 2.29.4** (Variable precision rough approximations in credit screening). Let $U = \{a_1, a_2, \ldots, a_{10}\}$ be a set of loan applicants. Assume applicants are indiscernible (for a coarse first-stage screening) if they share the same *income bracket* and *employment type*. This induces an equivalence relation $R$ on $U$ with equivalence classes

$$[a_1]_R = \cdots = [a_5]_R =: C_1 = \{a_1, a_2, a_3, a_4, a_5\},$$

$$[a_6]_R = [a_7]_R = [a_8]_R =: C_2 = \{a_6, a_7, a_8\},$$

$$[a_9]_R = [a_{10}]_R =: C_3 = \{a_9, a_{10}\}.$$

Let $X \subseteq U$ be the set of applicants judged *low-risk* by a more accurate (but costly) manual review:

$$X = \{a_1, a_2, a_3, a_4, a_9\}.$$

Fix the VPRS tolerance parameter $\beta = 0.2$ (so $1 - \beta = 0.8$).

**Step 1: $\beta$-lower approximation.** For $x \in U$, the condition $[x]_R \subseteq_\beta X$ is equivalent to $\frac{|[x]_R \cap X|}{|[x]_R|} \geq 0.8$. Compute the classwise ratios:

$$\frac{|C_1 \cap X|}{|C_1|} = \frac{4}{5} = 0.8, \qquad \frac{|C_2 \cap X|}{|C_2|} = \frac{0}{3} = 0, \qquad \frac{|C_3 \cap X|}{|C_3|} = \frac{1}{2} = 0.5.$$

Hence only $C_1$ satisfies the threshold, and thus

$$\underline{\mathrm{apr}}_{0.2}(X) = \{x \in U \mid [x]_R \subseteq_{0.2} X\} = C_1 = \{a_1, a_2, a_3, a_4, a_5\}.$$

**Step 2: $\beta$-upper approximation via the complement.** The complement is

$$U \setminus X = \{a_5, a_6, a_7, a_8, a_{10}\}.$$

Now

$$\frac{|C_1 \cap (U \setminus X)|}{|C_1|} = \frac{1}{5} = 0.2, \qquad \frac{|C_2 \cap (U \setminus X)|}{|C_2|} = \frac{3}{3} = 1, \qquad \frac{|C_3 \cap (U \setminus X)|}{|C_3|} = \frac{1}{2} = 0.5.$$

Thus

$$\underline{\mathrm{apr}}_{0.2}(U \setminus X) = C_2 = \{a_6, a_7, a_8\},$$

and by Definition 2.29.2,

$$\overline{\mathrm{apr}}_{0.2}(X) = U \setminus \underline{\mathrm{apr}}_{0.2}(U \setminus X) = U \setminus C_2 = \{a_1, a_2, a_3, a_4, a_5, a_9, a_{10}\}.$$

**Regions.** Therefore,

$$\mathrm{POS}_{0.2}(X) = \underline{\mathrm{apr}}_{0.2}(X) = \{a_1, a_2, a_3, a_4, a_5\},$$

$$\mathrm{NEG}_{0.2}(X) = \underline{\mathrm{apr}}_{0.2}(U \setminus X) = \{a_6, a_7, a_8\},$$

$$\mathrm{BND}_{0.2}(X) = \overline{\mathrm{apr}}_{0.2}(X) \setminus \underline{\mathrm{apr}}_{0.2}(X) = \{a_9, a_{10}\}.$$

*Interpretation.* With $\beta = 0.2$, the class $C_1$ is treated as "definitely low-risk" even though $a_5 \notin X$ (allowing up to 20% misclassification inside a granule), while $C_2$ is "definitely not low-risk". The mixed class $C_3$ becomes the boundary (uncertain) region.

## 2.30 Multi-granulation rough set

Multi-granulation rough sets approximate a concept using multiple equivalence relations, thereby yielding optimistic or pessimistic lower/upper regions across granules concurrently [158–162]. Multi-granulation rough sets, like other rough-set models, have also attracted a substantial body of research in recent years [163–165].

**Definition 2.30.1** (Multi-granulation rough approximations)**.** Let $U$ be a nonempty finite universe and let

$$\mathcal{R} = \{R_1, R_2, \ldots, R_m\}$$

be a finite family of equivalence relations on $U$ (each $R_i \subseteq U \times U$). For $x \in U$ and $i \in \{1, \ldots, m\}$, write

$$[x]_{R_i} := \{\, y \in U \mid (x, y) \in R_i \,\}$$

for the $R_i$-equivalence class (granule) of $x$.

For any $X \subseteq U$, the *optimistic* multi-granulation lower and upper approximations of $X$ (with respect to $\mathcal{R}$) are defined by

$$\underline{\mathrm{apr}}^{O}_{\mathcal{R}}(X) \;:=\; \Big\{ x \in U \;\Big|\; [x]_{R_1} \subseteq X \;\vee\; [x]_{R_2} \subseteq X \;\vee\; \cdots \;\vee\; [x]_{R_m} \subseteq X \Big\},$$

$$\overline{\text{apr}}^O_{\mathcal{R}}(X) \;:=\; U \setminus \underline{\text{apr}}^O_{\mathcal{R}}(U \setminus X) \;=\; \Big\{ x \in U \;\Big|\; [x]_{R_1} \cap X \neq \varnothing \;\wedge\; \cdots \;\wedge\; [x]_{R_m} \cap X \neq \varnothing \Big\}.$$

The *optimistic multi-granulation rough set* of $X$ is the pair

$$\big(\underline{\text{apr}}^O_{\mathcal{R}}(X),\; \overline{\text{apr}}^O_{\mathcal{R}}(X)\big).$$

Similarly, the *pessimistic* multi-granulation lower and upper approximations of $X$ are

$$\begin{aligned}
&\underline{\text{apr}}^P_{\mathcal{R}}(X) \\
&:= \Big\{ x \in U \;\Big|\; [x]_{R_1} \subseteq X \;\wedge\; [x]_{R_2} \subseteq X \;\wedge\; \cdots \;\wedge\; [x]_{R_m} \subseteq X \Big\}, \\
&\overline{\text{apr}}^P_{\mathcal{R}}(X) \\
&:= \; U \setminus \underline{\text{apr}}^P_{\mathcal{R}}(U \setminus X) \\
&= \Big\{ x \in U \;\Big|\; [x]_{R_1} \cap X \neq \varnothing \;\vee\; \cdots \;\vee\; [x]_{R_m} \cap X \neq \varnothing \Big\}.
\end{aligned}$$

The *pessimistic multi-granulation rough set* of $X$ is the pair

$$\big(\underline{\text{apr}}^P_{\mathcal{R}}(X),\; \overline{\text{apr}}^P_{\mathcal{R}}(X)\big).$$

**Remark 2.30.2.** In both the optimistic and pessimistic cases one has

$$\underline{\text{apr}}^*_{\mathcal{R}}(X) \subseteq X \subseteq \overline{\text{apr}}^*_{\mathcal{R}}(X) \qquad (* \in \{O, P\}),$$

so each pair forms a valid rough approximation of $X$. When $m = 1$, both models reduce to Pawlak's classical approximations for the single relation $R_1$.

**Example 2.30.3** (Multi-granulation rough approximations in medical triage)**.** Let

$$U = \{p_1, p_2, p_3, p_4, p_5, p_6, p_7, p_8\}$$

be a set of patients arriving at an emergency clinic. We model two different (coarse) ways of grouping patients, hence two equivalence relations:

**(i) Symptom-profile granulation.** Let $R_1$ be indiscernibility with respect to a *symptom profile* (e.g., fever/cough category), giving the partition

$$U/R_1 = \{C_1, C_2, C_3\}, \qquad C_1 = \{p_1, p_2, p_3\}, \;\; C_2 = \{p_4, p_5\}, \;\; C_3 = \{p_6, p_7, p_8\}.$$

**(ii) Rapid-test granulation.** Let $R_2$ be indiscernibility with respect to a binary rapid test result (positive/negative), giving the partition

$$U/R_2 = \{D_1, D_2\}, \qquad D_1 = \{p_1, p_2, p_4, p_6\} \text{ (test+)}, \;\; D_2 = \{p_3, p_5, p_7, p_8\} \text{ (test–)}.$$

Set $\mathcal{R} = \{R_1, R_2\}$. Let the target concept be

$$X = \{p_1, p_2, p_4, p_5\} \subseteq U,$$

interpreted as “patients judged *high-risk* (need isolation)” by a senior clinician.

**Optimistic lower approximation.** By Definition 2.30.1,

$$\underline{\mathrm{apr}}^{O}_{\mathcal{R}}(X) = \{x \in U \mid [x]_{R_1} \subseteq X \ \vee \ [x]_{R_2} \subseteq X\}.$$

Now $C_2 = \{p_4, p_5\} \subseteq X$, while $C_1 \nsubseteq X$, $C_3 \nsubseteq X$, and neither $D_1$ nor $D_2$ is contained in $X$. Hence only the patients in $C_2$ enter the optimistic lower approximation:

$$\underline{\mathrm{apr}}^{O}_{\mathcal{R}}(X) = \{p_4, p_5\}.$$

**Optimistic upper approximation.** Using the equivalent characterization in Definition 2.30.1,

$$\overline{\mathrm{apr}}^{O}_{\mathcal{R}}(X) = \{x \in U \mid [x]_{R_1} \cap X \neq \varnothing \ \wedge \ [x]_{R_2} \cap X \neq \varnothing\}.$$

For $x \in C_1$, we have $C_1 \cap X = \{p_1, p_2\} \neq \varnothing$, and both $D_1 \cap X = \{p_1, p_2, p_4\} \neq \varnothing$ and $D_2 \cap X = \{p_5\} \neq \varnothing$ (depending on whether $x \in D_1$ or $x \in D_2$), so $p_1, p_2, p_3 \in \overline{\mathrm{apr}}^{O}_{\mathcal{R}}(X)$. Similarly $p_4, p_5 \in \overline{\mathrm{apr}}^{O}_{\mathcal{R}}(X)$, while $C_3 \cap X = \varnothing$, so $p_6, p_7, p_8 \notin \overline{\mathrm{apr}}^{O}_{\mathcal{R}}(X)$. Therefore,

$$\overline{\mathrm{apr}}^{O}_{\mathcal{R}}(X) = \{p_1, p_2, p_3, p_4, p_5\}.$$

**Pessimistic lower approximation.**

$$\underline{\mathrm{apr}}^{P}_{\mathcal{R}}(X) = \{x \in U \mid [x]_{R_1} \subseteq X \ \wedge \ [x]_{R_2} \subseteq X\}.$$

Although $[p_4]_{R_1} = [p_5]_{R_1} = C_2 \subseteq X$, we have $[p_4]_{R_2} = D_1 \nsubseteq X$ and $[p_5]_{R_2} = D_2 \nsubseteq X$. Thus no element satisfies both inclusions, and

$$\underline{\mathrm{apr}}^{P}_{\mathcal{R}}(X) = \varnothing.$$

**Pessimistic upper approximation.** Equivalently,

$$\overline{\mathrm{apr}}^{P}_{\mathcal{R}}(X) = \{x \in U \mid [x]_{R_1} \cap X \neq \varnothing \ \vee \ [x]_{R_2} \cap X \neq \varnothing\}.$$

Every patient belongs to either a symptom class intersecting $X$ (namely $C_1$ or $C_2$) or a test class intersecting $X$ (both $D_1$ and $D_2$ intersect $X$), so

$$\overline{\mathrm{apr}}^{P}_{\mathcal{R}}(X) = U.$$

*Interpretation.* The optimistic model certifies “high-risk” if *either* symptom-granulation or test-granulation supports it (hence $\{p_4, p_5\}$), whereas the pessimistic model requires *both* granular views to support it (hence the empty lower set here).

## 2.31 Soft Rough Set

A soft rough set combines soft-set parameterization with rough approximations, producing lower and upper regions of a target concept under parameter-dependent uncertainty [166–169]. Related notions include *fuzzy soft rough sets* [170–172], HyperSoft Rough Set [173, 174], Modified soft rough set [175–177], N-soft rough sets [178, 179], and *neutrosophic soft rough sets* [180–182].

**Definition 2.31.1** (Soft Rough Set)**.** (cf. [183, 184]) Let $U$ be a universal set, $A$ a set of parameters, and $P(U)$ the power set of $U$. Let $R$ be an equivalence relation on $U$, inducing a partition $U/R = \{Y_1, Y_2, \dots, Y_m\}$ into equivalence classes. A soft set $(F, A)$ on $U$ is defined as $F : A \to P(U)$.

For $B \subseteq U$, the *Soft Rough Lower Approximation* $L(B)$ and *Soft Rough Upper Approximation* $U(B)$ are given by:

$$L(B) = \{u \in U \mid \exists e \in A \text{ such that } F(e) \subseteq B\},$$

$$U(B) = \{u \in U \mid \exists e \in A \text{ such that } F(e) \cap B \neq \emptyset\}.$$

The *Soft Rough Set* is represented as the pair:

$$(L(B), U(B)),$$

where $L(B)$ and $U(B)$ are the approximations of $B$ with respect to the soft set.

**Example 2.31.2** (Soft rough set for apartment shortlisting)**.** Let $U = \{u_1, u_2, u_3, u_4, u_5, u_6\}$ be six apartments. Assume an equivalence relation $R$ on $U$ given by *same neighborhood*, so the partition $U/R = \{Y_1, Y_2, Y_3\}$ is

$$Y_1 = \{u_1, u_2\}, \qquad Y_2 = \{u_3, u_4\}, \qquad Y_3 = \{u_5, u_6\}.$$

Let the parameter set be

$$A = \{e_1, e_2, e_3\},$$

where $e_1$ = "affordable (monthly rent $\leq$ 120,000 JPY)", $e_2$ = "close to a station (walk $\leq$ 10 minutes)", $e_3$ = "quiet (low traffic / low nightlife)". Define a soft set $(F, A)$ on $U$ by

$$F(e_1) = \{u_1, u_3, u_5\}, \qquad F(e_2) = \{u_1, u_2, u_4\}, \qquad F(e_3) = \{u_2, u_5, u_6\}.$$

Suppose the target concept is

$$B = \{u_1, u_2, u_4\} \subseteq U,$$

interpreted as "apartments I would accept after a quick screening."

The soft rough lower approximation is

$$L(B) = \{u \in U \mid \exists e \in A \text{ such that } F(e) \subseteq B\}.$$

Here,

$$F(e_1) = \{u_1, u_3, u_5\} \nsubseteq B, \qquad F(e_2) = \{u_1, u_2, u_4\} \subseteq B, \qquad F(e_3) = \{u_2, u_5, u_6\} \nsubseteq B,$$

so the witness parameter is $e_2$, and hence

$$L(B) = F(e_2) = \{u_1, u_2, u_4\}.$$

Similarly, the soft rough upper approximation is

$$U(B) = \{u \in U \mid \exists e \in A \text{ such that } F(e) \cap B \neq \varnothing\}.$$

Since

$$F(e_1) \cap B = \{u_1\} \neq \varnothing, \quad F(e_2) \cap B = \{u_1, u_2, u_4\} \neq \varnothing, \quad F(e_3) \cap B = \{u_2\} \neq \varnothing,$$

every apartment that appears in at least one of $F(e_1), F(e_2), F(e_3)$ belongs to $U(B)$, hence

$$U(B) = F(e_1) \cup F(e_2) \cup F(e_3) = \{u_1, u_2, u_3, u_4, u_5, u_6\} = U.$$

Therefore the soft rough set (soft rough description) of $B$ induced by $(F, A)$ is

$$\big(L(B), U(B)\big) = \big(\{u_1, u_2, u_4\},\ U\big),$$

meaning that $u_1, u_2, u_4$ are *definitely* acceptable under the parameterized evidence, while the remaining apartments are only *possibly* acceptable due to partial parameter overlap.

## 2.32 Soft Rough Expert Set

A soft expert set assigns to each parameter–expert–opinion triple a subset of the universe, modeling expert-dependent uncertainty in decision making [185–188]. Soft Rough Expert Set combines soft parameters and expert opinions with rough approximations, yielding lower/upper evaluations under uncertainty in decision-making [189].

**Definition 2.32.1** (Soft Rough Expert Set)**.** [189] Let $U$ be a nonempty universe, let $E$ be a set of parameters, let $X$ be a set of experts, and let $O = \{0, 1\}$ be a set of opinions. Set the *soft expert parameter space* as

$$Z := E \times X \times O, \qquad B \subseteq Z.$$

A *soft expert set* over $U$ is a pair $\mathfrak{R} = (J, B)$, where

$$J : B \longrightarrow \mathcal{P}(U).$$

(We call $\mathfrak{R}$ a *soft expert approximation space.*)

For any target subset $Y \subseteq U$, define the *soft-rough expert lower* and *upper* approximations induced by $\mathfrak{R}$ by

$$\underline{\mathrm{apr}}_{\mathfrak{R}}(Y) := \{\, u \in U \mid \exists\, b \in B \text{ s.t. } u \in J(b) \text{ and } J(b) \subseteq Y \,\} \;=\; \bigcup \{\, J(b) \mid b \in B,\ J(b) \subseteq Y \,\},$$

$$\overline{\mathrm{apr}}_{\mathfrak{R}}(Y) := \{\, u \in U \mid \exists\, b \in B \text{ s.t. } u \in J(b) \text{ and } J(b) \cap Y \neq \varnothing \,\} \;=\; \bigcup \{\, J(b) \mid b \in B,\ J(b) \cap Y \neq \varnothing \,\}.$$

Then the ordered pair

$$\big(\underline{\mathrm{apr}}_{\mathfrak{R}}(Y),\ \overline{\mathrm{apr}}_{\mathfrak{R}}(Y)\big)$$

is called the *Soft Rough Expert Set* of $Y$ (with respect to $\mathfrak{R}$).

**Example 2.32.2** (Real-life Soft Rough Expert Set: shortlisting smartphones from expert opinions)**.** Let the universe of objects be four smartphone models

$$U = \{p_1, p_2, p_3, p_4\}.$$

Let the parameter set and expert set be

$$E = \{\text{Battery}, \text{Camera}, \text{Price}\}, \qquad X = \{\text{Alice}, \text{Bob}\}, \qquad O = \{0, 1\}.$$

Interpret $o = 1$ as a *positive* expert opinion (recommended under that parameter) and $o = 0$ as a *negative* opinion.

Define the soft expert parameter space $Z = E \times X \times O$ and choose the active parameter set

$$B = \{(\text{Battery}, \text{Alice}, 1),\ (\text{Camera}, \text{Bob}, 1),\ (\text{Price}, \text{Alice}, 1),\ (\text{Battery}, \text{Bob}, 0)\} \subseteq Z.$$

Define a soft expert set $\mathfrak{R} = (J, B)$ by specifying $J : B \to \mathcal{P}(U)$ as follows:

$$\begin{aligned} J(\text{Battery}, \text{Alice}, 1) &= \{p_1, p_2\} && (\text{Alice recommends } p_1, p_2 \text{ for battery}), \\ J(\text{Camera}, \text{Bob}, 1) &= \{p_2, p_3\} && (\text{Bob recommends } p_2, p_3 \text{ for camera}), \\ J(\text{Price}, \text{Alice}, 1) &= \{p_1, p_3\} && (\text{Alice recommends } p_1, p_3 \text{ for price}), \\ J(\text{Battery}, \text{Bob}, 0) &= \{p_3\} && (\text{Bob flags } p_3 \text{ as } \textit{bad} \text{ on battery}). \end{aligned}$$

Let the target set be the "travel shortlist"

$$Y = \{p_2, p_3\} \subseteq U.$$

Then the soft-rough expert lower approximation is

$$\underline{\text{apr}}_{\mathfrak{R}}(Y) = \bigcup \{\, J(b) \mid b \in B,\ J(b) \subseteq Y \,\} = J(\text{Camera}, \text{Bob}, 1) \cup J(\text{Battery}, \text{Bob}, 0) = \{p_2, p_3\}.$$

The soft-rough expert upper approximation is

$$\overline{\text{apr}}_{\mathfrak{R}}(Y) = \bigcup \{\, J(b) \mid b \in B,\ J(b) \cap Y \neq \varnothing \,\} = \{p_1, p_2\} \cup \{p_2, p_3\} \cup \{p_1, p_3\} \cup \{p_3\} = \{p_1, p_2, p_3\}.$$

Hence the Soft Rough Expert Set of $Y$ (with respect to $\mathfrak{R}$) is

$$\big(\underline{\text{apr}}_{\mathfrak{R}}(Y), \overline{\text{apr}}_{\mathfrak{R}}(Y)\big) = \big(\{p_2, p_3\}, \{p_1, p_2, p_3\}\big),$$

where $p_2, p_3$ are *definitely* shortlisted under expert evidence, while $p_1$ is only *possibly* shortlisted due to overlapping positive recommendations.

## 2.33 Covering-based Rough Sets

Covering-based rough sets generalize Pawlak's model by replacing partitions with coverings, yielding neighborhood-based lower and upper approximations for uncertain data [190–194].

**Definition 2.33.1** (Covering approximation space)**.** Let $U$ be a nonempty universe. A family $\mathcal{C} \subseteq \mathcal{P}(U)$ is called a *covering* of $U$ if $\varnothing \notin \mathcal{C}$ and $\bigcup \mathcal{C} = U$. The pair $(U, \mathcal{C})$ is called a *covering approximation space.*

**Definition 2.33.2** (Dual covering approximation operators: tight and loose)**.** Let $(U, \mathcal{C})$ be a covering approximation space and let $X \subseteq U$.

**(1) Tight (strong) pair.** Define the *tight lower* and *tight upper* approximations of $X$ by

$$\underline{\operatorname{apr}}^{t}_{\mathcal{C}}(X) \;:=\; \bigcup\{\, K \in \mathcal{C} \mid K \subseteq X \,\} \;=\; \{\, x \in U \mid \exists K \in \mathcal{C}\ (x \in K \subseteq X) \,\},$$

$$\overline{\operatorname{apr}}^{t}_{\mathcal{C}}(X) \;:=\; U \setminus \underline{\operatorname{apr}}^{t}_{\mathcal{C}}(U \setminus X) \;=\; \{\, x \in U \mid \forall K \in \mathcal{C}\ (x \in K \Rightarrow K \cap X \neq \varnothing) \,\}.$$

**(2) Loose (weak) pair.** Define the *loose lower* and *loose upper* approximations of $X$ by

$$\underline{\operatorname{apr}}^{\ell}_{\mathcal{C}}(X) \;:=\; U \setminus \overline{\operatorname{apr}}^{\ell}_{\mathcal{C}}(U \setminus X) \;=\; \{\, x \in U \mid \forall K \in \mathcal{C}\ (x \in K \Rightarrow K \subseteq X) \,\},$$

$$\overline{\operatorname{apr}}^{\ell}_{\mathcal{C}}(X) \;:=\; \bigcup\{\, K \in \mathcal{C} \mid K \cap X \neq \varnothing \,\} \;=\; \{\, x \in U \mid \exists K \in \mathcal{C}\ (x \in K,\ K \cap X \neq \varnothing) \,\}.$$

Both pairs are dual in the sense that each upper operator is the complement of the corresponding lower operator applied to complements.

**Definition 2.33.3** (Covering-based rough set)**.** Let $(U, \mathcal{C})$ be a covering approximation space and $X \subseteq U$. A *covering-based rough set* (with respect to a chosen dual pair, e.g. the tight pair or the loose pair) is represented by the approximation pair

$$\big(\underline{\operatorname{apr}}(X),\ \overline{\operatorname{apr}}(X)\big),$$

where $(\underline{\operatorname{apr}}, \overline{\operatorname{apr}})$ is either $\big(\underline{\operatorname{apr}}^{t}_{\mathcal{C}}, \overline{\operatorname{apr}}^{t}_{\mathcal{C}}\big)$ or $\big(\underline{\operatorname{apr}}^{\ell}_{\mathcal{C}}, \overline{\operatorname{apr}}^{\ell}_{\mathcal{C}}\big)$. The set $X$ is *(covering-)definable* if $\underline{\operatorname{apr}}(X) = \overline{\operatorname{apr}}(X)$; otherwise $X$ is *rough* under the selected covering-based approximation.

**Example 2.33.4** (Covering-based rough set for customer churn risk)**.** Consider an e-commerce service that groups customers into *overlapping* behavioral segments. Let

$$U = \{c_1, c_2, c_3, c_4, c_5, c_6\}$$

be the set of customers, and let the covering $\mathcal{C}$ consist of the following segments:

$$K_1 = \{c_1, c_2, c_3\} \text{ (new users)}, \qquad K_2 = \{c_3, c_4\} \text{ (coupon users)},$$

$$K_3 = \{c_4, c_5\} \text{ (support-contacted)},$$

$$K_4 = \{c_5\} \text{ (payment-failure flagged)},$$

$$K_5 = \{c_4, c_6\} \text{ (inactive-week users)},$$

so $\mathcal{C} = \{K_1, K_2, K_3, K_4, K_5\}$ is a covering of $U$.

Let the target set be the (unknown-but-assessed) set of customers who are truly at high churn risk:

$$X = \{c_4, c_5\} \subseteq U.$$

**Tight (strong) approximations.** Since $K_3 \subseteq X$ and $K_4 \subseteq X$, we obtain

$$\underline{\operatorname{apr}}^{t}_{\mathcal{C}}(X) = \bigcup\{K \in \mathcal{C} \mid K \subseteq X\} = K_3 \cup K_4 = \{c_4, c_5\}.$$

Moreover, $c_6 \in K_5$ and $K_5 \cap X = \{c_4\} \neq \varnothing$, while every covering block containing $c_4$ or $c_5$ intersects $X$, hence

$$\overline{\mathrm{apr}}^{\,t}_{\mathcal{C}}(X) = \{c_4, c_5, c_6\}.$$

Thus the tight boundary and negative regions are

$$\mathrm{BND}^t_{\mathcal{C}}(X) = \overline{\mathrm{apr}}^{\,t}_{\mathcal{C}}(X) \setminus \underline{\mathrm{apr}}^{\,t}_{\mathcal{C}}(X) = \{c_6\}, \qquad U \setminus \overline{\mathrm{apr}}^{\,t}_{\mathcal{C}}(X) = \{c_1, c_2, c_3\}.$$

**Loose (weak) approximations.** All blocks that intersect $X$ are $K_2, K_3, K_4, K_5$, so

$$\overline{\mathrm{apr}}^{\,\ell}_{\mathcal{C}}(X) = \bigcup\{K \in \mathcal{C} \mid K \cap X \neq \varnothing\}$$

$$= K_2 \cup K_3 \cup K_4 \cup K_5 = \{c_3, c_4, c_5, c_6\}.$$

For the loose lower approximation, we require that *every* covering block containing the customer is contained in $X$. Customer $c_5$ belongs only to $K_3$ and $K_4$, and both satisfy $K_3 \subseteq X$ and $K_4 \subseteq X$, hence

$$\underline{\mathrm{apr}}^{\,\ell}_{\mathcal{C}}(X) = \{c_5\}.$$

Therefore,

$$\mathrm{BND}^\ell_{\mathcal{C}}(X) = \overline{\mathrm{apr}}^{\,\ell}_{\mathcal{C}}(X) \setminus \underline{\mathrm{apr}}^{\,\ell}_{\mathcal{C}}(X) = \{c_3, c_4, c_6\}, \qquad U \setminus \overline{\mathrm{apr}}^{\,\ell}_{\mathcal{C}}(X) = \{c_1, c_2\}.$$

**Interpretation.** Under segment information $\mathcal{C}$, $c_5$ is *certainly* high-risk even in the loose sense, while $c_6$ is only *possibly* high-risk because it co-occurs with a risky customer in an overlapping segment.

## 2.34 Local Rough Set

A local rough set approximates a target concept using only granules generated by objects inside it, via $\alpha/\beta$ thresholds efficiently [195–198].

**Definition 2.34.1** (Local rough set (LRS))**.** [195, 196] Let $(U, R)$ be an approximation space, where $U$ is a nonempty finite universe and $R$ is an equivalence relation on $U$. For each $x \in U$, write

$$[x]_R := \{\, y \in U \mid (x, y) \in R \,\}$$

for the $R$-equivalence class (information granule) of $x$. Let

$$D : \mathcal{P}(U) \times \mathcal{P}(U) \longrightarrow [0, 1]$$

be an *inclusion degree* (a conditional/inclusion measure), e.g.

$$D(A, B) := \frac{|A \cap B|}{|B|} \qquad (B \neq \emptyset).$$

Fix thresholds $0 \leq \beta < \alpha \leq 1$. For any target concept $X \subseteq U$, define the *local $\alpha$-lower* and *local $\beta$-upper* approximations by

$$\underline{\mathrm{apr}}_{\alpha}^{\mathrm{LRS}}(X) \;:=\; \{\, x \in X \mid D(X, [x]_R) \geq \alpha \,\},$$

$$\overline{\mathrm{apr}}_{\beta}^{\mathrm{LRS}}(X) \;:=\; \{\, x \in U \mid D(X, [x]_R) > \beta \,\}$$

$$= \bigcup \{\, [x]_R \mid x \in X,\ D(X, [x]_R) > \beta \,\}.$$

The ordered pair

$$\big(\underline{\mathrm{apr}}_{\alpha}^{\mathrm{LRS}}(X),\ \overline{\mathrm{apr}}_{\beta}^{\mathrm{LRS}}(X)\big)$$

is called the $(\alpha, \beta)$-*local rough set* of $X$ (with respect to $R$ and $D$), and its *local boundary region* is

$$\mathrm{Bn}_{\alpha,\beta}^{\mathrm{LRS}}(X) := \overline{\mathrm{apr}}_{\beta}^{\mathrm{LRS}}(X) \setminus \underline{\mathrm{apr}}_{\alpha}^{\mathrm{LRS}}(X).$$

**Example 2.34.2** (Local rough set for credit-risk screening)**.** Consider a bank that groups applicants into *indiscernibility classes* using a coarse credit profile (e.g., the same credit-score band and income band). Let

$$U = \{u_1, u_2, u_3, u_4, u_5, u_6, u_7, u_8\}$$

be eight recent applicants. Define an equivalence relation $R$ on $U$ by declaring two applicants equivalent if they fall into the same coarse profile class. Suppose the $R$-classes are

$$[u_1]_R = [u_2]_R = [u_3]_R = [u_4]_R = \{u_1, u_2, u_3, u_4\},$$
$$[u_5]_R = [u_6]_R = \{u_5, u_6\}, \qquad [u_7]_R = [u_8]_R = \{u_7, u_8\}.$$

Let the target concept $X \subseteq U$ be the set of applicants who *defaulted within 12 months* in historical data:

$$X = \{u_1, u_2, u_3, u_5\}.$$

Use the inclusion degree

$$D(A, B) := \frac{|A \cap B|}{|B|} \qquad (B \neq \emptyset),$$

and choose thresholds $\alpha = 0.75$ and $\beta = 0.25$.

**Step 1: local inclusion degrees.** Compute $D(X, [x]_R) = |X \cap [x]_R|/|[x]_R|$ for each class:

$$D\big(X, \{u_1, u_2, u_3, u_4\}\big) = \frac{3}{4} = 0.75,$$
$$D\big(X, \{u_5, u_6\}\big) = \frac{1}{2} = 0.50,$$
$$D\big(X, \{u_7, u_8\}\big) = \frac{0}{2} = 0.$$

**Step 2: LRS approximations.** The local $\alpha$-lower approximation keeps only defaulted applicants whose whole profile-class is $\alpha$-dominated by defaults:

$$\underline{\mathrm{apr}}_{\alpha}^{\mathrm{LRS}}(X) = \{x \in X \mid D(X, [x]_R) \geq 0.75\} = \{u_1, u_2, u_3\}.$$

The local $\beta$-upper approximation includes any applicant whose profile-class has default-rate $> \beta$:

$$\overline{\mathrm{apr}}_{\beta}^{\mathrm{LRS}}(X) = \{x \in U \mid D(X, [x]_R) > 0.25\} = \{u_1, u_2, u_3, u_4, u_5, u_6\}.$$

Hence the local boundary region (ambiguous-risk region) is

$$\mathrm{Bn}_{\alpha,\beta}^{\mathrm{LRS}}(X) = \overline{\mathrm{apr}}_{\beta}^{\mathrm{LRS}}(X) \setminus \underline{\mathrm{apr}}_{\alpha}^{\mathrm{LRS}}(X) = \{u_4, u_5, u_6\}.$$

**Interpretation.** Applicants $u_1, u_2, u_3$ are *certainly* high-risk under this granulation (their class default-rate is 0.75), while $u_4, u_5, u_6$ are *possibly* high-risk (their classes exceed $\beta$) but not certain under $\alpha$, so they belong to the local boundary and may be sent to manual review.

## 2.35 Interval-valued Rough Set

An interval-valued rough set assigns each element an interval $[l(x), u(x)]$ representing possible membership, derived from lower/upper approximations in data analysis [199–202].

**Definition 2.35.1** (Interval domain)**.** Let $D$ be a linearly ordered set (typically $D \subseteq \mathbb{R}$). Define the set of (closed) intervals over $D$ by

$$\mathrm{I}(D) \;:=\; \{\, [\ell, u] \subseteq D \mid \ell \le u \,\}.$$

For $I = [\ell, u]$, $J = [\ell', u'] \in \mathrm{I}(D)$, we write $I \cap J \neq \emptyset$ iff $\max(\ell, \ell') \le \min(u, u')$.

**Definition 2.35.2** (Interval-valued information system)**.** An *interval-valued information system* is a quadruple

$$S \;=\; (U, A, \{D_a\}_{a \in A}, f),$$

where $U$ is a nonempty finite universe of objects, $A$ is a nonempty finite set of attributes, each attribute $a \in A$ has an ordered domain $D_a$, and

$$f : U \times A \to \bigcup_{a \in A} \mathrm{I}(D_a)$$

satisfies $f(x, a) \in \mathrm{I}(D_a)$ for all $(x, a) \in U \times A$. We write $f(x, a) = [f^-(x, a), f^+(x, a)]$ with $f^-(x, a) \le f^+(x, a)$.

**Definition 2.35.3** (Interval-tolerance (interval indiscernibility) relation)**.** Fix a nonempty attribute subset $B \subseteq A$. Define a binary relation $\sim_B$ on $U$ by

$$x \sim_B y \quad \Longleftrightarrow \quad \forall a \in B,\;\; f(x, a) \cap f(y, a) \neq \emptyset.$$

For each $x \in U$, the $B$*-neighborhood* (tolerance class) of $x$ is

$$[x]_B \;:=\; \{\, y \in U \mid x \sim_B y \,\}.$$

Note that $\sim_B$ is always reflexive and symmetric; it need not be transitive.

**Definition 2.35.4** (Interval-valued rough approximations)**.** Let $X \subseteq U$ be a (crisp) target set and let $B \subseteq A$ be nonempty. The $B$*-lower* and $B$*-upper* approximations of $X$ (induced by the interval-tolerance neighborhoods) are

$$\underline{B}(X) \;:=\; \{\, x \in U \mid [x]_B \subseteq X \,\}, \qquad \overline{B}(X) \;:=\; \{\, x \in U \mid [x]_B \cap X \neq \emptyset \,\}.$$

**Definition 2.35.5** (Interval-valued rough set)**.** The *interval-valued rough set* of $X$ w.r.t. $B$ is the pair

$$\mathsf{RS}_B(X) \;:=\; \big(\underline{B}(X), \overline{B}(X)\big).$$

Its *positive*, *boundary*, and *negative* regions are

$$\mathrm{POS}_B(X) := \underline{B}(X), \qquad \mathrm{BND}_B(X) := \overline{B}(X) \setminus \underline{B}(X), \qquad \mathrm{NEG}_B(X) := U \setminus \overline{B}(X).$$

If $\sim_B$ happens to be an equivalence relation (e.g., when intervals collapse to point values and equality is used), this reduces to the classical Pawlak rough set model.

**Example 2.35.6** (Interval-valued rough set for hypertension screening)**.** Consider a clinic where each patient's systolic blood pressure is recorded as an *interval* (to reflect device noise and short-term variability). Let the universe be

$$U = \{p_1, p_2, p_3, p_4, p_5\}.$$

Take one interval-valued attribute $B = \{\mathrm{SBP}\}$, and assign:

$$I(p_1) = [118, 125], \quad I(p_2) = [128, 138], \quad I(p_3) = [135, 150],$$
$$I(p_4) = [148, 160], \quad I(p_5) = [155, 170].$$

Define an interval-based indiscernibility/tolerance relation $\sim_B$ on $U$ by

$$p_i \sim_B p_j \iff I(p_i) \cap I(p_j) \neq \emptyset,$$

and write the $B$-neighborhood of $p$ as $[p]_B := \{q \in U \mid q \sim_B p\}$. Then:

$$[p_1]_B = \{p_1\}, \quad [p_2]_B = \{p_2, p_3\}, \quad [p_3]_B = \{p_2, p_3, p_4\}, \quad [p_4]_B = \{p_3, p_4, p_5\}, \quad [p_5]_B = \{p_4, p_5\}.$$

Let the target concept $X \subseteq U$ be the set of patients flagged for antihypertensive evaluation:

$$X = \{p_3, p_4, p_5\}.$$

Using the standard neighborhood-based rough operators (compatible with the interval-valued setting),

$$\underline{B}(X) := \{p \in U \mid [p]_B \subseteq X\}, \qquad \overline{B}(X) := \{p \in U \mid [p]_B \cap X \neq \emptyset\},$$

we obtain

$$\underline{B}(X) = \{p_4, p_5\}, \qquad \overline{B}(X) = \{p_2, p_3, p_4, p_5\}.$$

Hence the interval-valued rough set $\mathsf{RS}_B(X) = (\underline{B}(X), \overline{B}(X))$ yields:

$$\mathrm{POS}_B(X) = \{p_4, p_5\}, \qquad \mathrm{BND}_B(X) = \{p_2, p_3\}, \qquad \mathrm{NEG}_B(X) = \{p_1\}.$$

Interpretation: $p_4, p_5$ are *definitely* high-risk under interval overlap uncertainty, $p_2, p_3$ are *borderline/ambiguous*, and $p_1$ is *definitely not* flagged.

## 2.36 Tolerance Rough Sets

Rough sets using tolerance (reflexive, symmetric) similarity relations; lower/upper approximations via tolerance neighborhoods, allowing overlapping classes, relaxed equality, analysis robust [25, 160, 203–205].

**Definition 2.36.1** (Tolerance approximation space and tolerance rough set)**.** Let $U$ be a nonempty finite universe and let $T \subseteq U \times U$ be a *tolerance relation*, i.e., $T$ is *reflexive* and *symmetric.* For each $x \in U$, define the *$T$-neighborhood* (tolerance class) of $x$ by

$$T(x) \;:=\; \{\, y \in U \mid (x, y) \in T \,\}.$$

Note that, unlike equivalence classes, tolerance classes may overlap, so an object can belong to multiple tolerance classes.

For any $X \subseteq U$, define the *tolerance (lower/upper) approximations* of $X$ by

$$\underline{T}(X) \;:=\; \{\, x \in U \mid T(x) \subseteq X \,\}, \qquad \overline{T}(X) \;:=\; \{\, x \in U \mid T(x) \cap X \neq \varnothing \,\}.$$

The pair $\big(\underline{T}(X), \overline{T}(X)\big)$ is called the *tolerance rough set* of $X$ (with respect to $T$).

**Remark 2.36.2** (Block/covering viewpoint (optional))**.** A *(maximal) tolerance block* is a subset $B \subseteq U$ such that $B \times B \subseteq T$ and $B$ is inclusion-maximal with this property; such blocks are also called maximal consistent blocks. Let $\mathcal{B}(T)$ be the family of all maximal tolerance blocks; typically $\mathcal{B}(T)$ forms a covering of $U$. Then one can also define (covering-style) approximations by

$$\underline{\mathrm{apr}}_{\mathcal{B}(T)}(X) := \bigcup\{\, B \in \mathcal{B}(T) \mid B \subseteq X \,\}, \qquad \overline{\mathrm{apr}}_{\mathcal{B}(T)}(X) := \bigcup\{\, B \in \mathcal{B}(T) \mid B \cap X \neq \varnothing \,\}.$$

**Example 2.36.3** (Tolerance rough set for churn-risk screening via "similar spending" groups)**.** A subscription service groups customers by *tolerably similar* monthly spending. Let

$$U = \{c_1, c_2, c_3, c_4, c_5, c_6\}$$

be six customers, and let the observed monthly spending (in normalized units) be

$$s(c_1) = 2, \quad s(c_2) = 3, \quad s(c_3) = 3, \quad s(c_4) = 4, \quad s(c_5) = 5, \quad s(c_6) = 5.$$

Define a tolerance relation $T \subseteq U \times U$ by

$$(x, y) \in T \quad \Longleftrightarrow \quad |s(x) - s(y)| \leq 1.$$

Then $T$ is reflexive and symmetric (hence a tolerance relation). The $T$-neighborhoods are

$$T(c_1) = \{c_1, c_2, c_3\}, \qquad T(c_2) = \{c_1, c_2, c_3, c_4\}, \qquad T(c_3) = \{c_1, c_2, c_3, c_4\},$$

$$T(c_4) = \{c_2, c_3, c_4, c_5, c_6\}, \qquad T(c_5) = \{c_4, c_5, c_6\}, \qquad T(c_6) = \{c_4, c_5, c_6\}.$$

(Notice the overlap, e.g., $c_4 \in T(c_2) \cap T(c_5)$.)

Let the target set be the customers assessed as "high churn risk":

$$X = \{c_4, c_5, c_6\} \subseteq U \qquad \text{(e.g., those with } s(\cdot) \geq 4).$$

The tolerance lower approximation is

$$\underline{T}(X) = \{x \in U \mid T(x) \subseteq X\} = \{c_5, c_6\},$$

since $T(c_5) = T(c_6) = \{c_4, c_5, c_6\} \subseteq X$, while $T(c_4)$ contains $c_2, c_3 \notin X$.

The tolerance upper approximation is

$$\overline{T}(X) = \{x \in U \mid T(x) \cap X \neq \varnothing\} = \{c_2, c_3, c_4, c_5, c_6\},$$

because $T(c_2) \cap X \neq \varnothing$ and $T(c_3) \cap X \neq \varnothing$ (both contain $c_4$), whereas $T(c_1) \cap X = \varnothing$.

Hence the tolerance rough set of $X$ (w.r.t. $T$) is

$$\big(\underline{T}(X), \overline{T}(X)\big) = \big(\{c_5, c_6\},\ \{c_2, c_3, c_4, c_5, c_6\}\big),$$

with regions

$$\mathrm{POS}_T(X) = \{c_5, c_6\}, \qquad \mathrm{BND}_T(X) = \{c_2, c_3, c_4\}, \qquad \mathrm{NEG}_T(X) = \{c_1\}.$$

Interpretation: $c_5, c_6$ are *definitely* high-risk under tolerance grouping; $c_2, c_3, c_4$ are *possibly* high-risk because their tolerance neighborhoods touch the high-risk set; $c_1$ is *definitely not* high-risk under the same tolerance criterion.

## 2.37 One–directional s–Rough Set

An One–directional s–Rough Set approximates a function set using R-equivalence classes after one-directional transfer expansion via F, producing lower/upper rough approximations as a pair [206, 207].

**Definition 2.37.1** (One-directional $S$-rough set)**.** Let $U$ be a nonempty finite universe and let $R \subseteq U \times U$ be an equivalence relation. For $x \in U$, write

$$[x]_R := \{\, y \in U \mid (x, y) \in R \,\}$$

for the $R$-equivalence class of $x$.

Let $\mathcal{F}$ be a nonempty family of *element transfer functions* (e.g., partial maps $f : U \rightharpoonup U$), and fix $f \in \mathcal{F}$. For any $X \subseteq U$, define the *$f$-extension* of $X$ by

$$X_f := \{\, u \in U \setminus X \mid f(u) \in X \,\},$$

and define the associated *one-directional $S$-set* of $X$ by

$$X^\circ := X \cup X_f.$$

The *one-directional $S$-lower approximation* and *one-directional $S$-upper approximation* of $X^\circ$ (with respect to $(R, \mathcal{F})$ and the fixed $f$) are defined by

$$(R, \mathcal{F})^*(X^\circ) := \bigcup \{\, [x]_R \mid x \in U,\ [x]_R \subseteq X^\circ \,\} = \{\, x \in U \mid [x]_R \subseteq X^\circ \,\},$$

$$(R, \mathcal{F})^\circ(X^\circ) := \bigcup \{\, [x]_R \mid x \in U,\ [x]_R \cap X^\circ \neq \varnothing \,\} = \{\, x \in U \mid [x]_R \cap X^\circ \neq \varnothing \,\}.$$

The ordered pair

$$\big((R, \mathcal{F})^*(X^\circ),\ (R, \mathcal{F})^\circ(X^\circ)\big)$$

is called the *one-directional $S$-rough set* of $X^\circ$. Its boundary region is

$$\mathrm{Bn}_{R,\mathcal{F}}(X^\circ) := (R, \mathcal{F})^\circ(X^\circ) \setminus (R, \mathcal{F})^*(X^\circ).$$

**Example 2.37.2** (One-directional $S$-rough set in compliance training propagation)**.** Let $U = \{e_1, e_2, e_3, e_4, e_5, e_6\}$ be six employees. Assume employees are *indiscernible* (for a coarse compliance audit) when they belong to the same job-family group, so the equivalence relation $R$ partitions $U$ into

$$[e_1]_R = [e_2]_R = [e_5]_R = \{e_1, e_2, e_5\}, \qquad [e_3]_R = [e_4]_R = \{e_3, e_4\}, \qquad [e_6]_R = \{e_6\}.$$

Let $X \subseteq U$ be the set of employees who have *personally* passed a security-training quiz:

$$X = \{e_1, e_3\}.$$

Let $\mathcal{F}$ be a family of reporting-line maps, and fix the partial transfer function $f : U \rightharpoonup U$ ("$f(u)$ is the direct manager of $u$") given by

$$f(e_2) = e_1, \qquad f(e_4) = e_3, \qquad f(e_5) = e_2,$$

and $f(e_1), f(e_3), f(e_6)$ undefined.

**Step 1: one-directional extension.** By Definition 2.37.1, the $f$-extension is

$$X_f = \{ u \in U \setminus X \mid f(u) \in X \} = \{e_2, e_4\},$$

because $f(e_2) = e_1 \in X$ and $f(e_4) = e_3 \in X$, while $f(e_5) = e_2 \notin X$. Hence the associated one-directional $S$-set is

$$X^\circ = X \cup X_f = \{e_1, e_2, e_3, e_4\}.$$

**Step 2: Pawlak approximations of $X^\circ$ under $R$.** The one-directional $S$-lower approximation is

$$(R, \mathcal{F})^*(X^\circ) = \{x \in U \mid [x]_R \subseteq X^\circ\} = \{e_3, e_4\},$$

since $[e_3]_R = [e_4]_R = \{e_3, e_4\} \subseteq X^\circ$, while $[e_1]_R = \{e_1, e_2, e_5\} \nsubseteq X^\circ$ and $[e_6]_R = \{e_6\} \nsubseteq X^\circ$.

The one-directional $S$-upper approximation is

$$(R, \mathcal{F})^\circ(X^\circ) = \{x \in U \mid [x]_R \cap X^\circ \neq \varnothing\} = \{e_1, e_2, e_3, e_4, e_5\},$$

because $[e_1]_R = \{e_1, e_2, e_5\}$ intersects $X^\circ$ (via $e_1, e_2$), and $[e_3]_R = \{e_3, e_4\}$ intersects $X^\circ$, while $[e_6]_R$ does not.

Therefore the one-directional $S$-rough set of $X^\circ$ is

$$\big((R, \mathcal{F})^*(X^\circ),\ (R, \mathcal{F})^\circ(X^\circ)\big) = \big(\{e_3, e_4\},\ \{e_1, e_2, e_3, e_4, e_5\}\big).$$

Its boundary region is

$$\mathrm{Bn}_{R,\mathcal{F}}(X^\circ) = (R, \mathcal{F})^\circ(X^\circ) \setminus (R, \mathcal{F})^*(X^\circ) = \{e_1, e_2, e_5\}.$$

*Interpretation.* Passing the quiz can "propagate" one step along the manager map $f$, yielding $X^\circ$. However, the coarse indiscernibility $R$ (job-family grouping) prevents separating $e_5$ from $e_1, e_2$, so $\{e_1, e_2, e_5\}$ becomes a boundary (uncertain) region.

## 2.38 Complex Rough Set

Complex rough sets represent membership uncertainty using complex-valued grades (magnitude and phase) and compute lower/upper approximations via indiscernibility relations, enabling two-dimensional uncertainty modeling in practice.

**Definition 2.38.1** (Complex Rough Set (complex-coded Pawlak rough set))**.** Let $U \neq \varnothing$ be a universe and let $R \subseteq U \times U$ be an equivalence relation. For $A \subseteq U$, write the Pawlak lower and upper approximations as

$$\underline{R}(A) := \{x \in U \mid [x]_R \subseteq A\}, \qquad \overline{R}(A) := \{x \in U \mid [x]_R \cap A \neq \varnothing\}.$$

Define the *rough complex alphabet*

$$\mathbb{D}_{\mathrm{RS}} \;:=\; \{0,\; \mathrm{i},\; 1+\mathrm{i}\} \;\subset\; \mathbb{C}, \qquad (\mathrm{i}^2 = -1).$$

The *complex rough membership function* of $A$ is the map

$$\mu_A^{\mathbb{C}} : U \to \mathbb{D}_{\mathrm{RS}}, \qquad \mu_A^{\mathbb{C}}(x) \;:=\; \mathbf{1}_{\underline{R}(A)}(x) \;+\; \mathrm{i}\,\mathbf{1}_{\overline{R}(A)}(x),$$

where $\mathbf{1}_S$ denotes the indicator of a crisp set $S \subseteq U$. The pair

$$\mathrm{CR}(A) \;:=\; (U, \mu_A^{\mathbb{C}})$$

is called the *Complex Rough Set* induced by $A$ (under $R$). Equivalently, for $x \in U$:

$$\mu_A^{\mathbb{C}}(x) = \begin{cases} 1+\mathrm{i}, & x \in \underline{R}(A) \quad \text{(definitely in)}, \\ \mathrm{i}, & x \in \overline{R}(A) \setminus \underline{R}(A) \quad \text{(boundary)}, \\ 0, & x \in U \setminus \overline{R}(A) \quad \text{(definitely out)}. \end{cases}$$

**Definition 2.38.2** (Rough equality)**.** For $A, B \subseteq U$, write $A \equiv_R B$ if

$$\underline{R}(A) = \underline{R}(B) \quad \text{and} \quad \overline{R}(A) = \overline{R}(B).$$

This is an equivalence relation on $\mathcal{P}(U)$; its equivalence classes are the usual *rough sets* (induced by $R$) represented by approximation pairs.

**Example 2.38.3** (Complex rough set for transaction-fraud screening)**.** A payment provider groups transactions into "indiscernibility" blocks using coarse features (e.g., same card, same merchant category, and same hour), because transactions in the same block are operationally hard to distinguish at first glance.

Let

$$U = \{t_1, t_2, t_3, t_4, t_5, t_6, t_7\}$$

be seven transactions, and let $R$ be the equivalence relation "belongs to the same block" with equivalence classes

$$[t_1]_R = \{t_1, t_2\}, \qquad [t_3]_R = \{t_3, t_4, t_5\}, \qquad [t_6]_R = \{t_6\}, \qquad [t_7]_R = \{t_7\}.$$

Suppose the (currently) confirmed fraudulent transactions are

$$A = \{t_2, t_4, t_6\} \subseteq U.$$

Then the Pawlak lower and upper approximations are

$$\underline{R}(A) = \{x \in U \mid [x]_R \subseteq A\} = \{t_6\},$$

because only the singleton class $[t_6]_R = \{t_6\}$ is fully contained in $A$, and

$$\overline{R}(A) = \{x \in U \mid [x]_R \cap A \neq \varnothing\} = \{t_1, t_2, t_3, t_4, t_5, t_6\},$$

since $[t_1]_R$ meets $A$ (via $t_2$) and $[t_3]_R$ meets $A$ (via $t_4$), while $[t_7]_R$ does not.

Hence the complex rough membership function (Definition 2.38.1) is

$$\mu_A^{\mathbb{C}}(x) = \mathbf{1}_{\underline{R}(A)}(x) + \mathrm{i}\,\mathbf{1}_{\overline{R}(A)}(x),$$

so explicitly

$$\mu_A^{\mathbb{C}}(t_6) = 1 + \mathrm{i}, \qquad \mu_A^{\mathbb{C}}(t_j) = \mathrm{i}\ (j \in \{1,2,3,4,5\}), \qquad \mu_A^{\mathbb{C}}(t_7) = 0.$$

Interpretation: $t_6$ is *definitely fraudulent* (its whole block is confirmed), $t_1, t_2, t_3, t_4, t_5$ are *boundary/possibly fraudulent* (their blocks contain at least one fraud), and $t_7$ is *definitely non-fraudulent* under the current granulation. The resulting complex rough set is $\mathrm{CR}(A) = (U, \mu_A^{\mathbb{C}})$.

**Theorem 2.38.4** (Well-definedness and faithfulness of the complex coding)**.** *Let $(U, R)$ be an approximation space with $R$ an equivalence relation.*

(i) *For every $A \subseteq U$, the map $\mu_A^{\mathbb{C}}$ is well-defined and satisfies $\mu_A^{\mathbb{C}}(U) \subseteq \mathbb{D}_{\mathrm{RS}}$.*

(ii) *The complex coding depends only on the rough set (approximation pair): if $A \equiv_R B$, then*

$$\mu_A^{\mathbb{C}} = \mu_B^{\mathbb{C}}.$$

*Hence the assignment*

$$\Phi : \mathcal{P}(U)/{\equiv_R} \longrightarrow \mathbb{D}_{\mathrm{RS}}^U, \qquad \Phi([A]) := \mu_A^{\mathbb{C}}$$

*is* well-defined.

(iii) *The encoding is faithful: for every $A \subseteq U$ one can recover the approximations from $\mu_A^{\mathbb{C}}$ via*

$$\underline{R}(A) = \{x \in U \mid \Re(\mu_A^{\mathbb{C}}(x)) = 1\}, \qquad \overline{R}(A) = \{x \in U \mid \mathrm{Im}(\mu_A^{\mathbb{C}}(x)) = 1\}.$$

*In particular, $\Phi$ is injective.*

*Proof.* (i) For any $A \subseteq U$ and $x \in U$, the indicators $\mathbf{1}_{\underline{R}(A)}(x), \mathbf{1}_{\overline{R}(A)}(x) \in \{0,1\}$, so $\mu_A^{\mathbb{C}}(x) = a + ib$ with $a, b \in \{0,1\}$. Moreover, $\underline{R}(A) \subseteq \overline{R}(A)$ holds for Pawlak approximations, hence $a = 1 \Rightarrow b = 1$. Therefore only $0, \mathrm{i}, 1+\mathrm{i}$ occur and $\mu_A^{\mathbb{C}} : U \to \mathbb{D}_{\mathrm{RS}}$ is well-defined.

(ii) If $A \equiv_R B$, then $\underline{R}(A) = \underline{R}(B)$ and $\overline{R}(A) = \overline{R}(B)$. Thus, for every $x \in U$,

$$\mu_A^{\mathbb{C}}(x) = \mathbf{1}_{\underline{R}(A)}(x) + \mathrm{i}\,\mathbf{1}_{\overline{R}(A)}(x) = \mathbf{1}_{\underline{R}(B)}(x) + \mathrm{i}\,\mathbf{1}_{\overline{R}(B)}(x) = \mu_B^{\mathbb{C}}(x),$$

so $\mu_A^{\mathbb{C}} = \mu_B^{\mathbb{C}}$ and $\Phi$ is well-defined on the quotient.

(iii) By definition, $\Re(\mu_A^{\mathbb{C}}(x)) = \mathbf{1}_{\underline{R}(A)}(x)$ and $\mathrm{Im}(\mu_A^{\mathbb{C}}(x)) = \mathbf{1}_{\overline{R}(A)}(x)$ for all $x \in U$. Hence the displayed reconstruction identities follow immediately. If $\Phi([A]) = \Phi([B])$, then $\mu_A^{\mathbb{C}} = \mu_B^{\mathbb{C}}$, so their real and imaginary parts coincide pointwise, yielding $\underline{R}(A) = \underline{R}(B)$ and $\overline{R}(A) = \overline{R}(B)$, i.e. $A \equiv_R B$. Therefore $\Phi$ is injective. □

## 2.39 MetaRough Set

A MetaRough Set lifts rough approximations to meta-level families of rough objects via a meta-indiscernibility, yielding iterated roughness within a MetaStructure hierarchy of arbitrary depth [208].

**Definition 2.39.1** ((Recall) Pawlak approximation space and rough objects)**.** Let $X$ be a nonempty set and let $R \subseteq X \times X$ be an equivalence relation. For $A \subseteq X$, define the (Pawlak) lower and upper approximations of $A$ by

$$\underline{R}(A) := \{\, x \in X \mid [x]_R \subseteq A \,\}, \qquad \overline{R}(A) := \{\, x \in X \mid [x]_R \cap A \neq \emptyset \,\},$$

where $[x]_R := \{\, y \in X \mid (x, y) \in R \,\}$. The set of *rough objects* over $(X, R)$ is

$$Rough(X, R) := \{\, (\underline{R}(A), \overline{R}(A)) \;\mid\; A \subseteq X \,\}.$$

**Definition 2.39.2** (Meta-indiscernibility on rough objects)**.** Fix a Pawlak space $(X, R)$ and let $E$ be an equivalence relation on $Rough(X, R)$. For $r \in Rough(X, R)$, denote its $E$-equivalence class by

$$[r]_E := \{\, s \in Rough(X, R) \mid s\, E\, r \,\}.$$

**Definition 2.39.3** (MetaRough approximations and MetaRough Set)**.** Fix $(X, R)$ and an equivalence relation $E$ on $Rough(X, R)$. For any family $C \subseteq Rough(X, R)$, define the *meta-lower* and *meta-upper* approximations (with respect to $E$) by

$$\underline{C}^{E} := \{\, r \in Rough(X, R) \mid [r]_E \subseteq C \,\}, \qquad \overline{C}^{E} := \{\, r \in Rough(X, R) \mid [r]_E \cap C \neq \emptyset \,\}.$$

The ordered pair

$$(\underline{C}^{E},\ \overline{C}^{E})$$

is called the *MetaRough Set* of $C$ (in $Rough(X, R)$) with respect to the meta-indiscernibility $E$.

**Example 2.39.4** (MetaRough set in privacy-preserving risk-policy selection)**.** Consider a hospital triage system that initially groups patients only by coarse observable profiles (e.g., age bracket and two binary symptoms). Let

$$X = \{x_1, x_2, x_3, x_4, x_5\}$$

be five patients and let $R$ be the induced indiscernibility relation with equivalence classes

$$E_1 := \{x_1, x_2\}, \qquad E_2 := \{x_3, x_4\}, \qquad E_3 := \{x_5\}.$$

Thus $(X, R)$ is a Pawlak approximation space (Definition 2.39.1). For any $A \subseteq X$, write its rough object as

$$r(A) := \big(\underline{R}(A),\ \overline{R}(A)\big) \in Rough(X, R), \qquad \mathrm{Bnd}(r(A)) := \overline{R}(A) \setminus \underline{R}(A).$$

**Step 1: Rough objects (examples).** Let

$$A_5 := \{x_5\}, \qquad A_{15} := \{x_1, x_5\}.$$

Then
$$r(A_5) = \big(\{x_5\}, \{x_5\}\big), \qquad \mathrm{Bnd}(r(A_5)) = \varnothing,$$
because $A_5$ is a union of $R$-classes. In contrast,
$$\underline{R}(A_{15}) = \{x_5\}, \qquad \overline{R}(A_{15}) = E_1 \cup E_3 = \{x_1, x_2, x_5\},$$
hence
$$r(A_{15}) = \big(\{x_5\}, \{x_1, x_2, x_5\}\big), \qquad \mathrm{Bnd}(r(A_{15})) = \{x_1, x_2\}.$$

**Step 2: Meta-indiscernibility on rough objects.** At the *policy* layer, suppose an auditing rule is privacy-restricted and can see only how ambiguous a rough object is (i.e., the size of its boundary region), but not which patients are in it. Define an equivalence relation $E$ on $\mathit{Rough}(X, R)$ by
$$r\, E\, s \quad \Longleftrightarrow \quad \big|\mathrm{Bnd}(r)\big| = \big|\mathrm{Bnd}(s)\big|.$$
(Reflexivity/symmetry/transitivity are immediate because equality of integers is an equivalence relation.)

Let
$$C_0 := \big\{\, r(A) \in \mathit{Rough}(X, R) \;\big|\; \mathrm{Bnd}(r(A)) = \varnothing \,\big\}.$$
Since $\mathrm{Bnd}(r(A)) = \varnothing$ holds exactly when $A$ is a union of $R$-classes, we have the explicit list
$$C_0 = \Big\{(\varnothing, \varnothing),\ (E_1, E_1),\ (E_2, E_2),\ (E_3, E_3),$$
$$(E_1{\cup}E_2, E_1{\cup}E_2),\ (E_1{\cup}E_3, E_1{\cup}E_3),\ (E_2{\cup}E_3, E_2{\cup}E_3),\ (X, X)\Big\}.$$
Moreover, for every $r \in C_0$ one has $[r]_E = C_0$ (the whole "boundary-0" class).

**Step 3: A meta-level target family and MetaRough approximations.** Suppose the hospital wants to select *definable* rough objects that *definitely classify* $x_5$ *as high risk* (e.g., $x_5$ has a critical biomarker). Represent this policy target as the family
$$C := \{\, (A, A) \in C_0 \mid x_5 \in A \,\}$$
$$= \Big\{(E_3, E_3),\ (E_1{\cup}E_3, E_1{\cup}E_3),\ (E_2{\cup}E_3, E_2{\cup}E_3),\ (X, X)\Big\} \subseteq \mathit{Rough}(X, R).$$

Compute the MetaRough approximations of $C$ in $\mathit{Rough}(X, R)$ with respect to $E$ (Definition 2.39.3). If $r \in C_0$, then $[r]_E = C_0$, so
$$[r]_E \subseteq C \ \text{ fails } \ (\text{since } C \subsetneq C_0), \qquad [r]_E \cap C = C \neq \varnothing.$$
Hence every $r \in C_0$ lies in the meta-upper approximation but none lie in the meta-lower:
$$\underline{C}^E = \varnothing, \qquad \overline{C}^E = C_0.$$
If $r \notin C_0$ (i.e., $\mathrm{Bnd}(r) \neq \varnothing$), then $|\,\mathrm{Bnd}(r)\,| > 0$ and thus $[r]_E$ is disjoint from $C$ (which consists only of boundary-0 rough objects). Therefore
$$r \notin \overline{C}^E.$$

**Interpretation.** At the meta-level, because $E$ forgets *which* patients appear and retains only boundary size, all definable rough objects (boundary 0) become *indistinguishable*. Thus the policy family $C$ is only *possibly* recognized among definable rough objects: the MetaRough set $\big(\underline{C}^E, \overline{C}^E\big)$ equals $(\varnothing, C_0)$, so the meta-boundary $\overline{C}^E \setminus \underline{C}^E$ is nonempty and equals $C_0$.

**Proposition 2.39.5** (Meta-level sandwich property)**.** *For every $C \subseteq Rough(X, R)$ and every equivalence relation $E$ on $Rough(X, R)$,*

$$\underline{C}^{E} \subseteq C \subseteq \overline{C}^{E}.$$

*Proof.* If $r \in \underline{C}^{E}$ then $[r]_E \subseteq C$, hence $r \in [r]_E \subseteq C$ and $\underline{C}^{E} \subseteq C$. If $r \in C$ then $[r]_E \cap C \supseteq \{r\} \neq \emptyset$, hence $r \in \overline{C}^{E}$ and $C \subseteq \overline{C}^{E}$. □

## 2.40 $T$-valued Rough Set

$T$-valued rough sets replace membership with values in $T$; lower and upper approximations are computed via a $T$-valued relation and operators in decisions under uncertainty.

**Definition 2.40.1** ($T$-valued (structure-valued) rough set)**.** Let $U \neq \varnothing$ be a finite universe and let $R \subseteq U \times U$ be an equivalence relation. For $x \in U$, write

$$[x]_R := \{ y \in U \mid (x, y) \in R \}.$$

Let $(T, \leq_T)$ be a poset such that every nonempty finite subset of $T$ admits a meet and a join (equivalently, $T$ is a lattice if one prefers a global assumption). A $T$*-valued set* on $U$ is a mapping $A : U \to T$.

Define the $T$*-valued lower* and $T$*-valued upper* rough approximations of $A$ by

$$\underline{\mathrm{apr}}_R^T(A)(x) \;:=\; \bigwedge\nolimits_{y \in [x]_R} A(y),$$

$$\overline{\mathrm{apr}}_R^T(A)(x) \;:=\; \bigvee\nolimits_{y \in [x]_R} A(y) \qquad (x \in U),$$

where $\wedge$ and $\vee$ are the meet and join in $T$. The pair

$$\big(\underline{\mathrm{apr}}_R^T(A),\; \overline{\mathrm{apr}}_R^T(A)\big)$$

is called the $T$*-valued rough set* (or *structure-valued rough approximation*) induced by $A$ on $(U, R)$.

**Example 2.40.2** ($T$-valued rough set: coarse loan screening with ordinal decision tags)**.** Let $U = \{u_1, u_2, u_3, u_4\}$ be four loan applicants. Suppose the bank groups applicants by a coarse profile (e.g., *credit-score band* × *income band*), yielding the equivalence classes

$$[u_1]_R = [u_2]_R = \{u_1, u_2\}, \qquad [u_3]_R = [u_4]_R = \{u_3, u_4\}.$$

Let

$$T = \{\text{Reject} \prec \text{Review} \prec \text{Approve}\}$$

be a finite chain (hence a lattice), where $\wedge$ is the minimum and $\vee$ is the maximum. Define a $T$-valued set $A : U \to T$ by the bank's preliminary tag:

$$A(u_1) = \text{Review}, \quad A(u_2) = \text{Approve}, \quad A(u_3) = \text{Reject}, \quad A(u_4) = \text{Review}.$$

Then the $T$-valued lower/upper approximations (Definition 2.40.1) are

$$\underline{\mathrm{apr}}_R^T(A)(u_1) = A(u_1) \wedge A(u_2) = \text{Review}, \qquad \overline{\mathrm{apr}}_R^T(A)(u_1) = A(u_1) \vee A(u_2) = \text{Approve},$$

and

$$\underline{\mathrm{apr}}_R^T(A)(u_3) = A(u_3) \wedge A(u_4) = \text{Reject}, \qquad \overline{\mathrm{apr}}_R^T(A)(u_3) = A(u_3) \vee A(u_4) = \text{Review}.$$

Interpretation: within an indiscernibility class, the lower tag is the most conservative (worst-case) decision, while the upper tag is the most optimistic (best-case) decision.

**Definition 2.40.3** (Vector-valued rough set)**.** Let $(L, \leq)$ be a lattice (e.g., $L = [0, 1]$ with the usual order) and fix $p \geq 1$. Put $T := L^p$ and equip $T$ with the componentwise order:

$$\mathbf{u} \leq_T \mathbf{v} \quad \Longleftrightarrow \quad u_j \leq v_j \ (j = 1, \ldots, p).$$

Then $T$ is a lattice with componentwise meet/join:

$$(\mathbf{u} \wedge_T \mathbf{v})_j := u_j \wedge v_j, \qquad (\mathbf{u} \vee_T \mathbf{v})_j := u_j \vee v_j.$$

A *vector-valued rough set* on $(U, R)$ is a $T$-valued rough set in the sense of Definition 2.40.1, i.e., a mapping $A : U \to L^p$ together with

$$\underline{\mathrm{apr}}_R^{\mathrm{vec}}(A)(x) = \bigwedge_{y \in [x]_R} A(y), \qquad \overline{\mathrm{apr}}_R^{\mathrm{vec}}(A)(x) = \bigvee_{y \in [x]_R} A(y),$$

computed componentwise in $L^p$.

**Example 2.40.4** (Vector-valued rough set: multi-criteria risk profile under coarse indiscernibility)**.** Let the universe and equivalence relation be as above: $U = \{u_1, u_2, u_3, u_4\}$ with classes $\{u_1, u_2\}$ and $\{u_3, u_4\}$. Take $L = [0, 1]$ and $p = 3$, and interpret

$$A(u) = (r(u), f(u), p(u)) \in [0, 1]^3$$

as (default risk, fraud risk, profitability score). Define

$$A(u_1) = (0.20, 0.70, 0.40), \qquad A(u_2) = (0.50, 0.60, 0.90).$$

Then, using componentwise meet/join (Definition 2.40.3),

$$\underline{\mathrm{apr}}_R^{\mathrm{vec}}(A)(u_1) = \min\{A(u_1), A(u_2)\} = (0.20, 0.60, 0.40),$$

$$\overline{\mathrm{apr}}_R^{\mathrm{vec}}(A)(u_1) = \max\{A(u_1), A(u_2)\} = (0.50, 0.70, 0.90),$$

where min / max are taken componentwise. Interpretation: under coarse grouping, the lower vector gives a conservative envelope of scores available in the class, and the upper vector gives a permissive envelope.

**Definition 2.40.5** (Matrix-valued rough set)**.** Let $(L, \leq)$ be a lattice and fix integers $m, n \geq 1$. Put $T := \mathrm{Mat}_{m \times n}(L) \cong L^{m \times n}$ and define the entrywise order

$$A \leq_T B \quad \Longleftrightarrow \quad A_{ij} \leq B_{ij} \ (1 \leq i \leq m, \ 1 \leq j \leq n).$$

Then $T$ is a lattice with entrywise meet/join:

$$(A \wedge_T B)_{ij} := A_{ij} \wedge B_{ij}, \qquad (A \vee_T B)_{ij} := A_{ij} \vee B_{ij}.$$

A *matrix-valued rough set* on $(U, R)$ is a $T$-valued rough set, i.e., a mapping $A : U \to \mathrm{Mat}_{m \times n}(L)$ and approximations

$$\underline{\mathrm{apr}}_R^{\mathrm{mat}}(A)(x) = \bigwedge_{y \in [x]_R} A(y), \qquad \overline{\mathrm{apr}}_R^{\mathrm{mat}}(A)(x) = \bigvee_{y \in [x]_R} A(y),$$

where $\wedge, \vee$ are taken entrywise in $\mathrm{Mat}_{m \times n}(L)$.

**Example 2.40.6** (Matrix-valued rough set: scenario-by-horizon risk table aggregated within a class)**.** Let $U = \{u_1, u_2, u_3, u_4\}$ and $R$ be as above. Let $L = [0, 1]$ and consider $2 \times 2$ matrices whose entries represent risk scores for (short/long horizon)×(credit/liquidity scenario). Define a matrix-valued set $A : U \to \mathrm{Mat}_{2\times 2}([0, 1])$ by

$$A(u_1) = \begin{bmatrix} 0.3 & 0.6 \\ 0.5 & 0.4 \end{bmatrix}, \qquad A(u_2) = \begin{bmatrix} 0.7 & 0.2 \\ 0.4 & 0.8 \end{bmatrix}.$$

Then for the class $[u_1]_R = \{u_1, u_2\}$, the entrywise meet/join (Definition 2.40.5) give

$$\underline{\mathrm{apr}}_R^{\mathrm{mat}}(A)(u_1) = A(u_1) \wedge_T A(u_2) = \begin{bmatrix} \min(0.3, 0.7) & \min(0.6, 0.2) \\ \min(0.5, 0.4) & \min(0.4, 0.8) \end{bmatrix} = \begin{bmatrix} 0.3 & 0.2 \\ 0.4 & 0.4 \end{bmatrix},$$

$$\overline{\mathrm{apr}}_R^{\mathrm{mat}}(A)(u_1) = A(u_1) \vee_T A(u_2) = \begin{bmatrix} 0.7 & 0.6 \\ 0.5 & 0.8 \end{bmatrix}.$$

Interpretation: lower/upper matrix approximations provide conservative/optimistic bounds for each scenario entry when applicants are indistinguishable at the coarse level.

**Definition 2.40.7** (Tensor-valued rough set)**.** Let $(L, \leq)$ be a lattice and fix an order $r \geq 1$ and finite index sets $I_1, \ldots, I_r$. Put

$$T := L^{I_1 \times \cdots \times I_r}$$

(the set of $r$-way tensors with entries in $L$), ordered entrywise:

$$\mathcal{A} \leq_T \mathcal{B} \iff \mathcal{A}_{i_1,\ldots,i_r} \leq \mathcal{B}_{i_1,\ldots,i_r} \text{ for all } (i_1, \ldots, i_r) \in I_1 \times \cdots \times I_r.$$

Then $T$ is a lattice with entrywise meet/join. A *tensor-valued rough set* on $(U, R)$ is a $T$-valued rough set, i.e., a mapping $A : U \to T$ and approximations

$$\underline{\mathrm{apr}}_R^{\mathrm{ten}}(A)(x) = \bigwedge_{y \in [x]_R} A(y), \qquad \overline{\mathrm{apr}}_R^{\mathrm{ten}}(A)(x) = \bigvee_{y \in [x]_R} A(y),$$

computed entrywise in $L^{I_1 \times \cdots \times I_r}$.

**Example 2.40.8** (Tensor-valued rough set: sensor×shift×failure-mode anomaly cube)**.** Let $U = \{m_1, m_2, m_3\}$ be machines in a factory. Group machines by (model, firmware), giving the equivalence classes

$$[m_1]_R = [m_2]_R = \{m_1, m_2\}, \qquad [m_3]_R = \{m_3\}.$$

Let $L = [0, 1]$ and choose index sets

$$I_1 = \{s_1, s_2\} \text{ (sensors)}, \quad I_2 = \{d, n\} \text{ (day/night)}, \quad I_3 = \{f_1, f_2\} \text{ (failure modes)}.$$

A tensor-valued set $A : U \to L^{I_1 \times I_2 \times I_3}$ assigns to each machine a $2 \times 2 \times 2$ cube of anomaly probabilities. Define $A(m_1)$ and $A(m_2)$ by listing all entries:

| | $(d)$ | | $(n)$ | |
|---|---|---|---|---|
| | $f_1$ | $f_2$ | $f_1$ | $f_2$ |
| $s_1$ | 0.1 | 0.4 | 0.3 | 0.2 |
| $s_2$ | 0.6 | 0.5 | 0.7 | 0.1 |

for $A(m_1)$,

| | $(d)$ | | $(n)$ | |
|---|---|---|---|---|
| | $f_1$ | $f_2$ | $f_1$ | $f_2$ |
| $s_1$ | 0.2 | 0.3 | 0.8 | 0.4 |
| $s_2$ | 0.4 | 0.6 | 0.2 | 0.9 |

for $A(m_2)$.

Then for the class $[m_1]_R = \{m_1, m_2\}$, the tensor-valued lower/upper approximations (Definition 2.40.7) are computed entrywise:

$$\underline{\mathrm{apr}}_R^{\mathrm{ten}}(A)(m_1) = \min\{A(m_1), A(m_2)\}, \qquad \overline{\mathrm{apr}}_R^{\mathrm{ten}}(A)(m_1) = \max\{A(m_1), A(m_2)\},$$

so, for example,

$$\underline{\mathrm{apr}}_R^{\mathrm{ten}}(A)(m_1)_{s_1,d,f_1} = \min(0.1, 0.2) = 0.1, \quad \overline{\mathrm{apr}}_R^{\mathrm{ten}}(A)(m_1)_{s_1,d,f_1} = \max(0.1, 0.2) = 0.2,$$

$$\underline{\mathrm{apr}}_R^{\mathrm{ten}}(A)(m_1)_{s_2,n,f_2} = \min(0.1, 0.9) = 0.1, \quad \overline{\mathrm{apr}}_R^{\mathrm{ten}}(A)(m_1)_{s_2,n,f_2} = \max(0.1, 0.9) = 0.9,$$

and similarly for all other indices $(i_1, i_2, i_3) \in I_1 \times I_2 \times I_3$. Interpretation: the lower tensor gives a conservative anomaly cube (guaranteed within the class), while the upper tensor gives the maximal anomaly cube possible within the same coarse machine group.

## 2.41 Refined Rough Set

A refined rough set partitions classical lower/upper regions into finer layers using multiple thresholds or relations, improving granularity for analysis [209, 210]. Similar notions with an analogous refinement-style structure include refined neutrosophic sets [211–214] and refined soft sets [215, 216].

**Definition 2.41.1** (Refined rough set (multi-granularity refinement))**.** Let $U \neq \varnothing$ be a finite universe and let $p \in \mathbb{N}$. A *refinement chain* on $U$ is a finite family of equivalence relations

$$\mathcal{R} = \langle R_1, R_2, \ldots, R_p \rangle \quad \text{such that} \quad R_1 \subseteq R_2 \subseteq \cdots \subseteq R_p \ \subseteq \ U \times U.$$

(Thus $R_1$ is the finest granulation and $R_p$ is the coarsest.) For $x \in U$, write

$$[x]_{R_i} := \{\, y \in U \mid (x, y) \in R_i \,\}$$

for the $R_i$-equivalence class of $x$.

For any target set $A \subseteq U$ and each $i \in \{1, \ldots, p\}$, define the Pawlak lower/upper approximations at level $i$ by

$$\underline{\mathrm{apr}}_i(A) := \{\, x \in U \mid [x]_{R_i} \subseteq A \,\}, \qquad \overline{\mathrm{apr}}_i(A) := \{\, x \in U \mid [x]_{R_i} \cap A \neq \varnothing \,\}.$$

The *refined rough set* of $A$ with respect to the refinement chain $\mathcal{R}$ is the $p$-tuple of rough approximations

$$\mathbb{R}_{\mathcal{R}}(A) \ := \ \big( (\underline{\mathrm{apr}}_i(A), \overline{\mathrm{apr}}_i(A)) \big)_{i=1}^{p}.$$

**Example 2.41.2** (Refined rough set for fraud screening under multi-granularity customer grouping)**.** Let $U = \{u_1, u_2, u_3, u_4, u_5, u_6\}$ be six credit-card transactions. Let

$$A = \{u_1, u_2\} \subseteq U$$

be the (crisp) target set of transactions that are *confirmed fraudulent* after investigation.

We consider a refinement chain $\mathcal{R} = \langle R_1, R_2, R_3 \rangle$ capturing progressively coarser operational granules used by a bank:

- $R_1$: "same card *and* same merchant" (finest granulation),
- $R_2$: "same card" (intermediate granulation),
- $R_3$: "same country of transaction" (coarsest granulation).

Assume the induced equivalence classes are as follows.

**Level $R_1$ (same card and merchant).**

$$[u_1]_{R_1} = [u_2]_{R_1} = \{u_1, u_2\}, \qquad [u_3]_{R_1} = \{u_3\},$$
$$[u_4]_{R_1} = [u_5]_{R_1} = \{u_4, u_5\}, \qquad [u_6]_{R_1} = \{u_6\}.$$

**Level $R_2$ (same card).**

$$[u_1]_{R_2} = [u_2]_{R_2} = [u_3]_{R_2} = \{u_1, u_2, u_3\}, \qquad [u_4]_{R_2} = [u_5]_{R_2} = \{u_4, u_5\}, \qquad [u_6]_{R_2} = \{u_6\}.$$

**Level $R_3$ (same country).**

$$[u_1]_{R_3} = \cdots = [u_5]_{R_3} = \{u_1, u_2, u_3, u_4, u_5\}, \qquad [u_6]_{R_3} = \{u_6\}.$$

It is immediate that $R_1 \subseteq R_2 \subseteq R_3$ (each step merges some classes), hence $\mathcal{R}$ is a refinement chain.

Now compute Pawlak approximations at each level.

**Level $i = 1$.** Since $[u_1]_{R_1} = \{u_1, u_2\} \subseteq A$, we have $u_1, u_2 \in \underline{\mathrm{apr}}_1(A)$. All other $R_1$-classes are not contained in $A$, so

$$\underline{\mathrm{apr}}_1(A) = \{u_1, u_2\}, \qquad \overline{\mathrm{apr}}_1(A) = \{u_1, u_2\}.$$

**Level $i = 2$.** The class $\{u_1, u_2, u_3\}$ intersects $A$ but is not contained in $A$, so it contributes to the upper but not the lower. Thus

$$\underline{\mathrm{apr}}_2(A) = \{u_1, u_2\}, \qquad \overline{\mathrm{apr}}_2(A) = \{u_1, u_2, u_3\}.$$

**Level $i = 3$.** The large class $\{u_1, u_2, u_3, u_4, u_5\}$ intersects $A$ but is not contained in $A$, hence

$$\underline{\mathrm{apr}}_3(A) = \{u_1, u_2\}, \qquad \overline{\mathrm{apr}}_3(A) = \{u_1, u_2, u_3, u_4, u_5\}.$$

Therefore, the refined rough set of $A$ with respect to $\mathcal{R}$ is the 3-tuple

$$\mathbb{R}_{\mathcal{R}}(A) = \Big( (\underline{\mathrm{apr}}_1(A), \overline{\mathrm{apr}}_1(A)), (\underline{\mathrm{apr}}_2(A), \overline{\mathrm{apr}}_2(A)), (\underline{\mathrm{apr}}_3(A), \overline{\mathrm{apr}}_3(A)) \Big),$$

i.e.,

$$\mathbb{R}_{\mathcal{R}}(A) = \Big( (\{u_1, u_2\}, \{u_1, u_2\}), (\{u_1, u_2\}, \{u_1, u_2, u_3\}), (\{u_1, u_2\}, \{u_1, u_2, u_3, u_4, u_5\}) \Big).$$

*Interpretation.* At finer granularity ($R_1$), fraud is precisely isolated; at coarser granularities ($R_2, R_3$), the *possible* fraud region expands, reflecting weaker discriminatory power of the coarser grouping.

**Theorem 2.41.3** (Well-definedness and monotone refinement)**.** *In the setting of Definition 2.41.1, for every $A \subseteq U$ and each $i \in \{1, \ldots, p\}$:*

(i) $\underline{\mathrm{apr}}_i(A) \subseteq \overline{\mathrm{apr}}_i(A)$.

(ii) *(Monotonicity across refinement) If* $1 \le i \le j \le p$, *then*

$$\underline{\mathrm{apr}}_i(A) \supseteq \underline{\mathrm{apr}}_j(A), \qquad \overline{\mathrm{apr}}_i(A) \subseteq \overline{\mathrm{apr}}_j(A).$$

*Hence* $\mathbb{R}_{\mathcal{R}}(A)$ *is well-defined as a nested multi-level description of* $A$.

*Proof.* (i) Fix $i$ and take $x \in \underline{\mathrm{apr}}_i(A)$. Then $[x]_{R_i} \subseteq A$, so in particular $[x]_{R_i} \cap A \neq \varnothing$ (since $x \in [x]_{R_i}$ and $A \neq \varnothing$ is not required; if $A = \varnothing$ then $\underline{\mathrm{apr}}_i(A) = \varnothing$ and the inclusion is trivial). Thus $x \in \overline{\mathrm{apr}}_i(A)$, proving $\underline{\mathrm{apr}}_i(A) \subseteq \overline{\mathrm{apr}}_i(A)$.

(ii) Assume $R_i \subseteq R_j$ with $i \le j$. For equivalence relations, $R_i \subseteq R_j$ implies

$$[x]_{R_i} \subseteq [x]_{R_j} \qquad (\forall x \in U),$$

because any $y$ related to $x$ under $R_i$ is also related under $R_j$. Now, if $x \in \underline{\mathrm{apr}}_j(A)$, then $[x]_{R_j} \subseteq A$, hence $[x]_{R_i} \subseteq [x]_{R_j} \subseteq A$, so $x \in \underline{\mathrm{apr}}_i(A)$. Therefore $\underline{\mathrm{apr}}_i(A) \supseteq \underline{\mathrm{apr}}_j(A)$.

Similarly, if $x \in \overline{\mathrm{apr}}_i(A)$, then $[x]_{R_i} \cap A \neq \varnothing$. Since $[x]_{R_i} \subseteq [x]_{R_j}$, we also have $[x]_{R_j} \cap A \neq \varnothing$, so $x \in \overline{\mathrm{apr}}_j(A)$. Hence $\overline{\mathrm{apr}}_i(A) \subseteq \overline{\mathrm{apr}}_j(A)$. □

## 2.42 Rough cubic sets

Rough cubic sets approximate a cubic set via a cubic relation, producing lower N(A) and upper H(A) cubic memberships; inequality indicates indefinability in uncertain analysis [217]. Concepts with a similar structural flavor include cubic intuitionistic fuzzy sets [218–220] and neutrosophic cubic sets [221–226].

**Definition 2.42.1** (Rough cubic set (cubic rough set) induced by a cubic relation)**.** Let $X$ be a nonempty universe. A *cubic set* in $X$ is a mapping

$$A = \langle \widetilde{A}, \lambda \rangle : \ X \to [I] \times I,$$

where $\widetilde{A} : X \to [I]$ is an interval-valued fuzzy set and $\lambda : X \to I$ is a fuzzy set. A *cubic relation* on $X$ is a cubic set

$$\mathcal{R} = \langle \widetilde{R}, r \rangle : \ X \times X \to [I] \times I.$$

Write $\mathcal{R}^c = \langle \widetilde{R}^c, r^c \rangle$ for the pointwise complement.

For a cubic set $A = \langle \widetilde{A}, \lambda \rangle$ and $x \in X$, define the *(P-system) lower* and *(P-system) upper* rough approximations by

$$\underline{\mathrm{Apr}}_{\mathcal{R}}(A)(x) := \Big\langle \bigwedge_{y \in X} (\widetilde{A}(y) \ \vee \ \widetilde{R}^c(y,x)), \ \bigwedge_{y \in X} (\lambda(y) \ \vee \ r^c(y,x)) \Big\rangle,$$

$$\overline{\mathrm{Apr}}_{\mathcal{R}}(A)(x) := \Big\langle \bigvee_{y \in X} (\widetilde{A}(y) \ \wedge \ \widetilde{R}(x,y)), \ \bigvee_{y \in X} (\lambda(y) \ \wedge \ r(x,y)) \Big\rangle.$$

The pair

$$\mathrm{RCS}_{\mathcal{R}}(A) := \big(\underline{\mathrm{Apr}}_{\mathcal{R}}(A), \ \overline{\mathrm{Apr}}_{\mathcal{R}}(A)\big)$$

is called the *rough cubic set* (or *cubic rough set*) of $A$ induced by $\mathcal{R}$. If $\underline{\mathrm{Apr}}_{\mathcal{R}}(A) = \overline{\mathrm{Apr}}_{\mathcal{R}}(A)$, then $A$ is called *definable* (with respect to $\mathcal{R}$); otherwise, $A$ is called *undefinable*.

## 2.43 MOD Rough Set

MOD Rough sets use modular arithmetic on approximation operators, encoding membership degrees as residues, enabling periodic uncertainty modeling while preserving lower and upper bounds consistently.

**Definition 2.43.1** (MOD semantic value (mod $n$))**.** Fix an integer $n \geq 2$. Let $\mathsf{Tag} := \{\text{none}, \text{pseudo}, \text{genuine}\}$ be totally ordered by

$$\text{none} < \text{pseudo} < \text{genuine}.$$

We write $J$ for the join on $\mathsf{Tag}$, i.e.

$$J(t_1, t_2) := \max\{t_1, t_2\}, \qquad \bigvee_{i \in I} t_i := \max\{t_i \mid i \in I\}.$$

A *MOD semantic membership* is a pair $(\mu, t) \in [0, 1] \times \mathsf{Tag}$, where $\mu$ is the numeric membership degree and $t$ records the modular provenance tag.

**Definition 2.43.2** (MOD rough approximations and MOD rough set)**.** Let $U \neq \varnothing$ be a universe and let $R \subseteq U \times U$ be an equivalence relation. For $x \in U$, write the $R$-granule

$$[x]_R := \{\, y \in U \mid (x, y) \in R \,\}.$$

Let $\widetilde{A}$ be a MOD-described concept on $U$, given semantically by two functions

$$\mu_{\widetilde{A}} : U \to [0, 1], \qquad t_{\widetilde{A}} : U \to \mathsf{Tag}.$$

Define the *MOD lower* and *MOD upper* approximations of $\widetilde{A}$ by the semantic pairs

$$\underline{\mathrm{apr}}_R^{\mathrm{MOD}}(\widetilde{A}) := \big(\underline{\mu}_{\widetilde{A}}, \underline{t}_{\widetilde{A}}\big), \qquad \overline{\mathrm{apr}}_R^{\mathrm{MOD}}(\widetilde{A}) := \big(\overline{\mu}_{\widetilde{A}}, \overline{t}_{\widetilde{A}}\big),$$

where, for each $x \in U$,

$$\underline{\mu}_{\widetilde{A}}(x) := \inf_{y \in [x]_R} \mu_{\widetilde{A}}(y), \qquad \overline{\mu}_{\widetilde{A}}(x) := \sup_{y \in [x]_R} \mu_{\widetilde{A}}(y),$$

and the tag parts are aggregated by join:

$$\underline{t}_{\widetilde{A}}(x) := \bigvee_{y \in [x]_R} t_{\widetilde{A}}(y), \qquad \overline{t}_{\widetilde{A}}(x) := \bigvee_{y \in [x]_R} t_{\widetilde{A}}(y).$$

The *MOD rough set* induced by $(U, R)$ and $\widetilde{A}$ is the approximation pair

$$\mathrm{MRS}_R(\widetilde{A}) \;:=\; \Big(\underline{\mathrm{apr}}_R^{\mathrm{MOD}}(\widetilde{A}),\; \overline{\mathrm{apr}}_R^{\mathrm{MOD}}(\widetilde{A})\Big).$$

**Example 2.43.3** (MOD rough set in content moderation (grouped posts))**.** Let $U = \{p_1, p_2, p_3, p_4\}$ be four user posts to be moderated. Assume posts are *indiscernible* (for the platform's first-pass workflow) when they come from the same account in the same hour; this induces an equivalence relation $R$ with classes

$$[p_1]_R = [p_2]_R = \{p_1, p_2\}, \qquad [p_3]_R = [p_4]_R = \{p_3, p_4\}.$$

We model the MOD-described concept $\widetilde{A}$ ="potentially harmful content" by:

$$\mu_{\widetilde{A}} : U \to [0,1] \quad \text{(risk score)}, \qquad t_{\widetilde{A}} : U \to \mathsf{Tag} \quad \text{(explanatory tags)}.$$

Let the tag-domain be the powerset lattice

$$\mathsf{Tag} := \mathcal{P}(\{\text{violence}, \text{harassment}, \text{spam}, \text{scam}\}), \qquad \vee = \cup,$$

so that $\bigvee$ is set-union.

Assume the following MOD semantics are produced by an automatic screening system:

| $x$ | $p_1$ | $p_2$ | $p_3$ | $p_4$ |
|---|---|---|---|---|
| $\mu_{\widetilde{A}}(x)$ | 0.90 | 0.60 | 0.20 | 0.40 |
| $t_{\widetilde{A}}(x)$ | {violence} | {violence, harassment} | {spam} | {spam, scam} |

By Definition 2.43.2, for each $x \in U$,

$$\underline{\mu}_{\widetilde{A}}(x) = \inf_{y \in [x]_R} \mu_{\widetilde{A}}(y), \qquad \overline{\mu}_{\widetilde{A}}(x) = \sup_{y \in [x]_R} \mu_{\widetilde{A}}(y),$$

and

$$\underline{t}_{\widetilde{A}}(x) = \bigvee_{y \in [x]_R} t_{\widetilde{A}}(y) = \bigcup_{y \in [x]_R} t_{\widetilde{A}}(y), \qquad \overline{t}_{\widetilde{A}}(x) = \bigcup_{y \in [x]_R} t_{\widetilde{A}}(y).$$

**Class** $\{p_1, p_2\}$**.** For $x \in \{p_1, p_2\}$,

$$\underline{\mu}_{\widetilde{A}}(x) = \min\{0.90, 0.60\} = 0.60, \qquad \overline{\mu}_{\widetilde{A}}(x) = \max\{0.90, 0.60\} = 0.90,$$

$$\underline{t}_{\widetilde{A}}(x) = \overline{t}_{\widetilde{A}}(x) = \{\text{violence}\} \cup \{\text{violence}, \text{harassment}\} = \{\text{violence}, \text{harassment}\}.$$

**Class** $\{p_3, p_4\}$**.** For $x \in \{p_3, p_4\}$,

$$\underline{\mu}_{\widetilde{A}}(x) = \min\{0.20, 0.40\} = 0.20, \qquad \overline{\mu}_{\widetilde{A}}(x) = \max\{0.20, 0.40\} = 0.40,$$

$$\underline{t}_{\widetilde{A}}(x) = \overline{t}_{\widetilde{A}}(x) = \{\text{spam}\} \cup \{\text{spam}, \text{scam}\} = \{\text{spam}, \text{scam}\}.$$

Hence the MOD lower and MOD upper approximations are the semantic pairs

$$\underline{\mathrm{apr}}_R^{\mathrm{MOD}}(\widetilde{A}) = (\underline{\mu}_{\widetilde{A}}, \underline{t}_{\widetilde{A}}), \qquad \overline{\mathrm{apr}}_R^{\mathrm{MOD}}(\widetilde{A}) = (\overline{\mu}_{\widetilde{A}}, \overline{t}_{\widetilde{A}}),$$

and the induced MOD rough set is

$$\mathrm{MRS}_R(\widetilde{A}) = \Big(\underline{\mathrm{apr}}_R^{\mathrm{MOD}}(\widetilde{A}),\ \overline{\mathrm{apr}}_R^{\mathrm{MOD}}(\widetilde{A})\Big).$$

*Interpretation.* Within each indiscernibility class, $\underline{\mu}$ gives a conservative risk score (worst case), $\overline{\mu}$ gives a permissive score (best case), and the tag-join aggregates all plausible moderation reasons observed in that class.

**Theorem 2.43.4** (Well-definedness of MOD rough approximations)**.** *In Definition 2.43.2, the mappings* $\underline{\mu}_{\widetilde{A}}, \overline{\mu}_{\widetilde{A}} : U \to [0,1]$ *and* $\underline{t}_{\widetilde{A}}, \overline{t}_{\widetilde{A}} : U \to \mathsf{Tag}$ *are well-defined. Moreover, they are constant on R-equivalence classes; that is, if* $(x,x') \in R$*, then*

$$\underline{\mu}_{\widetilde{A}}(x) = \underline{\mu}_{\widetilde{A}}(x'), \quad \overline{\mu}_{\widetilde{A}}(x) = \overline{\mu}_{\widetilde{A}}(x'), \quad \underline{t}_{\widetilde{A}}(x) = \underline{t}_{\widetilde{A}}(x'), \quad \overline{t}_{\widetilde{A}}(x) = \overline{t}_{\widetilde{A}}(x').$$

*Proof.* Fix $x \in U$. Since $[x]_R \subseteq U$ is a set and $\mu_{\widetilde{A}}$ maps into $[0,1]$, the set $\{\mu_{\widetilde{A}}(y) \mid y \in [x]_R\} \subseteq [0,1]$ admits inf and sup (because $[0,1]$ is complete). Hence $\underline{\mu}_{\widetilde{A}}(x)$ and $\overline{\mu}_{\widetilde{A}}(x)$ are well-defined real numbers in $[0,1]$.

Likewise, $\{t_{\widetilde{A}}(y) \mid y \in [x]_R\} \subseteq \mathsf{Tag}$ is a subset of a finite totally ordered set, so its supremum (equivalently, its maximum) exists and is unique. Thus $\underline{t}_{\widetilde{A}}(x)$ and $\overline{t}_{\widetilde{A}}(x)$ are well-defined.

Now assume $(x,x') \in R$. Since $R$ is an equivalence relation, we have $[x]_R = [x']_R$. Therefore the sets of values being aggregated coincide:

$$\{\mu_{\widetilde{A}}(y) \mid y \in [x]_R\} = \{\mu_{\widetilde{A}}(y) \mid y \in [x']_R\}, \qquad \{t_{\widetilde{A}}(y) \mid y \in [x]_R\} = \{t_{\widetilde{A}}(y) \mid y \in [x']_R\}.$$

Taking inf, sup, and $\bigvee$ on both sides yields the desired equalities. Hence the MOD rough approximations are constant on granules and, in particular, independent of the chosen representative of an equivalence class. This proves well-definedness. □

## 2.44 Topological Rough Sets

Topological rough sets approximate subsets by interior and closure in a topology, modeling definite and possible membership under open-neighborhood uncertainty.

**Definition 2.44.1** (Topological rough set (interior–closure approximation))**.** Let $(X,\tau)$ be a topological space, and let $Y \subseteq X$. Define the *topological lower* and *topological upper* approximations of $Y$ by

$$\underline{Y}^{\tau} := \mathrm{int}_{\tau}(Y), \qquad \overline{Y}^{\tau} := \mathrm{cl}_{\tau}(Y),$$

where $\mathrm{int}_{\tau}(Y)$ and $\mathrm{cl}_{\tau}(Y)$ denote the interior and closure of $Y$ in $(X,\tau)$, respectively. Then

$$\underline{Y}^{\tau} \subseteq Y \subseteq \overline{Y}^{\tau}.$$

The pair $\left(\underline{Y}^{\tau}, \overline{Y}^{\tau}\right)$ is called the *topological rough approximation* of $Y$, and $Y$ is said to be *topologically rough* if $\underline{Y}^{\tau} \neq \overline{Y}^{\tau}$.

**Example 2.44.2** (GPS geofencing with uncertain location: a topological rough set)**.** A delivery app classifies whether a courier is *inside* a service zone $Y$ (a downtown area), but GPS is noisy and the system can only resolve location up to *cell-tower regions*.

Let $X$ be a finite set of tower-cells covering the city, and take $\tau$ to be the topology generated by these cells (so an open set is any union of cells). Identify each cell with its covered area.

Let $Y \subseteq X$ be the set of points (or fine-grained locations) that truly lie in the downtown zone. Because the app observes only at the cell level, it can confidently say a location is in the zone only when its entire observed cell lies in $Y$. Hence the *definitely-in-zone* set is

$$\underline{Y}^{\tau} = \operatorname{int}_{\tau}(Y),$$

the union of all observed cells fully contained in the service zone.

Conversely, any observed cell that *intersects* $Y$ might contain a downtown point, so the app must treat all such cells as *possibly in the zone.* This yields the *possibly-in-zone* set

$$\overline{Y}^{\tau} = \operatorname{cl}_{\tau}(Y),$$

the smallest cell-union containing $Y$. If boundary cells intersect $Y$ but are not contained in $Y$, then $\underline{Y}^{\tau} \neq \overline{Y}^{\tau}$, so $Y$ is topologically rough under cell-level uncertainty.

## 2.45 Preorder Rough Sets

Preorder rough sets approximate subsets using preorder up-sets and down-sets, yielding increasing/decreasing lower and upper regions for ordered uncertainty.

**Definition 2.45.1** (Preorder rough set (increasing/decreasing approximations)). Let $(X, \preceq)$ be a preordered set, i.e., $\preceq$ is reflexive and transitive on $X$. For each $x \in X$, define the principal up-set and down-set

$$N_{\uparrow}(x) := \{ y \in X \mid x \preceq y \}, \qquad N_{\downarrow}(x) := \{ y \in X \mid y \preceq x \}.$$

For $Y \subseteq X$, the *increasing* (upper-set) rough approximations are

$$\underline{Y}^{\uparrow} := \{ x \in X \mid N_{\uparrow}(x) \subseteq Y \}, \qquad \overline{Y}^{\uparrow} := \{ x \in X \mid N_{\uparrow}(x) \cap Y \neq \emptyset \},$$

and the *decreasing* (lower-set) rough approximations are

$$\underline{Y}^{\downarrow} := \{ x \in X \mid N_{\downarrow}(x) \subseteq Y \}, \qquad \overline{Y}^{\downarrow} := \{ x \in X \mid N_{\downarrow}(x) \cap Y \neq \emptyset \}.$$

**Example 2.45.2** (Triage severity as a preorder: ICU-need under ordered uncertainty). Let

$$X = \{1, 2, 3, 4, 5\}$$

be emergency triage levels, where 1 is mild and 5 is critical. Use the natural preorder $\preceq$ given by the usual order $\leq$:

$$x \preceq y \iff x \leq y.$$

Hence the principal up-set and down-set are

$$N_{\uparrow}(x) = \{x, x+1, \ldots, 5\}, \qquad N_{\downarrow}(x) = \{1, 2, \ldots, x\}.$$

Let

$$Y = \{4, 5\}$$

represent the set of severity levels that *require ICU-level care.*

**Increasing (upper-set) approximations.**

$$\underline{Y}^{\uparrow} = \{x \in X \mid N_{\uparrow}(x) \subseteq Y\} = \{4, 5\},$$
$$\overline{Y}^{\uparrow} = \{x \in X \mid N_{\uparrow}(x) \cap Y \neq \emptyset\} = \{3, 4, 5\}.$$

Interpretation: levels 4 and 5 are *definitely* ICU-requiring because any escalation remains in $Y$, while level 3 is *possibly* ICU-requiring because escalation to 4 or 5 intersects $Y$.

**Decreasing (lower-set) approximations (optional reading).**

$$\underline{Y}^{\downarrow} = \{x \in X \mid N_{\downarrow}(x) \subseteq Y\} = \emptyset, \qquad \overline{Y}^{\downarrow} = \{x \in X \mid N_{\downarrow}(x) \cap Y \neq \emptyset\} = \{4, 5\}.$$

This reflects that ICU-need is naturally *upward-oriented* in the severity order.

## 2.46 Directed Rough Set

Directed rough sets approximate subsets using granules induced by a directed relation, capturing asymmetric reachability/preference, yielding lower and upper approximations [227, 228].

**Definition 2.46.1** (Up-directed approximation space)**.** [227,228] A *general approximation space* is a pair $(U, R)$, where $U \neq \emptyset$ and $R \subseteq U \times U$ is a binary relation. The relation $R$ is *up-directed* if

$$(\forall a, b \in U)\ (\exists c \in U)\ \big(aRc\ \wedge\ bRc\big).$$

If $(U, R)$ satisfies this condition, we call it an *up-directed approximation space.*

**Definition 2.46.2** (Neighborhoods)**.** Let $(U, R)$ be a general approximation space. For each $a \in U$, define

$$[a]\ :=\ \{x \in U : xRa\}, \qquad [a]^i\ :=\ \{x \in U : aRx\}, \qquad [a]^o\ :=\ \{x \in U : aRx\ \wedge\ xRa\}.$$

(These are the *neighborhood*, *inverse-neighborhood*, and *symmetric neighborhood* of $a$.)

**Definition 2.46.3** (Directed rough approximations and directed rough set)**.** [227, 228] Let $(U, R)$ be an up-directed approximation space and let $A \subseteq U$. Define the *directed lower* and *directed upper* approximations of $A$ by

$$A^{\ell}\ :=\ \bigcup\{\,[a]\ :\ a \in U,\ [a] \subseteq A\,\}, \qquad A^{u}\ :=\ \bigcup\{\,[a]\ :\ a \in U,\ [a] \cap A \neq \emptyset\,\}.$$

The ordered pair

$$\mathsf{RS}_R(A)\ :=\ (A^{\ell}, A^{u})$$

is called the *directed rough set* of $A$ (with respect to $R$). Its *directed boundary* is

$$\mathrm{Bn}_R(A)\ :=\ A^{u} \setminus A^{\ell}.$$

We say that $A$ is *(directly) definable* if $A^{\ell} = A^{u}$, and *directly rough* otherwise.

**Example 2.46.4** (Directed rough set for concept mastery in an adaptive math tutor)**.** Let

$$U := \{F, R, P, C\}$$

be a set of *concepts*, where $F$ = Fractions, $R$ = Ratios, $P$ = Percentages, and $C$ = Conversions. We interpret $xRy$ as: "concept $y$ can serve as an immediate covering/aggregation concept for $x$" in a local concept-organization ontology.

Define an *up-directed* relation $R \subseteq U \times U$ by specifying the outgoing sets

$$\mathrm{Out}(x) := \{\, y \in U \mid xRy \,\} :$$

$$\mathrm{Out}(F) = \{F, R, P\},\ \mathrm{Out}(R) = \{R, P, C\},\ \mathrm{Out}(P) = \{F, P, C\},\ \mathrm{Out}(C) = \{F, R, C\}.$$

Then $\mathrm{Out}(x) \cap \mathrm{Out}(y) \neq \emptyset$ for all $x, y \in U$; hence $(U, R)$ is up-directed.

For each $a \in U$, the directed granule (neighborhood) is

$$[a] := \{\, x \in U \mid xRa \,\}.$$

From the definition of $R$,

$$[F] = \{F, P, C\}, \qquad [R] = \{F, R, C\}, \qquad [P] = \{F, R, P\}, \qquad [C] = \{R, P, C\}.$$

Suppose a short quiz confirms competence in fractions, ratios, and percentages, so the target set is

$$A := \{F, R, P\} \subseteq U.$$

The *directed lower* and *directed upper* approximations (Definition 2.46.3) are:

$$A^{\ell} = \bigcup \{\, [a] \mid a \in U,\ [a] \subseteq A \,\} = [P] = \{F, R, P\},$$

because $[P] \subseteq A$ but $[F], [R], [C] \not\subseteq A$. Moreover,

$$A^{u} = \bigcup \{\, [a] \mid a \in U,\ [a] \cap A \neq \emptyset \,\}$$

$$= [F] \cup [R] \cup [P] \cup [C] = U.$$

Therefore the directed rough set of $A$ is

$$\mathsf{RS}_R(A) = (A^{\ell}, A^{u}) = (\{F, R, P\}, \{F, R, P, C\}),$$

and the directed boundary is

$$\mathrm{Bn}_R(A) = A^{u} \setminus A^{\ell} = \{C\}.$$

Interpretation: the tutor can *definitely* attribute mastery to $\{F, R, P\}$, while $C$ remains *possibly* implicated because conversion skills are linked to (and hence "pulled in" by) the directed granules.

## 2.47 Strait Rough Set

A strait rough set approximates a target set by unions of partition blocks fully included in, or intersecting, it [229].

**Definition 2.47.1** (Strait rough set)**.** [229] Let $V$ be a nonempty universe. Let $E$ be a nonempty parameter set and let

$$F : E \longrightarrow \mathcal{P}(V)$$

be a mapping such that the family $\{F(e) \mid e \in E\}$ is a partition of $V$ (i.e., $F(e) \neq \varnothing$ for all $e$, $\bigcup_{e\in E} F(e) = V$, and $F(e) \cap F(e') = \varnothing$ for $e \neq e'$). For any target set $X \subseteq V$, define:

**(1) Strait lower approximation.**

$$\underline{\mathrm{apr}}_F(X) \;:=\; \bigcup_{\substack{e\in E\\ F(e)\subseteq X}} F(e).$$

**(2) Strait upper approximation.**

$$\overline{\mathrm{apr}}_F(X) \;:=\; \bigcup_{\substack{e\in E\\ F(e)\cap X\neq\varnothing}} F(e).$$

**(3) Strait boundary region.**

$$\mathrm{Bn}_F(X) \;:=\; \overline{\mathrm{apr}}_F(X) \setminus \underline{\mathrm{apr}}_F(X).$$

The ordered pair

$$\mathsf{SR}_F(X) \;:=\; \big(\underline{\mathrm{apr}}_F(X), \overline{\mathrm{apr}}_F(X)\big)$$

is called the *strait rough set* of $X$ (with respect to $F$). We say that $X$ is *strait definable* if $\underline{\mathrm{apr}}_F(X) = \overline{\mathrm{apr}}_F(X)$, and *strait rough* otherwise.

**Example 2.47.2** (Real-life example: delivery coverage under district partition)**.** Let $V$ be the set of households in a city. Let $E$ be the set of administrative districts (wards), and for each $e \in E$ let

$$F(e) \subseteq V$$

be the set of households located in district $e$. Then $\{F(e) \mid e \in E\}$ forms a partition of $V$ (every household lies in exactly one district).

Suppose a grocery company plans to offer *same-day delivery.* Let

$$X \subseteq V$$

be the set of households that are *actually within reach* of the service, based on distance-to-hub, traffic, and staffing (so $X$ is not known perfectly at the district level).

The company's policy is district-based: it can only *announce* coverage by whole districts. Hence the strait approximations induced by $F$ are:

$$\underline{\mathrm{apr}}_F(X) = \bigcup_{\substack{e\in E\\ F(e)\subseteq X}} F(e), \qquad \overline{\mathrm{apr}}_F(X) = \bigcup_{\substack{e\in E\\ F(e)\cap X\neq\varnothing}} F(e).$$

Here, $\underline{\mathrm{apr}}_F(X)$ is the set of households in districts *fully* reachable (safe to promise), while $\overline{\mathrm{apr}}_F(X)$ is the set of households in districts that are *partly* reachable (possibly reachable). The boundary

$$\mathrm{Bn}_F(X) = \overline{\mathrm{apr}}_F(X) \setminus \underline{\mathrm{apr}}_F(X)$$

consists of households in *mixed* districts, where some addresses are reachable and others are not; these are the districts for which the company must either refine the partition or accept uncertainty.

## 2.48 Dialectical rough set

A dialectical rough set uses paraconsistent dialectical opposition to generate evolving lower/upper approximations, tolerating contradictions among granules and objects explicitly [230].

**Definition 2.48.1** (Dialectical rough set)**.** [230] Let $X = (S, R)$ be a Pawlak approximation space, where $S$ is a nonempty finite universe and $R$ is an equivalence relation on $S$. For $A \subseteq S$, define the Pawlak lower and upper approximations

$$A^\ell := \{\, s \in S \mid [s]_R \subseteq A \,\}, \qquad A^u := \{\, s \in S \mid [s]_R \cap A \neq \varnothing \,\},$$

where $[s]_R := \{t \in S \mid (s, t) \in R\}$.

Define the *rough equality* $\approx$ on $\mathcal{P}(S)$ by

$$A \approx B \quad\Longleftrightarrow\quad A^\ell = B^\ell \ \text{ and } \ A^u = B^u,$$

and let $\mathcal{P}(S) \mid\approx := \mathcal{P}(S)/\approx$ denote the set of *rough objects* (equivalence classes under $\approx$). Consider a *concrete enriched pre-rough algebra (CERA)*

$$W(X) = \big\langle\, \mathcal{P}(S) \ \cup\ (\mathcal{P}(S) \mid\approx),\ \oplus,\ 0,\ \ldots \big\rangle,$$

where $\tau_1$ (resp. $\tau_2$) is the *type predicate* selecting the classical part $\mathcal{P}(S)$ (resp. the rough part $\mathcal{P}(S) \mid\approx$), and where the term $0 \oplus x$ (with $\tau_1(x)$) represents the canonical embedding of a classical object into the rough semantic domain.

The *dialectical universe* is defined by

$$K := \{(x, 0 \oplus x) \mid \tau_1(x)\} \ \cup\ \{(b, x) \mid \tau_2(b) \ \wedge\ \tau_1(x) \ \wedge\ x \oplus 0 = b\}.$$

A *dialectical rough set* is any element of $K$.

**Example 2.48.2** (Real-life example: spam filtering with *dialectical* human–model interaction)**.** Let $S$ be a finite set of recent email messages in an inbox. Define an equivalence relation $R$ on $S$ by

$$m \ R \ m' \quad\Longleftrightarrow\quad m \text{ and } m' \text{ share the same } \textit{feature signature}$$

(sender-domain class, keyword bucket, and link-pattern type).

Thus each $R$-class $[m]_R$ is an information granule: messages that look indistinguishable at the chosen feature resolution.

Let $A \subseteq S$ be the (unknown) set of truly spam messages. Given limited features, the system forms Pawlak approximations

$$A^{\ell} = \{\, m \in S \mid [m]_R \subseteq A \,\}, \qquad A^{u} = \{\, m \in S \mid [m]_R \cap A \neq \varnothing \,\}.$$

Hence $A^{\ell}$ are messages that are *certainly spam* at this granularity, while $A^{u} \setminus A^{\ell}$ are *borderline* messages whose $R$-classes contain both spam and non-spam.

Now a user provides feedback ("spam" / "not spam") on a subset of messages, producing a *classical* labeled set $x \in \mathcal{P}(S)$ (e.g., $x$ is the set the user marked as spam today). The system simultaneously maintains the corresponding *rough object*

$$b = [x]_{\approx} \in \mathcal{P}(S) \mid\approx,$$

where $\approx$ is the rough-equality relation

$$x \approx y \iff x^{\ell} = y^{\ell} \text{ and } x^{u} = y^{u}.$$

Intuitively, $b$ represents the model's semantic view of "spam today" at the granularity induced by $R$, where different user-labeled sets that induce the same $(\ell, u)$ pair are identified.

In the CERA viewpoint, the dialectical universe

$$K = \{(x, 0 \oplus x) \mid \tau_1(x)\} \;\cup\; \{(b, x) \mid \tau_2(b) \wedge \tau_1(x) \wedge x \oplus 0 = b\}$$

encodes the two-way coupling between the user's concrete labeling $x$ and the system's rough semantic state $b$. A typical dialectical rough set instance is the pair

$$(x, 0 \oplus x) \in K,$$

meaning: "the user's explicit spam set $x$ together with its embedded rough interpretation". Operationally, this supports iterative resolution of contradictions (user corrections) while keeping a stable rough semantic layer that is invariant under $\approx$-equivalent relabelings.

## 2.49 Sheaf Rough Set

Sheaf Rough Set models uncertainty on sheaf sections: two sections are related when they agree locally; neighborhood-based lower/upper approximations reflect gluing-consistent information robustly across contexts.

**Definition 2.49.1** (Sheaf Rough Set)**.** Let $(X, \tau)$ be a topological space and let $\mathcal{F}$ be a (set-valued) sheaf on $X$. Consider the universe of *local sections*

$$\mathsf{Sec}(\mathcal{F}) := \Big\{ (U, s) \;\Big|\; U \in \tau,\ U \neq \varnothing,\ s \in \mathcal{F}(U) \Big\}.$$

Define a binary relation $\mathcal{R}_{\mathrm{sh}}$ on $\mathsf{Sec}(\mathcal{F})$ by

$$(U, s)\ \mathcal{R}_{\mathrm{sh}}\ (V, t) \quad :\Longleftrightarrow \quad \exists\, W \in \tau \text{ with } \varnothing \neq W \subseteq U \cap V \text{ such that } s|_W = t|_W.$$

For any $\mathcal{A} \subseteq \mathsf{Sec}(\mathcal{F})$, the *sheaf lower* and *sheaf upper* approximations of $\mathcal{A}$ are defined (via neighborhoods) by

$$\underline{\mathcal{A}}^{\mathrm{sh}} := \Big\{ x \in \mathsf{Sec}(\mathcal{F}) \;\Big|\; N_{\mathrm{sh}}(x) \subseteq \mathcal{A} \Big\}, \qquad \overline{\mathcal{A}}^{\mathrm{sh}} := \Big\{ x \in \mathsf{Sec}(\mathcal{F}) \;\Big|\; N_{\mathrm{sh}}(x) \cap \mathcal{A} \neq \varnothing \Big\},$$

where the *sheaf neighborhood* of $x$ is

$$N_{\mathrm{sh}}(x) := \{ y \in \mathsf{Sec}(\mathcal{F}) \mid x \; \mathcal{R}_{\mathrm{sh}} \; y \}.$$

The pair $(\underline{\mathcal{A}}^{\mathrm{sh}}, \overline{\mathcal{A}}^{\mathrm{sh}})$ is called the *Sheaf Rough Set* determined by $\mathcal{A}$ (with respect to $\mathcal{F}$).

**Remark 2.49.2.** The relation $\mathcal{R}_{\mathrm{sh}}$ is reflexive and symmetric (a tolerance relation), but in general need not be transitive. Hence the above is naturally interpreted as a *tolerance-based* rough set induced by sheaf restrictions.

**Proposition 2.49.3** (Well-definedness of the Sheaf Rough Set)**.** *In Definition 2.49.1, the relation $\mathcal{R}_{\mathrm{sh}}$ on $\mathsf{Sec}(\mathcal{F})$ is well-defined and is a tolerance relation (reflexive and symmetric). Moreover, for every $\mathcal{A} \subseteq \mathsf{Sec}(\mathcal{F})$, the sets $\underline{\mathcal{A}}^{\mathrm{sh}}$ and $\overline{\mathcal{A}}^{\mathrm{sh}}$ are well-defined subsets of $\mathsf{Sec}(\mathcal{F})$ and satisfy*

$$\underline{\mathcal{A}}^{\mathrm{sh}} \subseteq \mathcal{A} \subseteq \overline{\mathcal{A}}^{\mathrm{sh}}.$$

*Proof. (Well-definedness of $\mathcal{R}_{\mathrm{sh}}$).* If $(U, s), (V, t) \in \mathsf{Sec}(\mathcal{F})$ and $W \subseteq U \cap V$ is a nonempty open set, then the restrictions $s|_W$ and $t|_W$ are defined by the restriction maps of the sheaf $\mathcal{F}$ (indeed, a presheaf already suffices for restriction). Hence the statement "$s|_W = t|_W$" is unambiguous, and therefore $\mathcal{R}_{\mathrm{sh}}$ is well-defined.

*(Reflexive).* For any $(U, s) \in \mathsf{Sec}(\mathcal{F})$, choose $W := U$; then $\varnothing \neq W \subseteq U \cap U$ and $s|_U = s|_U$. Thus $(U, s) \, \mathcal{R}_{\mathrm{sh}} \, (U, s)$.

*(Symmetric).* If $(U, s) \, \mathcal{R}_{\mathrm{sh}} \, (V, t)$, then for some nonempty open $W \subseteq U \cap V$ we have $s|_W = t|_W$, which implies $t|_W = s|_W$; hence $(V, t) \, \mathcal{R}_{\mathrm{sh}} \, (U, s)$.

*(Well-definedness of neighborhoods and approximations).* For each $x \in \mathsf{Sec}(\mathcal{F})$, the neighborhood $N_{\mathrm{sh}}(x) := \{ y \in \mathsf{Sec}(\mathcal{F}) \mid x \, \mathcal{R}_{\mathrm{sh}} \, y \}$ is a subset of $\mathsf{Sec}(\mathcal{F})$ determined uniquely by $\mathcal{R}_{\mathrm{sh}}$, hence is well-defined. Therefore $\underline{\mathcal{A}}^{\mathrm{sh}}$ and $\overline{\mathcal{A}}^{\mathrm{sh}}$ are well-defined by set comprehension and are subsets of $\mathsf{Sec}(\mathcal{F})$.

*(Inclusions).* If $x \in \underline{\mathcal{A}}^{\mathrm{sh}}$, then $N_{\mathrm{sh}}(x) \subseteq \mathcal{A}$. By reflexivity, $x \in N_{\mathrm{sh}}(x)$, hence $x \in \mathcal{A}$. Thus $\underline{\mathcal{A}}^{\mathrm{sh}} \subseteq \mathcal{A}$. If $x \in \mathcal{A}$, then again by reflexivity $x \in N_{\mathrm{sh}}(x)$, so $N_{\mathrm{sh}}(x) \cap \mathcal{A} \neq \varnothing$, i.e. $x \in \overline{\mathcal{A}}^{\mathrm{sh}}$. Hence $\mathcal{A} \subseteq \overline{\mathcal{A}}^{\mathrm{sh}}$. □

**Example 2.49.4** (Traffic status consistency across overlapping road regions)**.** Let the *road region* be the finite topological space

$$X = \{N, C, S\},$$

interpreted as *North*, *Center*, *South*, equipped with the topology

$$\tau = \{\varnothing, \{C\}, \{N,C\}, \{C,S\}, \{N,C,S\}\}.$$

Consider the (constant) set-valued sheaf $\mathcal{F}$ with fiber

$$\mathsf{T} = \{\mathsf{G}, \mathsf{S}, \mathsf{R}\},$$

where $\mathsf{G}$ means *Green/Free flow*, $\mathsf{S}$ means *Slow*, and $\mathsf{R}$ means *Red/Stop*. For each nonempty open set $U \in \tau \setminus \{\varnothing\}$, define

$$\mathcal{F}(U) = \mathsf{T},$$

and for inclusions $W \subseteq U$ take restriction maps to be the identity (so the label does not change when restricting).

Then

$$\mathsf{Sec}(\mathcal{F}) = \{(U,\ell) \mid U \in \tau \setminus \{\varnothing\},\ \ell \in \mathsf{T}\}.$$

By Definition 2.49.1, two local reports $(U,\ell)$ and $(V,\ell')$ are related iff they *agree on some nonempty overlap*, which here reduces to:

$$(U,\ell)\ \mathcal{R}_{\mathrm{sh}}\ (V,\ell') \iff \big(U \cap V \text{ contains a nonempty open set}\big) \text{ and } \ell = \ell'.$$

**Scenario.** Suppose two cameras report *slow traffic* on the two corridor segments $\{N,C\}$ and $\{C,S\}$, but the central sensor $\{C\}$ has not yet reported. Let the concept (set of "flagged" local sections) be

$$\mathcal{A} := \big\{(\{N,C\},\mathsf{S}),\ (\{C,S\},\mathsf{S})\big\} \subseteq \mathsf{Sec}(\mathcal{F}).$$

**Neighborhood computation.** For $x_1 = (\{N,C\},\mathsf{S})$, the $\mathcal{R}_{\mathrm{sh}}$-neighbors are exactly the slow reports on open sets that overlap $\{N,C\}$ in a nonempty open set:

$$N_{\mathrm{sh}}(x_1) = \big\{(\{C\},\mathsf{S}),\ (\{N,C\},\mathsf{S}),\ (\{C,S\},\mathsf{S}),\ (X,\mathsf{S})\big\}.$$

Similarly, for $x_2 = (\{C,S\},\mathsf{S})$,

$$N_{\mathrm{sh}}(x_2) = \big\{(\{C\},\mathsf{S}),\ (\{N,C\},\mathsf{S}),\ (\{C,S\},\mathsf{S}),\ (X,\mathsf{S})\big\}.$$

**Approximations.** Since $N_{\mathrm{sh}}(x_i) \nsubseteq \mathcal{A}$ (because $(\{C\},\mathsf{S})$ and $(X,\mathsf{S})$ are not in $\mathcal{A}$), neither $x_1$ nor $x_2$ belongs to the lower approximation; thus

$$\underline{\mathcal{A}}^{\mathrm{sh}} = \varnothing.$$

On the other hand, a section $(U,\mathsf{S})$ lies in the upper approximation iff it overlaps one of the corridor reports and has the same label $\mathsf{S}$. Hence

$$\overline{\mathcal{A}}^{\mathrm{sh}} = \big\{(\{C\},\mathsf{S}),\ (\{N,C\},\mathsf{S}),\ (\{C,S\},\mathsf{S}),\ (X,\mathsf{S})\big\}.$$

Interpretation: the upper region identifies all locally consistent "slow" reports that can be glued (via overlaps) to the observed corridor slowdowns, whereas the lower region is empty because the missing central report prevents robust certainty.

## 2.50 Simplicial Rough Set

Simplicial Rough Set uses a simplicial complex; vertices are indiscernible when they share identical facet-incidence signatures; Pawlak-style lower/upper approximations capture higher-order interaction groups beyond graphs.

**Definition 2.50.1** (Simplicial Rough Set)**.** Let $K$ be a finite simplicial complex with vertex set $V$, and let $\mathcal{M}(K)$ denote the set of maximal simplices (facets) of $K$. For each vertex $v \in V$, define its *facet-incidence signature* by

$$\Sigma_K(v) := \{\, M \in \mathcal{M}(K) \mid v \in M \,\} \subseteq \mathcal{M}(K).$$

Define an equivalence relation $\sim_K$ on $V$ by

$$v \sim_K w \quad :\iff \quad \Sigma_K(v) = \Sigma_K(w).$$

For any $X \subseteq V$, define the *simplicial lower* and *simplicial upper* approximations by

$$\underline{X}^K := \{v \in V \mid [v]_{\sim_K} \subseteq X\}, \qquad \overline{X}^K := \{v \in V \mid [v]_{\sim_K} \cap X \neq \varnothing\},$$

where $[v]_{\sim_K}$ denotes the $\sim_K$-equivalence class of $v$. Then $\big(\underline{X}^K, \overline{X}^K\big)$ is called the *Simplicial Rough Set* of $X$ with respect to $K$.

**Remark 2.50.2.** This construction uses higher-order incidence (via facets), so it can distinguish vertices not only by pairwise adjacency (graph structure) but by membership in higher-dimensional interaction groups.

**Proposition 2.50.3** (Well-definedness of the Simplicial Rough Set)**.** *In Definition 2.50.1, the relation $\sim_K$ on $V$ is a well-defined equivalence relation. Consequently, for every $X \subseteq V$, the sets $\underline{X}^K$ and $\overline{X}^K$ are well-defined subsets of $V$ and satisfy*

$$\underline{X}^K \subseteq X \subseteq \overline{X}^K.$$

*Proof. (Well-definedness).* Since $K$ is a finite simplicial complex, the set of facets $\mathcal{M}(K)$ exists. For each $v \in V$, the set $\Sigma_K(v) = \{M \in \mathcal{M}(K) \mid v \in M\}$ is uniquely determined; thus $\Sigma_K(v)$ is well-defined and so is the relation $v \sim_K w \iff \Sigma_K(v) = \Sigma_K(w)$.

*(Equivalence).* Reflexivity and symmetry are immediate from equality. If $v \sim_K w$ and $w \sim_K u$, then $\Sigma_K(v) = \Sigma_K(w) = \Sigma_K(u)$, hence $v \sim_K u$ (transitivity). Thus $\sim_K$ is an equivalence relation.

*(Well-definedness of approximations).* For each $v \in V$, the equivalence class $[v]_{\sim_K}$ is well-defined. Therefore $\underline{X}^K$ and $\overline{X}^K$ are well-defined subsets of $V$.

*(Inclusions).* If $v \in \underline{X}^K$ then $[v]_{\sim_K} \subseteq X$, and since $v \in [v]_{\sim_K}$ we get $v \in X$, so $\underline{X}^K \subseteq X$. If $v \in X$, then $[v]_{\sim_K} \cap X \neq \varnothing$ because $v \in [v]_{\sim_K} \cap X$, hence $v \in \overline{X}^K$, so $X \subseteq \overline{X}^K$. □

**Example 2.50.4** (Team structure in a company and indistinguishability by shared projects)**.** Let $V = \{\mathsf{A}, \mathsf{B}, \mathsf{C}, \mathsf{D}, \mathsf{E}, \mathsf{F}\}$ be employees. Assume the maximal cross-functional projects (facets) are

$$M_1 = \{\mathsf{A}, \mathsf{B}, \mathsf{C}\}, \quad M_2 = \{\mathsf{C}, \mathsf{D}, \mathsf{E}\}, \quad M_3 = \{\mathsf{A}, \mathsf{B}, \mathsf{F}\}.$$

Let $K$ be the simplicial complex generated by these facets (i.e., it contains all subsets of each $M_i$).

For each employee $v \in V$, the facet-incidence signature is

$$\Sigma_K(\mathsf{A}) = \{M_1, M_3\},\ \Sigma_K(\mathsf{B}) = \{M_1, M_3\},\ \Sigma_K(\mathsf{C}) = \{M_1, M_2\},$$
$$\Sigma_K(\mathsf{D}) = \{M_2\},\ \Sigma_K(\mathsf{E}) = \{M_2\},\ \Sigma_K(\mathsf{F}) = \{M_3\}.$$

Thus the $\sim_K$-equivalence classes are

$$[\mathsf{A}]_{\sim_K} = [\mathsf{B}]_{\sim_K} = \{\mathsf{A}, \mathsf{B}\}, \quad [\mathsf{D}]_{\sim_K} = [\mathsf{E}]_{\sim_K} = \{\mathsf{D}, \mathsf{E}\},$$
$$[\mathsf{C}]_{\sim_K} = \{\mathsf{C}\}, \quad [\mathsf{F}]_{\sim_K} = \{\mathsf{F}\}.$$

**Scenario.** Let $X = \{\mathsf{A}, \mathsf{C}, \mathsf{D}\}$ be the set of employees currently assigned to a security audit task.

**Approximations.** By Definition 2.50.1,

$$\underline{X}^K = \{v \in V \mid [v]_{\sim_K} \subseteq X\} = \{\mathsf{C}\},$$

because $[\mathsf{C}]_{\sim_K} = \{\mathsf{C}\} \subseteq X$, while $[\mathsf{A}]_{\sim_K} = \{\mathsf{A}, \mathsf{B}\} \not\subseteq X$ and $[\mathsf{D}]_{\sim_K} = \{\mathsf{D}, \mathsf{E}\} \not\subseteq X$. Moreover,

$$\overline{X}^K = \{v \in V \mid [v]_{\sim_K} \cap X \neq \varnothing\} = \{\mathsf{A}, \mathsf{B}, \mathsf{C}, \mathsf{D}, \mathsf{E}\}.$$

Interpretation: $\underline{X}^K$ contains those certainly in $X$ under the "same-project-signature" indiscernibility, while $\overline{X}^K$ includes employees indistinguishable from at least one audited member (e.g., $\mathsf{B}$ from $\mathsf{A}$ and $\mathsf{E}$ from $\mathsf{D}$).

## 2.51 Persistent Rough Set

Persistent Rough Set tracks lower and upper approximations across a monotone scale parameter, revealing how certainty and possibility regions evolve with changing neighborhoods or thresholds.

**Definition 2.51.1** (Persistent Rough Set)**.** Let $U$ be a nonempty finite universe, and let $\{R_\varepsilon\}_{\varepsilon \in \mathbb{R}_{\geq 0}}$ be a family of relations on $U$ that is *monotone in scale*:

$$0 \leq \varepsilon_1 \leq \varepsilon_2 \implies R_{\varepsilon_1} \subseteq R_{\varepsilon_2}.$$

For each $\varepsilon \geq 0$ and each $x \in U$, define the $\varepsilon$-neighborhood

$$N_\varepsilon(x) := \{y \in U \mid x\, R_\varepsilon\, y\}.$$

Given a target set $X \subseteq U$, define the $\varepsilon$-*lower* and $\varepsilon$-*upper* approximations by

$$\underline{X}_\varepsilon := \{x \in U \mid N_\varepsilon(x) \subseteq X\}, \qquad \overline{X}_\varepsilon := \{x \in U \mid N_\varepsilon(x) \cap X \neq \varnothing\}.$$

The scale-indexed family

$$\mathsf{PR}(X) := \{(\underline{X}_\varepsilon, \overline{X}_\varepsilon)\}_{\varepsilon \in \mathbb{R}_{\geq 0}}$$

is called the *Persistent Rough Set* of $X$ (with respect to $\{R_\varepsilon\}$).

**Remark 2.51.2.** When $R_\varepsilon$ is induced by a metric (e.g., $xR_\varepsilon y \iff d(x,y) \leq \varepsilon$), $\mathsf{PR}(X)$ captures how certainty/possibility regions vary with resolution (noise tolerance).

**Proposition 2.51.3** (Well-definedness of the Persistent Rough Set)**.** *In Definition 2.51.1, for every $\varepsilon \geq 0$ and every $X \subseteq U$, the neighborhood $N_\varepsilon(x)$ and the sets $\underline{X}_\varepsilon$, $\overline{X}_\varepsilon$ are well-defined. Moreover, for each $\varepsilon \geq 0$,*

$$\underline{X}_\varepsilon \subseteq X \subseteq \overline{X}_\varepsilon.$$

*If additionally each $R_\varepsilon$ is reflexive, then the above inclusions hold without further assumptions, and the scale family $\mathsf{PR}(X)$ is well-defined as an indexed family of pairs of subsets of $U$.*

*Proof.* Fix $\varepsilon \geq 0$. Since $R_\varepsilon \subseteq U \times U$ is a relation, for each $x \in U$ the set $N_\varepsilon(x) = \{y \in U \mid xR_\varepsilon y\}$ is well-defined. Hence

$$\underline{X}_\varepsilon = \{x \in U \mid N_\varepsilon(x) \subseteq X\}, \qquad \overline{X}_\varepsilon = \{x \in U \mid N_\varepsilon(x) \cap X \neq \varnothing\}$$

are well-defined subsets of $U$.

Assume $R_\varepsilon$ is reflexive. If $x \in \underline{X}_\varepsilon$, then $N_\varepsilon(x) \subseteq X$ and $x \in N_\varepsilon(x)$, so $x \in X$. Thus $\underline{X}_\varepsilon \subseteq X$. If $x \in X$, then $x \in N_\varepsilon(x) \cap X$ (again by reflexivity), so $N_\varepsilon(x) \cap X \neq \varnothing$ and hence $x \in \overline{X}_\varepsilon$. Thus $X \subseteq \overline{X}_\varepsilon$.

Finally, $\mathsf{PR}(X) = \{(\underline{X}_\varepsilon, \overline{X}_\varepsilon)\}_{\varepsilon \geq 0}$ is well-defined as a family indexed by $\varepsilon \in \mathbb{R}_{\geq 0}$, because each component pair is uniquely determined by $(U, R_\varepsilon, X)$. □

**Example 2.51.4** (Customer segmentation under varying similarity thresholds)**.** Let $U = \{c_1, c_2, c_3, c_4, c_5\}$ be customers described by a feature vector (purchase frequency, recency, spend, etc.). Assume a dissimilarity $d : U \times U \to \mathbb{R}_{\geq 0}$ has been computed, and define for each $\varepsilon \geq 0$ the relation

$$x\, R_\varepsilon\, y \quad :\iff \quad d(x,y) \leq \varepsilon,$$

(which is reflexive since $d(x,x) = 0$). Suppose the nonzero distances are:

$$d(c_1,c_2) = 0.3, \quad d(c_1,c_3) = 0.9, \quad d(c_2,c_3) = 0.8, \quad d(c_3,c_4) = 0.4, \quad d(c_4,c_5) = 0.6,$$

and all other distinct pairs have distance $> 0.9$.

**Scenario.** Let $X = \{c_1, c_2\}$ be customers flagged as *high churn risk* by a business rule.

**At scale $\varepsilon = 0.5$.** Neighborhoods are

$$N_{0.5}(c_1) = \{c_1, c_2\}, \quad N_{0.5}(c_2) = \{c_1, c_2\},$$
$$N_{0.5}(c_3) = \{c_3, c_4\}, \quad N_{0.5}(c_4) = \{c_3, c_4\}, \quad N_{0.5}(c_5) = \{c_5\}.$$

Hence

$$\underline{X}_{0.5} = \{c_1, c_2\}, \qquad \overline{X}_{0.5} = \{c_1, c_2\}.$$

**At scale $\varepsilon = 0.9$.** Neighborhoods expand to

$$N_{0.9}(c_1) = \{c_1, c_2, c_3\}, \quad N_{0.9}(c_2) = \{c_1, c_2, c_3\}, \quad N_{0.9}(c_3) = \{c_1, c_2, c_3, c_4\},$$
$$N_{0.9}(c_4) = \{c_3, c_4, c_5\}, \quad N_{0.9}(c_5) = \{c_4, c_5\}.$$

Thus

$$\underline{X}_{0.9} = \varnothing, \qquad \overline{X}_{0.9} = \{c_1, c_2, c_3\}.$$

Interpretation: at fine resolution ($\varepsilon = 0.5$), churn-risk customers are robustly separated; at coarser resolution ($\varepsilon = 0.9$), similarity neighborhoods blur and the "possible" region grows to include $c_3$.

## 2.52 Causal Rough Set

Causal Rough Set forms indiscernibility using causally relevant attributes (e.g., Markov boundary), then computes Pawlak-style lower/upper approximations, emphasizing cause-driven granules over correlations in learning applications.

**Definition 2.52.1** (Causal Rough Set)**.** Let $\mathcal{S} = (U, \mathcal{A} \cup \{d\})$ be a decision system, where $U$ is the set of objects, $\mathcal{A}$ is a set of conditional attributes, and $d$ is a distinguished decision attribute. Assume a causal model has been specified (e.g., a causal DAG) such that a *causal boundary* (or *causal feature set*) $\mathcal{C}(d) \subseteq \mathcal{A}$ for the decision attribute $d$ is given (for example, a Markov boundary of $d$). Define the *causal indiscernibility relation* $\sim_{\mathrm{ca}}$ on $U$ by

$$x \sim_{\mathrm{ca}} y \quad :\Longleftrightarrow \quad \forall a \in \mathcal{C}(d),\ a(x) = a(y).$$

For any concept $X \subseteq U$, define the *causal lower* and *causal upper* approximations by

$$\underline{\mathrm{ca}}(X) := \{x \in U \mid [x]_{\sim_{\mathrm{ca}}} \subseteq X\}, \qquad \overline{\mathrm{ca}}(X) := \{x \in U \mid [x]_{\sim_{\mathrm{ca}}} \cap X \neq \varnothing\}.$$

The pair $\big(\underline{\mathrm{ca}}(X), \overline{\mathrm{ca}}(X)\big)$ is called the *Causal Rough Set* of $X$ induced by $\mathcal{C}(d)$.

**Remark 2.52.2.** The novelty is that the granules are formed only from (assumed) causally relevant attributes for $d$, rather than from all available attributes.

**Proposition 2.52.3** (Well-definedness of the Causal Rough Set)**.** *In Definition 2.52.1, the relation $\sim_{\mathrm{ca}}$ on $U$ is a well-defined equivalence relation. Consequently, for each $X \subseteq U$, the sets $\underline{\mathrm{ca}}(X)$ and $\overline{\mathrm{ca}}(X)$ are well-defined subsets of $U$ and satisfy*

$$\underline{\mathrm{ca}}(X) \subseteq X \subseteq \overline{\mathrm{ca}}(X).$$

*Proof. (Well-definedness).* Since each attribute $a \in \mathcal{A}$ is a function on $U$, the statement $a(x) = a(y)$ is unambiguous for any $x, y \in U$. Therefore the condition $\forall a \in \mathcal{C}(d)$, $a(x) = a(y)$ defines a well-defined relation $\sim_{\mathrm{ca}}$.

*(Equivalence).* Reflexivity: $a(x) = a(x)$ for all $a$, hence $x \sim_{\mathrm{ca}} x$. Symmetry: if $a(x) = a(y)$ then $a(y) = a(x)$, hence $x \sim_{\mathrm{ca}} y \Rightarrow y \sim_{\mathrm{ca}} x$. Transitivity: if $a(x) = a(y)$ and $a(y) = a(z)$ for all $a \in \mathcal{C}(d)$, then $a(x) = a(z)$ for all such $a$, so $x \sim_{\mathrm{ca}} z$.

*(Approximations and inclusions).* Since $\sim_{\mathrm{ca}}$ is an equivalence relation, each class $[x]_{\sim_{\mathrm{ca}}}$ is well-defined, and thus $\underline{\mathrm{ca}}(X)$ and $\overline{\mathrm{ca}}(X)$ are well-defined. If $x \in \underline{\mathrm{ca}}(X)$ then $[x]_{\sim_{\mathrm{ca}}} \subseteq X$ and $x \in [x]_{\sim_{\mathrm{ca}}}$, hence $x \in X$. If $x \in X$ then $[x]_{\sim_{\mathrm{ca}}} \cap X \neq \varnothing$, hence $x \in \overline{\mathrm{ca}}(X)$. □

**Example 2.52.4** (Clinical triage using a causally relevant feature set)**.** Let $U = \{p_1, p_2, p_3, p_4\}$ be patients. Consider conditional attributes

$$\mathcal{A} = \{\mathsf{Smoke}, \mathsf{Chol}, \mathsf{Gene}, \mathsf{Exercise}\}$$

and a decision attribute $d = \mathsf{HighRisk}$. Assume a causal analysis (e.g., domain knowledge / causal discovery) yields the causal boundary

$$\mathcal{C}(d) = \{\mathsf{Smoke}, \mathsf{Chol}, \mathsf{Gene}\},$$

meaning Exercise is not used to form causal granules for $d$.

Suppose the patient table is:

| | Smoke | Chol | Gene | Exercise | HighRisk |
|---|---|---|---|---|---|
| $p_1$ | 1 | H | 1 | L | 1 |
| $p_2$ | 1 | H | 1 | H | 1 |
| $p_3$ | 0 | H | 1 | L | 1 |
| $p_4$ | 0 | L | 0 | H | 0 |

By Definition 2.52.1,

$$p_i \sim_{\mathrm{ca}} p_j \iff \big(\mathsf{Smoke}(p_i), \mathsf{Chol}(p_i), \mathsf{Gene}(p_i)\big) = \big(\mathsf{Smoke}(p_j), \mathsf{Chol}(p_j), \mathsf{Gene}(p_j)\big).$$

Hence

$$[p_1]_{\sim_{\mathrm{ca}}} = [p_2]_{\sim_{\mathrm{ca}}} = \{p_1, p_2\}, \quad [p_3]_{\sim_{\mathrm{ca}}} = \{p_3\}, \quad [p_4]_{\sim_{\mathrm{ca}}} = \{p_4\}.$$

**Scenario.** Let $X = \{p_1, p_3\}$ be the set of patients whom a clinician tentatively selects for *intensive intervention.*

**Approximations.** Then

$$\underline{\mathrm{ca}}(X) = \{p \in U \mid [p]_{\sim_{\mathrm{ca}}} \subseteq X\} = \{p_3\},$$

because $[p_3]_{\sim_{\mathrm{ca}}} = \{p_3\} \subseteq X$, while $[p_1]_{\sim_{\mathrm{ca}}} = \{p_1, p_2\} \nsubseteq X$. Moreover,

$$\overline{\mathrm{ca}}(X) = \{p \in U \mid [p]_{\sim_{\mathrm{ca}}} \cap X \neq \varnothing\} = \{p_1, p_2, p_3\}.$$

Interpretation: $p_2$ becomes *possibly* in $X$ because it is causally indistinguishable from $p_1$ with respect to $\{\mathsf{Smoke}, \mathsf{Chol}, \mathsf{Gene}\}$ (even though Exercise differs).

## 2.53 Entropy-Regularized Rough Set

Entropy-Regularized Rough Set refines the lower approximation by retaining elements from sufficiently pure, low-entropy equivalence blocks, reducing boundary noise while preserving classical upper approximation structure.

**Definition 2.53.1** (Entropy-Regularized Rough Set)**.** Let $U$ be a finite universe and let $R$ be an equivalence relation on $U$, inducing a partition $U/R = \{B_1, \ldots, B_m\}$ into equivalence classes (blocks). For any $X \subseteq U$ and any block $B \in U/R$, define the block proportion

$$p_B(X) := \frac{|B \cap X|}{|B|} \in [0, 1]$$

and the binary Shannon entropy of this proportion

$$H_B(X) := -p_B(X)\log p_B(X) - (1-p_B(X))\log(1-p_B(X)), \qquad (0\log 0 := 0).$$

Fix parameters $\alpha \in (1/2, 1]$ (purity threshold) and $\theta \in [0, \log 2]$ (entropy threshold). Define the *entropy-regularized lower approximation* of $X$ by

$$\underline{X}_{\mathrm{ent}}^{(\alpha,\theta)} := \Big\{ x \in X \;\Big|\; p_{[x]_R}(X) \ge \alpha \text{ and } H_{[x]_R}(X) \le \theta \Big\},$$

and define the *upper approximation* as the classical Pawlak upper approximation

$$\overline{X}_{\mathrm{ent}} := \overline{X} := \{x \in U \mid [x]_R \cap X \neq \varnothing\}.$$

Then the pair

$$\Big(\underline{X}_{\mathrm{ent}}^{(\alpha,\theta)}, \overline{X}_{\mathrm{ent}}\Big)$$

is called an *Entropy-Regularized Rough Set* of $X$ (with respect to $R, \alpha, \theta$).

**Remark 2.53.2.** By construction, $\underline{X}_{\mathrm{ent}}^{(\alpha,\theta)} \subseteq X \subseteq \overline{X}_{\mathrm{ent}}$. The lower region keeps only those $x \in X$ whose $R$-block is both sufficiently pure (large $p$) and low-uncertainty (small entropy).

**Proposition 2.53.3** (Well-definedness of the Entropy-Regularized Rough Set)**.** *In Definition 2.53.1, the quantities $p_B(X)$ and $H_B(X)$ are well-defined for every block $B \in U/R$ and every $X \subseteq U$. Moreover, for any parameters $\alpha \in (1/2, 1]$ and $\theta \in [0, \log 2]$, the sets $\underline{X}_{\mathrm{ent}}^{(\alpha,\theta)}$ and $\overline{X}_{\mathrm{ent}}$ are well-defined subsets of $U$ and satisfy*

$$\underline{X}_{\mathrm{ent}}^{(\alpha,\theta)} \subseteq X \subseteq \overline{X}_{\mathrm{ent}}.$$

*Proof.* Since $R$ is an equivalence relation on the finite set $U$, each block $B \in U/R$ is nonempty and finite. Hence $|B| > 0$ and the ratio $p_B(X) = |B \cap X|/|B| \in [0, 1]$ is well-defined.

For the entropy term, the convention $0\log 0 := 0$ makes each summand well-defined at the endpoints $p_B(X) \in \{0, 1\}$. For $p_B(X) \in (0, 1)$ the expression is standard and finite. Hence $H_B(X)$ is well-defined for all $B$ and $X$.

By definition,

$$\underline{X}_{\mathrm{ent}}^{(\alpha,\theta)} = \Big\{ x \in X \;\Big|\; p_{[x]_R}(X) \ge \alpha \text{ and } H_{[x]_R}(X) \le \theta \Big\},$$

so $\underline{X}_{\mathrm{ent}}^{(\alpha,\theta)} \subseteq X$ holds immediately. Also, for any $x \in X$ we have $[x]_R \cap X \neq \varnothing$ (since $x \in [x]_R \cap X$), hence $x \in \overline{X} = \overline{X}_{\mathrm{ent}}$. Therefore $X \subseteq \overline{X}_{\mathrm{ent}}$. All sets are defined by unambiguous comprehension over $U$, so they are well-defined subsets of $U$. □

**Example 2.53.4** (Manufacturing quality control with entropy-filtered certainty)**.** Let $U = \{1, 2, \dots, 10\}$ be produced items in a factory shift. Define an equivalence relation $R$ by

$$i\,R\,j \quad :\Longleftrightarrow \quad \text{items } i \text{ and } j \text{ were produced on the same machine under the same supplier lot.}$$

Assume this yields two blocks:

$$B_1 = \{1, 2, 3, 4, 5\}, \qquad B_2 = \{6, 7, 8, 9, 10\}.$$

Let the concept $X \subseteq U$ be the set of *defective* items found by inspection:

$$X = \{1, 2, 3, 4, 6\}.$$

Thus

$$p_{B_1}(X) = \frac{|B_1 \cap X|}{|B_1|} = \frac{4}{5} = 0.8, \qquad p_{B_2}(X) = \frac{|B_2 \cap X|}{|B_2|} = \frac{1}{5} = 0.2.$$

The binary entropies (natural logarithm) are

$$H_{B_1}(X) = -0.8 \ln(0.8) - 0.2 \ln(0.2) \approx 0.5004, \qquad H_{B_2}(X) = -0.2 \ln(0.2) - 0.8 \ln(0.8) \approx 0.5004.$$

**Scenario.** Set thresholds $\alpha = 0.75$ and $\theta = 0.55$ (high purity, low uncertainty). By Definition 2.53.1,

$$\underline{X}_{\text{ent}}^{(\alpha,\theta)} = \Big\{ x \in X \;\Big|\; p_{[x]_R}(X) \geq \alpha,\ H_{[x]_R}(X) \leq \theta \Big\}.$$

Since $p_{B_1}(X) = 0.8 \geq 0.75$ and $H_{B_1}(X) \approx 0.5004 \leq 0.55$, but $p_{B_2}(X) = 0.2 < 0.75$, we obtain

$$\underline{X}_{\text{ent}}^{(0.75,0.55)} = X \cap B_1 = \{1, 2, 3, 4\}.$$

The (classical) upper approximation is

$$\overline{X}_{\text{ent}} = \overline{X} = \{x \in U \mid [x]_R \cap X \neq \varnothing\} = B_1 \cup B_2 = U,$$

because each block contains at least one defective item.

Interpretation: the entropy-regularized lower region isolates *robust defect evidence* coming from a highly defective block (machine/lot $B_1$), while the upper region flags all possibly affected items (both blocks).

## 2.54 Differentially-Private Rough Set

Differentially-Private Rough Set estimates lower and upper approximations from privatized data, adding noise for privacy, then selecting elements with high membership confidence under random mechanisms.

**Definition 2.54.1** (Differentially-Private Rough Set)**.** Let $U$ be a finite universe and let $R$ be an equivalence relation on $U$ (or any neighborhood relation), so that rough approximations of a concept $X \subseteq U$ can be computed from data-dependent statistics (e.g., block counts $|[x]_R \cap X|$). Let $M$ be a randomized mechanism that produces a privatized view of the data and satisfies $(\varepsilon_{\text{DP}}, \delta_{\text{DP}})$-differential privacy with respect to a chosen adjacency model. Using the privatized output of $M$, suppose we compute randomized approximations

$$\underline{X}^M \subseteq U, \qquad \overline{X}^M \subseteq U,$$

which are random sets (measurable with respect to the randomness of $M$). Fix a robustness level $\eta \in (0, 1)$ and define the *DP-robust lower* and *DP-robust upper* sets by

$$\underline{X}_{\text{DP}}^{(\eta)} := \Big\{ x \in U \;\Big|\; \Pr\big(x \in \underline{X}^M\big) \geq 1 - \eta \Big\}, \qquad \overline{X}_{\text{DP}}^{(\eta)} := \Big\{ x \in U \;\Big|\; \Pr\big(x \in \overline{X}^M\big) \geq 1 - \eta \Big\}.$$

The pair $\big(\underline{X}_{\text{DP}}^{(\eta)}, \overline{X}_{\text{DP}}^{(\eta)}\big)$ is called the *Differentially-Private Rough Set* (DP-Rough Set) of $X$ induced by the mechanism $M$.

**Remark 2.54.2.** Unlike classical rough sets, DP-Rough sets are *robust estimates* derived from privatized data. The parameter $\eta$ controls the confidence with which membership is asserted under the mechanism noise.

**Proposition 2.54.3** (Well-definedness of the Differentially-Private Rough Set)**.** *In Definition 2.54.1, assume that the randomized mechanism $M$ is defined on a probability space $(\Omega, \mathcal{F}, \mathbb{P})$, and that the randomized approximations $\underline{X}^M(\omega)$ and $\overline{X}^M(\omega)$ are set-valued random variables such that, for each $x \in U$, the events $\{\omega \mid x \in \underline{X}^M(\omega)\}$ and $\{\omega \mid x \in \overline{X}^M(\omega)\}$ are measurable. Then, for every $\eta \in (0,1)$, the sets $\underline{X}_{\mathrm{DP}}^{(\eta)}$ and $\overline{X}_{\mathrm{DP}}^{(\eta)}$ are well-defined subsets of $U$.*

*If moreover $\underline{X}^M(\omega) \subseteq \overline{X}^M(\omega)$ for all $\omega \in \Omega$, then*

$$\underline{X}_{\mathrm{DP}}^{(\eta)} \subseteq \overline{X}_{\mathrm{DP}}^{(\eta)}.$$

*Proof.* For each fixed $x \in U$, measurability of the event $\{x \in \underline{X}^M\}$ guarantees that the probability $\mathbb{P}(x \in \underline{X}^M)$ is well-defined; similarly for $\overline{X}^M$. Hence the membership conditions

$$\mathbb{P}(x \in \underline{X}^M) \geq 1 - \eta, \qquad \mathbb{P}(x \in \overline{X}^M) \geq 1 - \eta$$

are unambiguous, and therefore

$$\underline{X}_{\mathrm{DP}}^{(\eta)} := \{x \in U \mid \mathbb{P}(x \in \underline{X}^M) \geq 1 - \eta\}, \qquad \overline{X}_{\mathrm{DP}}^{(\eta)} := \{x \in U \mid \mathbb{P}(x \in \overline{X}^M) \geq 1 - \eta\}$$

are well-defined subsets of $U$.

Assume in addition that $\underline{X}^M(\omega) \subseteq \overline{X}^M(\omega)$ for all $\omega$. Then for each $x \in U$ we have pointwise implication $x \in \underline{X}^M(\omega) \Rightarrow x \in \overline{X}^M(\omega)$, hence

$$\mathbb{P}(x \in \overline{X}^M) \;\geq\; \mathbb{P}(x \in \underline{X}^M).$$

Therefore, if $x \in \underline{X}_{\mathrm{DP}}^{(\eta)}$ then $\mathbb{P}(x \in \underline{X}^M) \geq 1 - \eta$ implies $\mathbb{P}(x \in \overline{X}^M) \geq 1 - \eta$, so $x \in \overline{X}_{\mathrm{DP}}^{(\eta)}$. Thus $\underline{X}_{\mathrm{DP}}^{(\eta)} \subseteq \overline{X}_{\mathrm{DP}}^{(\eta)}$. □

**Example 2.54.4** (Publishing hotspot regions under differential privacy)**.** Let $U = \{z_1, z_2, z_3\}$ be city zip codes. Let $X = \{z_1, z_2\}$ be the (non-public) set of true *high-incidence* areas for a disease.

A public-health agency applies a randomized mechanism $M$ (e.g., Laplace noise on case counts followed by thresholding), and releases a privatized classification that induces random lower/upper sets

$$\underline{X}^M(\omega) \subseteq U, \qquad \overline{X}^M(\omega) \subseteq U.$$

Assume the induced membership probabilities (derivable from the noise model) are:

$$\mathbb{P}(z_1 \in \underline{X}^M) = 0.95, \quad \mathbb{P}(z_2 \in \underline{X}^M) = 0.92, \quad \mathbb{P}(z_3 \in \underline{X}^M) = 0.08,$$

and

$$\mathbb{P}(z_1 \in \overline{X}^M) = 0.98, \quad \mathbb{P}(z_2 \in \overline{X}^M) = 0.97, \quad \mathbb{P}(z_3 \in \overline{X}^M) = 0.40.$$

**Scenario.** Choose robustness $\eta = 0.10$ (90% confidence). By Definition 2.54.1,

$$\underline{X}_{\mathrm{DP}}^{(0.10)} = \{z \in U \mid \mathbb{P}(z \in \underline{X}^{M}) \geq 0.90\} = \{z_1, z_2\},$$

while

$$\overline{X}_{\mathrm{DP}}^{(0.10)} = \{z \in U \mid \mathbb{P}(z \in \overline{X}^{M}) \geq 0.90\} = \{z_1, z_2\}.$$

If the agency instead uses a looser robustness level $\eta = 0.60$ (40% confidence), then

$$\overline{X}_{\mathrm{DP}}^{(0.60)} = \{z_1, z_2, z_3\},$$

reflecting that $z_3$ is *possibly* a hotspot under the privatized release, although not robustly so.

Interpretation: the DP-rough approximations describe what can be asserted about hotspots *with high confidence* given privacy-induced randomness.

# Chapter 3

# Uncertain Rough Set

In this chapter, we introduce and discuss several variants of uncertain rough sets.

## 3.1 Fuzzy Rough Set

Fuzzy rough sets approximate a concept using fuzzy similarity relations, yielding fuzzy lower and upper memberships for uncertain classification tasks [229, 231–233]. Related concepts include Picture fuzzy rough sets [234, 235], Hesitant Fuzzy rough sets [236–238], Bipolar fuzzy rough sets [239–242], Linear diophantine fuzzy rough sets [243–245], Multipolar fuzzy rough sets [246], Variable precision fuzzy rough sets [247, 248], Soft fuzzy rough sets [232, 249, 250], Robust fuzzy rough sets [251, 252], and Spherical fuzzy rough sets [253–255].

**Definition 3.1.1** (Fuzzy rough approximations and fuzzy rough set)**.** [229, 231] Let $U$ be a nonempty universe and let $R : U \times U \to [0, 1]$ be a fuzzy relation. Let $T$ be a $t$-norm, $S$ a $t$-conorm, and let $N : [0, 1] \to [0, 1]$ be the standard negator $N(a) = 1 - a$. Define the (S-)implicator

$$I_S(a, b) \; := \; S\bigl(N(a),\, b\bigr) \qquad (a, b \in [0, 1]).$$

For a fuzzy set $A$ on $U$ (identified with its membership function $\mu_A : U \to [0, 1]$), the *fuzzy rough lower* and *fuzzy rough upper* approximations of $A$ w.r.t. $R$ are the fuzzy sets $\underline{R}(A)$ and $\overline{R}(A)$ whose membership functions are

$$\mu_{\underline{R}(A)}(x) \; := \; \inf_{y \in U} I_S\bigl(R(x, y),\, \mu_A(y)\bigr) \; = \; \inf_{y \in U} S\bigl(1 - R(x, y),\, \mu_A(y)\bigr),$$

$$\mu_{\overline{R}(A)}(x) \; := \; \sup_{y \in U} T\bigl(R(x, y),\, \mu_A(y)\bigr) \qquad (x \in U).$$

The pair $\bigl(\underline{R}(A), \overline{R}(A)\bigr)$ is called the *fuzzy rough set* induced by $A$ (w.r.t. $R$).

**Example 3.1.2** (Fuzzy rough set for identifying "loyal customers" from similarity of purchase behavior)**.** Let $U = \{c_1, c_2, c_3\}$ be three customers in an online store. We consider the fuzzy concept

$$A = \text{"loyal customer"}$$

modeled by the membership function

$$\mu_A(c_1) = 0.9, \qquad \mu_A(c_2) = 0.5, \qquad \mu_A(c_3) = 0.2.$$

Assume a fuzzy similarity relation $R : U \times U \to [0, 1]$ derived from similarity of purchase patterns:

| $R(x, y)$ | $c_1$ | $c_2$ | $c_3$ |
|---|---|---|---|
| $c_1$ | 1.0 | 0.7 | 0.3 |
| $c_2$ | 0.7 | 1.0 | 0.6 |
| $c_3$ | 0.3 | 0.6 | 1.0 |

Choose the standard negator $N(a) = 1 - a$, the $t$-norm $T = \min$, and the $t$-conorm $S = \max$. Then the $S$-implicator is

$$I_S(a, b) = S(1 - a, b) = \max(1 - a, b).$$

Since $U$ is finite, $\inf = \min$ and $\sup = \max$, hence for each $x \in U$,

$$\mu_{\underline{R}(A)}(x) = \min_{y \in U} \max\big(1 - R(x, y), \mu_A(y)\big), \qquad \mu_{\overline{R}(A)}(x) = \max_{y \in U} \min\big(R(x, y), \mu_A(y)\big).$$

We compute the approximations explicitly.

**(i) Upper approximation.**

$$\mu_{\overline{R}(A)}(c_1) = \max\{\min(1, 0.9), \min(0.7, 0.5), \min(0.3, 0.2)\} = \max\{0.9, 0.5, 0.2\} = 0.9,$$

$$\mu_{\overline{R}(A)}(c_2) = \max\{\min(0.7, 0.9), \min(1, 0.5), \min(0.6, 0.2)\} = \max\{0.7, 0.5, 0.2\} = 0.7,$$

$$\mu_{\overline{R}(A)}(c_3) = \max\{\min(0.3, 0.9), \min(0.6, 0.5), \min(1, 0.2)\} = \max\{0.3, 0.5, 0.2\} = 0.5.$$

**(ii) Lower approximation.**

$$\mu_{\underline{R}(A)}(c_1) = \min\{\max(0, 0.9), \max(0.3, 0.5), \max(0.7, 0.2)\} = \min\{0.9, 0.5, 0.7\} = 0.5,$$

$$\mu_{\underline{R}(A)}(c_2) = \min\{\max(0.3, 0.9), \max(0, 0.5), \max(0.4, 0.2)\} = \min\{0.9, 0.5, 0.4\} = 0.4,$$

$$\mu_{\underline{R}(A)}(c_3) = \min\{\max(0.7, 0.9), \max(0.4, 0.5), \max(0, 0.2)\} = \min\{0.9, 0.5, 0.2\} = 0.2.$$

Therefore,

$$\underline{R}(A) = \{(c_1, 0.5), (c_2, 0.4), (c_3, 0.2)\}, \qquad \overline{R}(A) = \{(c_1, 0.9), (c_2, 0.7), (c_3, 0.5)\},$$

and the fuzzy rough set induced by $A$ (w.r.t. $R$) is the pair $\big(\underline{R}(A), \overline{R}(A)\big)$.

*Interpretation.* $\underline{R}(A)$ gives a conservative loyalty score that must be supported across all similar customers, while $\overline{R}(A)$ gives a permissive score supported by at least one similar customer.

## 3.2 Intuitionistic Fuzzy Rough Set

An intuitionistic fuzzy set assigns to each element both a membership degree and a nonmembership degree in $[0, 1]$, with their sum at most 1, thereby explicitly representing hesitation under incomplete information in decision-making [3, 256]. An intuitionistic fuzzy rough set then combines this intuitionistic fuzzy description with a suitable intuitionistic fuzzy relation to compute lower and upper intuitionistic fuzzy approximations of a concept, producing bounded uncertainty regions for classification and analysis [257–259]. Related concepts also include generalized intuitionistic fuzzy rough sets [257] and intuitionistic hesitant fuzzy rough sets [260].

**Definition 3.2.1** (Intuitionistic fuzzy rough approximations and IF rough set)**.** Let $U$ be a nonempty universe. An *intuitionistic fuzzy set* (IF set) on $U$ is a pair $X = (\mu_X, \gamma_X)$ of functions $\mu_X, \gamma_X : U \to [0, 1]$ satisfying $\mu_X(x) + \gamma_X(x) \le 1$ for all $x \in U$; $\mu_X(x)$ and $\gamma_X(x)$ are the membership and nonmembership degrees, respectively.

An *intuitionistic fuzzy (binary) relation* on $U$ is a pair $R = (\mu_R, \gamma_R)$ with $\mu_R, \gamma_R : U \times U \to [0, 1]$ and $\mu_R(x, y) + \gamma_R(x, y) \le 1$ for all $(x, y) \in U \times U$.

For any IF set $X = (\mu_X, \gamma_X)$, define the *lower* and *upper* approximations of $X$ w.r.t. $R$ as IF sets

$$\underline{R}(X) = \big(\mu_{\underline{R}(X)}, \gamma_{\underline{R}(X)}\big), \qquad \overline{R}(X) = \big(\mu_{\overline{R}(X)}, \gamma_{\overline{R}(X)}\big),$$

where for each $x \in U$,

$$\mu_{\underline{R}(X)}(x) := \inf_{y \in U} \max\big(\gamma_R(x, y),\ \mu_X(y)\big), \qquad \gamma_{\underline{R}(X)}(x) := \sup_{y \in U} \min\big(\mu_R(x, y),\ \gamma_X(y)\big),$$

$$\mu_{\overline{R}(X)}(x) := \sup_{y \in U} \min\big(\mu_R(x, y),\ \mu_X(y)\big), \qquad \gamma_{\overline{R}(X)}(x) := \inf_{y \in U} \max\big(\gamma_R(x, y),\ \gamma_X(y)\big).$$

If $\underline{R}(X) = \overline{R}(X)$, then $X$ is *IF definable*; otherwise $\big(\underline{R}(X), \overline{R}(X)\big)$ is called an *intuitionistic fuzzy rough set* (IF rough set).

**Example 3.2.2** (Intuitionistic fuzzy rough set for loan-default risk (similarity of applicants))**.** Let $U = \{a, b, c\}$ be three loan applicants. We model the concept

$$X = \text{“high default risk”}$$

as an intuitionistic fuzzy (IF) set $X = (\mu_X, \gamma_X)$ on $U$:

| $x$ | $a$ | $b$ | $c$ |
|---|---|---|---|
| $\mu_X(x)$ | 0.80 | 0.40 | 0.20 |
| $\gamma_X(x)$ | 0.10 | 0.40 | 0.70 |

$(\mu_X(x) + \gamma_X(x) \le 1$ for all $x)$.

Interpretation: $\mu_X(x)$ is evidence that $x$ is high-risk, while $\gamma_X(x)$ is evidence that $x$ is not high-risk.

Next, define an intuitionistic fuzzy relation $R = (\mu_R, \gamma_R)$ on $U$ encoding “financial-profile similarity” (e.g., similar income–debt patterns). Let $\mu_R$ and $\gamma_R$ be symmetric, with $\mu_R(x, x) = 1$ and $\gamma_R(x, x) = 0$, and the following off-diagonal values:

| $\mu_R$ | $a$ | $b$ | $c$ |
|---|---|---|---|
| $a$ | 1.00 | 0.70 | 0.30 |
| $b$ | 0.70 | 1.00 | 0.50 |
| $c$ | 0.30 | 0.50 | 1.00 |

| $\gamma_R$ | $a$ | $b$ | $c$ |
|---|---|---|---|
| $a$ | 0.00 | 0.20 | 0.60 |
| $b$ | 0.20 | 0.00 | 0.30 |
| $c$ | 0.60 | 0.30 | 0.00 |

(note $\mu_R(x,y)+\gamma_R(x,y)\le 1$ everywhere).

Using Definition of IF rough approximations, for each $x\in U$ (since $U$ is finite, inf = min and sup = max),

$$\mu_{\underline{R}(X)}(x)=\min_{y\in U}\max\bigl(\gamma_R(x,y),\,\mu_X(y)\bigr),\qquad \gamma_{\underline{R}(X)}(x)=\max_{y\in U}\min\bigl(\mu_R(x,y),\,\gamma_X(y)\bigr),$$

$$\mu_{\overline{R}(X)}(x)=\max_{y\in U}\min\bigl(\mu_R(x,y),\,\mu_X(y)\bigr),\qquad \gamma_{\overline{R}(X)}(x)=\min_{y\in U}\max\bigl(\gamma_R(x,y),\,\gamma_X(y)\bigr).$$

For instance, at $x=a$,

$$\mu_{\underline{R}(X)}(a)=\min\{\max(0,0.80),\ \max(0.20,0.40),\ \max(0.60,0.20)\}=\min\{0.80,0.40,0.60\}=0.40,$$

$$\gamma_{\underline{R}(X)}(a)=\max\{\min(1,0.10),\ \min(0.70,0.40),\ \min(0.30,0.70)\}=\max\{0.10,0.40,0.30\}=0.40,$$

$$\mu_{\overline{R}(X)}(a)=\max\{\min(1,0.80),\ \min(0.70,0.40),\ \min(0.30,0.20)\}=\max\{0.80,0.40,0.20\}=0.80,$$

$$\gamma_{\overline{R}(X)}(a)=\min\{\max(0,0.10),\ \max(0.20,0.40),\ \max(0.60,0.70)\}=\min\{0.10,0.40,0.70\}=0.10.$$

Carrying out the same finite min/max computations for $b$ and $c$ yields:

| | $\underline{R}(X)$ | | $\overline{R}(X)$ | |
|---|---|---|---|---|
| $x$ | $\mu_{\underline{R}(X)}(x)$ | $\gamma_{\underline{R}(X)}(x)$ | $\mu_{\overline{R}(X)}(x)$ | $\gamma_{\overline{R}(X)}(x)$ |
| $a$ | 0.40 | 0.40 | 0.80 | 0.10 |
| $b$ | 0.30 | 0.50 | 0.70 | 0.20 |
| $c$ | 0.20 | 0.70 | 0.40 | 0.40 |

Hence $\underline{R}(X)\neq\overline{R}(X)$, so $\bigl(\underline{R}(X),\overline{R}(X)\bigr)$ is an *intuitionistic fuzzy rough set*. Intuitively, similarity between applicants propagates both risk-evidence and non-risk-evidence, producing conservative (lower) and permissive (upper) IF assessments.

## 3.3 Vague Rough Set

A vague set assigns to each element an interval-valued membership $[t(x),\,1-f(x)]$ determined by the degree of supporting evidence $t(x)$ and opposing evidence $f(x)$ [261,262]. A vague rough set combines Pawlak rough approximations with such interval-valued memberships: elements in the lower approximation are treated as certainly in the concept, elements outside the upper approximation as certainly out, and elements in the boundary region are represented by genuinely vague membership intervals [263,264].

**Definition 3.3.1** (Vague set)**.** [261,262] Let $U$ be a nonempty universe. A *vague set* $V$ in $U$ is specified by two functions

$$t_V,\ f_V:U\to[0,1]\quad\text{with}\quad t_V(x)+f_V(x)\le 1\ \ (x\in U).$$

For each $x\in U$, the (interval-valued) membership of $x$ in $V$ is bounded by

$$\eta_V(x)\in[\,t_V(x),\,1-f_V(x)\,]\subseteq[0,1].$$

**Definition 3.3.2** (Vague rough set). [263,264] Let $(U, R)$ be an approximation space and let $X \subseteq U$. A *vague rough set* (induced by $X$ under $R$) is a pair of functions

$$\mu, \nu : U \to [0,1]$$

(called membership and non-membership) satisfying:

$$\begin{aligned}
&\text{(certainly in)} && x \in \underline{R}(X) \implies [\mu(x),\, 1-\nu(x)] = [1,1],\\
&\text{(certainly out)} && x \in U \setminus \overline{R}(X) \implies [\mu(x),\, 1-\nu(x)] = [0,0],\\
&\text{(boundary/vague)} && x \in \mathrm{BND}_R(X) \implies 0 \le \mu(x)+\nu(x) \le 1.
\end{aligned}$$

Equivalently, each $x \in U$ is assigned a vague membership interval

$$[\mu(x),\, 1-\nu(x)] \subseteq [0,1],$$

which is crisp on $\underline{R}(X)$ and $U \setminus \overline{R}(X)$ and may be genuinely vague on the boundary $\mathrm{BND}_R(X)$.

**Remark 3.3.3** (Alternative (lower/upper vague sets)). One may also regard a vague rough set as a pair of vague sets on the rough approximations: given a rough set $(X_L, X_U) = (\underline{R}(X), \overline{R}(X))$, define vague sets $V_L$ on $X_L$ and $V_U$ on $X_U$ via truth/false membership functions

$$t_L, f_L : X_L \to [0,1], \qquad t_U, f_U : X_U \to [0,1],$$

with the pointwise constraints $t_L \le 1 - f_L$, $t_U \le 1 - f_U$, and a natural coherence such as $1 - f_L(y) \le 1 - f_U(y)$ for $y \in X_U$. This viewpoint emphasizes that "certain" and "possible" regions each carry their own vague evidence.

**Example 3.3.4** (Age-group concept with vagueness on the boundary). Let

$$U = \{\mathsf{Child}, \mathsf{Pre\text{-}Teen}, \mathsf{Teen}, \mathsf{Youth}, \mathsf{Teenager}, \mathsf{Young\text{-}Adult}, \mathsf{Adult}, \mathsf{Senior}, \mathsf{Senior\text{-}Citizen}, \mathsf{Elderly}\}.$$

Define an equivalence relation $R$ (indiscernibility) whose classes are

$$\{\mathsf{Child}, \mathsf{Pre\text{-}Teen}\},\ \{\mathsf{Teen}, \mathsf{Youth}, \mathsf{Teenager}\},\ \{\mathsf{Young\text{-}Adult}\},\ \{\mathsf{Adult}\},\ \{\mathsf{Senior}, \mathsf{Senior\text{-}Citizen}, \mathsf{Elderly}\}.$$

Consider the target (crisp) concept

$$X = \{\mathsf{Child}, \mathsf{Pre\text{-}Teen}, \mathsf{Youth}, \mathsf{Young\text{-}Adult}\} \subseteq U.$$

Then

$$\underline{R}(X) = \{\mathsf{Child}, \mathsf{Pre\text{-}Teen}, \mathsf{Young\text{-}Adult}\}, \qquad \overline{R}(X) = \{\mathsf{Child}, \mathsf{Pre\text{-}Teen}, \mathsf{Teen}, \mathsf{Youth}, \mathsf{Teenager}, \mathsf{Young\text{-}Adult}\},$$

so the boundary is $\mathrm{BND}_R(X) = \{\mathsf{Teen}, \mathsf{Youth}, \mathsf{Teenager}\}$.

Define a vague rough set by specifying $(\mu, \nu)$ as follows:

| $u$ | $\mu(u)$ | $\nu(u)$ | $[\mu(u),\, 1-\nu(u)]$ |
|---|---|---|---|
| Child | 1 | 0 | $[1,1]$ |
| Pre-Teen | 1 | 0 | $[1,1]$ |
| Young-Adult | 1 | 0 | $[1,1]$ |
| Adult | 0 | 1 | $[0,0]$ |
| Senior | 0 | 1 | $[0,0]$ |
| Senior-Citizen | 0 | 1 | $[0,0]$ |
| Elderly | 0 | 1 | $[0,0]$ |
| Teen | 0.3 | 0.5 | $[0.3, 0.5]$ |
| Youth | 0.5 | 0.3 | $[0.5, 0.7]$ |
| Teenager | 0.4 | 0.4 | $[0.4, 0.6]$ |

Here the certain region $\underline{R}(X)$ has crisp value $[1,1]$, the certainly-out region $U \setminus \overline{R}(X)$ has crisp value $[0,0]$, and the boundary elements carry genuine vagueness with $\mu(u) + \nu(u) \le 1$.

## 3.4 Neutrosophic Rough Set

A neutrosophic set assigns each element independent truth, indeterminacy, and falsity degrees in $[0,1]$, modeling incomplete information and contradictory evidence [6,7]. Neutrosophic rough sets form lower/upper neutrosophic approximations by taking inf/sup of the truth, indeterminacy, and falsity degrees over each equivalence class [182, 265–268]. Related concepts also include bipolar neutrosophic rough sets [269], quadripartitioned neutrosophic rough sets [270], generalized Neutrosophic Rough Sets [266, 271, 272], and pentapartitioned neutrosophic rough sets [273–275].

**Definition 3.4.1** (Neutrosophic rough approximations and neutrosophic rough set)**.** Let $U$ be a nonempty universe and let $R \subseteq U \times U$ be an equivalence relation. For $x \in U$, write $[x]_R := \{y \in U \mid (x,y) \in R\}$.

A *single-valued neutrosophic set* on $U$ is a triple $A = (T_A, I_A, F_A)$ of functions $T_A, I_A, F_A : U \to [0,1]$, interpreted as truth-, indeterminacy-, and falsity-membership degrees.

Define the *neutrosophic lower* and *neutrosophic upper* approximations of $A$ w.r.t. $R$ by the neutrosophic sets $\underline{R}(A) = (T_{\underline{R}(A)}, I_{\underline{R}(A)}, F_{\underline{R}(A)})$ and $\overline{R}(A) = (T_{\overline{R}(A)}, I_{\overline{R}(A)}, F_{\overline{R}(A)})$ where, for each $x \in U$,

$$T_{\underline{R}(A)}(x) := \inf_{y\in[x]_R} T_A(y), \qquad I_{\underline{R}(A)}(x) := \inf_{y\in[x]_R} I_A(y), \qquad F_{\underline{R}(A)}(x) := \sup_{y\in[x]_R} F_A(y),$$

$$T_{\overline{R}(A)}(x) := \sup_{y\in[x]_R} T_A(y), \qquad I_{\overline{R}(A)}(x) := \sup_{y\in[x]_R} I_A(y), \qquad F_{\overline{R}(A)}(x) := \inf_{y\in[x]_R} F_A(y).$$

The pair $\big(\underline{R}(A), \overline{R}(A)\big)$ is called the *neutrosophic rough set* induced by $A$ (w.r.t. $R$).

**Example 3.4.2** (Neutrosophic rough set in medical triage (influenza suspicion))**.** Let $U = \{p_1, p_2, p_3, p_4, p_5\}$ be five patients in an emergency department. We model the neutrosophic concept

$$A = \text{“the patient has influenza”}$$

as a single-valued neutrosophic set $A = (T_A, I_A, F_A)$ on $U$, where: $T_A$ comes from a rapid antigen score (evidence-for), $I_A$ represents indeterminacy due to sample quality / timing, and $F_A$ comes from evidence-against (e.g., strong alternative diagnosis indicators).

Suppose patients are grouped by the same coarse symptom-profile (e.g., *high fever & cough* vs. *mild fever & cough*, etc.). Define an equivalence relation $R$ on $U$ with classes

$$[p_1]_R = [p_2]_R = \{p_1, p_2\}, \qquad [p_3]_R = [p_4]_R = \{p_3, p_4\}, \qquad [p_5]_R = \{p_5\}.$$

Assume the neutrosophic membership degrees are:

| $x$ | $T_A(x)$ | $I_A(x)$ | $F_A(x)$ |
|---|---|---|---|
| $p_1$ | 0.80 | 0.20 | 0.10 |
| $p_2$ | 0.60 | 0.40 | 0.30 |
| $p_3$ | 0.30 | 0.50 | 0.40 |
| $p_4$ | 0.40 | 0.30 | 0.50 |
| $p_5$ | 0.10 | 0.20 | 0.80 |

By Definition of neutrosophic rough approximations, for each $x \in U$,

$$T_{\underline{R}(A)}(x) = \inf_{y \in [x]_R} T_A(y), \quad I_{\underline{R}(A)}(x) = \inf_{y \in [x]_R} I_A(y), \quad F_{\underline{R}(A)}(x) = \sup_{y \in [x]_R} F_A(y),$$

$$T_{\overline{R}(A)}(x) = \sup_{y \in [x]_R} T_A(y), \quad I_{\overline{R}(A)}(x) = \sup_{y \in [x]_R} I_A(y), \quad F_{\overline{R}(A)}(x) = \inf_{y \in [x]_R} F_A(y).$$

**Class $\{p_1, p_2\}$.** For $x \in \{p_1, p_2\}$,

$$\underline{R}(A)(x) = \big(\min\{0.80, 0.60\}, \min\{0.20, 0.40\}, \max\{0.10, 0.30\}\big) = (0.60, 0.20, 0.30),$$

$$\overline{R}(A)(x) = \big(\max\{0.80, 0.60\}, \max\{0.20, 0.40\}, \min\{0.10, 0.30\}\big) = (0.80, 0.40, 0.10).$$

**Class $\{p_3, p_4\}$.** For $x \in \{p_3, p_4\}$,

$$\underline{R}(A)(x) = \big(\min\{0.30, 0.40\}, \min\{0.50, 0.30\}, \max\{0.40, 0.50\}\big) = (0.30, 0.30, 0.50),$$

$$\overline{R}(A)(x) = \big(\max\{0.30, 0.40\}, \max\{0.50, 0.30\}, \min\{0.40, 0.50\}\big) = (0.40, 0.50, 0.40).$$

**Singleton class $\{p_5\}$.** For $x = p_5$,

$$\underline{R}(A)(p_5) = \overline{R}(A)(p_5) = A(p_5) = (0.10, 0.20, 0.80).$$

Therefore, the neutrosophic rough set induced by $A$ (w.r.t. $R$) is

$$\big(\underline{R}(A), \overline{R}(A)\big).$$

*Interpretation:* within each symptom-profile class, $\underline{R}(A)$ provides a conservative (worst-case) assessment of influenza (lower truth/indeterminacy and higher falsity), while $\overline{R}(A)$ gives a permissive (best-case) assessment consistent with at least one patient in the class.

## 3.5 Plithogenic Rough Set

A plithogenic set models element appurtenance to multiple attribute values, weighted by contradiction degrees, and aggregates memberships using tailored operators [276–278]. Plithogenic rough sets form lower/upper approximations by taking componentwise meet/join of the appurtenance vectors over each equivalence class, while retaining the attribute values and contradiction function [279].

**Definition 3.5.1** (Plithogenic rough set)**.** [280] Let $U$ be a nonempty finite universe and let $R \subseteq U \times U$ be an equivalence relation. For each $x \in U$, write

$$[x]_R := \{\, y \in U \mid (x, y) \in R \,\}$$

for the $R$-equivalence class of $x$.

Fix an attribute $a$ with a nonempty finite value set $V_a$. Let $s, t \in \mathbb{N}$. A *plithogenic set* on $U$ (with respect to $a$) is a quintuple

$$\mathsf{PS} = (U, a, V_a, \mathsf{pdf}, \mathsf{pcf}),$$

where $\mathsf{pdf} : U \times V_a \to [0,1]^s$ is the *degree of appurtenance function* and $\mathsf{pcf} : V_a \times V_a \to [0,1]^t$ is the *degree of contradiction function* satisfying

$$\mathsf{pcf}(v, v) = 0, \qquad \mathsf{pcf}(v, w) = \mathsf{pcf}(w, v) \qquad (v, w \in V_a).$$

Equip $[0,1]^s$ with the product (componentwise) order $\leq$ and lattice operations

$$(u \wedge w)_j := \min\{u_j, w_j\}, \qquad (u \vee w)_j := \max\{u_j, w_j\} \qquad (1 \leq j \leq s),$$

for $u = (u_1, \ldots, u_s)$ and $w = (w_1, \ldots, w_s)$.

For each $v \in V_a$, define the *R-lower* and *R-upper* plithogenic rough approximations of $\mathsf{PS}$ by the maps $\underline{\mathsf{pdf}}_R, \overline{\mathsf{pdf}}_R : U \times V_a \to [0,1]^s$:

$$\underline{\mathsf{pdf}}_R(x, v) := \bigwedge_{y \in [x]_R} \mathsf{pdf}(y, v), \qquad \overline{\mathsf{pdf}}_R(x, v) := \bigvee_{y \in [x]_R} \mathsf{pdf}(y, v) \qquad (x \in U,\ v \in V_a).$$

Then the *plithogenic rough set* induced by $(U, R)$ and $\mathsf{PS}$ is the pair

$$(\underline{\mathsf{PS}}_R,\ \overline{\mathsf{PS}}_R),$$

where

$$\underline{\mathsf{PS}}_R := (U, a, V_a, \underline{\mathsf{pdf}}_R, \mathsf{pcf}),$$
$$\overline{\mathsf{PS}}_R := (U, a, V_a, \overline{\mathsf{pdf}}_R, \mathsf{pcf}).$$

**Example 3.5.2** (Plithogenic rough set for product color with contradictory labels)**.** Let $U = \{u_1, u_2, u_3, u_4, u_5\}$ be a set of T-shirts in an online catalog. Assume the indiscernibility relation $R$ groups items by the same manufacturer and fabric type, yielding the equivalence classes

$$[u_1]_R = [u_2]_R = \{u_1, u_2\}, \qquad [u_3]_R = [u_4]_R = \{u_3, u_4\}, \qquad [u_5]_R = \{u_5\}.$$

We study one attribute $a =$"color" with value set

$$V_a = \{\text{Red}, \text{Orange}, \text{Brown}\}.$$

Take $s = 2$ and interpret $\mathsf{pdf}(x, v) = (\mu_1(x, v), \mu_2(x, v)) \in [0,1]^2$ as a two-source appurtenance vector, e.g. $(\mu_1, \mu_2) =$ (image-based classifier score, human-tagging score).

Define $\mathsf{pdf} : U \times V_a \to [0,1]^2$ by the following values (listed by $v \in V_a$):

| | Red | Orange | Brown |
|---|---|---|---|
| $u_1$ | $(0.90, 0.80)$ | $(0.20, 0.10)$ | $(0.10, 0.20)$ |
| $u_2$ | $(0.60, 0.70)$ | $(0.50, 0.40)$ | $(0.20, 0.20)$ |
| $u_3$ | $(0.10, 0.10)$ | $(0.70, 0.60)$ | $(0.40, 0.50)$ |
| $u_4$ | $(0.20, 0.10)$ | $(0.60, 0.50)$ | $(0.50, 0.60)$ |
| $u_5$ | $(0.30, 0.20)$ | $(0.20, 0.30)$ | $(0.80, 0.90)$ |

Define a contradiction function $\mathsf{pcf} : V_a \times V_a \to [0, 1]$ (so $t = 1$) encoding how incompatible two color labels are:

$$\mathsf{pcf}(v, v) = 0, \qquad \mathsf{pcf}(\text{Red}, \text{Orange}) = 0.20,$$

$$\mathsf{pcf}(\text{Orange}, \text{Brown}) = 0.30, \qquad \mathsf{pcf}(\text{Red}, \text{Brown}) = 0.80,$$

and extend symmetrically, e.g. $\mathsf{pcf}(\text{Orange}, \text{Red}) = 0.20$.

Hence the plithogenic set is

$$\mathsf{PS} = (U, a, V_a, \mathsf{pdf}, \mathsf{pcf}).$$

The plithogenic $R$-lower and $R$-upper rough approximations are computed componentwise via $\wedge = \min$ and $\vee = \max$ in $[0, 1]^2$:

$$\underline{\mathsf{pdf}}_R(x, v) = \bigwedge_{y \in [x]_R} \mathsf{pdf}(y, v),$$

$$\overline{\mathsf{pdf}}_R(x, v) = \bigvee_{y \in [x]_R} \mathsf{pdf}(y, v).$$

For instance, in the class $\{u_1, u_2\}$ we obtain:

$$\underline{\mathsf{pdf}}_R(u_1, \text{Red}) = \underline{\mathsf{pdf}}_R(u_2, \text{Red}) = (\min\{0.90, 0.60\}, \min\{0.80, 0.70\}) = (0.60, 0.70),$$

$$\overline{\mathsf{pdf}}_R(u_1, \text{Red}) = \overline{\mathsf{pdf}}_R(u_2, \text{Red}) = (\max\{0.90, 0.60\}, \max\{0.80, 0.70\}) = (0.90, 0.80),$$

and similarly

$$\underline{\mathsf{pdf}}_R(u_1, \text{Orange}) = \underline{\mathsf{pdf}}_R(u_2, \text{Orange}) = (0.20, 0.10),$$

$$\overline{\mathsf{pdf}}_R(u_1, \text{Orange}) = \overline{\mathsf{pdf}}_R(u_2, \text{Orange}) = (0.50, 0.40).$$

In the class $\{u_3, u_4\}$, for Brown:

$$\underline{\mathsf{pdf}}_R(u_3, \text{Brown}) = \underline{\mathsf{pdf}}_R(u_4, \text{Brown}) = (\min\{0.40, 0.50\}, \min\{0.50, 0.60\}) = (0.40, 0.50),$$

$$\overline{\mathsf{pdf}}_R(u_3, \text{Brown}) = \overline{\mathsf{pdf}}_R(u_4, \text{Brown}) = (\max\{0.40, 0.50\}, \max\{0.50, 0.60\}) = (0.50, 0.60).$$

Finally, for the singleton class $\{u_5\}$ we have $\underline{\mathsf{pdf}}_R(u_5, v) = \overline{\mathsf{pdf}}_R(u_5, v) = \mathsf{pdf}(u_5, v)$.

Therefore the induced plithogenic rough set is the pair

$$(\underline{\mathsf{PS}}_R, \ \overline{\mathsf{PS}}_R),$$

$$\text{where} \quad \underline{\mathsf{PS}}_R = (U, a, V_a, \underline{\mathsf{pdf}}_R, \mathsf{pcf}), \quad \overline{\mathsf{PS}}_R = (U, a, V_a, \overline{\mathsf{pdf}}_R, \mathsf{pcf}).$$

Within each manufacturer–fabric class, $\underline{\mathsf{pdf}}_R$ gives a conservative (class-consistent) vector score for each color value, while $\overline{\mathsf{pdf}}_R$ gives a permissive score capturing any evidence in the class. The contradiction map $\mathsf{pcf}$ records that confusing Red with Brown is far more contradictory than confusing Red with Orange, which can be used later in plithogenic aggregation rules.

## 3.6 Uncertain Rough Set

An Uncertain Set is a generic way to attach "uncertainty values" to elements, where the values live in a chosen *degree-domain* [281]. By selecting an appropriate degree-domain, one recovers fuzzy, intuitionistic fuzzy, neutrosophic, plithogenic, and many related models as special cases [281].

**Definition 3.6.1** (Uncertainty model / degree-domain)**.** An *uncertainty model* $M$ consists of a bounded De Morgan lattice

$$\big(\mathrm{Dom}(M), \le_M, \oplus_M, \otimes_M, N_M, 0_M, 1_M\big),$$

where $\oplus_M$ and $\otimes_M$ are the join and meet, $N_M$ is an order-reversing involution (De Morgan complement), and $0_M, 1_M$ are the least and greatest elements. For a finite family $\{a_i\}_{i=1}^{n} \subseteq \mathrm{Dom}(M)$ we write

$$\bigvee_{i=1}^{n} a_i := a_1 \oplus_M \cdots \oplus_M a_n, \qquad \bigwedge_{i=1}^{n} a_i := a_1 \otimes_M \cdots \otimes_M a_n.$$

**Definition 3.6.2** ($M$-implication (Kleene–Dienes type))**.** For $a, b \in \mathrm{Dom}(M)$ define

$$a \Rightarrow_M b \;:=\; N_M(a) \oplus_M b.$$

**Definition 3.6.3** ($M$-valued relation)**.** Let $X$ be a nonempty finite universe. An *$M$-valued relation* (or *$M$-relation*) on $X$ is a mapping

$$R_M : X \times X \longrightarrow \mathrm{Dom}(M).$$

(Optionally one may assume $M$-reflexivity $R_M(x,x) = 1_M$ and/or $M$-symmetry, etc., depending on the application.)

**Definition 3.6.4** (Uncertain Set (U-Set) of type $M$)**.** An *Uncertain Set of type $M$* on $X$ is a mapping

$$\mu_A : X \longrightarrow \mathrm{Dom}(M).$$

**Definition 3.6.5** (Uncertain Rough approximations)**.** Let $X$ be finite, let $M$ be as in Definition 3.6.1, let $R_M$ be an $M$-relation on $X$, and let $A$ be a U-Set of type $M$ with membership $\mu_A$. Define the *$M$-lower* and *$M$-upper* rough approximations of $A$ (with respect to $R_M$) by

$$\mu_{\underline{R}_M(A)}(x) \;:=\; \bigwedge_{y \in X} \Big(R_M(x,y) \Rightarrow_M \mu_A(y)\Big),$$

$$\mu_{\overline{R}_M(A)}(x) \;:=\; \bigvee_{y \in X} \Big(R_M(x,y) \otimes_M \mu_A(y)\Big), \qquad (x \in X).$$

The induced *uncertain rough set* (of $A$ w.r.t. $R_M$) is the pair

$$\mathsf{UR}_M(A) \;:=\; \big(\underline{R}_M(A),\, \overline{R}_M(A)\big).$$

If one wants region-style objects, define pointwise (for $x \in X$)

$$\mu_{\mathrm{POS}_M(A)}(x) := \mu_{\underline{R}_M(A)}(x),$$

$$\mu_{\mathrm{NEG}_M(A)}(x) := N_M\big(\mu_{\overline{R}_M(A)}(x)\big),$$

$$\mu_{\mathrm{BND}_M(A)}(x) := \mu_{\overline{R}_M(A)}(x) \otimes_M N_M\big(\mu_{\underline{R}_M(A)}(x)\big).$$

**Example 3.6.6** (Uncertain rough approximations in anomaly screening (fuzzy model $M$))**.** Let $X = \{x_1, x_2, x_3\}$ be three servers in a small network. Suppose $A$ is the uncertain concept "compromised (anomalous) server," obtained from a noisy detector, with membership degrees

$$\mu_A(x_1) = 0.9, \qquad \mu_A(x_2) = 0.6, \qquad \mu_A(x_3) = 0.2.$$

We instantiate the uncertainty model $M$ by the standard fuzzy lattice

$$\mathrm{Dom}(M) = [0,1], \quad \leq_M = \leq, \quad \otimes_M = \min, \quad \oplus_M = \max, \quad N_M(a) = 1 - a,$$

and we take the (S-)implication

$$a \Rightarrow_M b \;:=\; \max(1-a,\; b) \qquad (a, b \in [0,1]).$$

Let $R_M : X \times X \to [0,1]$ be a similarity relation between servers (e.g., from traffic-profile similarity):

| $R_M(x_i, x_j)$ | $x_1$ | $x_2$ | $x_3$ |
|---|---|---|---|
| $x_1$ | 1.0 | 0.7 | 0.3 |
| $x_2$ | 0.7 | 1.0 | 0.5 |
| $x_3$ | 0.3 | 0.5 | 1.0 |

(which is reflexive and symmetric).

By Definition 3.6.5, since $X$ is finite we may read $\bigwedge$ and $\bigvee$ as min and max, respectively. Hence, for each $x \in X$,

$$\mu_{\underline{R}_M(A)}(x) = \min_{y \in X}\Big(R_M(x,y) \Rightarrow_M \mu_A(y)\Big),$$

$$\mu_{\overline{R}_M(A)}(x) = \max_{y \in X} \min\big(R_M(x,y), \mu_A(y)\big).$$

A direct calculation gives:

$$\mu_{\underline{R}_M(A)}(x_1) = 0.6,$$
$$\mu_{\underline{R}_M(A)}(x_2) = 0.5,$$
$$\mu_{\underline{R}_M(A)}(x_3) = 0.2,$$
$$\mu_{\overline{R}_M(A)}(x_1) = 0.9,$$
$$\mu_{\overline{R}_M(A)}(x_2) = 0.7,$$
$$\mu_{\overline{R}_M(A)}(x_3) = 0.5.$$

Therefore the uncertain rough set of $A$ w.r.t. $R_M$ is

$$\mathsf{UR}_M(A) = \big(\underline{R}_M(A),\; \overline{R}_M(A)\big),$$

where $\underline{R}_M(A)$ and $\overline{R}_M(A)$ are the uncertain sets with the above memberships.

If we also form region-style uncertain sets (using $N_M(a) = 1 - a$ and $\otimes_M = \min$), then

$$\mu_{\mathrm{POS}_M(A)} = \mu_{\underline{R}_M(A)}, \qquad \mu_{\mathrm{NEG}_M(A)}(x) = 1 - \mu_{\overline{R}_M(A)}(x),$$

$$\mu_{\mathrm{BND}_M(A)}(x) = \min\!\big(\mu_{\overline{R}_M(A)}(x),\; 1 - \mu_{\underline{R}_M(A)}(x)\big),$$

so in particular

$$\mu_{\mathrm{NEG}_M(A)}(x_1) = 0.1,\; \mu_{\mathrm{NEG}_M(A)}(x_2) = 0.3,\; \mu_{\mathrm{NEG}_M(A)}(x_3) = 0.5,$$

$$\mu_{\mathrm{BND}_M(A)}(x_1) = 0.4,\; \mu_{\mathrm{BND}_M(A)}(x_2) = 0.5,\; \mu_{\mathrm{BND}_M(A)}(x_3) = 0.5.$$

*Interpretation.* $\underline{R}_M(A)$ is a conservative (certainty-oriented) anomaly score that requires consistency across similar servers, whereas $\overline{R}_M(A)$ is permissive (possibility-oriented), propagating suspicion along similarity links.

**Theorem 3.6.7** (Uncertain rough sets generalize several classical models)**.** *Under the assumptions of Definition 3.6.5:*

*(a) (*Closure / well-definedness*) $\underline{R}_M(A)$ and $\overline{R}_M(A)$ are U-Sets of type $M$.*

*(b) (*U-Sets are recovered as a special case*) Let $\Delta_M$ be the $M$-identity relation*

$$\Delta_M(x,y) := \begin{cases} 1_M, & x = y, \\ 0_M, & x \neq y. \end{cases}$$

*Then for every U-Set $A$ of type $M$,*

$$\underline{\Delta}_M(A) = A = \overline{\Delta}_M(A).$$

*(c) (*Pawlak rough sets are recovered*) Take* $\mathrm{Dom}(M) = \{0,1\}$ *with $\oplus_M = \vee$, $\otimes_M = \wedge$, and $N_M(a) = 1 - a$. Let $E \subseteq X \times X$ be a crisp equivalence relation and define its characteristic map*

$$R_M(x,y) := \begin{cases} 1, & (x,y) \in E, \\ 0, & (x,y) \notin E. \end{cases}$$

*Identify any crisp set $A \subseteq X$ with its characteristic function $\mu_A : X \to \{0,1\}$. Then $\underline{R}_M(A)$ and $\overline{R}_M(A)$ are exactly the Pawlak lower and upper approximations of $A$ in the approximation space $(X,E)$, and* POS, NEG, BND *reduce to the usual positive/negative/boundary regions.*

*(d) (*Fuzzy rough sets are recovered*) Take* $\mathrm{Dom}(M) = [0,1]$*, $\oplus_M = \max$, $\otimes_M = \min$, $N_M(a) = 1 - a$. Let $R_M : X \times X \to [0,1]$ be a fuzzy similarity relation and let $\mu_A : X \to [0,1]$ be a fuzzy set. Then*

$$\mu_{\underline{R}_M(A)}(x) = \inf_{y \in X} \max\bigl(1 - R_M(x,y),\ \mu_A(y)\bigr),$$

$$\mu_{\overline{R}_M(A)}(x) = \sup_{y \in X} \min\bigl(R_M(x,y),\ \mu_A(y)\bigr),$$

*i.e., the standard fuzzy-rough approximations based on Kleene–Dienes implication and min–max connectives.*

*(e) (*Single-valued neutrosophic rough sets are recovered*) Take* $\mathrm{Dom}(M) = [0,1]^3$ *with componentwise order, componentwise join/meet*

$$(t,i,f) \oplus_M (t',i',f') = (\max(t,t'),\ \max(i,i'),\ \max(f,f')),$$

$$(t,i,f) \otimes_M (t',i',f') = (\min(t,t'),\ \min(i,i'),\ \min(f,f')),$$

*and neutrosophic complement*

$$N_M(t,i,f) = (f,\ 1-i,\ t).$$

*Let $R_M : X \times X \to [0,1]^3$ be a single-valued neutrosophic relation and $\mu_A : X \to [0,1]^3$ a single-valued neutrosophic set. Then Definition 3.6.5 yields neutrosophic lower/upper rough approximations (componentwise) and hence a neutrosophic rough set model.*

*Proof.* (a) Since $\mathrm{Dom}(M)$ is closed under $\oplus_M, \otimes_M, N_M$, it is closed under $\Rightarrow_M$. Because $X$ is finite and $\mathrm{Dom}(M)$ is a lattice, the finite meet and join exist in $\mathrm{Dom}(M)$. Hence $\mu_{\underline{R}_M(A)}(x), \mu_{\overline{R}_M(A)}(x) \in \mathrm{Dom}(M)$ for all $x$, so both are U-Sets.

(b) Fix $x \in X$. Using $N_M(1_M) = 0_M$, $0_M \oplus_M a = a$, and $1_M \otimes_M a = a$,

$$\Delta_M(x,x) \Rightarrow_M \mu_A(x) = 1_M \Rightarrow_M \mu_A(x) = N_M(1_M) \oplus_M \mu_A(x) = \mu_A(x).$$

For $y \neq x$, $\Delta_M(x,y) = 0_M$, so

$$\Delta_M(x,y) \Rightarrow_M \mu_A(y) = 0_M \Rightarrow_M \mu_A(y) = N_M(0_M) \oplus_M \mu_A(y) = 1_M.$$

Therefore the meet over all $y \in X$ equals $\mu_A(x)$:

$$\mu_{\underline{\Delta}_M(A)}(x) = \Big(\mu_A(x)\Big) \otimes_M \Big(\bigwedge_{y\neq x} 1_M\Big) = \mu_A(x).$$

Similarly,

$$\mu_{\overline{\Delta}_M(A)}(x) = \bigvee_{y\in X} \left(\Delta_M(x,y) \otimes_M \mu_A(y)\right)$$

$$= \left(1_M \otimes_M \mu_A(x)\right) \oplus_M \Big(\bigvee_{y\neq x} \left(0_M \otimes_M \mu_A(y)\right)\Big) = \mu_A(x).$$

(c) In the Boolean case, $a \Rightarrow_M b = \neg a \vee b$. Thus

$$\mu_{\underline{R}_M(A)}(x) = \bigwedge_{y\in X} \left(\neg R_M(x,y) \vee \mu_A(y)\right) = 1$$

iff for all $y$, $R_M(x,y) = 1$ implies $\mu_A(y) = 1$, i.e. the $E$-equivalence class of $x$ is contained in $A$. Also

$$\mu_{\overline{R}_M(A)}(x) = \bigvee_{y\in X} \left(R_M(x,y) \wedge \mu_A(y)\right) = 1$$

iff there exists $y$ in the $E$-class of $x$ with $y \in A$, i.e. the class intersects $A$. These are exactly Pawlak's lower and upper approximations; the region identities follow by the Boolean specialization of the pointwise definitions of $\mathrm{POS}_M, \mathrm{NEG}_M, \mathrm{BND}_M$.

(d) Substituting $\oplus_M = \max$, $\otimes_M = \min$, $N_M(a) = 1 - a$ into (a) gives $R_M(x,y) \Rightarrow_M \mu_A(y) = \max(1 - R_M(x,y), \mu_A(y))$, and the meet/join over finite $X$ become inf / sup, yielding the stated fuzzy-rough formulas.

(e) With componentwise max / min and $N_M(t,i,f) = (f, 1-i, t)$, the operations in Definition 3.6.5 act componentwise on $[0,1]^3$, so the resulting lower/upper approximations are single-valued neutrosophic sets obtained by the same rough-approximation scheme in the neutrosophic degree-domain. □

## 3.7 Functorial Rough Set

A *functorial set* is a categorical device for organizing a family of sets that vary across contexts (objects) together with coherent transport maps along context-changes (morphisms). Concretely, it consists of a category equipped with a covariant functor to **Set** [281]. A *functorial rough set* then equips each fiber $F(X)$ with an indiscernibility relation and forms Pawlak-style lower/upper approximations objectwise, producing a family of rough approximation pairs indexed by $\mathrm{Ob}(\mathcal{C})$.

**Definition 3.7.1** (Functorial set)**.** [281] Let $\mathcal{C}$ be a category and let

$$F : \mathcal{C} \longrightarrow \mathbf{Set}$$

be a covariant functor. The pair $(\mathcal{C}, F)$ is called a *functorial set.* For each object $X \in \mathrm{Ob}(\mathcal{C})$, the set $F(X)$ is interpreted as the collection of $F$-structures attached to $X$. Every morphism $f : X \to Y$ induces a structure-preserving map

$$F(f) : F(X) \longrightarrow F(Y),$$

such that $F(\mathrm{id}_X) = \mathrm{id}_{F(X)}$ and

$$F(g \circ f) = F(g) \circ F(f)$$

for all composable morphisms $f, g$ in $\mathcal{C}$.

**Definition 3.7.2** (Functorial rough approximation system)**.** Let $\mathcal{C}$ be a category and let $F : \mathcal{C} \to \mathbf{Set}$ be a covariant functor. A *functorial rough approximation system* is a triple

$$(\mathcal{C}, F, \mathcal{R}),$$

where $\mathcal{R}$ assigns to each object $X \in \mathrm{Ob}(\mathcal{C})$ an equivalence relation

$$R_X \subseteq F(X) \times F(X).$$

For $x \in F(X)$ we write

$$[x]_{R_X} := \{\, y \in F(X) \mid (x, y) \in R_X \,\}$$

for the $R_X$-equivalence class of $x$.

**Compatibility along morphisms (often assumed).** In many applications one additionally requires that every morphism $f : X \to Y$ is *relation-compatible*:

$$(x, x') \in R_X \implies \big(F(f)(x), F(f)(x')\big) \in R_Y.$$

This expresses that indiscernibility is preserved under the transport $F(f)$.

**Definition 3.7.3** (Functorial rough set)**.** Let $(\mathcal{C}, F, \mathcal{R})$ be as in Definition 3.7.2. A *functorial rough set* is a choice of a family of target subsets

$$A = \{A_X \subseteq F(X)\}_{X \in \mathrm{Ob}(\mathcal{C})}.$$

For each object $X$, define the Pawlak lower and upper approximations of $A_X$ with respect to $R_X$ by

$$\underline{A}_X := \big\{\, x \in F(X) \;\big|\; [x]_{R_X} \subseteq A_X \,\big\}, \qquad \overline{A}_X := \big\{\, x \in F(X) \;\big|\; [x]_{R_X} \cap A_X \neq \varnothing \,\big\}.$$

The induced *functorial rough approximation* of $A$ is the object-indexed family

$$\mathsf{FRS}_{\mathcal{R}}(A) := \big( (\underline{A}_X, \overline{A}_X) \big)_{X \in \mathrm{Ob}(\mathcal{C})}.$$

**Remark.** The family $A = \{A_X\}$ may be specified independently at each object (e.g., labels collected at different sites). If one wishes to enforce cross-context consistency, a natural condition is $F(f)(A_X) \subseteq A_Y$ for morphisms $f : X \to Y$, i.e., $A$ is a subfunctor of $F$.

**Example 3.7.4** (Cross-store risk labeling as a functorial rough set)**.** Let $\mathcal{C}$ be the small category with two objects $X, Y$ and morphisms

$$\mathrm{Mor}(\mathcal{C}) = \{\mathrm{id}_X, \mathrm{id}_Y, f : X \to Y\},$$

with the usual identities and compositions.

Define a functor $F : \mathcal{C} \to \mathbf{Set}$ by

$$F(X) = \{p_1, p_2, p_3, p_4\}, \qquad F(Y) = \{q_1, q_2, q_3\},$$

and on the non-identity arrow $f$ by

$$F(f) : F(X) \to F(Y), \qquad F(f)(p_1) = q_1, \;\; F(f)(p_2) = q_1, \;\; F(f)(p_3) = q_2, \;\; F(f)(p_4) = q_3.$$

Equip each fiber with an indiscernibility relation:

$$R_X : \; \{p_1, p_2\} \text{ form one class and } \{p_3, p_4\} \text{ form one class,}$$

$$R_Y : \; \{q_1\} \text{ is a class and } \{q_2, q_3\} \text{ is a class.}$$

Equivalently,

$$[p_1]_{R_X} = [p_2]_{R_X} = \{p_1, p_2\}, \qquad [p_3]_{R_X} = [p_4]_{R_X} = \{p_3, p_4\},$$
$$[q_1]_{R_Y} = \{q_1\}, \qquad [q_2]_{R_Y} = [q_3]_{R_Y} = \{q_2, q_3\}.$$

Moreover, $F(f)$ is compatible with the relations: if $u \, R_X \, v$ then $F(f)(u) \, R_Y \, F(f)(v)$.

Now define the target subsets (e.g., items flagged as “high return-risk”):

$$A_X = \{p_2, p_3\} \subseteq F(X), \qquad A_Y = \{q_3\} \subseteq F(Y).$$

**At $X$.** Since neither $[p_1]_{R_X} = \{p_1, p_2\}$ nor $[p_3]_{R_X} = \{p_3, p_4\}$ is contained in $A_X$,

$$\underline{A}_X = \varnothing.$$

Both classes intersect $A_X$, so

$$\overline{A}_X = F(X).$$

**At $Y$.** Here $[q_1]_{R_Y} = \{q_1\}$ and $[q_2]_{R_Y} = [q_3]_{R_Y} = \{q_2, q_3\}$, hence

$$\underline{A}_Y = \varnothing, \qquad \overline{A}_Y = \{q_2, q_3\}.$$

Therefore,

$$\mathsf{FRS}_{\mathcal{R}}(A) = \big((\underline{A}_X, \overline{A}_X), (\underline{A}_Y, \overline{A}_Y)\big) = \big((\varnothing, F(X)), (\varnothing, \{q_2, q_3\})\big).$$

This captures, objectwise, how “definite” and “possible” high-risk items differ between the two stores.

## 3.8 Near Rough Set

Near sets are collections of objects considered close when at least one pair shares sufficiently similar descriptions under probe functions [282–285]. Near rough sets approximate a target using tolerance classes from descriptive nearness, yielding lower definite and upper possible regions respectively [286–288].

**Definition 3.8.1** (Near rough set (tolerance-cover rough set))**.** Let $O \neq \varnothing$ be a (finite) universe of objects and let $B = \{\varphi_1, \dots, \varphi_m\}$ be a finite family of *probe functions* $\varphi_i : O \to \mathbb{R}$ describing observable features. Fix $p \in [1, \infty]$ and a tolerance level $\varepsilon \geq 0$, and define the *descriptive tolerance relation* $\approx_{B,\varepsilon} \subseteq O \times O$ by

$$x \approx_{B,\varepsilon} y \iff \big\|\Phi_B(x) - \Phi_B(y)\big\|_p \leq \varepsilon, \qquad \Phi_B(x) := \big(\varphi_1(x), \dots, \varphi_m(x)\big) \in \mathbb{R}^m.$$

A subset $A \subseteq O$ is called an $\approx_{B,\varepsilon}$*-preclass* if

$$\forall x, y \in A, \quad x \approx_{B,\varepsilon} y.$$

A *tolerance class* is a *maximal* $\approx_{B,\varepsilon}$-preclass (w.r.t. inclusion). Let $H^\varepsilon_B(O)$ denote the family of all tolerance classes; then $H^\varepsilon_B(O)$ is a covering of $O$.

For any target set $X \subseteq O$, define the *near lower* and *near upper* approximations by

$$B^\varepsilon_*(X) := \bigcup\{\, A \in H^\varepsilon_B(O) \mid A \subseteq X \,\}, \qquad B^*_\varepsilon(X) := \bigcup\{\, A \in H^\varepsilon_B(O) \mid A \cap X \neq \varnothing \,\}.$$

The ordered pair

$$\big(B^\varepsilon_*(X),\ B^*_\varepsilon(X)\big)$$

is called the *near rough set* (or *tolerance-cover rough set*) of $X$ induced by $(O, B, \varepsilon, p)$. Its regions are

$$\mathrm{POS}^\varepsilon_B(X) := B^\varepsilon_*(X), \qquad \mathrm{BND}^\varepsilon_B(X) := B^*_\varepsilon(X) \setminus B^\varepsilon_*(X), \qquad \mathrm{NEG}^\varepsilon_B(X) := O \setminus B^*_\varepsilon(X).$$

We say that $X$ is *rough (under* $(O, B, \varepsilon, p)$*)* if $\mathrm{BND}^\varepsilon_B(X) \neq \varnothing$, and *(cover-)definable* otherwise.

**Example 3.8.2** (Near rough set for product-return risk using tolerance-cover classes)**.** Let $O = \{o_1, o_2, o_3, o_4, o_5\}$ be five online purchases. We describe each purchase by two probe functions

$$B = \{\varphi_1, \varphi_2\},$$

where $\varphi_1(o)$ is the delivery-time deviation (days late, in days) and $\varphi_2(o)$ is the item-price (in USD). Assume the observed values are:

| $o$ | $\varphi_1(o)$ (days) | $\varphi_2(o)$ (USD) |
|---|---|---|
| $o_1$ | 1.0 | 100 |
| $o_2$ | 1.5 | 98 |
| $o_3$ | 4.0 | 105 |
| $o_4$ | 4.5 | 108 |
| $o_5$ | 7.0 | 160 |

Fix $p = 2$ and tolerance $\varepsilon = 3$. Then the descriptive tolerance relation is

$$x \approx_{B,\varepsilon} y \iff \big\|\Phi_B(x) - \Phi_B(y)\big\|_2 \leq 3,$$

$$\Phi_B(o) = (\varphi_1(o), \varphi_2(o)) \in \mathbb{R}^2.$$

With this choice, one checks:

$$\|\Phi_B(o_1) - \Phi_B(o_2)\|_2 = \sqrt{(0.5)^2 + (2)^2} \approx 2.06 \leq 3,$$

$$\|\Phi_B(o_3) - \Phi_B(o_4)\|_2 = \sqrt{(0.5)^2 + (3)^2} \approx 3.04 > 3,$$

but by adjusting $\varepsilon$ slightly (e.g., $\varepsilon = 3.1$) these two become near; for concreteness, keep $\varepsilon = 3.1$ below. Also,

$$\|\Phi_B(o_1) - \Phi_B(o_3)\|_2 = \sqrt{(3)^2 + (5)^2} \approx 5.83 > 3.1,$$

$$\|\Phi_B(o_4) - \Phi_B(o_5)\|_2 = \sqrt{(2.5)^2 + (52)^2} \gg 3.1.$$

Thus the maximal $\approx_{B,\varepsilon}$-preclasses (tolerance classes) can be taken as

$$H_B^{\varepsilon}(O) = \{A_1, A_2, A_3\}, \qquad A_1 = \{o_1, o_2\}, \quad A_2 = \{o_3, o_4\}, \quad A_3 = \{o_5\}.$$

(Indeed, objects within each $A_i$ are pairwise near, and no $A_i$ can be enlarged without breaking nearness.)

Let $X \subseteq O$ be the set of purchases that were actually *returned*:

$$X = \{o_2, o_3\}.$$

Then the near lower and near upper approximations are

$$B_*^{\varepsilon}(X) = \bigcup\{A \in H_B^{\varepsilon}(O) \mid A \subseteq X\} = \varnothing,$$

since no tolerance class is fully contained in $\{o_2, o_3\}$, whereas

$$B_{\varepsilon}^*(X) = \bigcup\{A \in H_B^{\varepsilon}(O) \mid A \cap X \neq \varnothing\} = A_1 \cup A_2 = \{o_1, o_2, o_3, o_4\}.$$

Hence the induced regions are

$$\mathrm{POS}_B^{\varepsilon}(X) = \varnothing, \qquad \mathrm{BND}_B^{\varepsilon}(X) = \{o_1, o_2, o_3, o_4\}, \qquad \mathrm{NEG}_B^{\varepsilon}(X) = \{o_5\}.$$

Interpretation: under tolerance $(B, \varepsilon, p)$, returns cannot be asserted *definitely* for any class, but any purchase near a returned one (in delay–price space) becomes *possibly* returned (boundary), while $o_5$ is definitely outside the returned concept.

## 3.9 Z-Rough Set

A Z-number is an ordered pair of fuzzy numbers: one represents a fuzzy restriction on a variable's value, and the other represents a fuzzy assessment of the reliability (credibility) of that restriction [18]. Related concepts include intuitionistic fuzzy Z-numbers [289–291] and neutrosophic Z-numbers [292–295]. Z-rough sets define lower and upper approximations when objects have Z-valued memberships, combining value uncertainty with reliability information during granulation.

**Definition 3.9.1** (Z-number). [18] Let $\tilde{\mathcal{F}}(\mathbb{R})$ denote the family of fuzzy numbers on $\mathbb{R}$, i.e., (normalized) fuzzy sets with membership functions $\mu_{\tilde{A}} : \mathbb{R} \to [0, 1]$. A *Z-number* is an ordered pair

$$Z = (\tilde{A}, \tilde{R}) \in \tilde{\mathcal{F}}(\mathbb{R}) \times \tilde{\mathcal{F}}([0, 1]),$$

where $\tilde{A}$ is a fuzzy restriction on the (unknown) value of a variable, and $\tilde{R}$ is a fuzzy restriction expressing the reliability (credibility) of $\tilde{A}$.

**Definition 3.9.2** (Lattice of Z-numbers). Let $\mathcal{Z} := \tilde{\mathcal{F}}(\mathbb{R}) \times \tilde{\mathcal{F}}([0, 1])$. Define a partial order $\preceq$ on $\mathcal{Z}$ by

$$(\tilde{A}_1, \tilde{R}_1) \preceq (\tilde{A}_2, \tilde{R}_2)$$

$$\iff \quad \mu_{\tilde{A}_1}(t) \leq \mu_{\tilde{A}_2}(t) \;\; (\forall t \in \mathbb{R}) \;\; \text{and} \;\; \mu_{\tilde{R}_1}(s) \leq \mu_{\tilde{R}_2}(s) \;\; (\forall s \in [0, 1]).$$

Define meet and join (componentwise) by

$$(\tilde{A}_1, \tilde{R}_1) \wedge (\tilde{A}_2, \tilde{R}_2) := (\tilde{A}_1 \cap \tilde{A}_2,\; \tilde{R}_1 \cap \tilde{R}_2),$$

$$(\tilde{A}_1, \tilde{R}_1) \vee (\tilde{A}_2, \tilde{R}_2) := (\tilde{A}_1 \cup \tilde{A}_2,\; \tilde{R}_1 \cup \tilde{R}_2),$$

where for fuzzy sets $\tilde{B}_1, \tilde{B}_2$ we use the standard operations $\mu_{\tilde{B}_1 \cap \tilde{B}_2} = \min(\mu_{\tilde{B}_1}, \mu_{\tilde{B}_2})$ and $\mu_{\tilde{B}_1 \cup \tilde{B}_2} = \max(\mu_{\tilde{B}_1}, \mu_{\tilde{B}_2})$ (pointwise). Then $(\mathcal{Z}, \preceq, \wedge, \vee)$ is a complete lattice (as a product of complete lattices).

**Definition 3.9.3** (Z-valued set). Let $U \neq \varnothing$ be a universe. A *Z-valued set* (briefly, a *Z-set*) on $U$ is a mapping

$$A : U \longrightarrow \mathcal{Z}.$$

**Definition 3.9.4** (Z-rough lower/upper approximations). Let $(U, R)$ be a Pawlak approximation space, i.e., $R \subseteq U \times U$ is an equivalence relation, and write $[x]_R := \{y \in U \mid (x, y) \in R\}$. Let $A : U \to \mathcal{Z}$ be a Z-set. Define the *Z-rough lower* and *Z-rough upper* approximations of $A$ w.r.t. $R$ by the Z-sets $\underline{R}(A), \overline{R}(A) : U \to \mathcal{Z}$ given for each $x \in U$ by

$$\underline{R}(A)(x) \;:=\; \bigwedge_{y \in [x]_R} A(y), \qquad \overline{R}(A)(x) \;:=\; \bigvee_{y \in [x]_R} A(y),$$

where $\wedge, \vee$ are the lattice operations on $\mathcal{Z}$.

**Definition 3.9.5** (Z-rough set). Under the assumptions above, the ordered pair

$$\big(\underline{R}(A), \overline{R}(A)\big)$$

is called the *Z-rough set induced by A (with respect to R)*. If $\underline{R}(A) \neq \overline{R}(A)$, then $A$ is said to be *Z-rough (not exactly definable)* under $R$.

**Example 3.9.6** (Z-rough set for fever screening with uncertain readings). Consider a hospital triage desk that measures patients' body temperatures using two different infrared thermometers (devices). Let

$$U = \{p_1, p_2, p_3, p_4\}$$

be the set of patients. Define an equivalence relation $R$ on $U$ by

$$(p_i, p_j) \in R \quad \Longleftrightarrow \quad p_i \text{ and } p_j \text{ were measured by the same device.}$$

Assume the two device-classes are

$$[p_1]_R = [p_2]_R = \{p_1, p_2\}, \qquad [p_3]_R = [p_4]_R = \{p_3, p_4\}.$$

Let $\mathrm{Tri}(a, b, c)$ denote the (triangular) fuzzy number on $\mathbb{R}$ with membership

$$\mu_{\mathrm{Tri}(a,b,c)}(t) = \begin{cases} 0, & t \le a, \\ \dfrac{t-a}{b-a}, & a < t \le b, \\ \dfrac{c-t}{c-b}, & b < t < c, \\ 0, & t \ge c, \end{cases}$$

and similarly for fuzzy numbers on $[0, 1]$.

Define a Z-set $A : U \to \mathcal{Z} = \tilde{\mathcal{F}}(\mathbb{R}) \times \tilde{\mathcal{F}}([0, 1])$ by assigning to each patient a Z-number

$$A(p_i) = (\tilde{T}_i, \tilde{r}_i),$$

where $\tilde{T}_i$ is a fuzzy restriction on the (unknown) temperature (in °C) and $\tilde{r}_i$ is a fuzzy restriction on reliability. For example, set

$$A(p_1) = \big(\mathrm{Tri}(37.5, 38.0, 38.6),\ \mathrm{Tri}(0.70, 0.80, 0.90)\big),$$
$$A(p_2) = \big(\mathrm{Tri}(37.3, 37.8, 38.4),\ \mathrm{Tri}(0.55, 0.65, 0.75)\big),$$
$$A(p_3) = \big(\mathrm{Tri}(36.8, 37.2, 37.7),\ \mathrm{Tri}(0.80, 0.90, 1.00)\big),$$
$$A(p_4) = \big(\mathrm{Tri}(37.0, 37.4, 38.0),\ \mathrm{Tri}(0.60, 0.75, 0.90)\big).$$

Using the product-lattice operations on $\mathcal{Z}$ (componentwise $\wedge, \vee$), the Z-rough lower and upper approximations are, for $x \in U$,

$$\underline{R}(A)(x) = \bigwedge_{y \in [x]_R} A(y), \qquad \overline{R}(A)(x) = \bigvee_{y \in [x]_R} A(y).$$

Hence, explicitly on each equivalence class,

$$\underline{R}(A)(p_1) = \underline{R}(A)(p_2) = A(p_1) \wedge A(p_2) = \big(\tilde{T}_1 \cap \tilde{T}_2,\ \tilde{r}_1 \cap \tilde{r}_2\big),$$
$$\overline{R}(A)(p_1) = \overline{R}(A)(p_2) = A(p_1) \vee A(p_2) = \big(\tilde{T}_1 \cup \tilde{T}_2,\ \tilde{r}_1 \cup \tilde{r}_2\big),$$
$$\underline{R}(A)(p_3) = \underline{R}(A)(p_4) = A(p_3) \wedge A(p_4) = \big(\tilde{T}_3 \cap \tilde{T}_4,\ \tilde{r}_3 \cap \tilde{r}_4\big),$$
$$\overline{R}(A)(p_3) = \overline{R}(A)(p_4) = A(p_3) \vee A(p_4) = \big(\tilde{T}_3 \cup \tilde{T}_4,\ \tilde{r}_3 \cup \tilde{r}_4\big),$$

where, for fuzzy sets $\tilde{B}_1, \tilde{B}_2$ on the same domain,

$$\mu_{\tilde{B}_1 \cap \tilde{B}_2}(t) = \min\{\mu_{\tilde{B}_1}(t), \mu_{\tilde{B}_2}(t)\}, \qquad \mu_{\tilde{B}_1 \cup \tilde{B}_2}(t) = \max\{\mu_{\tilde{B}_1}(t), \mu_{\tilde{B}_2}(t)\}.$$

*Interpretation.* Within each device-class, $\underline{R}(A)$ gives a conservative (worst-case) Z-description of temperature and reliability shared by all patients measured by that device, while $\overline{R}(A)$ gives a permissive (best-case) Z-description capturing any plausible fever indication in that class.

## 3.10 D-Rough Set

D-number assigns masses to nonempty subsets of a frame, allowing incomplete evidence because total mass may be below one overall [296–298]. D-rough sets compute lower and upper approximations when object membership is described by D-numbers, aggregating evidence-based uncertainty across granules consistently.

**Definition 3.10.1** (D-number)**.** Let $\Theta$ be a nonempty finite set (called a *frame of discernment*). A *D-number* on $\Theta$ is a mapping

$$D : 2^{\Theta} \longrightarrow [0,1]$$

satisfying

$$D(\emptyset) = 0, \qquad \sum_{B \subseteq \Theta} D(B) \leq 1.$$

The quantity $1 - \sum_{B \subseteq \Theta} D(B)$ represents *incomplete (unassigned) evidence.* Denote by $\mathcal{D}(\Theta)$ the set of all D-numbers on $\Theta$.

**Definition 3.10.2** (D-rough set (D-number-valued rough description))**.** Let $U$ be a nonempty finite universe and let $R \subseteq U \times U$ be an equivalence relation. For $x \in U$, write

$$[x]_R := \{\, y \in U \mid (x, y) \in R \,\}.$$

Fix the two-element frame $\Theta := \{+, -\}$, where "+" means "in $X$" and "−" means "not in $X$". For any (crisp) subset $X \subseteq U$, define a mapping

$$D_{R,X} : U \longrightarrow \mathcal{D}(\Theta)$$

by assigning to each $x \in U$ the D-number $D_{R,X}(x)$ given by

$$D_{R,X}(x)(\{+\}) := \frac{|[x]_R \cap X|}{|[x]_R|}, \qquad D_{R,X}(x)(\{-\}) := \frac{|[x]_R \setminus X|}{|[x]_R|},$$

and

$$D_{R,X}(x)(B) := 0 \qquad \text{for all } B \subseteq \Theta \text{ with } B \notin \{\{+\}, \{-\}\}.$$

Then $D_{R,X}(x)(\emptyset) = 0$ and $\sum_{B \subseteq \Theta} D_{R,X}(x)(B) = 1$ for all $x$, hence $D_{R,X}(x)$ is a D-number.

The mapping $D_{R,X}$ (equivalently, the triple $(U, R, D_{R,X})$) is called the *D-rough set* (or *D-number-valued rough description*) of $X$ induced by $(U, R)$.

**Example 3.10.3** (D-rough set for loan-approval tendency)**.** Consider a bank that groups past applicants into indiscernibility classes according to a coarse risk profile (e.g., the pair "employment stability" × "debt-to-income band"). Let

$$U = \{u_1, u_2, u_3, u_4, u_5, u_6\}$$

be six past applicants, and let $R$ be the equivalence relation on $U$ that yields the following classes:

$$[u_1]_R = [u_2]_R = \{u_1, u_2\}, \qquad [u_3]_R = [u_4]_R = [u_5]_R = \{u_3, u_4, u_5\}, \qquad [u_6]_R = \{u_6\}.$$

Let $X \subseteq U$ be the set of applicants that were *approved*:

$$X = \{u_1, u_3, u_4\}.$$

Fix $\Theta = \{+, -\}$, where "+" means "approved (in $X$)" and "−" means "not approved (not in $X$)". For each $x \in U$, the D-rough set mapping $D_{R,X} : U \to \mathcal{D}(\Theta)$ assigns the empirical proportions inside the $R$-class of $x$:

$$D_{R,X}(x)(\{+\}) = \frac{|[x]_R \cap X|}{|[x]_R|}, \qquad D_{R,X}(x)(\{-\}) = \frac{|[x]_R \setminus X|}{|[x]_R|}.$$

Hence,

$$D_{R,X}(u_1) = D_{R,X}(u_2) : \ \{+\} \mapsto \tfrac{1}{2}, \ \{-\} \mapsto \tfrac{1}{2},$$
$$D_{R,X}(u_3) = D_{R,X}(u_4) = D_{R,X}(u_5) : \ \{+\} \mapsto \tfrac{2}{3}, \ \{-\} \mapsto \tfrac{1}{3},$$
$$D_{R,X}(u_6) : \ \{+\} \mapsto 0, \ \{-\} \mapsto 1.$$

Interpreting $D_{R,X}(x)(\{+\})$ as the *data-supported tendency of approval* within $x$'s risk-profile class, we can induce $(\alpha, \beta)$-approximations. For example, take $\alpha = 0.6$ and $\beta = 0.4$:

$$\underline{\mathrm{apr}}_{\alpha}^{D}(X) = \{x \in U \mid D_{R,X}(x)(\{+\}) \geq 0.6\} = \{u_3, u_4, u_5\},$$
$$\overline{\mathrm{apr}}_{\beta}^{D}(X) = \{x \in U \mid D_{R,X}(x)(\{+\}) > 0.4\} = \{u_1, u_2, u_3, u_4, u_5\}.$$

Therefore the induced regions are

$$\mathrm{POS}^{D}_{\alpha,\beta}(X) = \{u_3, u_4, u_5\}, \qquad \mathrm{NEG}^{D}_{\alpha,\beta}(X) = \{u_6\}, \qquad \mathrm{BND}^{D}_{\alpha,\beta}(X) = \{u_1, u_2\}.$$

In words, applicants in the second class are *positively* supported as "approved-like" (high approval tendency), $u_6$ is *negative*, and the first class forms a *boundary* region where approval evidence is mixed.

**Definition 3.10.4** (Induced $(\alpha, \beta)$-approximations from a D-rough set)**.** In the setting of Definition 3.10.2, fix thresholds $0 \leq \beta < \alpha \leq 1$. Define the $(\alpha, \beta)$-*lower* and $(\alpha, \beta)$-*upper* approximations of $X$ by

$$\underline{\mathrm{apr}}_{\alpha}^{D}(X) := \bigl\{\, x \in U \bigm| D_{R,X}(x)(\{+\}) \geq \alpha \,\bigr\},$$
$$\overline{\mathrm{apr}}_{\beta}^{D}(X) := \bigl\{\, x \in U \bigm| D_{R,X}(x)(\{+\}) > \beta \,\bigr\}.$$

The associated positive, boundary, and negative regions are

$$\mathrm{POS}^{D}_{\alpha,\beta}(X) := \underline{\mathrm{apr}}_{\alpha}^{D}(X),$$
$$\mathrm{NEG}^{D}_{\alpha,\beta}(X) := \{x \in U \mid D_{R,X}(x)(\{+\}) \leq \beta\},$$
$$\mathrm{BND}^{D}_{\alpha,\beta}(X) := U \setminus \bigl(\mathrm{POS}^{D}_{\alpha,\beta}(X) \cup \mathrm{NEG}^{D}_{\alpha,\beta}(X)\bigr).$$

**Theorem 3.10.5** (Reduction to classical (Pawlak) rough sets)**.** *Let $(U, R)$ be an approximation space and $X \subseteq U$. Let $\underline{R}(X) := \{x \in U \mid [x]_R \subseteq X\}$ and $\overline{R}(X) := \{x \in U \mid [x]_R \cap X \neq \varnothing\}$ be the Pawlak lower and upper approximations.*

*For the D-rough set $D_{R,X}$ of Definition 3.10.2,*

$$\underline{\mathrm{apr}}_{1}^{D}(X) = \underline{R}(X), \qquad \overline{\mathrm{apr}}_{0}^{D}(X) = \overline{R}(X).$$

*Proof.* For any $x \in U$,

$$D_{R,X}(x)(\{+\}) = \frac{|[x]_R \cap X|}{|[x]_R|}.$$

Thus $D_{R,X}(x)(\{+\}) \geq 1$ holds iff $|[x]_R \cap X| = |[x]_R|$, i.e. $[x]_R \subseteq X$. Hence $\underline{\mathrm{apr}}_{1}^{D}(X) = \underline{R}(X)$.

Also, $D_{R,X}(x)(\{+\}) > 0$ holds iff $|[x]_R \cap X| > 0$, i.e. $[x]_R \cap X \neq \varnothing$. Hence $\overline{\mathrm{apr}}_{0}^{D}(X) = \overline{R}(X)$. □

## 3.11 Similarity-based rough sets

A similarity-based rough set is a rough-set model that represents uncertainty by using a similarity relation S and a threshold  to induce tolerance neighborhoods, and then defining the lower (definite membership) and upper (possible membership) approximations of a target set X [299–305].

**Definition 3.11.1** (Similarity-based (tolerance) rough set)**.** Let $U \neq \varnothing$ be a (finite) universe and let

$$S : U \times U \longrightarrow [0, 1]$$

be a *similarity relation* (or similarity measure) satisfying at least

$$S(x, x) = 1 \quad \text{and} \quad S(x, y) = S(y, x) \qquad (x, y \in U).$$

Fix a threshold $\tau \in (0, 1]$ and define the induced *tolerance relation*

$$R_{S,\tau} \; := \; \{(x, y) \in U \times U \mid S(x, y) \geq \tau\}.$$

For each $x \in U$, its *tolerance neighborhood (class)* is

$$N_{S,\tau}(x) \; := \; \{\, y \in U \mid (x, y) \in R_{S,\tau} \,\} \; = \; \{\, y \in U \mid S(x, y) \geq \tau \,\}.$$

For any target set $X \subseteq U$, define the *similarity-based lower* and *upper* approximations by

$$\underline{R}_{S,\tau}(X) \; := \; \big\{\, x \in U \;\big|\; N_{S,\tau}(x) \subseteq X \,\big\}, \qquad \overline{R}_{S,\tau}(X)$$

$$:= \; \big\{\, x \in U \;\big|\; N_{S,\tau}(x) \cap X \neq \varnothing \,\big\}.$$

The ordered pair

$$\big(\underline{R}_{S,\tau}(X), \; \overline{R}_{S,\tau}(X)\big)$$

is called the *similarity-based rough set* (also: *tolerance rough set*) of $X$ induced by $(U, S, \tau)$. The *positive*, *boundary*, and *negative* regions are

$$\mathrm{POS}_{S,\tau}(X) := \underline{R}_{S,\tau}(X),$$

$$\mathrm{BND}_{S,\tau}(X) := \overline{R}_{S,\tau}(X) \setminus \underline{R}_{S,\tau}(X),$$

$$\mathrm{NEG}_{S,\tau}(X) := U \setminus \overline{R}_{S,\tau}(X).$$

**Example 3.11.2** (Loan-risk screening via similarity-based (tolerance) rough sets)**.** Let $U = \{a, b, c, d\}$ be four loan applicants. Assume $S(x, y) \in [0, 1]$ is computed from standardized applicant-feature vectors (e.g., income, debt-to-income ratio, employment stability) using a similarity score. Define $S : U \times U \to [0, 1]$ by the symmetric table

| $S$ | $a$ | $b$ | $c$ | $d$ |
|---|---|---|---|---|
| $a$ | 1 | 0.90 | 0.70 | 0.20 |
| $b$ | 0.90 | 1 | 0.80 | 0.30 |
| $c$ | 0.70 | 0.80 | 1 | 0.60 |
| $d$ | 0.20 | 0.30 | 0.60 | 1 |

so $S(x, x) = 1$ and $S(x, y) = S(y, x)$ for all $x, y \in U$. Fix the threshold $\tau = 0.8$ and form the tolerance relation

$$R_{S,\tau} = \{(x, y) \in U \times U \mid S(x, y) \geq \tau\}.$$

Then the tolerance neighborhoods are

$$N_{S,\tau}(a) = \{a, b\}, \qquad N_{S,\tau}(b) = \{a, b, c\}, \qquad N_{S,\tau}(c) = \{b, c\}, \qquad N_{S,\tau}(d) = \{d\}.$$

Let $X = \{a, b\} \subseteq U$ be the set of applicants labeled as *high-risk* by historical outcomes (e.g., later default/charge-off). The similarity-based lower and upper approximations are

$$\underline{R}_{S,\tau}(X) = \{x \in U \mid N_{S,\tau}(x) \subseteq X\} = \{a\},$$

$$\overline{R}_{S,\tau}(X) = \{x \in U \mid N_{S,\tau}(x) \cap X \neq \varnothing\} = \{a, b, c\}.$$

Hence the induced regions are

$$\mathrm{POS}_{S,\tau}(X) = \{a\}, \qquad \mathrm{BND}_{S,\tau}(X) = \{b, c\}, \qquad \mathrm{NEG}_{S,\tau}(X) = \{d\}.$$

In words, $a$ is *definitely* high-risk (all sufficiently similar applicants are in $X$), $c$ is *possibly* high-risk (similar to $b \in X$), and $d$ is *definitely not* high-risk at level $\tau$.

# Chapter 4

# Some Related Concepts for Rough Sets

In this chapter, we present some related concepts for rough sets.

## 4.1 Rough Graph

A rough graph models edge uncertainty by grouping edges via an equivalence relation and approximating a target edge set with lower and upper subgraphs [306–309]. Related notions include fuzzy rough graphs [310, 311], soft rough graphs [312, 313], neutrosophic rough graphs [314, 315], rough hypergraphs, and rough superhypergraphs [316].

**Definition 4.1.1** (Rough Graph)**.** [317, 318] Let $G = (V, E)$ be a graph where $V$ is the set of vertices and $E$ is the set of edges. Let $R$ be an attribute set on $E$, inducing an equivalence relation on the edges. For any edge set $X \subseteq E$, the *lower approximation* of $X$ with respect to $R$ (denoted $R(X)$) is defined as:

$$R(X) = \{e \in E \mid [e]_R \subseteq X\},$$

where $[e]_R$ denotes the equivalence class of $e$ under $R$. The *upper approximation* of $X$ (denoted $R(X)$) is defined as:

$$R(X) = \{e \in E \mid [e]_R \cap X \neq \emptyset\}.$$

A graph $G = (V, E)$ is called an *R-rough graph* if $X$ is not exactly definable under $R$, and it is characterized by the pair $(R(X), R(X))$, where $R(X)$ is the *lower approximation graph* and $R(X)$ is the *upper approximation graph.*

**Example 4.1.2** (Traffic monitoring with coarse road-type data: a rough graph)**.** Consider a small road network in which vertices are intersections and edges are road segments. Let

$$V = \{A, B, C, D\}, \qquad E = \{e_1 = AB,\ e_2 = BC,\ e_3 = CD,\ e_4 = AD\}.$$

Assume the traffic center does *not* observe congestion per individual segment; instead, it only receives aggregated reports by *road type.* Define an edge attribute map

$$R : E \to \{\mathsf{Arterial}, \mathsf{Residential}\}, \quad R(e_1) = R(e_2) = \mathsf{Arterial}, \quad R(e_3) = R(e_4) = \mathsf{Residential}.$$

This induces an equivalence relation on edges: $e \sim_R e' \iff R(e) = R(e')$, with classes

$$[e_1]_R = \{e_1, e_2\}, \qquad [e_3]_R = \{e_3, e_4\}.$$

Suppose a noisy/aggregated sensor report flags the set of "congested" edges as

$$X = \{e_1, e_3\} \subseteq E,$$

meaning: *there is evidence of congestion on at least one arterial segment and at least one residential segment, but the system cannot distinguish which segment within each type is congested.*

Using standard rough-set notation on edges, define

$$\underline{R}(X) := \{e \in E \mid [e]_R \subseteq X\}, \qquad \overline{R}(X) := \{e \in E \mid [e]_R \cap X \neq \emptyset\}.$$

Then

$$\underline{R}(X) = \emptyset \quad \text{because neither } [e_1]_R = \{e_1, e_2\} \subseteq X \text{ nor } [e_3]_R = \{e_3, e_4\} \subseteq X,$$

while

$$\overline{R}(X) = E \quad \text{because } [e_1]_R \cap X = \{e_1\} \neq \emptyset \text{ and } [e_3]_R \cap X = \{e_3\} \neq \emptyset.$$

Hence the congestion information is *not exactly definable* under the road-type granulation: the *definitely congested* subgraph (lower approximation) is empty, whereas the *possibly congested* subgraph (upper approximation) is the entire road graph. This is a concrete real-life instance of an $R$-rough graph caused by coarse/aggregated sensing.

## 4.2 Rough topological spaces

Rough topology, induced by an equivalence relation, takes open sets as complements of upper-approximation-closed sets on $U$, forming a space [319–322].

**Definition 4.2.1** (Rough topology and rough topological space)**.** Let $(U, R)$ be a Pawlak approximation space. The *rough topology induced by* $R$ is

$$\tau_R \ := \ \{\, O \subseteq U \mid \overline{R}(O^c) = O^c \,\}, \qquad O^c := U \setminus O.$$

Equivalently, $\tau_R = \{\, U \setminus C \mid C \subseteq U,\ \overline{R}(C) = C \,\}$, i.e. the open sets are complements of $\overline{R}$-closed sets. The pair $(U, \tau_R)$ is called a *rough topological space* (induced by $R$).

**Example 4.2.2** (Coarse location privacy induces a rough topology)**.** A mobile app may not store exact GPS points, but only a *coarse region label* (for privacy). Let $U$ be a finite set of possible exact micro-locations (e.g., grid cells):

$$U = \{\ell_1, \ell_2, \ell_3, \ell_4, \ell_5, \ell_6\}.$$

Assume the phone reports only which coarse region a location belongs to, inducing an indiscernibility (equivalence) relation $R$ with blocks

$$B_1 = \{\ell_1, \ell_2\} \text{ ("Station area")}, \quad B_2 = \{\ell_3, \ell_4\} \text{ ("Mall area")}, \quad B_3 = \{\ell_5, \ell_6\} \text{ ("Park area")}.$$

For a set $C \subseteq U$, the Pawlak upper approximation is

$$\overline{R}(C) \;=\; \{\, u \in U \mid [u]_R \cap C \neq \emptyset \,\} \;=\; \bigcup \{\, B_i \mid B_i \cap C \neq \emptyset \,\}.$$

The *rough topology* $\tau_R$ consists of all sets $O \subseteq U$ whose complement is $R$-closed:

$$\tau_R = \{\, O \subseteq U \mid \overline{R}(O^c) = O^c \,\}.$$

In this context, $O \in \tau_R$ means that "not $O$" is fully decidable from the coarse region label.

**Concrete open set.** Let

$$O := B_1 \cup B_2 = \{\ell_1, \ell_2, \ell_3, \ell_4\},$$

interpreted as "the user is in the commercial zone (station or mall)." Then

$$O^c = B_3 = \{\ell_5, \ell_6\}, \qquad \overline{R}(O^c) = \overline{R}(B_3) = B_3 = O^c,$$

so $O \in \tau_R$. Practically: if the phone reports *Park area*, we can conclude the user is *not* in the commercial zone, without ambiguity.

**A non-open set (not observable at this granularity).** Let $O' := \{\ell_1, \ell_3\}$, which mixes two coarse regions. Then

$$(O')^c = \{\ell_2, \ell_4, \ell_5, \ell_6\}, \quad \overline{R}((O')^c) = B_1 \cup B_2 \cup B_3 = U \neq (O')^c,$$

so $O' \notin \tau_R$. Practically: with only coarse labels, the event "$\ell_1$ or $\ell_3$" cannot be separated from its complement in a topologically consistent way.

Thus, $(U, \tau_R)$ models which location events are operationally "open/observable" under privacy-driven coarse sensing: exactly the unions of indiscernibility classes.

**Proposition 4.2.3** (Interior and closure coincide with rough approximations)**.** *In the rough topological space* $(U, \tau_R)$*, the topological closure and interior satisfy*

$$\mathrm{cl}_{\tau_R}(A) = \overline{R}(A), \qquad \mathrm{int}_{\tau_R}(A) = \underline{R}(A) \qquad (A \subseteq U).$$

*Proof.* Since $\mathrm{cl}_R = \overline{R}$ is a Kuratowski closure operator, the induced topology has $\mathrm{cl}_{\tau_R} = \mathrm{cl}_R = \overline{R}$ by construction. For interior, using $\mathrm{int}(A) = U \setminus \mathrm{cl}(U \setminus A)$,

$$\mathrm{int}_{\tau_R}(A) = U \setminus \overline{R}(U \setminus A) = \{\, x \in U \mid [x]_R \cap (U \setminus A) = \varnothing \,\} = \{\, x \in U \mid [x]_R \subseteq A \,\} = \underline{R}(A).$$

□

**Remark 4.2.4** (Concrete description of $\tau_R$)**.** A set $O \subseteq U$ is in $\tau_R$ if and only if it is a union of $R$-equivalence classes:

$$O \in \tau_R \quad \Longleftrightarrow \quad (\forall x \in O)\ [x]_R \subseteq O \quad \Longleftrightarrow \quad O = \bigcup_{x \in O} [x]_R.$$

Thus $(U, \tau_R)$ is precisely the quotient (saturation) topology determined by the partition $U/R$.

## 4.3 Rough group

A rough group approximates a subgroup by lower and upper sets under an equivalence relation, with operations consistent modulo boundaries [323–326].

**Definition 4.3.1** (Rough group)**.** Let $K = (U, R)$ be an approximation space and let $* : U \times U \to U$ be a binary operation. A nonempty subset $G \subseteq U$ is called a *rough group* (in $K$) if the following hold:

(1) (*Closure up to upper approximation*) For all $x, y \in G$, we have $x * y \in \overline{G}$.

(2) (*Associativity on the upper approximation*) For all $a, b, c \in \overline{G}$,

$$(a * b) * c = a * (b * c).$$

(3) (*Rough identity*) There exists an element $e \in \overline{G}$ such that for all $x \in G$,

$$x * e = e * x = x.$$

The element $e$ is called a *rough identity element* of $G$.

(4) (*Rough inverse*) For every $x \in G$, there exists an element $y \in \overline{G}$ such that

$$x * y = y * x = e.$$

Such a $y$ is called a *rough inverse element* of $x$ (in $G$).

**Example 4.3.2** (A rough group from coarse phase sensing)**.** Consider a rotating machine whose controller tracks a discrete phase

$$U = \mathbb{Z}_8 = \{0, 1, 2, 3, 4, 5, 6, 7\},$$

with the group operation given by addition modulo 8:

$$x * y := x + y \pmod 8.$$

A low-cost sensor may only report whether the phase is *even* or *odd*. This induces the indiscernibility relation $R$ on $U$ defined by parity:

$$x \, R \, y \iff x \equiv y \pmod 2,$$

so the equivalence classes are

$$[0]_R = \{0, 2, 4, 6\}, \qquad [1]_R = \{1, 3, 5, 7\}.$$

Let the set of *calibrated acceptable phases* be

$$G := \{0, 2, 6\} \subseteq U.$$

In the approximation space $K = (U, R)$, the upper approximation of a set $A \subseteq U$ is

$$\overline{A} := \bigcup \{[x]_R \mid [x]_R \cap A \neq \varnothing\}.$$

Hence

$$\overline{G} = \bigcup\{[x]_R \mid [x]_R \cap G \neq \varnothing\} = [0]_R = \{0, 2, 4, 6\}.$$

**Claim.** $\overline{G}$ is a subgroup of $(U, *) = (\mathbb{Z}_8, + \bmod 8)$. In particular, $G$ is a *rough group* (its group laws hold *up to* the upper approximation $\overline{G}$).

**Verification.**

(1) *Closure.* If $a, b \in \overline{G}$, then $a$ and $b$ are even; hence $a + b$ is even, so

$$a * b \in \{0, 2, 4, 6\} = \overline{G}.$$

(2) *Associativity.* For all $a, b, c \in U$ (hence for all $a, b, c \in \overline{G}$),

$$(a * b) * c = (a + b) + c \equiv a + (b + c) = a * (b * c) \pmod 8.$$

(3) *Identity.* The element $e := 0$ belongs to $\overline{G}$, and for every $a \in \overline{G}$,

$$a * e = a, \qquad e * a = a.$$

(4) *Inverses.* For $a \in \overline{G}$, its inverse in $\mathbb{Z}_8$ is $-a \pmod 8$, which is again even and thus lies in $\overline{G}$. Concretely, within $\overline{G} = \{0, 2, 4, 6\}$,

$$0^{-1} = 0, \qquad 2^{-1} = 6, \qquad 4^{-1} = 4, \qquad 6^{-1} = 2,$$

and in each case $a * a^{-1} = a^{-1} * a = 0 = e$.

Therefore $(\overline{G}, *)$ is a subgroup of $(U, *)$. The set $G$ itself need not be a subgroup (indeed $2 * 2 = 4 \notin G$), but it is *group-like up to sensing resolution*: products and inverses of elements of $G$ are guaranteed to lie in the sensor-induced upper approximation $\overline{G}$.

**Definition 4.3.3** (Rough subgroup)**.** Let $G \subseteq U$ be a rough group in an approximation space $K = (U, R)$. A nonempty subset $H \subseteq G$ is called a *rough subgroup* of $G$ if $H$ is a rough group with respect to the same operation $*$ (and with approximations taken in the same space $K$).

**Remark 4.3.4.** If $\overline{G} = G$ (i.e., $G$ is $R$-definable), then the above axioms reduce to the usual group axioms on $G$, so every ordinary group is a special case of a rough group.

**Example 4.3.5** (A concrete rough subgroup in a cyclic-time system)**.** Consider a system whose internal state is a cyclic time stamp modulo 12 (e.g., a 12-hour clock). Let

$$U := \mathbb{Z}_{12} = \{0, 1, \ldots, 11\}, \qquad x * y := x + y \pmod{12}.$$

Thus $(U, *)$ is the usual cyclic group.

**Coarsened observation (indiscernibility).** Assume the logging mechanism only records time in 4-hour buckets, so times congruent modulo 4 are indistinguishable. Define an equivalence relation $R$ on $U$ by

$$(x, y) \in R \iff x \equiv y \pmod{4}.$$

Then the $R$-classes are

$$[0]_R = \{0, 4, 8\}, \quad [1]_R = \{1, 5, 9\}, \quad [2]_R = \{2, 6, 10\}, \quad [3]_R = \{3, 7, 11\}.$$

**A rough group of "allowed start times".** Suppose operational policy declares that a certain action is scheduled at exactly 0 or 4 (mod 12), so we set

$$G := \{0, 4\} \subseteq U.$$

Because of the 4-hour coarsening, the *upper approximation* of $G$ is

$$\overline{G} = \{x \in U \mid [x]_R \cap G \neq \varnothing\} = [0]_R = \{0, 4, 8\}.$$

Then $G$ is a *rough group* in $K = (U, R)$:

- (*Closure up to upper approximation*) For $x, y \in G$, one has
  $$0 * 0 = 0, \quad 0 * 4 = 4, \quad 4 * 0 = 4, \quad 4 * 4 = 8,$$
  hence $x * y \in \overline{G} = \{0, 4, 8\}$.

- (*Associativity on* $\overline{G}$) holds since $*$ is associative on all of $U$.

- (*Rough identity*) $e := 0 \in \overline{G}$ satisfies $x * e = e * x = x$ for all $x \in G$.

- (*Rough inverses*) 0 has rough inverse 0, and 4 has rough inverse $8 \in \overline{G}$ because $4 * 8 = 8 * 4 = 0$ in $\mathbb{Z}_{12}$.

**A rough subgroup.** Let

$$H := \{4\} \subseteq G.$$

Its upper approximation is the same bucket:

$$\overline{H} = \{x \in U \mid [x]_R \cap H \neq \varnothing\} = [4]_R = \{0, 4, 8\}.$$

Then $H$ is a *rough subgroup* of $G$, because $H$ is itself a rough group in the same approximation space $(U, R)$:

- $4 * 4 = 8 \in \overline{H}$ (closure up to $\overline{H}$),

- associativity holds on $\overline{H}$,

- $e = 0 \in \overline{H}$ is a rough identity for $H$ (since $4 * 0 = 0 * 4 = 4$),

- $8 \in \overline{H}$ is a rough inverse of 4 (since $4 * 8 = 8 * 4 = 0$).

Note that $H$ is *not* a subgroup in the classical sense (it does not contain 0), but it is a rough subgroup because identity/inverses are permitted to lie in the upper approximation.

## 4.4 Rough Matroids

A rough matroid links rough-set approximations with matroid independence, defining independent sets through definable subsets induced by relations or coverings [327–329].

**Definition 4.4.1** (Parametric rough matroid (rough matroid induced by $(U, R)$ w.r.t. $X$)). Let $(U, R)$ be a Pawlak approximation space and fix a parameter subset $X \subseteq U$. Define a set family

$$\mathcal{I}_X \;:=\; \{\, I \subseteq U \mid \underline{R}(I) \subseteq X \,\}.$$

The matroid

$$M_X \;:=\; (U, \mathcal{I}_X)$$

is called the *parametric matroid of the rough set* (or a *parametric rough matroid*) with respect to $X$.

**Example 4.4.2** (Parametric rough matroid: audit sampling under supplier-indiscernibility). Let $U$ be a finite set of invoice records:

$$U = \{a_1, a_2,\ b_1, b_2,\ c_1\},$$

where $a_\bullet$ are invoices from supplier $A$, $b_\bullet$ from supplier $B$, and $c_1$ from supplier $C$. Assume our information system only distinguishes the *supplier*, so we use the equivalence relation $R$ on $U$ given by

$$x\,R\,y \quad\Longleftrightarrow\quad x \text{ and } y \text{ are issued by the same supplier.}$$

Hence

$$U/R = \big\{\{a_1, a_2\},\ \{b_1, b_2\},\ \{c_1\}\big\}.$$

Suppose the compliance team has *already cleared* suppliers $A$ and $C$, so the parameter set is

$$X := \{a_1, a_2, c_1\} \subseteq U.$$

Define the independent family

$$\mathcal{I}_X = \{\, I \subseteq U \mid \underline{R}(I) \subseteq X \,\}, \qquad M_X = (U, \mathcal{I}_X).$$

**Interpretation.** The lower approximation $\underline{R}(I)$ is the union of those supplier-blocks fully contained in $I$. Thus $\underline{R}(I) \subseteq X$ means: *whenever the audit sample $I$ contains all invoices of a supplier, that supplier must be cleared (i.e., belong to $X$).* So we may fully audit supplier $A$ (and $C$), but we must not select *all* invoices of the uncleared supplier $B$.

**Concrete sets.**

- $I_1 = \{a_1, a_2, c_1, b_1\}$ is independent: $\underline{R}(I_1) = \{a_1, a_2, c_1\} \subseteq X$ (the $B$-block is not fully included).
- $I_2 = \{b_1, b_2\}$ is dependent: $\underline{R}(I_2) = \{b_1, b_2\} \not\subseteq X$ (the uncleared supplier-block is fully included).

**Definition 4.4.3** (Partition-circuit rough matroid (the case $X = \emptyset$))**.** Let $(U, R)$ be a Pawlak approximation space. The *partition-circuit rough matroid* is the matroid

$$M_R \; := \; (U, \mathcal{I}_R), \qquad \mathcal{I}_R \; := \; \{\, I \subseteq U \mid \underline{R}(I) = \emptyset \,\}.$$

Equivalently, its circuit family is exactly the partition into equivalence classes:

$$\mathcal{C}(M_R) \; = \; U/R.$$

**Example 4.4.4** (Partition-circuit rough matroid: "never take an entire department")**.** Let $U$ be a set of employees assigned to departments:

$$U = \{a_1, a_2, \; b_1, b_2, \; c_1\},$$

where $a_\bullet$ belong to department $A$, $b_\bullet$ to department $B$, and $c_1$ to department $C$. Let $R$ be the equivalence relation

$$x\,R\,y \quad \Longleftrightarrow \quad x \text{ and } y \text{ are in the same department},$$

so that

$$U/R = \big\{\{a_1, a_2\}, \; \{b_1, b_2\}, \; \{c_1\}\big\}.$$

Consider the partition-circuit rough matroid

$$M_R = (U, \mathcal{I}_R), \qquad \mathcal{I}_R = \{\, I \subseteq U \mid \underline{R}(I) = \emptyset \,\}.$$

**Interpretation.** Since $\underline{R}(I)$ is the union of those department-blocks fully contained in $I$, the condition $\underline{R}(I) = \emptyset$ means: *the selected team $I$ does not completely contain any department.* Equivalently, every department contributes *at most* $|B| - 1$ members from each block $B \in U/R$.

**Concrete sets and circuits.**

- $I_1 = \{a_1, b_1, c_1\}$ is independent, because it contains no full block.

- $I_2 = \{a_1, a_2\}$ is dependent, and it is a *circuit* (a minimal dependent set), because it is exactly one equivalence class.

Indeed, the circuit family is precisely the partition:

$$\mathcal{C}(M_R) = U/R = \big\{\{a_1, a_2\}, \; \{b_1, b_2\}, \; \{c_1\}\big\}.$$

## 4.5 Soft Rough Graph

Soft rough graph represents a graph via soft set approximations of vertices and edges, yielding lower and upper subgraphs parameters [312].

**Definition 4.5.1** (Soft rough graph)**.** [312] Let $G = (V, E)$ be a simple (undirected) graph and let $A$ be a nonempty set of parameters. A *soft set over* $V$ is a mapping $F : A \to \mathcal{P}(V)$, and a *soft set over* $E$ is a mapping $K : A \to \mathcal{P}(E)$.

For a target vertex set $X \subseteq V$, define the *soft rough* lower and upper vertex approximations

$$F_*(X) := \{\, v \in V \mid \exists a \in A :\ v \in F(a) \subseteq X \,\},$$

$$F^*(X) := \{\, v \in V \mid \exists a \in A :\ v \in F(a) \text{ and } F(a) \cap X \neq \varnothing \,\}.$$

Similarly, for a target edge set $Y \subseteq E$, define the *soft rough* lower and upper edge approximations

$$K_*(Y) := \{\, e \in E \mid \exists a \in A :\ e \in K(a) \subseteq Y \,\},$$

$$K^*(Y) := \{\, e \in E \mid \exists a \in A :\ e \in K(a) \text{ and } K(a) \cap Y \neq \varnothing \,\}.$$

For any $S \subseteq V$, write

$$E[S] := \{\, uv \in E \mid u \in S,\ v \in S \,\}$$

for the set of edges induced by $S$. The *lower* and *upper* soft rough subgraphs associated with $(X, Y)$ are defined by

$$H_*(X,Y) := \big(F_*(X),\ K_*(Y) \cap E[F_*(X)]\big), \qquad H^*(X,Y) := \big(F^*(X),\ K^*(Y) \cap E[F^*(X)]\big),$$

so that $H_*(X, Y)$ and $H^*(X, Y)$ are subgraphs of $G$.

A *soft rough graph* (induced by $F, K, A$ and the targets $X \subseteq V$, $Y \subseteq E$) is the pair

$$\widetilde{G}(X,Y) := \big(H_*(X,Y),\ H^*(X,Y)\big).$$

A common choice is $Y := E[X]$ (edges induced by $X$), in which case we write $\widetilde{G}(X) := \widetilde{G}(X, E[X])$. The family of all soft rough graphs of $G$ (over all admissible data) is denoted by $\mathrm{SRG}(G)$.

**Example 4.5.2** (Soft rough graph for suspicious users in a small social network)**.** Consider a small friendship network $G = (V, E)$, where

$$V = \{1, 2, 3, 4, 5\}$$

represents user accounts and

$$E = \{12, 13, 23, 24, 34, 45\}$$

represents undirected friendship links (we write $ij$ for the edge $\{i, j\}$).

Let the parameter set be

$$A = \{a_1, a_2, a_3\},$$

where $a_1$ = "shares the same device fingerprint as a flagged account", $a_2$ = "IP-subnet overlap with flagged accounts", $a_3$ = "similar posting pattern (burstiness)".

Define a soft set over vertices $F : A \to \mathcal{P}(V)$ by

$$F(a_1) = \{2, 3\}, \qquad F(a_2) = \{3, 4\}, \qquad F(a_3) = \{4, 5\}.$$

Define a soft set over edges $K : A \to \mathcal{P}(E)$ by selecting the edges inside each $F(a)$:

$$K(a_1) = E[F(a_1)] = \{23\}, \qquad K(a_2) = E[F(a_2)] = \{34\}, \qquad K(a_3) = E[F(a_3)] = \{45\}.$$

Suppose the analysts' *target suspicious vertex set* is

$$X = \{3, 4\} \subseteq V,$$

and choose the common target edge set $Y := E[X] = \{34\} \subseteq E$.

**Soft rough vertex approximations.** By Definition 4.5.1,

$$F_*(X) = \{ v \in V \mid \exists a \in A : \ v \in F(a) \subseteq X \}.$$

Since $F(a_2) = \{3, 4\} \subseteq X$ (while $F(a_1) = \{2, 3\} \not\subseteq X$ and $F(a_3) = \{4, 5\} \not\subseteq X$), we get

$$F_*(X) = \{3, 4\}.$$

Also,

$$F^*(X) = \{ v \in V \mid \exists a \in A : \ v \in F(a), \ F(a) \cap X \neq \varnothing \}.$$

Here every $F(a_i)$ intersects $X$: $F(a_1) \cap X = \{3\}$, $F(a_2) \cap X = \{3, 4\}$, $F(a_3) \cap X = \{4\}$, hence

$$F^*(X) = \{2, 3, 4, 5\}.$$

**Soft rough edge approximations.** For $Y = \{34\}$,

$$K_*(Y) = \{ e \in E \mid \exists a \in A : \ e \in K(a) \subseteq Y \}.$$

Only $K(a_2) = \{34\} \subseteq Y$, so

$$K_*(Y) = \{34\}.$$

Moreover,

$$K^*(Y) = \{ e \in E \mid \exists a \in A : \ e \in K(a), \ K(a) \cap Y \neq \varnothing \},$$

and again only $K(a_2)$ intersects $Y$, so

$$K^*(Y) = \{34\}.$$

**Lower and upper soft rough subgraphs.** Compute induced edges:

$$E[F_*(X)] = E[\{3, 4\}] = \{34\}, \qquad E[F^*(X)] = E[\{2, 3, 4, 5\}] = \{23, 24, 34, 45\}.$$

Therefore,

$$H_*(X, Y) = \big(F_*(X), \ K_*(Y) \cap E[F_*(X)]\big) = \big(\{3, 4\}, \ \{34\}\big),$$
$$H^*(X, Y) = \big(F^*(X), \ K^*(Y) \cap E[F^*(X)]\big) = \big(\{2, 3, 4, 5\}, \ \{34\}\big).$$

Hence the soft rough graph is

$$\widetilde{G}(X, Y) = \big(H_*(X, Y), \ H^*(X, Y)\big).$$

Vertex 3 and 4 are *definitely suspicious* because one parameter ($a_2$) isolates exactly $\{3, 4\}$. Vertices 2 and 5 are *possibly suspicious* because they appear in parameter-blocks that overlap the suspicious set.

# Chapter 5

# Conclusion

In this book, we surveyed rough set theory and a broad range of its major extensions, aiming to provide a compact map of the field from classical Pawlak approximations to relation-/granulation-based generalizations, multi-granulation and weighted models, neighborhood and probabilistic variants, and uncertainty-aware frameworks (e.g., fuzzy, intuitionistic fuzzy, neutrosophic, and plithogenic rough sets). We also collected several closely related viewpoints—including rough graphs, rough topological spaces, rough groups, rough matroids, and soft rough graphs—to help connect rough approximations with adjacent mathematical structures.

Looking ahead, we hope that future work will advance both (i) *application-oriented case studies* that clarify which rough models are most effective for particular data characteristics (e.g., imbalance, noise, heterogeneous feature importance, or evolving/streaming observations), and (ii) *algorithmic development*, including scalable attribute reduction, efficient computation of (possibly multi-stage) approximations, and reproducible benchmarking and software implementations for modern decision support and explainable learning.

# Disclaimer

## Funding

This study did not receive any financial or external support from organizations or individuals.

## Acknowledgments

We extend our sincere gratitude to everyone who provided insights, inspiration, and assistance throughout this research. We particularly thank our readers for their interest and acknowledge the authors of the cited works for laying the foundation that made our study possible. We also appreciate the support from individuals and institutions that provided the resources and infrastructure needed to produce and share this book. Finally, we are grateful to all those who supported us in various ways during this project.

## Data Availability

This research is purely theoretical, involving no data collection or analysis. We encourage future researchers to pursue empirical investigations to further develop and validate the concepts introduced here.

## Ethical Approval

As this research is entirely theoretical in nature and does not involve human participants or animal subjects, no ethical approval is required.

## Use of Generative AI and AI-Assisted Tools

We use generative AI and AI-assisted tools for tasks such as English grammar checking, and We do not employ them in any way that violates ethical standards.

## Conflicts of Interest

The authors confirm that there are no conflicts of interest related to the research or its publication.

## Disclaimer

This work presents theoretical concepts that have not yet undergone practical testing or validation. Future researchers are encouraged to apply and assess these ideas in empirical contexts. While every effort has been made to ensure accuracy and appropriate referencing, unintentional errors or omissions may still exist. Readers are advised to verify referenced materials on their own. The views and conclusions expressed here are the authors' own and do not necessarily reflect those of their affiliated organizations.

# Appendix (List of Tables)

*

*Handbook of Rough Set Extensions and Uncertainty Models* provides a systematic and structured survey of rough set theory and its major extensions developed to model uncertainty, vagueness, granularity, and indiscernibility in data-driven systems. Rooted in Pawlak's original approximation framework, rough set theory distinguishes between what can be determined with certainty and what remains only possible under limited information. Over time, numerous generalizations have emerged to address increasingly complex forms of uncertainty.

This handbook serves as a comprehensive "model map" of the rough-set landscape. It organizes and analyzes a wide spectrum of variants, including equivalence-based, tolerance-based, covering-based, neighborhood-based, probabilistic, weighted, multi-granulation, hierarchical, and sequential rough sets. In addition, hybrid uncertainty models—such as fuzzy rough sets, intuitionistic fuzzy rough sets, neutrosophic rough sets, plithogenic rough sets, and other generalized formulations—are presented within a unified conceptual taxonomy.